\documentclass[a4paper,12pt,times,PageStyleII,twoside,custommargin,custombib,print]{PhDThesisPSnPDF}
\graphicspath{{AcceleratingDNNs/AnalogHardwareAccelerators/HardwareModels/non-associativity/imgs/} {AcceleratingDNNs/AnalogHardwareAccelerators/HardwareModels/whitebox/imgs/} {AcceleratingDNNs/AnalogHardwareAccelerators/robustness/hardening/imgs/} {AcceleratingDNNs/AnalogHardwareAccelerators/robustness/vant/imgs/} {AcceleratingDNNs/AnalogHardwareAccelerators/robustness/walking-noise/imgs/} {AcceleratingDNNs/ModelCompression/Galen/imgs/} {AcceleratingDNNs/ModelCompression/rennfes/imgs/} {BayesianNeuralNetworks/bnns/imgs/} {BayesianNeuralNetworks/ensembles/imgs/} {BayesianNeuralNetworks/pfp/imgs/} {BayesianNeuralNetworks/photonics/imgs/}}

\ifsetCustomMargin
  \RequirePackage[left=37mm,right=30mm,top=35mm,bottom=30mm]{geometry}
  \setFancyHdr %
\fi

\ifsetCustomFont
\fi

\RequirePackage[labelsep=space,tableposition=top]{caption}
\usepackage{booktabs} %
\usepackage{multirow}

\usepackage{tabularx}

\usepackage{siunitx} %

\usepackage{xstring}   %
\usepackage{pgffor}    %

\usepackage{setspace} %

\RequirePackage[backend=biber, style=ieee, citestyle=ieee, sorting=ydnt, natbib=false, backref=false]{biblatex}
\DeclareLanguageMapping{english}{english-apa}

\renewbibmacro*{doi+eprint+url}{%
  \printfield{doi}%
  \newunit\newblock
  \printfield[eprint:arxiv]{eprint}%
  \newunit\newblock
  \iffieldundef{doi}{\printfield{url}}{}%
}

\DeclareFieldFormat{eprint:arxiv}{%
  arXiv\addcolon\space
  \ifhyperref{\href{https://arxiv.org/abs/#1}{#1}}{#1}%
}

\AtEveryBibitem{%
  \iffieldundef{doi}{}{\clearfield{url}}%
  \clearfield{eprintclass}%
  \clearfield{primaryclass}%
}

\DeclareFieldFormat{url}{\url{#1}}

\usepackage{enumitem}

\usepackage[acronym,nomain,toc]{glossaries}

\usepackage{svg}
\usepackage{subcaption}
\usepackage[export]{adjustbox}

\usepackage[utf8]{inputenc}
\usepackage[english,ngerman]{babel} %
\usepackage{pifont}

\usepackage{algorithm,algpseudocode}
\algdef{SE}[DOWHILE]{Do}{doWhile}{\algorithmicdo}[1]{\algorithmicwhile\ #1}%

\usepackage[flushleft]{threeparttable}

\usepackage[super]{nth}

\definecolor{gray75}{gray}{0.75}

\usepackage{titlesec}

\usepackage{mathtools}

\usepackage{listings}
\definecolor{lstgray}{rgb}{0.95,0.95,0.95}

\usepackage{lmodern}
\usepackage{titlesec}
\usepackage{microtype}
\usepackage{tikz}

\usetikzlibrary{arrows.meta,positioning}

\usepackage[palatino]{quotchap}
\usepackage{todonotes}
\usepackage{import}
\usepackage{amsmath}

\usepackage{threeparttable}
\usepackage{bm}
\usepackage{adjustbox}
\usepackage{pgfplots} %
\usepackage{pgf-pie} %
\usepackage{graphicx}
\usepackage{subcaption}
\usepackage[normalem]{ulem}
\usepackage{multirow}
\usepackage{pifont} %
\usepackage{environ}
\usetikzlibrary{spy,positioning,calc,shapes.geometric}

\usepackage{epigraph} %

\newcommand{\matB}{\mathbf{B}}

\newcommand{\matW}{\mathbf{W}}

\newcommand{\underlineitalic}[1]{\underline{#1}}

\PassOptionsToPackage{hyphens}{url}\usepackage{hyperref}
\usepackage{etoolbox}

\newtoggle{bbx@minimal}

\AtEveryBibitem{%
  \iftoggle{bbx@minimal}{%
    \clearfield{url}%
    \clearfield{doi}%
    \clearfield{eprint}%
    \clearfield{isbn}%
    \clearfield{issn}%
    \clearlist{publisher}%
    \clearfield{note}%
    \clearfield{pagetotal}%
    \clearfield{address}%
  }{}%
}

\usepackage{etoolbox}
\newtoggle{ownpubs@intro}

\newcommand{\reprofrom}[1]{Reproduced with permission from~\cite{#1}.}
\newcommand{\adjfrom}[1]{Adjusted from~\cite{#1}.}

\makeglossaries

\newacronym{iot}{IoT}{Internet of Things}
\newacronym{ml}{ML}{machine learning}
\newacronym{dl}{DL}{deep learning}
\newacronym{nn}{NN}{neural network}
\newacronym{dnn}{DNN}{deep neural network}
\newacronym{bnn}{BNN}{Bayesian neural network}
\newacronym{cnn}{CNN}{convolutional neural network}
\newacronym{rnn}{RNN}{recurrent neural network}
\newacronym{snn}{SNN}{spiking neural network}
\newacronym{ann}{ANN}{artificial neural network}
\newacronym{llm}{LLM}{large language model}
\newacronym{nlp}{NLP}{natural language processing}

\newacronym{sgd}{SGD}{stochastic gradient descent}
\newacronym{vi}{VI}{variational inference}
\newacronym{svi}{SVI}{stochastic variational inference}
\newacronym{mcmc}{MCMC}{Markov chain Monte Carlo}
\newacronym{hmc}{HMC}{Hamiltonian Monte Carlo}
\newacronym{nuts}{NUTS}{No-U-Turn Sampler}
\newacronym{elbo}{ELBO}{evidence lower bound}
\newacronym{pfp}{PFP}{Probabilistic Forward Pass}
\newacronym{ste}{STE}{straight-through estimator}
\newacronym{sam}{SAM}{Sharpness-Aware Minimization}
\newacronym{ddpg}{DDPG}{Deep Deterministic Policy Gradient}

\newacronym{qat}{QAT}{quantization-aware training}
\newacronym{ptq}{PTQ}{post-training quantization}
\newacronym{mac}{MAC}{multiply–accumulate}
\newacronym{bops}{BOPs}{bit operations}
\newacronym{cmp}{CMP}{compression method parameter}
\newacronym{nas}{NAS}{neural architecture search}
\newacronym{amc}{AMC}{AutoML for Model Compression}
\newacronym{haq}{HAQ}{Hardware-Aware Quantization}
\newacronym{hil}{HIL}{hardware-in-the-loop}

\newacronym{de}{DE}{Deep Ensemble}
\newacronym{mcdo}{MCDO}{Monte Carlo Dropout}
\newacronym{rde}{RDE}{Repulsive Deep Ensemble}
\newacronym{rlle}{RLLE}{Repulsive Last-Layer Ensemble}
\newacronym{svgd}{SVGD}{Stein Variational Gradient Descent}
\newacronym{povi}{POVI}{Particle-Optimization Variational Inference}
\newacronym{LLPOVI}{LL-POVI}{Last-Layer Particle-Optimization Variational Inference}
\newacronym{fLLPOVI}{fLL-POVI}{Function-Space Last-Layer Particle-Optimization Variational Inference}
\newacronym{RLLPOVI}{RLL-POVI}{Repulsive Last-Layer Particle-Optimization Variational Inference}

\newacronym{nll}{NLL}{negative log-likelihood}
\newacronym{ece}{ECE}{expected calibration error}
\newacronym{sme}{SME}{softmax entropy}
\newacronym{mi}{MI}{mutual information}
\newacronym{ess}{ESS}{effective sample size}
\newacronym{euc}{EUC}{Epistemic Uncertainty Coefficient}
\newacronym{auroc}{AUROC}{area under the receiver operating characteristic curve}
\newacronym{rauc}{rAUC}{relative area under the accuracy–noise curve}
\newacronym{roc}{ROC}{receiver operating characteristic}
\newacronym{tpr}{TPR}{true positive rate}
\newacronym{fpr}{FPR}{false positive rate}
\newacronym{eu}{EU}{epistemic uncertainty}
\newacronym{au}{AU}{aleatoric uncertainty}

\newacronym{re}{RE}{representational efficiency}
\newacronym{ce}{CE}{computational efficiency}
\newacronym{pq}{PQ}{prediction quality}
\newacronym{bs}{BS}{batch size}
\newacronym{flops}{FLOPs}{floating-point operations}

\newacronym{cpu}{CPU}{Central Processing Unit}
\newacronym{gpu}{GPU}{Graphics Processing Unit}
\newacronym{fpga}{FPGA}{Field-Programmable Gate Array}
\newacronym{tpu}{TPU}{Tensor Processing Unit}
\newacronym{npu}{NPU}{Neural Processing Unit}
\newacronym{asic}{ASIC}{Application-Specific Integrated Circuit}
\newacronym{soc}{SoC}{System-on-Chip}
\newacronym{arm}{ARM}{Advanced RISC Machines}
\newacronym{simd}{SIMD}{Single Instruction Multiple Data}
\newacronym{isa}{ISA}{Instruction Set Architecture}
\newacronym{lut}{LUT}{lookup table}

\newacronym{bss2}{BSS-2}{BrainScaleS-2}
\newacronym{ota}{OTA}{operational transconductance amplifier}
\newacronym{adc}{ADC}{analog-to-digital converter}
\newacronym{dac}{DAC}{digital-to-analog converter}
\newacronym{rram}{RRAM}{resistive random-access memory}
\newacronym{pcm}{PCM}{phase-change material}
\newacronym{mzi}{MZI}{Mach–Zehnder interferometer}
\newacronym{wdm}{WDM}{wavelength-division multiplexing}
\newacronym{thz}{THz}{terahertz}
\newacronym{ase}{ASE}{amplified spontaneous emission}
\newacronym{gst}{GST}{germanium–antimony–telluride}
\newacronym{cmos}{CMOS}{complementary metal–oxide–semiconductor}

\newacronym{tvm}{TVM}{Tensor Virtual Machine}
\newacronym{mlir}{MLIR}{Multi-Level Intermediate Representation}
\newacronym{xla}{XLA}{Accelerated Linear Algebra}
\newacronym{ir}{IR}{intermediate representation}
\newacronym{tir}{TensorIR}{tensor intermediate representation}
\newacronym{te}{TE}{tensor expression}
\newacronym{api}{API}{Application Programming Interface}
\newacronym{jit}{JIT}{just-in-time}
\newacronym{ppl}{PPL}{Probabilistic Programming Language}

\newacronym{kl}{KL}{Kullback–Leibler divergence}
\newacronym{pdf}{PDF}{probability density function}
\newacronym{mse}{MSE}{mean squared error}
\newacronym{std}{STD}{standard deviation}
\newacronym{rbf}{RBF}{radial basis function}
\newacronym{sde}{SDE}{stochastic differential equation}

\newacronym{gsc}{GSC}{Google Speech Commands}
\newacronym{wrn}{WRN}{Wide Residual Network}
\newacronym{vgg}{VGG}{Visual Geometry Group}
\newacronym{mlp}{MLP}{multi-layer perceptron}

\newacronym{vant}{VANT}{Variance-Aware Noisy Training}
\newacronym{bn}{BN}{batch normalization}
\newacronym{af}{AF}{activation function}
\newacronym{ood}{OOD}{out-of-distribution}
\newacronym{id}{ID}{in-distribution}
\newacronym{kde}{KDE}{Kernel Density Estimation}
\newacronym{kan}{KAN}{Kolmogorov–Arnold Network}
\newacronym{relu}{ReLU}{Rectified Linear Unit}

\title{Resource-Efficient and Robust Inference of Deep and Bayesian Neural Networks on Embedded and Analog Computing Platforms}

\subtitle{A Study of Compression, Robustness, and Bayesian Inference from Embedded Processors to Analog Photonic Accelerators}

\author{Bernhard Klein}

\dept{Institute of Computer Engineering}

\university{Heidelberg University}
\supervisor{Prof. Dr. Holger Fröning}

\supervisorlinewidth{0.45\textwidth}

\degreetitle{Doctor of Natural Sciences}

\subject{Doctoral Dissertation — Computer Science, Heidelberg University}
\keywords{Bayesian Neural Networks, Probabilistic Machine Learning, 
Analog Computing, Robustness, Energy-Efficient AI, Hardware-Aware Training, 
Heidelberg University}

\ifdefineAbstract
\fi

\ifdefineChapter
\fi

\begin{document}

\frontmatter
\pagestyle{empty}      

\begin{titlepage}
  
	\pagestyle{empty}
 
  	\centering
	\LARGE{INAUGURAL-DISSERTATION}  \\
	\vspace{2cm}
	\large{
	\normalsize{submitted to the} \\[5mm]
	\normalsize{Combined Faculty of Mathematics, Engineering and Natural Sciences} \\[5mm]
	\normalsize{of  the} \\[5mm]
	\Large{Ruprecht–Karls-University} \\
	\Large{Heidelberg} \\[5mm]
	\normalsize{for the degree of} \\[5mm]
	\Large{Doctor of Natural Sciences}
	} \\
	\vspace{5cm}
	\normalsize{put forward by} \\
	\Large{Bernhard Klein, M.Sc.}\\
	\vspace{1cm}
	\normalsize{born in \\Böblingen, Baden-Württemberg} \\
	\vfill
	\normalsize{Date of oral exam: ...........................}
	
	\newpage
	
	\vspace*{15cm}

	\vspace*{1cm}

  	\newpage
	\maketitle  
  
\end{titlepage}

\clearpage{}%

\begin{dedication} 

\begin{center}
\itshape
To my family and friends,\\
and especially to my loving parents and sister,\\
for their unwavering support and belief in me.
\end{center}

\end{dedication}

\clearpage{}%
\clearpage{}%
\begin{abstract}
\begingroup
\setstretch{1.00}

While modern machine learning has transformed numerous application domains, its growing computational demands increasingly constrain scalability and efficiency, particularly on embedded and resource-constrained platforms.  
In practical deployments, neural networks must not only operate efficiently but also provide reliable predictions when faced with distributional changes or previously unseen data.  
Bayesian neural networks offer a principled framework for quantifying uncertainty, but their higher computational demands further compound these challenges.

This work advances \emph{resource-efficient and robust inference} for both conventional and Bayesian neural networks through the joint pursuit of algorithmic and hardware efficiency.  
The former reduces computational cost through model compression and approximate Bayesian inference, while the latter optimizes mapping to digital accelerators and explores novel analog hardware platforms, bridging algorithmic optimization and physical realization.  

The first contribution introduces the \emph{Galen} framework, which performs automatic, layer-specific compression guided by sensitivity analysis and hardware-in-the-loop feedback, jointly optimizing quantization and pruning to balance accuracy and efficiency on embedded devices.  
As analog accelerators offer additional efficiency gains at the cost of noise, their modeling exposes device imperfections, while a layer-wise analysis reveals how networks learn to tolerate such effects during training.  
This work extends noisy training to nonstationary conditions, thereby enhancing robustness and stability in analog hardware.

A complementary line of work advances probabilistic inference.  
Building on insights into Bayesian-neural-network design and training, this work develops analytic and ensemble-based approximations that replace costly sampling, integrates them into a compiler stack, and optimizes them for probabilistic inference on embedded hardware.
Finally, \emph{probabilistic photonic computing} introduces a novel paradigm in which controlled analog noise serves as an intrinsic entropy source, enabling ultrafast and energy-efficient probabilistic inference directly in hardware.  

Together, these studies demonstrate how \emph{efficiency and reliability can be advanced jointly} through the co-design of algorithms, compilers, and hardware, laying the foundation for the next generation of \emph{trustworthy and energy-efficient} machine-learning systems.  

\endgroup
\end{abstract}

\begin{otherlanguage}{ngerman}
\begin{abstractde}
\begingroup
\setstretch{1.00}

Moderne Verfahren des maschinellen Lernens haben zahlreiche Anwendungsfelder grundlegend verändert.  
Mit dem stetig wachsenden Rechenbedarf stoßen sie jedoch zunehmend an Grenzen hinsichtlich Skalierbarkeit und Effizienz – insbesondere auf eingebetteten und ressourcenbeschränkten Plattformen.
Bei der Anwendung in realen Systemen müssen neuronale Netze nicht nur effizient arbeiten, sondern auch unter sich verändernden Datenverteilungen oder bei bislang unbekannten Datenpunkten verlässliche Vorhersagen liefern.  
Bayessche neuronale Netze bieten hierfür einen konsistenten theoretischen Rahmen zur Quantifizierung von Unsicherheiten, ihr zusätzlicher Rechenaufwand verstärkt diese Herausforderungen jedoch weiter.  

Diese Arbeit verfolgt das Ziel einer \emph{ressourceneffizienten und robusten Inferenz} sowohl für konventionelle als auch für bayessche neuronale Netze durch die gemeinsame Optimierung von Algorithmen und Hardware.  
Die algorithmische Effizienz wird durch Modellkompression und approximative bayessche Verfahren verbessert, während die Hardwareeffizienz sowohl die Abbildung auf digitale Beschleuniger als auch die Erforschung neuartiger analoger Plattformen umfasst und damit eine Brücke zwischen algorithmischer Optimierung und hardwareseitiger Realisierung schlägt.  

Den ersten Beitrag stellt das \emph{Galen}-Framework dar, das eine automatische, feinaufgelöste Kompression auf Grundlage von Sensitivitätsanalysen und Hardware-in-the-Loop-Rückkopplung durchführt.  
Quantisierung und Pruning werden dabei gemeinsam optimiert, um Genauigkeit und Effizienz auf eingebetteten Systemen in Einklang zu bringen.  
Da analoge Beschleuniger zusätzliche Effizienzgewinne auf Kosten von Rechenrauschen bieten, werden ihre Nichtidealitäten modelliert.  
Eine Analyse auf Ebene der Netzwerkschichten zeigt, wie neuronale Netze lernen, solche Störungen während des Trainings zu tolerieren.  
Darauf aufbauend erweitert diese Arbeit das Training mit Rauschinjektion auf nichtstationäre Bedingungen, wodurch Robustheit und Stabilität in analogen Beschleunigern gesteigert werden.  

Ein weiterer Schwerpunkt liegt auf der probabilistischen Inferenz.  
Aufbauend auf Erkenntnissen zum Entwurf und Training bayesscher neuronaler Netze werden effiziente analytische und ensemblebasierte Approximationen entwickelt, die aufwändiges Sampling ersetzen und in einer Compiler-Infrastruktur mit optimierten probabilistischen Operatoren für eingebettete Hardware umgesetzt sind.  
Schließlich wird mit dem \emph{probabilistischen photonischen Rechnen} ein neuartiges Paradigma eingeführt, bei dem kontrolliertes analoges Rauschen als intrinsische Entropiequelle dient und ultraschnelle, energieeffiziente probabilistische Inferenz direkt in photonischer Hardware ermöglicht.  

Zusammenfassend zeigt diese Arbeit, dass \emph{Effizienz und Zuverlässigkeit gemeinsam gesteigert werden können}, wenn Algorithmen, Compiler und Hardware als integriertes System konzipiert werden.  
Damit wird das Fundament für die nächste Generation vertrauenswürdiger und energieeffizienter Systeme des maschinellen Lernens gelegt.

\endgroup
\end{abstractde}
\end{otherlanguage}
\selectlanguage{english}    %

\clearpage{}%

\setcounter{tocdepth}{1}
\tableofcontents

\printnomenclature

\mainmatter
\pagestyle{fancy}    
\clearpage{}%
\chapter{Introduction}  %

\noindent
The extraordinary progress of \acrfull{ml} over the past decade has been driven by increasingly large models, massive datasets, and powerful compute infrastructures.  
While these advances have led to remarkable performance across domains such as computer vision~\cite{krizhevsky2012imagenet}, speech recognition~\cite{hannun2014deep}, and \acrlong{nlp}~\cite{devlin2019bert}, they come at the cost of substantial computational and energy demands.  
The growing size and complexity of modern \acrfullpl{dnn}~\cite{villalobos2022modelsizes,jordan2015mltrends}, together with the end of \emph{Dennard scaling}---the breakdown of power efficiency gains from transistor miniaturization~\cite{esmaeilzadeh2011dennardscaling}---have made throughput and energy efficiency critical bottlenecks for both large-scale training~\cite{narayanan2021efficient} and on-device inference~\cite{howard2017mobilenets,han2016deepcompression}.  
These efficiency constraints have intensified the need for methods that preserve predictive performance under limited computational and energy budgets, driving progress in both algorithmic optimization and hardware specialization.  

At the same time, \acrshortpl{dnn}, despite their empirical success, fundamentally lack a principled mechanism to quantify uncertainty.  
In classification tasks, for example, the softmax function is typically used to transform output logits into normalized scores that are often interpreted as probabilities.  
However, these softmax-derived values do not represent true probabilities in a Bayesian sense; rather, they are uncalibrated confidence scores without a theoretical grounding in probability theory~\cite{guo2017calibration}.  
This shortcoming becomes particularly evident when models are confronted with inputs that differ from their training distribution, commonly referred to as \acrfull{ood} data~\cite{ovadia2019can}.  
In such cases, neural networks often produce highly confident yet incorrect predictions, lacking any mechanism to signal elevated uncertainty or to acknowledge that an input lies outside their domain of competence.  

A theoretical framework to equip neural architectures with this capability is offered by Bayesian statistics, giving rise to \emph{\acrfullpl{bnn}}~\cite{jospin2022handsonBNN,blundell2015}.
By treating model parameters as probability distributions rather than fixed values, \acrshortpl{bnn} yield predictive distributions that quantify the uncertainty associated with each prediction, providing a principled measure of model confidence.
Such uncertainty estimates enable more informed decision-making, improve robustness under distributional shift, and are essential for deploying neural networks in safety-critical applications.
However, \acrshortpl{bnn} are considerably more computationally demanding than conventional neural networks, since both the estimation of parameter distributions during training and the computation of predictive uncertainty during inference require multiple stochastic forward passes.
This repeated sampling makes their use on embedded and energy-constrained systems particularly challenging.

Consequently, the central question motivating this work is how to realize \emph{resource-efficient and reliable} neural network inference that combines high computational efficiency with trustworthy predictive behavior, encompassing both deterministic and probabilistic models.

\vspace{1.5em}
This work approaches the goal of resource-efficient inference from two complementary directions. 
The first focuses on \emph{reducing the computational workload through algorithmic simplification}, achieving comparable performance with fewer operations through techniques such as model compression and probabilistic approximation. 
The second addresses the \emph{efficient realization of computations} by improving execution efficiency on existing hardware through optimized mapping of computations to hardware resources and by employing specialized architectures---such as analog matrix–multiply accelerators---that execute fundamental operations with much higher energy efficiency.
These two directions are closely connected: algorithmic simplifications can ease hardware demands, while hardware characteristics in turn influence algorithm design and the need for robustness. 
This work therefore applies this combined perspective first to deterministic \acrlongpl{nn}, developing methods for automatic compression on embedded processors and, in parallel, analyzing the execution characteristics of analog matrix–multiply accelerators.
It then extends this approach to \acrlongpl{bnn}, where Bayesian approximations, efficient operator–to–hardware mappings, and photonic computing architectures are explored to realize scalable and uncertainty-aware inference.
Together, these studies support a unified view of neural network design in which algorithms, compilers, and hardware are co-optimized as elements of a single integrated system.
Building on this integrated perspective, the following discussion examines how efficiency and reliability can be advanced in practice.

\vspace{0.5em}
A well-established approach to enhancing efficiency is the simplification of neural networks through compression techniques. 
Methods such as quantization~\cite{jacob2018quantization,Zhang2018}, pruning~\cite{LeCun1989,Hassibi1992,Han2015}, and knowledge distillation~\cite{hinton2015knowledgedistillation}---discussed in more detail in Chapter~\ref{ch:rennfes}---are widely used to reduce model size, latency, and power consumption in embedded systems.
In practice, it is often unclear how much compression an individual layer can tolerate and how these choices translate into measurable performance gains on the target hardware; consequently, conventional compression pipelines typically apply uniform compression policies across layers, leaving a large degree of layer-specific flexibility unused. 
Building on this groundwork, \emph{Galen}~\cite{krieger2023galen} introduces an automatic, hardware-aware compression framework that combines layer-wise sensitivity analysis with measured latencies on embedded processors.
By adapting compression strength per layer and per hardware architecture, Galen leverages this flexibility to generate heterogeneous pruning and quantization strategies optimized jointly for accuracy preservation and inference speed.

Beyond reducing the computational cost of neural network inference, further efficiency gains can be achieved through alternative computing paradigms.
Analog accelerators promise substantial improvements in performance per watt, as they perform the fundamental \acrfull{mac} operation of neural networks at orders of magnitude lower energy than digital implementations~\cite{Murmann2021}. 
This advantage arises from executing computation directly in the physics of the device rather than through digital abstraction---for example, by charge accumulation in electrical \acrshort{cmos}-based systems~\cite{schemmel2021accelerated} or by optical interference in photonic circuits~\cite{Shen2017photonic}. 
However, analog computation is inherently affected by circuit-level imperfections such as nonlinearities, saturation effects, leakage, crosstalk, and various sources of noise. 
Moreover, these characteristics vary across devices due to manufacturing variations and evolve over time under the influence of environmental factors such as temperature changes.
Unlike digital processors, analog hardware does not abstract away these imperfections, and their effects accumulate during computation, requiring algorithms that can tolerate or adapt to them. 
To achieve such adaptation, neural networks are often trained directly on the hardware in a process known as hardware-in-the-loop training, where forward passes are executed on the physical device to expose the model to real analog imperfections. 
While this approach effectively compensates for device-specific distortions, its practicality is often limited by the throughput and availability of the hardware.
To better understand these effects and to accelerate this process, this work develops a white-box model of analog neural accelerators~\cite{klein2021bss2whitebox}, exemplified by the \acrfull{bss2}~\cite{schemmel2021accelerated} platform, which serves as a representative analog matrix–multiply accelerator.
During this modeling process, it was observed that arithmetic operations implemented in analog hardware do not generally preserve associativity~\cite{kuhn2023nonassociativity}, a fundamental property assumed in mathematics and relied upon across all domains of computer science---discussed in more detail in~Chapter~\ref{ch:analog:hardware_models}.

Building on these models, this work systematically investigates the robustness of neural networks to analog nonidealities through controlled noise injection. 
The resulting \emph{Walking Noise} framework~\cite{borras2024walkingnoiseECML} provides a methodology to map layer-wise sensitivity to additive and multiplicative noise, enabling the identification of particularly robust or vulnerable layers and offering insight into how neural networks learn to tolerate such disturbances through noisy training.
In a comparative study of hardening strategies~\cite{wang2025hardening}, various approaches were evaluated to improve the robustness of neural networks against noisy computations. 
Among them, noisy training proved to be the most effective technique, but only when the noise characteristics during training were consistent with those encountered during inference---a condition that is difficult to maintain in systems subject to dynamically changing factors such as temperature-induced drift.
To address this challenge and further improve resilience under realistic operating conditions, \emph{\acrfull{vant}}~\cite{wang2025vant} introduces adaptive noise scaling that enables stable performance under environmental fluctuations. 
Together, these results demonstrate that robustness to analog imperfections requires incorporating hardware characteristics and noise behavior directly into the training process rather than compensating for them after deployment.

\vspace{0.5em}
While these approaches focus on efficient and reliable execution of deterministic models, many applications also require principled uncertainty estimation to quantify the reliability of predictions. 
\acrlongpl{bnn} address this by representing network weights as probability distributions, producing predictive posteriors whose variance decomposes into epistemic and aleatoric uncertainty~\cite{jospin2022handsonBNN,roth2024jmlr}. 
However, exact Bayesian inference is computationally prohibitive: sampling-based methods such as \acrfull{mcmc} scale poorly, and even variational approaches like \acrfull{svi} require repeated stochastic forward passes. 

This work therefore investigates Bayesian approximation methods and their hardware-efficient implementation to enable \acrshort{bnn} inference on resource-constrained embedded systems. 
First, the training of \acrshortpl{bnn} using \acrshort{mcmc} and \acrshort{svi} methods is examined, highlighting that the choice of activation function strongly influences convergence, predictive accuracy, and uncertainty estimation. 

Second, the \emph{\acrfull{pfp}}~\cite{roth2016pfp} is employed as a closed-form approximation of \acrshort{svi} in \acrshortpl{bnn}. 
By assuming both weights and activations to follow Gaussian distributions, it enables analytic propagation of means and variances through all layers of the network, including nonlinearities handled via moment-matching. 
This removes the need for repeated sampling, replacing multiple stochastic forward passes with a single analytic forward computation.
To enable practical deployment, a dedicated operator library implementing these Gaussian-propagating operations is integrated into the \acrfull{tvm}~\cite{chen2018tvm} compiler framework and optimized for embedded \acrshort{arm} processors, achieving speedups of several orders of magnitude over naive \acrshort{svi}-based \acrshort{bnn} inference~\cite{klein2025pfp}.

Complementary to this analytically grounded approach, ensemble-based methods offer a more practical route to uncertainty estimation by approximating Bayesian diversity through multiple deterministic predictors.
Finally, ensemble-based methods for uncertainty estimation are compared, including \emph{\acrfull{mcdo}} and \emph{\acrfullpl{de}}, alongside the recently introduced \emph{\acrfullpl{rlle}}~\cite{steger2024rlle}. 
\acrshortpl{rlle} share a common deterministic backbone and replace the final layer with an ensemble of prediction heads, greatly reducing the number of trainable parameters while maintaining prediction diversity. 
A function-space repulsion term promotes diversity between ensemble heads, enabling efficient fine-tuning on pretrained networks and producing calibrated uncertainty estimates at minimal additional computational cost. 
Implemented and evaluated using the \acrshort{tvm} compiler stack, \acrshortpl{rlle} achieve the fastest inference among all evaluated approaches. 
Together, these studies demonstrate how algorithmic approximations and hardware-aware mapping can jointly reduce the computational cost of \acrshort{bnn} inference while maintaining reliable uncertainty estimation.

Similar to the investigation of deterministic neural networks, this work extends the exploration beyond algorithmic approximations and optimizations of established architectures toward emerging analog hardware technologies. 
In photonic accelerators, intrinsic fluctuations---such as photon shot noise and chaotic light dynamics---introduce stochasticity that is typically regarded as a limitation to precision. 
Here, this physical noise is leveraged as a controllable source of randomness for probabilistic inference, allowing stochastic sampling to emerge directly from the hardware. 
In the presented photonic experiments~\cite{brueckerhoff2024chaoticlight,brueckerhoff2025photonicAI}, a probabilistic model of the photonic hardware is integrated into Bayesian training and inference loops, enabling the constructive use of stochastic fluctuations characteristic of chaotic light for probabilistic computation.
To make this possible, a controllable noise encoding scheme was developed that maps desired activation variances onto optical intensity distributions following Bose–Einstein statistics. 
The \acrshort{bnn} architecture was co-designed with the photonic prototype to account for its physical constraints enabling stable training and reliable uncertainty estimation under real hardware conditions. 
This approach extends the algorithmic–hardware co-design principle beyond robustness toward the deliberate utilization of analog noise as a foundation for efficient probabilistic computation.

\newpage
\section*{Contributions}
This thesis makes the following primary contributions:
\begin{itemize}

\item \textbf{Compressing \acrshortpl{dnn} for efficient inference on embedded systems:}  
This contribution advances resource-efficient deterministic inference through both analytical insight and automated optimization.  
A comprehensive study of compression methods~\cite{roth2024jmlr}—including quantization, pruning, and knowledge distillation—demonstrates that reductions in parameters or \acrshort{flops} do not reliably predict runtime efficiency, emphasizing the need for empirical, hardware-specific evaluation.  
Building on this foundation, the \emph{Galen} framework~\cite{krieger2023galen} performs automatic, reinforcement-learning-based compression that adapts pruning and quantization per layer and per compute architecture, guided by \emph{sensitivity analysis} and \emph{hardware-in-the-loop} latency measurements, achieving superior accuracy–efficiency trade-offs.

\item \textbf{Modeling and mitigating analog nonidealities in neural accelerators:}  
This work addresses the challenges of analog imperfections in neural hardware through complementary advances in modeling and robustness.  
Detailed white-box and data-driven models of the \acrshort{bss2} analog accelerator~\cite{klein2021bss2whitebox,kuhn2023nonassociativity} capture device-specific nonlinearities, saturation effects, noise, and ordering dependencies, revealing the non-associativity of analog accumulation and enabling efficient, hardware-aware training without continuous device access.  
Building on this foundation, the \emph{Walking Noise} framework~\cite{borras2024walkingnoiseECML} introduces a methodology to quantify layer-wise robustness to additive and multiplicative disturbances, while a systematic comparison of hardening strategies~\cite{wang2025hardening} reveals noisy training as the most effective approach to improve robustness against noisy analog computations.  
To further sustain resilience under dynamically varying conditions, such as temperature-induced drift, \emph{\acrshort{vant}}~\cite{wang2025vant} extends this approach through adaptive noise scaling, demonstrating that robust analog computation requires integrating hardware characteristics directly into the training process.

\newpage
\item \textbf{Efficient \acrshort{bnn} inference on embedded systems:}  
This work addresses the computational challenges of Bayesian neural networks by developing scalable approximation and implementation techniques that enable efficient inference on embedded hardware.  
It begins by analyzing \acrshortpl{bnn} and their core inference methods, \acrshort{mcmc} and \acrshort{svi}, evaluating their scalability and uncertainty estimation quality, and revealing how activation functions critically affect convergence and predictive behavior.  
Building on these insights, the \emph{\acrlong{pfp}}~\cite{roth2016pfp} is implemented as a closed-form approximation to variational inference, analytically propagating Gaussian means and variances through neural layers.  
A dedicated operator library integrated into the \acrshort{tvm} compiler enables optimized execution on embedded \acrshort{arm} processors, achieving substantial speedups over sampling-based approaches~\cite{klein2025pfp}.  
In addition to these analytic methods, ensemble-based Bayesian approximations are explored, including \emph{\acrlongpl{rlle}}~\cite{steger2024rlle}, which promote diversity between prediction heads via a function-space repulsion term while sharing a common deterministic backbone.  
Complementing the analytic acceleration achieved by the \acrlong{pfp}, the \acrshort{tvm}-based implementation of \acrshortpl{rlle} attains further substantial speedups—particularly at larger batch sizes—while maintaining high-quality uncertainty estimation, establishing \acrshortpl{rlle} as an efficient and scalable method for embedded \acrshort{bnn} inference.

\item \textbf{Harnessing photonic noise for probabilistic inference:}  
This work explores photonic accelerators as a hardware platform for efficient Bayesian neural network inference, leveraging intrinsic optical noise as a controllable source of stochasticity.  
Through modeling and hardware–algorithm co-design of chaotic-light photonic systems~\cite{brueckerhoff2024chaoticlight,brueckerhoff2025photonicAI}, the inherent randomness of light interference is transformed from a limiting factor into a functional entropy source for probabilistic computation.  
A dedicated noise-encoding scheme and co-adapted \acrshort{bnn} architecture enable stable training and accurate uncertainty estimation under real hardware constraints, transforming analog noise from a source of error into an active computational resource.

\end{itemize}

\newpage
\section*{Publications}
The work builds upon and extends the following publications:  
\begingroup
\setlength\bibitemsep{0.3\baselineskip}
\small
\setlist[itemize]{leftmargin=*}

\subsection*{Journal Articles}
\begin{itemize}
  \item Wolfgang Roth\textsuperscript{*}, Günther Schindler\textsuperscript{*}, \uline{Bernhard Klein}\textsuperscript{*}, Robert Peharz, Sebastian Tschiatschek, Holger Fröning, Franz Pernkopf, and Zoubin Ghahramani,  
  “Resource-Efficient Neural Networks for Embedded Systems,” \emph{Journal of Machine Learning Research (JMLR)}, 2024.  
  \href{https://jmlr.org/papers/volume25/18-566/18-566.pdf}{[JMLR]} \href{https://arxiv.org/abs/2001.03048}{[arXiv]}

  \item Frank Brückerhoff-Plückelmann, Hendrik Borras, \uline{Bernhard Klein}, Akhil Varri, Marlon Becker, Jelle Dijkstra, Martin Brückerhoff, C. David Wright, Martin Salinga, Harish Bhaskaran, Benjamin Risse, Holger Fröning, and Wolfram Pernice,  
  “Probabilistic Photonic Computing with Chaotic Light,” \emph{Nature Communications}, 2024.  
  \href{https://doi.org/10.1038/s41467-024-54931-6}{[DOI]}

  \item Frank Brückerhoff-Plückelmann, Anna P. Ovvyan, Akhil Varri, Hendrik Borras, \uline{Bernhard Klein}, Lennart Meyer, C. David Wright, Harish Bhaskaran, Ghazi Sarwat Syed, Abu Sebastian, Holger Fröning, and Wolfram Pernice,  
  “Probabilistic Photonic Computing for AI,” \emph{Nature Computational Science}, May 2025.  
  \href{https://doi.org/10.1038/s43588-025-00800-1}{[DOI]}
\end{itemize}

\subsection*{Conference Papers}
\begin{itemize}
  \item Hendrik Borras\textsuperscript{*}, \uline{Bernhard Klein}\textsuperscript{*}, and Holger Fröning,  
  “Walking Noise: On Layer-Specific Robustness of Neural Architectures Against Noisy Computations and Associated Characteristic Learning Dynamics,”  
  in \emph{European Conference on Machine Learning and Principles and Practice of Knowledge Discovery in Databases (ECML-PKDD)}, 2024.  
  \href{https://doi.org/10.1007/978-3-031-70359-1_3}{[DOI]} \href{https://arxiv.org/abs/2212.10430}{[arXiv]}

  \item Xiao Wang, Hendrik Borras, \uline{Bernhard Klein}, and Holger Fröning,  
  “Variance-Aware Noisy Training: Hardening DNNs Against Unstable Analog Computations,”  
  in \emph{European Conference on Machine Learning and Principles and Practice of Knowledge Discovery in Databases (ECML-PKDD)}, 2025.  
  \href{https://doi.org/10.1007/978-3-032-06109-6_9}{[DOI]} \href{https://arxiv.org/abs/2503.16183}{[arXiv]}
\end{itemize}

\subsection*{Workshop Papers}
\begin{itemize}
  \item Torben Krieger\textsuperscript{*}, \uline{Bernhard Klein}\textsuperscript{*}, and Holger Fröning,  
  “Towards Hardware-Specific Automatic Compression of Neural Networks,”  
  \emph{AAAI Workshop on Practical Deep Learning in the Wild}, 2023.  
  \emph{Best Paper Award.} \href{https://doi.org/10.48550/arXiv.2212.07818}{[DOI]}

  \item Lisa Kuhn\textsuperscript{*}, \uline{Bernhard Klein}\textsuperscript{*}, and Holger Fröning,  
  “On the Non-Associativity of Analog Computations,”  
  in \emph{European Conference on Machine Learning and Principles and Practice of Knowledge Discovery in Databases (ECML-PKDD) Workshops (ITEM)}, 2025.  
  \href{https://doi.org/10.1007/978-3-031-74643-7_15}{[DOI]} \href{https://arxiv.org/abs/2309.14292}{[arXiv]}

  \item Sophie Steger, Christian Knoll, \uline{Bernhard Klein}, Holger Fröning, and Franz Pernkopf,  
  “Function-Space Diversity for Uncertainty Prediction via Repulsive Last-Layer Ensembles,”  
  \emph{ICML Workshop on Structured Probabilistic Inference \& Generative Modeling}, 2024.  
  \href{https://openreview.net/forum?id=FbMN9HjgHI}{[Link]} \href{https://arxiv.org/abs/2412.15758}{[arXiv]}

  \item Hendrik Borras\textsuperscript{*}, \uline{Bernhard Klein}\textsuperscript{*}, and Holger Fröning,  
  “Walking Noise: Understanding Implications of Noisy Computations on Classification Tasks,”  
  \emph{HiPEAC Workshop on Accelerated Machine Learning (AccML)}, 2023. (Earlier version of the ECML-PKDD 2024 paper.)  
  \href{https://accml.dcs.gla.ac.uk/papers/2023/5th_AccML_paper_2.pdf}{[Link]}

  \item Xiao Wang, Hendrik Borras, \uline{Bernhard Klein}, and Holger Fröning,  
  “On Hardening DNNs Against Noisy Computations,”  
  \emph{HiPEAC Workshop on Accelerated Machine Learning (AccML)}, 2025.  
  \href{https://arxiv.org/abs/2501.14531}{[arXiv]} \href{https://accml.dcs.gla.ac.uk/papers/2025/7th_AccML_paper_1.pdf}{[Link]}

  \item \uline{Bernhard Klein}, Christoph Gratl, Manfred Mücke, and Holger Fröning,  
  “Understanding Cache Boundness of ML Operators on ARM Processors,”  
  \emph{HiPEAC Workshop on Accelerated Machine Learning (AccML)}, 2021.  
  \href{https://arxiv.org/abs/2102.00932}{[arXiv]}

  \item \uline{Bernhard Klein}, Lisa Kuhn, Johannes Weis, Arne Emmel, Yannik Stradmann, Johannes Schemmel, and Holger Fröning,  
  “Towards Addressing Noise and Static Variations of Analog Computations Using Efficient Retraining,”  
  in \emph{European Conference on Machine Learning and Principles and Practice of Knowledge Discovery in Databases (ECML-PKDD) Workshops (ITEM)}, 2021.  
  \href{https://doi.org/10.1007/978-3-030-93736-2_32}{[DOI]}
\end{itemize}

\subsection*{Submitted for Review}
\begin{itemize}
  \item \uline{Bernhard Klein}, Falk Selker, Hendrik Borras, Sophie Steger, Franz Pernkopf, and Holger Fröning,  
  “Accelerated Execution of Bayesian Neural Networks Using a Single Probabilistic Forward Pass and Code Generation,”  
  under review at \emph{ACM Transactions on Architecture and Code Optimization (TACO)}, 2025.
\end{itemize}

\par\smallskip
\noindent\footnotesize\textsuperscript{*}\,Shared first authorship.
\endgroup
 
\section*{Thesis Structure}  
The remainder of this thesis is organized into two parts that together advance the goal of resource-efficient and reliable neural network inference.  

\begin{figure}[b]
  \centering
  \hspace*{-0.5cm} %
  \begin{tikzpicture}[x=7.5cm,y=5.0cm] %
    \tikzstyle{title}=[font=\bfseries\small,align=left]
    \tikzstyle{item}=[font=\footnotesize,align=left,text width=6.8cm]

    \draw[thin,gray!60] (0,0.5) -- (2,0.5);
    \draw[thin,gray!60] (1,0) -- (1,1);

    \node[title,anchor=west] at (0.05,0.94) {Digital Hardware · Deterministic};
    \node[title,anchor=west] at (0.05,0.44) {Analog Hardware · Deterministic};
    \node[title,anchor=west] at (1.05,0.94) {Digital Hardware · Probabilistic};
    \node[title,anchor=west] at (1.05,0.44) {Analog Hardware · Probabilistic};

    \node[item,anchor=north west] at (0.05,0.88) {%
      Chapter~\ref{ch:rennfes}:\; Resource-Efficient Inference\\
      Chapter~\ref{ch:galen}:\; Automatic Model Compression%
    };

    \node[item,anchor=north west] at (0.05,0.38) {%
      Chapter~\ref{ch:analog:hardware_models}:\; Modeling Analog Hardware\\
      Chapter~\ref{ch:robustness}:\; Robustness Against Noisy\\%
	  \phantom{Chapter~\ref{ch:robustness}:\; }Computations
    };

    \node[item,anchor=north west] at (1.05,0.88) {%
      Chapter~\ref{ch:bnns}:\; Bayesian Neural Networks\\
      Chapter~\ref{ch:pfp}:\; Compiling \acrshort{pfp}-based \acrshortpl{bnn}\\
      Chapter~\ref{ch:ensembles}:\; Ensemble-based \acrshortpl{bnn}%
    };

    \node[item,anchor=north west] at (1.05,0.38) {%
      Chapter~\ref{ch:photonics}:\; Probabilistic Photonic\\
	  \phantom{Chapter~\ref{ch:photonics}:\; }Computing for \acrshortpl{bnn}%
    };

	\end{tikzpicture}
	\caption{Thesis contributions organized along hardware technologies (vertical: digital vs.\ analog) and modeling paradigms (horizontal: deterministic vs.\ probabilistic).}
  \label{fig:intro:thesis-map}
\end{figure}

\textbf{Part~I} focuses on deterministic models and explores both algorithmic and hardware-oriented strategies for efficient execution.  
Chapter~\ref{ch:rennfes} surveys established compression techniques and analyzes their effectiveness across embedded computing platforms.  
Building on these foundations, Chapter~\ref{ch:galen} presents the \emph{Galen} framework for automatic, hardware-aware compression that jointly optimizes pruning and quantization for accuracy and latency.  
Chapters~\ref{ch:analog:hardware_models} and~\ref{ch:robustness} jointly address analog neural accelerators.  
Chapter~\ref{ch:analog:hardware_models} develops hardware models that capture device imperfections of analog accelerators, providing a foundation for efficient hardware-aware training.  
Building on these models, Chapter~\ref{ch:robustness} investigates the resilience of neural networks to such noisy computations, introducing the \emph{Walking Noise} framework for layer-wise robustness analysis and \emph{\acrshort{vant}} to harden \acrshortpl{dnn} for dynamic noise conditions.

\textbf{Part~II} extends these principles to probabilistic machine learning, emphasizing scalable Bayesian inference and uncertainty estimation.  
Chapter~\ref{ch:bnns} introduces Bayesian neural networks and compares inference methods such as \acrshort{mcmc} and \acrshort{svi}, highlighting how activation functions influence uncertainty and convergence.  
Chapter~\ref{ch:pfp} implements the \acrlong{pfp} as a compiler-integrated operator library for embedded inference, while Chapter~\ref{ch:ensembles} investigates ensemble-based approximations, including the \acrlongpl{rlle}, demonstrating their efficiency and calibration trade-offs.  
Finally, Chapter~\ref{ch:photonics} explores photonic accelerators as a hardware platform for probabilistic computation, showing how intrinsic optical noise can serve as a controllable stochastic resource for \acrshort{bnn} inference.

Chapter~\ref{ch:conclusion} concludes the thesis by summarizing the overarching findings, synthesizing deterministic and probabilistic efficiency concepts, and outlining future research directions in integrated algorithm–hardware co-design for energy-efficient machine learning.

The overall organization of this work is summarized in Figure~\ref{fig:intro:thesis-map}.  
The diagram arranges the chapters along two axes—hardware domain (digital to analog) and modeling paradigm (deterministic to probabilistic)—mirroring the transition from Part~I to Part~II.  
Together, they illustrate how the thesis progresses from digital model compression and analog robustness toward probabilistic and photonic computation as complementary routes to efficient and reliable neural inference.

\clearpage{}%
\clearpage{}%
\chapter{Background}
\label{ch:background}

\section{Deep Neural Networks}
\label{sec:background:dnns}

\Acrlongpl{dnn} have become the predominant model class in modern machine learning~\cite{LeCun2015DeepLearning,goodfellow2016deep}.
They consist of multiple layers of parameterized transformations, combined with nonlinear activation functions, that together approximate highly complex mappings from inputs to outputs.
The learnable parameters, often referred to as \emph{weights}, are adjusted during training such that the network minimizes a given loss function on available data.

\paragraph{Layer Types.}
At their core, \acrshortpl{dnn} alternate between linear operations and nonlinear activation functions.
Common choices for the activation function $\sigma(\cdot)$ include the \acrfull{relu}, sigmoid, or hyperbolic tangent, each shaping the representational capacity of the network differently.
The simplest linear operation is the dense (fully-connected) layer, which maps an input vector $\mathbf{x} \in \mathbb{R}^d$ to an output vector $\mathbf{y} \in \mathbb{R}^m$:
\begin{equation}
    \mathbf{y} = \sigma(\mathbf{W}\mathbf{x} + \mathbf{b}),
		\label{eq:background:dense}
\end{equation}
where $\mathbf{W} \in \mathbb{R}^{m \times d}$ and $\mathbf{b} \in \mathbb{R}^m$ are learnable parameters.
Convolutional layers generalize this operation to exploit the spatial structure of input data while reducing the number of parameters.
Let $\mathbf{I}$ denote the input feature map with $C$ channels, and let $\mathbf{W}$ denote a set of filter kernels with spatial dimensions $R \times S$ and $U$ output channels.
For an input position $(x,y)$, input channel $z$ and output channel $u$, the output feature map $\mathbf{O}$ is computed as
\begin{equation}
    O[z][u][x][y] = 
    \sum_{k=0}^{C-1}\sum_{i=0}^{R-1}\sum_{j=0}^{S-1}
        I[z][k][s x+i][s y+j] \cdot W[u][k][i][j] + B[u],
		\label{eq:background:conv}
\end{equation}
where $B[u]$ denotes the bias term for output channel $u$ and $s$ is the stride parameter controlling the step size.
The resulting output dimensions are
\begin{align}
    E &= \frac{H - R + s}{s}, \\
    F &= \frac{W - S + s}{s},
\end{align}
for an input feature map of height $H$ and width $W$.
This formulation reflects the convolution operation as implemented in modern deep learning frameworks, including multi-channel inputs, multiple filters, stride, and biases.

\paragraph{Training and Inference.}
Neural networks are trained by gradient-based optimization methods, most prominently stochastic gradient descent and its variants~\cite{Robbins1951SGD,Kingma2015Adam}.
Gradients are computed efficiently using backpropagation~\cite{Rumelhart1986BP}, which recursively applies the chain rule across the layered structure.
Once trained, inference corresponds to applying the learned transformations to unseen inputs, requiring only the forward pass.
This distinction is particularly important in practice: training is usually carried out on large-scale compute clusters, while inference is also often executed on embedded, mobile or specialized hardware where resources are constrained.

\paragraph{Impact and Limitations.}
The scalability and expressiveness of \acrshortpl{dnn} have driven breakthroughs across domains such as computer vision, natural language processing, and speech recognition~\cite{Krizhevsky2012AlexNet,Vaswani2017Attention}.
At the same time, the growing depth and parameter counts of state-of-the-art models result in high computational demand and memory consumption.
Balancing prediction quality with computational and energy efficiency is therefore a central challenge, motivating the resource-aware, robust, and probabilistic inference approaches that form the core contributions of this work.
\section{ML Frameworks and Compilers}
\label{sec:background:frameworks}

\subsection{Training Frameworks}
Modern deep learning practice is dominated by frameworks such as PyTorch~\cite{Paszke2019PyTorch}, TensorFlow~\cite{Abadi2016TensorFlow}, and JAX~\cite{Bradbury2018JAX}.
They provide user-friendly abstractions to define models, automatic differentiation to compute gradients, and efficient utilization of hardware accelerators.
These frameworks have become the standard entry point for designing and training \acrlongpl{dnn}, offering both flexibility for research and scalability for large-scale production.

\subsection{Probabilistic Programming Frameworks}
Classical machine learning frameworks are extended by \acrfullpl{ppl} to model random variables directly.
They provide support for sampling, probability distributions, and a range of Bayesian inference algorithms, including variational inference and Markov chain Monte Carlo (see Chapter~\ref{ch:bnns} for a detailed introduction to Bayesian inference methods).
Prominent \acrshortpl{ppl} include Pyro~\cite{bingham2019pyro}, NumPyro~\cite{Phan2019NumPyro}, and TensorFlow Probability~\cite{Dillon2017TFP}.

Pyro is built on top of PyTorch and emphasizes scalable variational inference, making it a natural choice for amortized inference and stochastic variational inference (SVI).
NumPyro, by contrast, builds on JAX and leverages \acrfull{jit} compilation to achieve high performance.
It places a stronger emphasis on \acrshort{mcmc} methods, providing highly optimized implementations of \acrlong{hmc} and the \acrlong{nuts}~\cite{hoffman2014nuts}.
In practice, NumPyro is often considered faster for large-scale \acrshort{mcmc}-based Bayesian inference due to its compilation-based workflow, while Pyro is regarded as more flexible and easier to customize, making it well suited for prototyping new probabilistic models and inference algorithms.
TensorFlow Probability extends TensorFlow with a large collection of probabilistic building blocks, covering both inference methods and probabilistic layers.

These frameworks are a necessary component of the modern probabilistic deep learning ecosystem, providing the modeling tools required to train \acrlongpl{bnn} with uncertainty-awareness while remaining compatible with standard deep learning workflows.

\subsection{Deep Learning Compilers}
\label{subsec:background:compilers}

Training frameworks excel at model specification and optimization, but efficient deployment on heterogeneous targets often requires compiler support.
Hand-tuned vendor libraries deliver strong performance for common operators on popular hardware architectures, yet they are difficult to retarget and cannot cover all combinations of hardware, data layouts, and custom operators.
\Acrlong{ml} compilers address this gap by lowering high-level models to hardware-aware implementations while applying graph- and kernel-level optimizations that improve latency and energy efficiency across diverse backends.

\paragraph{MLIR and XLA.}
The \acrlong{mlir} (\acrshort{mlir}) project provides a modular, multi-level intermediate representation with extensible \emph{dialects} for tensors, linear algebra, and device backends~\cite{Lattner2021mlir}.
Progressive lowering enables reuse of transformations across ecosystems: models can start from framework-level \acrfullpl{ir}, pass through domain-specific dialects, and end as target \acrfull{isa} or runtime code.
This design eases support for non-standard data types, dynamic shapes, and emerging accelerators without invasive changes to front-ends.
\acrshort{xla}~\cite{XLA2017} is a domain-specific compiler stack developed for TensorFlow and JAX, providing graph fusion, buffer reuse, and code generation.
It remains the primary compiler for Google's \acrshortpl{tpu} and increasingly relies on \acrshort{mlir} as its underlying compiler infrastructure.
Together, \acrshort{mlir} and \acrshort{xla} illustrate how compiler infrastructures can unify optimization passes across frameworks and backends while enabling efficient execution on specialized hardware.

\paragraph{TVM.}
The \acrfull{tvm} stack is an end-to-end optimizing compiler for \acrlongpl{dnn} that targets \acrshortpl{cpu}, \acrshortpl{gpu}, \acrshortpl{fpga}, and custom accelerators~\cite{chen2018tvm}.
Its \acrshort{ir} family (\acrshort{te}/TensorIR/Relax) separates \emph{what} to compute from \emph{how} to schedule it.
Crucially, \acrshort{tvm} is designed to support \emph{custom} operators and \emph{novel} hardware where vendor libraries are unavailable or insufficient, making it suitable for research settings and embedded deployments alike.
In this work, \acrshort{tvm} plays a central role across several chapters, serving as the common path from high-level models to hardware-optimized binaries.

\paragraph{Auto-tuning.}
To illustrate the scheduling challenge, consider the example of a 2D convolution kernel.
Its performance depends strongly on how the nested loops over output channels, spatial positions, and kernel indices are ordered, tiled, vectorized, and parallelized.
Unfavorable choices—such as iterating over strided dimensions or very small kernels like $3{\times}3$ in the innermost loop—can prevent effective vectorization and waste memory bandwidth, while good schedules expose cache locality, \acrshort{simd} efficiency, and balanced parallel work.
Since the optimal decisions vary across tensor shapes and hardware, manual scheduling is tedious and hardware-specific.
To address this challenge, \acrshort{tvm} integrates auto-tuning frameworks---first Ansor~\cite{Zheng2020ansor} and later the Meta-Scheduler~\cite{Shao2022tvmMetaScheduler}---that automatically explore the scheduling space by generating candidates, benchmarking them on the target device, and learning from results.
This process often discovers implementations that rival or surpass expert-crafted kernels, and in this work proved particularly effective for compressed and probabilistic operators where no vendor libraries exist.

\paragraph{Summary.}
Training frameworks, probabilistic programming environments, and deep learning compilers together form the toolchain that spans the full workflow.
Within this ecosystem, \acrshort{tvm} offers end-to-end code generation and powerful auto-tuning to achieve high performance for standard and non-standard operators on heterogeneous targets.
These capabilities are central whenever hand-tuned kernels are unavailable or suboptimal, enabling efficient deployment for \acrlong{nn} compression (Chapter~\ref{ch:galen}), probabilistic inference (Chapter~\ref{ch:pfp}), and ensembles (Chapter~\ref{ch:ensembles}) on resource-constrained embedded hardware.
\section{Hardware Platforms for Neural Networks}
\label{sec:background:hardware}

Advances in hardware have been a key driver of modern deep learning.  
Both training and inference impose high computational demands, and the choice of hardware platform often determines whether an application is feasible.  
This section provides a short overview of the most relevant processor classes for \acrlongpl{dnn}, highlighting their strengths and limitations for inference.  
All platforms appear in a wide range of scales, from mobile to data-center deployments.  

\paragraph{\acrshortpl{cpu}.}
\Acrfullpl{cpu} maximize general-purpose programmability and single-thread performance.  
Modern designs rely on multithreading and vectorization to achieve throughput, which makes them flexible and particularly suitable for handling irregular workloads.  
Support for low-precision arithmetic (e.g., 8-bit integer instructions) further improves efficiency in typical inference pipelines.  

\paragraph{\acrshortpl{gpu}.}
\Acrfullpl{gpu} contain large numbers of lightweight cores with high memory bandwidth, offering massive throughput for dense tensor operations.  
They follow a block-parallel execution paradigm, where many threads are grouped into warps and thread blocks, enabling massive fine-grained parallelism~\cite{Nickolls2008CUDA}.  
This design makes \acrshortpl{gpu} highly efficient for regular, highly parallel computations but less effective on irregular workloads.  
Their native support for reduced-precision arithmetic, such as 16-bit floating point and 8-bit integer, has been central to the deployment of modern deep learning, and recent architectures even extend this support to 4-bit formats for further efficiency gains.  
Efficient hardware support for dense matrix multiplications has made \acrshortpl{gpu} the main driver of modern \acrlongpl{dnn} during both training and inference.  

\paragraph{\acrshortpl{fpga}.}
\Acrfullpl{fpga} consist of configurable logic blocks that can be programmed into custom compute units.  
They operate at lower frequency and with limited on-chip memory compared to \acrshortpl{cpu} and \acrshortpl{gpu}, but allow arbitrary data formats and fine-grained parallelism.  
Their flexibility makes them attractive for latency-critical deployments with strict real-time constraints, where extreme low inference latencies are required.  
Moreover, \acrshortpl{fpga} are often employed as prototyping platforms for \acrfullpl{asic}, offering a balance between programmability and the ability to evaluate custom hardware designs.  

\paragraph{Domain-Specific Accelerators.}
Dedicated accelerators such as Google’s \acrshort{tpu}~\cite{jouppi2023tpuv4} implement systolic arrays optimized for dense linear algebra.  
They achieve high utilization on structured workloads with limited programmability.  
Support is typically restricted to standard low-precision formats such as 16-bit floating point or 8-bit integer, but their throughput and energy efficiency make them attractive for large-scale deployments.  

\paragraph{Analog Accelerators.}
Beyond digital processors, analog computing devices have gained attention as promising alternatives for neural network inference.  
Examples include resistive memory arrays~\cite{Nandakumar2018pcm}, electronic CMOS-based systems~\cite{Murmann2021,schemmel2021accelerated}, and optical or photonic processors~\cite{Lin2018optical,Shen2017photonic,brueckerhoff2025photonicAI}.  
By carrying out matrix multiplications directly in the analog domain, these devices promise orders-of-magnitude improvements in energy efficiency compared to digital hardware.  
However, analog computations are inherently noisy due to device variability, nonlinearities, and stochastic physical processes~\cite{klein2021bss2whitebox,kuhn2023nonassociativity}.  
This variability poses challenges for maintaining accuracy, but also creates opportunities for algorithmic co-design.  
In this work, analog noise and variability are analyzed through analog hardware models (Chapter~\ref{ch:analog:hardware_models}), addressed with robustness-enhancing training methods (Chapter~\ref{ch:robustness}), and eventually exploited as a source of stochasticity for probabilistic inference on photonic accelerators (Chapter~\ref{ch:photonics}).

\clearpage{}%
\clearpage{}%
\part{Accelerating Deep Neural Networks}
\label{pt:acceleratingdnns}
\chapter{Foundations of Resource-Efficient Inference of \acrshort{nn} for Embedded Systems}
\label{ch:rennfes}

\epigraph{Any intelligent fool can make things bigger and more complex... It takes a touch of genius---and a lot of courage---\\to move in the opposite direction.}{\textnormal{--- E. F. Schumacher, \textit{Small is Beautiful} (1973)}}

\noindent
Modern machine learning has advanced in parallel with the increasing availability of large-scale computational resources.  
Training and tuning contemporary \acrshortpl{dnn} is both computationally intensive and methodologically challenging, often requiring massive parallelism and sophisticated systems engineering.  
State-of-the-art models are typically trained in data centers equipped with abundant \acrshortpl{gpu} or specialized neural processors (\acrshortpl{npu}, \acrshortpl{tpu}), where energy consumption and latency are secondary considerations.  
Deployment scenarios, however, present a fundamentally different reality.  
Embedded systems such as smartphones, autonomous robots, or sensor nodes in the \acrlong{iot} operate under strict resource budgets: available memory is limited, energy is constrained by batteries, and latency must remain within tight bounds to ensure real-time responsiveness.  

This discrepancy creates a persistent tension: how can models that are developed and trained in nearly unconstrained laboratory environments be deployed on hardware with orders of magnitude fewer resources?  
Naïve model reduction often leads to unacceptable drops in prediction quality.  
Instead, resource-efficient inference requires systematic approaches that balance representational and computational demands against accuracy in a controlled fashion.  

Among the wide variety of proposed strategies, three broad categories have proven particularly effective for \acrshortpl{dnn} compression and acceleration: \emph{quantization}, \emph{pruning}, and \emph{structural efficiency}.  
These techniques reduce the memory footprint and inference cost while preserving competitive predictive performance.  
They are relevant across embedded hardware platforms ranging from general-purpose \acrshortpl{cpu} and \acrshortpl{gpu} to domain-specific accelerators and \acrshortpl{fpga}.  
At the same time, they differ markedly in terms of hardware friendliness, achievable compression ratios, and their impact on prediction quality.  
As we will see throughout this chapter, the effectiveness of compression techniques is highly hardware-dependent, underscoring the need for co-design of algorithms and deployment platforms.  

This chapter builds on our work \citeauthor{roth2024jmlr} -- \textit{Resource-Efficient Neural Networks for Embedded Systems}~\cite{roth2024jmlr}, which provides a comprehensive survey of embedded hardware platforms and efficiency-enhancing methods.  
For details beyond the scope of this chapter we refer the reader to the original work, while here we focus on the concepts and techniques most relevant for the subsequent chapters.  

By consolidating these foundations, this chapter prepares the ground for the Galen framework introduced in the following chapter, which employs hardware-in-the-loop optimization to automate compression strategies.

\section{Foundations of Efficiency}

\begin{figure}
	\centering
	\includegraphics[width=0.7\textwidth]{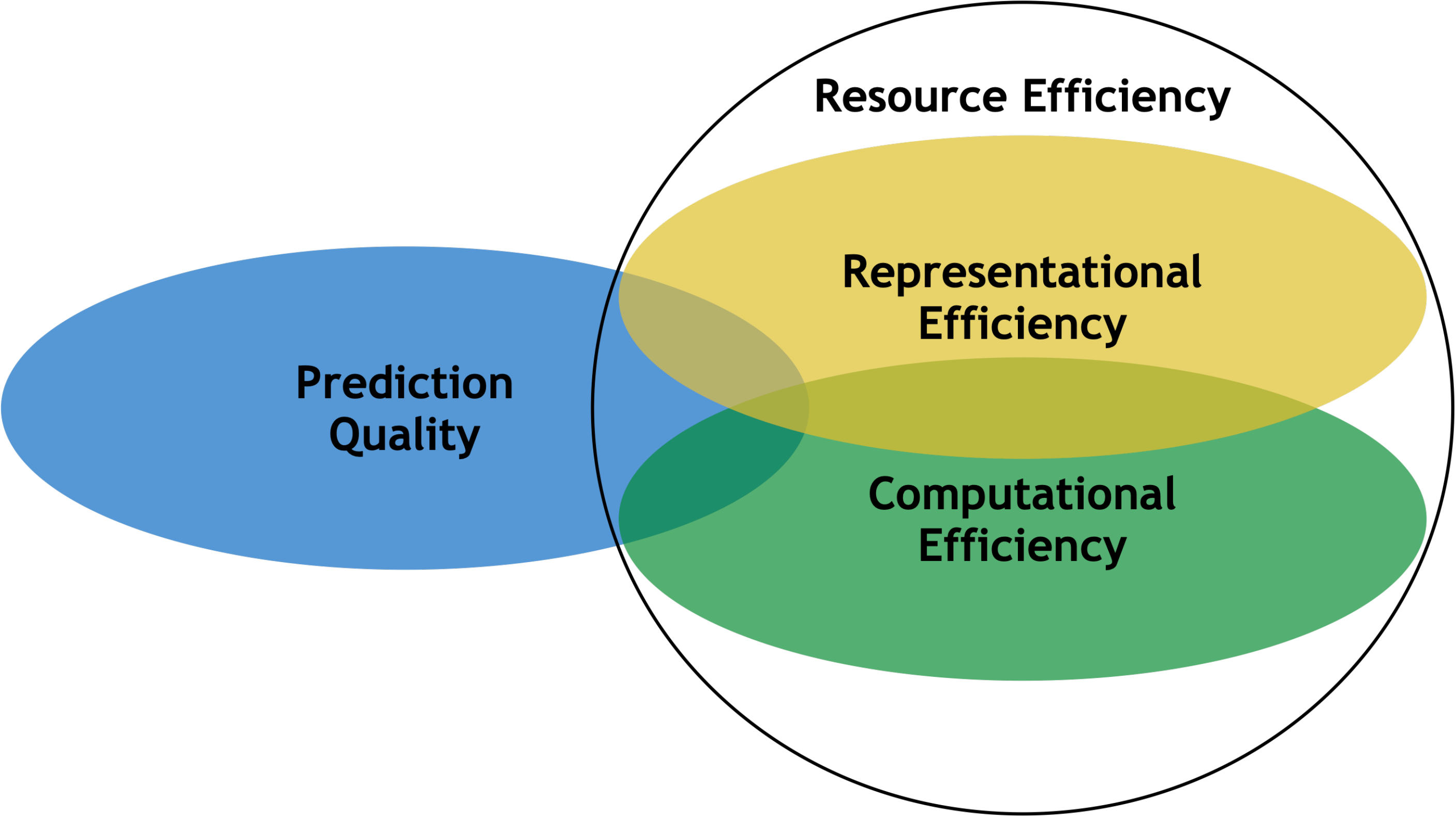}
	\caption{Systematic perspective on resource-efficient machine learning.  
	Representational efficiency, computational efficiency, and prediction quality form the triad that structures the discussion in this chapter.  
	\reprofrom{roth2024jmlr}}
	\label{fig:rennfes:systematic}
\end{figure}

Resource-efficient inference of \acrshortpl{dnn} is naturally a multi-objective optimization problem.  
On the one hand, the model must fit into the tight resource budgets of embedded systems.  
On the other hand, it must deliver sufficient prediction quality for the target application.  
As proposed by \citeauthor{roth2024jmlr}~\cite{roth2024jmlr} and illustrated in Figure~\ref{fig:rennfes:systematic}, efficiency can be structured into three complementary dimensions: representational efficiency, computational efficiency, and prediction quality.  

\textbf{Representational efficiency} captures how compactly parameters and activations are stored, measured by memory footprint and storage format.  
It is determined by the number of parameters, their sparsity, and the chosen numerical precision, for example floating-point versus quantized integer representations.  
Techniques such as quantization and pruning directly target this dimension.  

\textbf{Computational efficiency} refers to the time, throughput, and energy required to execute the model on the target hardware.  
It accounts not only for arithmetic operations but also, critically, for the movement of data through the memory hierarchy.  
Simple theoretical metrics such as \acrshort{flops} provide a rough indication of computational cost, but they do not reliably capture actual latency or energy consumption.  
This underlines the importance of evaluating efficiency directly on the deployed device.  

\textbf{Prediction quality} captures the predictive performance of the compressed model.  
While classical machine learning focuses almost exclusively on accuracy, in embedded deployment prediction quality must be considered jointly with efficiency demands to ensure that gains in efficiency do not come at the expense of essential predictive performance.  

Taken together, these three dimensions provide a systematic lens for analyzing and comparing efficiency techniques.  
Throughout the remainder of this chapter we will use them as a recurring reference point to evaluate quantization, pruning, and structural efficiency.

\section{Quantization}
\label{sc:rennfes:quantization}

Quantization reduces the bit-width used to represent weights and activations in \acrshortpl{dnn}, shrinking model size and activation memory while, on suitable hardware, also accelerating inference and lowering energy cost.  
At the extreme, binary weights $w \in \{-1,1\}$ and activations $x \in \{-1,1\}$ turn multiplications into efficient XNOR and bitcount operations, effectively reducing a network to a logical circuit.  

Training such discrete-valued networks is difficult because quantization is non-differentiable.  
The central challenge is to lower precision as much as possible without sacrificing the accuracy of a full-precision baseline.  
Over the past three decades, a wide range of techniques have been developed to address this challenge.  
The following sections review these approaches, beginning with early attempts at reduced-precision training and moving toward modern quantization-aware training and large-scale applications.
Early works such as Höhfeld and Fahlman~\cite{Hoehfeld1992a,Hoehfeld1992b} introduced stochastic rounding to prevent training stalls at low precision, a principle later shown effective on modern architectures~\cite{Gupta2015}.  

Lin et al.~\cite{Lin2015} eliminated most multiplications during training by quantizing weights stochastically to binary or ternary values in the forward pass and quantizing activations to powers of two in the backward pass, reducing multiplications to bit shifts.  
This yields significant training speedups, although inference still depends on full-precision weights.  

Subsequent work broadened the scope.  
Courbariaux et al.~\cite{Courbariaux2015binaryconnect} systematically compared floating-, fixed-, and dynamic fixed-point formats across bit-widths.  
Lin et al.~\cite{Lin2016} cast layer-wise bit allocation as a convex optimization problem that minimizes storage under a signal-to-quantization-noise constraint, yielding closed-form solutions for optimal bit-widths.  

\subsection{Quantization-aware Training}

\begin{figure}
	\centering
	\includegraphics[width=0.8\textwidth]{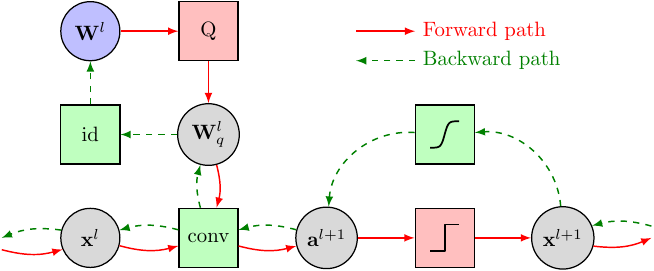}
	\caption{Schematic illustration of quantization with the \acrlong{std}.  
	The forward pass applies quantization to weights and activations, while the backward pass approximates the gradient as the identity function.  
	\reprofrom{roth2024jmlr}}
	\label{fig:rennfes:quantization-ste}
\end{figure}

Quantization functions are piecewise constant with zero or undefined derivatives, which breaks standard backpropagation.  
The \acrfull{ste}~\cite{Bengio2013}, illustrated in Figure~\ref{fig:rennfes:quantization-ste}, has therefore become the default workaround: weights are stored in full precision, quantized in the forward pass, and updated in the backward pass as if the quantizer were the identity.  
At test time, only the quantized weights are kept, and the same principle applies to activations.  

Early \acrfull{qat} methods applied this scheme to binary networks.  
Courbariaux et al.~\cite{Courbariaux2015binaryconnect} trained binary-weight models using either deterministic sign rounding or stochastic rounding via a hard--sigmoid, and Hubara et al.~\cite{Hubara2016} extended the approach to activations, reducing multiplications and additions to efficient XNOR and bitcount operations.  
Li et al.~\cite{Li2016} advanced to ternary weights $w \in \{-a,0,a\}$ with thresholds chosen to minimize quantization error, while Zhu et al.~\cite{Zhu2017} generalized to asymmetric ternary $w \in \{-a,0,b\}$ with trainable parameters and per-layer thresholds, often outperforming symmetric schemes.  

Filter quantization soon followed.  
Rastegari et al.~\cite{Rastegari2016} approximated convolutional filters as $\matW=\alpha\matB$, replacing most multiplications with additions and requiring only one multiplication per channel, with input quantization further enabling XNOR and bitcount convolutions.  
Lin et al.~\cite{Lin2017} improved accuracy by expressing filters as linear combinations of multiple binary bases.  

Other approaches tailored quantizers to data distributions.  
Cai et al.~\cite{Cai2017} introduced half-wave Gaussian quantization to better match ReLU activations.  
Miyashita et al.~\cite{Miyashita2016} quantized to powers of two, eliminating multiplications and improving robustness, while Zhou et al.~\cite{Zhou2017} proposed incremental quantization, alternately quantizing subsets of weights to $\{0\}\cup\{2^k\}$ and retraining until all layers were quantized.  

Deployment-oriented schemes also emerged.  
Jacob et al.~\cite{Jacob2018} introduced integer-only inference, emulating quantization during training and deploying 8-bit weights for inference, thereby reducing the model size by a factor of four.
Liu et al.~\cite{Liu2018} presented Bi-Real Net, a ResNet variant with binary convolutions in the residual path and real-valued shortcuts to preserve expressiveness.  

More flexible quantizers were later learned end-to-end.  
Zhang et al.~\cite{Zhang2018} proposed \emph{LQ-Net}, where quantization codebooks are optimized during training, with layer-wise activation quantizers and channel-wise weight quantizers improving flexibility and efficiency.  
Louizos et al.~\cite{Louizos2019} introduced \emph{Relaxed Quantization}, a differentiable soft-rounding scheme that injects noise before rounding and employs Gumbel--softmax~\cite{Jang2017} to approximate discrete levels, with a hard \acrshort{ste} variant for exact quantization.  

\begin{figure}
	\centering
	\includegraphics[width=0.5\textwidth]{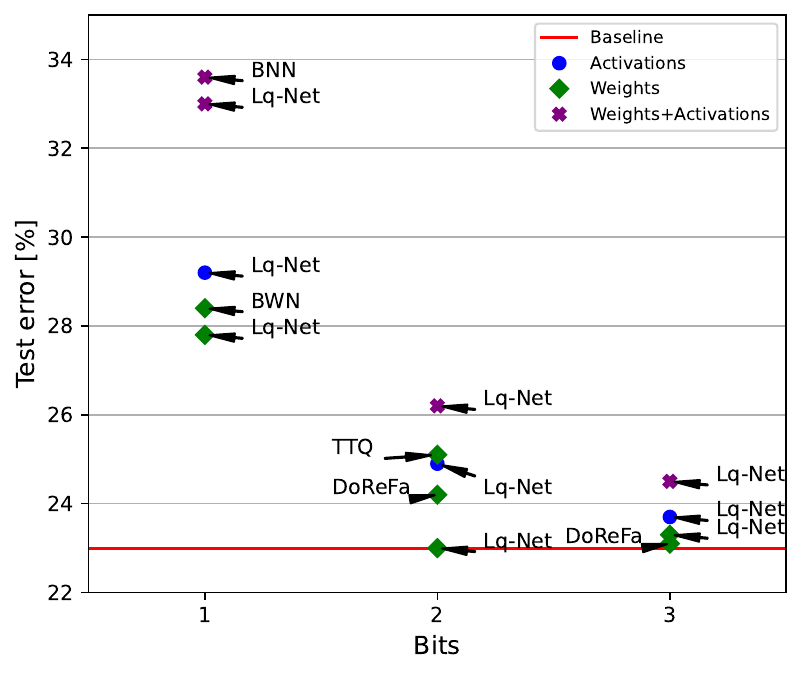}
	\caption{Comparison of several popular quantization methods using the DenseNet-BC-100 architecture on the CIFAR-100 dataset.  
    Test error is shown as a function of bit-width for weight and activation quantization.  
    As expected, lower bit-widths lead to larger errors: weight-only quantization degrades accuracy moderately, activation quantization incurs larger losses, and quantizing both amplifies the effect.  
    \reprofrom{roth2024jmlr}}
	\label{fig:rennfes:quantization-bitwidth}
\end{figure}

These methods illustrate the spectrum from binary to low-bit quantization and highlight the trade-off between efficiency and accuracy.  
As shown in Figure~\ref{fig:rennfes:quantization-bitwidth}, accuracy decreases nonlinearly with lower bit-widths: weight-only quantization impacts performance, but activation quantization leads to larger errors, and combining both amplifies the effect.  

Mixed-precision quantization further refined these ideas.  
Dong et al.~\cite{dong2019hawq} ranked network blocks by Hessian eigenvalues to assign bit-widths and defined an order for sequential quantization and fine-tuning with \acrshort{qat}.  

In most practical implementations, quantization follows a \emph{linear} scheme, in which the continuous value range is uniformly partitioned into discrete levels.  
Such linear quantizers are characterized by the quantization step size $Q_d$, the dynamic range $Q_{\mathrm{max}}$, and the bit-width $Q_b$, which are related by
\begin{align}
Q_{\mathrm{max}} = (2^{Q_b-1} - 1)\, Q_d .
\label{eq:linear_quantizer_relation}
\end{align}
Esser et al.~\cite{Esser2020} proposed to make the step sizes $Q_d^l$ trainable and learn them with the \acrshort{ste}, in contrast to fixed statistics as in XNOR-Net~\cite{Rastegari2016}.  
Uhlich et al.~\cite{Uhlich2020} extended this to mixed precision and compared alternative parameterizations of Eq.~\eqref{eq:linear_quantizer_relation}, finding that optimizing $(Q_d^l,Q_{\mathrm{max}}^l)$ by backpropagation yields stable training and implicitly determines the effective bit-widths $Q_b^l$.  

\paragraph{Quantization during backpropagation.}
Several methods also quantize gradients to further reduce training cost.  
Zhou et al.~\cite{Zhou2016} proposed flexible bit-width schemes for weights, activations, and backpropagation, highlighting the importance of stochastic quantization.  
Wu et al.~\cite{Wu2018} extended this idea with customized quantizers for weights, activations, and their gradients, enabling integer arithmetic throughout training and inference and accumulating updates directly in low precision.  

\paragraph{Theory and large-model results.}
Beyond empirical studies, theoretical work has analyzed convergence and approximation in quantized training~\cite{Li2017,Anderson2018}.  
Shekhovtsov and Yanush~\cite{Shekhovtsov2020} further showed that the \acrshort{ste} can be derived from linearization approximations in stochastic binary networks.  
Quantization is also increasingly applied to large language models~\cite{touvron2023llama,DBLP:journals/corr/abs-2005-14165,DBLP:journals/corr/abs-1909-08053}, where \acrshort{qat} is often impractical.  
Frantar et al.~\cite{frantar2023optq} addressed this with an efficient one-shot post-training quantization method for transformers, while Lin et al.~\cite{Lin2023AWQ} proposed activation-aware weight quantization that preserves salient weights via activation magnitudes and per-channel scaling.  

Most research on quantization emphasizes algorithmic methods and accuracy trade-offs, but hardware studies reveal additional critical factors.  
In our previous work~\cite{klein2021cacheboundness}, we showed that matrix multiplication and convolution on embedded \acrshort{arm} \acrshortpl{cpu} are typically \emph{cache-bound}: performance is limited by L1 bandwidth rather than arithmetic throughput.  
Quantization can mitigate this by reducing memory traffic, but the gains depend heavily on data layout and packing overhead, with bit-serial approaches particularly sensitive to bit-width.  
This analysis highlights that quantization must be co-designed with memory hierarchies and software stacks to achieve real speedups on embedded platforms.  

\subsection{Bayesian Approaches for Quantization}
\label{sec:literature_bayesian_quantization_approaches}

Several quantization methods connect naturally to Bayesian and variational inference.  
Achterhold et al.~\cite{Achterhold2018} extended the variational-dropout pruning approach of Louizos et al.~\cite{louizos2017} with mixtures of log-uniform priors centered at quantization levels, concentrating the posterior on discrete values and enabling quantization without fine-tuning.  

Other works model discrete weight distributions directly.  
Soudry et al.~\cite{Soudry2014} approximated the posterior with expectation propagation, while Shayer et al.~\cite{Shayer2018} optimized distributions over binary or ternary weights via the local reparameterization trick~\cite{Kingma2015a}.  
Peters and Welling~\cite{Peters2018} adapted this to sign activations, and Roth and Pernkopf~\cite{Roth2019} extended it to more than three discrete values.  

Van Baalen et al.~\cite{vanBaalen2020} introduced Bayesian mixed-precision quantization for power-of-two bit-widths, using recursive quantization with gates trained by variational inference; a zero-bit gate simultaneously enabled pruning.  

Havasi et al.~\cite{Havasi2019} proposed a Bayesian compression scheme that learns a mean-field variational posterior and encodes samples via importance-sampled atomic approximations, allowing efficient coding and recovery with a shared random seed.  

\paragraph{Summary.}
Quantization is one of the most effective and versatile compression techniques for \acrshortpl{dnn}.  
It spans a spectrum from simple 8-bit integer quantization to advanced mixed-precision and Bayesian formulations.  
Recent progress in learned quantizers, Hessian-aware policies, and Bayesian perspectives has improved robustness and flexibility.  
At scale, quantization is essential for deploying large models efficiently on both data-center accelerators and resource-constrained embedded systems.  
Overall, it provides a principled trade-off between efficiency and accuracy and remains a cornerstone of hardware-aware inference across model types and computing platforms.

\section{Pruning}
\label{sc:rennfes:pruning}

Pruning reduces the size and cost of \acrshortpl{dnn} by enforcing parameter sparsity: weights judged unimportant are set to zero, and the resulting sparsity is leveraged for efficiency.  
Two main paradigms are distinguished.  
\emph{Unstructured pruning} removes individual weights irrespective of their tensor location, typically preserving accuracy but requiring extreme sparsity and specialized structures to yield speedups.  
\emph{Structured pruning}, in contrast, eliminates entire neurons, channels, or filters, directly reducing tensor dimensions and remaining compatible with standard dense operations.  

The following sections review both unstructured and structured approaches, discuss Bayesian formulations, and conclude with methods that adapt pruning dynamically at inference time.  

\subsection{Unstructured Pruning}
Early work on pruning already explored sophisticated techniques beyond simple magnitude thresholds.  
LeCun et al.~\cite{LeCun1989} introduced \emph{optimal brain damage}, which uses a second-order Taylor expansion with a diagonal Hessian to estimate the loss increase from pruning individual weights.  
Weights with minimal estimated impact are removed, alternating with retraining to recover performance.  
Hassibi and Stork~\cite{Hassibi1992} refined this idea with \emph{optimal brain surgeon}, which approximates the full covariance matrix to prune low-impact weights while simultaneously adapting the remaining parameters.  
Although effective on small networks, these methods do not scale to modern architectures with millions of parameters, as computing and inverting the Hessian introduces prohibitive computational and memory costs.  

As a result, most later approaches returned to simpler magnitude-based pruning.  
Han et al.~\cite{Han2015} demonstrated that iteratively removing small weights and retraining can shrink networks substantially with little loss in accuracy.  
Their follow-up, \emph{Deep Compression}~\cite{han2016deepcompression}, extended this idea by combining pruning with quantization, weight sharing, and Huffman coding, achieving reductions of up to $49\times$ in memory footprint and 3–5$\times$ in energy use.  

Guo et al.~\cite{Guo2016} argued that pruning should be reversible.  
They introduced binary masks to track pruned weights and updated them with the \acrshort{ste}, allowing previously removed connections to reappear when beneficial.  

More recent work has also explored random sparsity.  
Gadhikar et al.~\cite{Gadhikar2023randompruning} showed that fixed random pruning patterns can induce effective sparsity at very low cost, making them attractive as initialization masks even if they are not optimal in isolation.  

\subsection{Structured Pruning}

Structured pruning removes entire groups of weights—such as neurons, channels, or filters—so that the resulting models remain compatible with dense tensor operations and yield hardware-level efficiency.  
Mariet and Sra~\cite{Mariet2016} used determinantal point processes to identify diverse, non-redundant neurons, pruning the rest and adjusting outgoing weights to minimize changes in the next layer.  
Wen et al.~\cite{Wen2016} incorporated group lasso regularization into training to induce sparsity at the level of filters, channels, or even whole layers.  
Liu et al.~\cite{Liu2017} exploited batch normalization, pruning channels by thresholding learned scaling factors, while Huang and Wang~\cite{Huang2018} introduced trainable scaling coefficients with $\ell^1$ regularization, driving entire structures to zero.  

Other methods adopt more data-driven criteria.  
ThiNet~\cite{Luo2017} pruned channels that minimally affected the output of the subsequent layer, followed by least-squares reconstruction of the remaining activations.  

Bayesian perspectives have also influenced structured pruning.  
Louizos et al.~\cite{Louizos2018} attached stochastic binary gates with trainable probabilities to weights or structures, encouraging sparsity with an $\ell^0$ regularizer and enabling differentiable training via Gumbel–softmax relaxations~\cite{Jang2017}.  
Li et al.~\cite{Li2019} refined this approach using an unbiased gradient estimator~\cite{Yin2019}.  

Finally, Liu et al.~\cite{Liu2019} argued that retraining pruned dense models is not always necessary: training sparse architectures from scratch can achieve comparable or better accuracy, suggesting that pruning is closely related to \acrlong{nas}.  

\paragraph{Bayesian Pruning.}
Bayesian approaches cast pruning as posterior inference over sparse weights.  
Graves~\cite{Graves2011} and Blundell et al.~\cite{blundell2015} applied mean-field variational inference with Gaussian posteriors and pruned weights with low signal-to-noise ratios.  
Variational dropout~\cite{Kingma2015a} provided another route: Molchanov et al.~\cite{Molchanov2017} optimized dropout rates per weight, removing those with high rates.  
Louizos et al.~\cite{louizos2017} extended this idea to structured groups using sparsity-promoting priors, linking pruning decisions to bit-width allocation via the minimum description length principle~\cite{Gruenwald2007}.  

\paragraph{Summary.}
Pruning reduces network size and computational cost by eliminating weights or structures with little impact on predictive accuracy.  
Unstructured methods can achieve very high sparsity but often fail to deliver speedups in practice, while structured pruning removes whole units such as neurons or channels, aligning with dense operations and yielding tangible efficiency gains.  
Together, these results highlight the heavy over-parameterization of modern networks and establish sparsity as a key mechanism for resource-efficient inference.  

\section{Neural Architecture Search} 

\Acrfull{nas} automates the search for effective \acrshort{dnn} architectures.  
Rather than relying on manual design, \acrshort{nas} explores a search space of candidate architectures and seeks models that optimize validation accuracy while increasingly incorporating resource efficiency objectives.  
This makes \acrshort{nas} particularly relevant for embedded deployment, where efficiency is as important as accuracy.  

The main difficulty lies in the high cost of evaluating candidate architectures, since each requires training and only yields noisy performance estimates, combined with the exponential growth of the search space.  
Zoph et al.~\cite{Zoph2017} cast \acrshort{nas} as a reinforcement learning task, where a controller \acrshort{rnn} generated architectures and received validation accuracy as reward.  
While effective, this approach required thousands of full training runs, making it impractical at larger scale.  
Follow-up work reduced cost through proxy tasks, parameter sharing, or differentiable relaxations~\cite{Zoph2018}.  

Several approaches incorporated hardware-awareness directly into the objective.  
MnasNet~\cite{tan2019mnasnet} extended reinforcement learning \acrshort{nas} with latency measurements on mobile devices, producing models that respect device-specific runtime constraints.  
Proxyless\acrshort{nas}~\cite{Cai2019} avoided proxy tasks by training over-parameterized networks in which each layer contained multiple candidate blocks; block-selection probabilities were optimized by gradient descent with the \acrshort{ste}, and predicted device latencies were included as a differentiable regularizer.  
Single-pass \acrshort{nas}~\cite{Stamoulis2019} further simplified this by consolidating all operations into a shared superblock, enabling efficient gradient-based training of both architecture and weights.  

EfficientNet~\cite{Tan2019} represents a widely adopted outcome of \acrshort{nas}.  
The method first searches for a small, resource-efficient baseline model and then enlarges it systematically using compound scaling of depth, width, and resolution, achieving state-of-the-art performance on ImageNet with comparatively modest model sizes.  

Beyond architecture design, \acrshort{nas} principles have also been applied to compression.  
Wang et al.~\cite{wang2019haq} used reinforcement learning to assign layer-wise bit widths for mixed-precision quantization, incorporating hardware-specific latency and energy constraints estimated from device-specific lookup tables.  
Similarly, Wu et al.~\cite{Wu2018b} optimized layer-wise quantization gates via differentiable stochastic optimization.  
These approaches demonstrate how search-based methods can jointly reason about architecture and compression to produce hardware-aware models.  

As discussed previously, pruning can also be interpreted as a form of implicit \acrshort{nas}.  
Liu et al.~\cite{Liu2019} showed that training sparse networks directly from scratch often matches or surpasses iterative prune–retrain pipelines, suggesting that pruning primarily identifies effective architectures rather than merely compressing weights.  

Building on the ideas of quantization, pruning, and \acrshort{nas}, the proposed \emph{Galen} framework~\cite{krieger2023galen} extends automatic compression by jointly searching over quantization levels and pruning ratios.  
The search is driven by reinforcement learning, while hardware-in-the-loop latency measurements and sensitivity analyses provide guidance on efficiency and accuracy.  
In this way, \emph{Galen} unifies multiple compression strategies within a single, hardware-aware optimization framework.  
A detailed discussion of this contribution follows in the next chapter.  

For a broader overview of \acrshort{nas}, including search spaces, optimization strategies, and performance estimation techniques, see Elsken et al.~\cite{elsken2019neural}.  
Together, these works highlight \acrshort{nas} as a flexible paradigm that can target not only accuracy but also resource efficiency and compression.  
\section{Hardware Platforms under Compression}
\label{sec:rennfes:hardware}

The efficiency of compression methods such as quantization and pruning strongly depends on the hardware platform on which the model is deployed.  
While the general characteristics of CPUs, GPUs, FPGAs, and domain-specific accelerators were introduced in Chapter~\ref{sec:background:hardware}, we here highlight how compression interacts with these platforms in practice.  
The discussion is guided by the results of \cite{roth2024jmlr}, which compare throughput–accuracy trade-offs across multiple hardware targets.

\paragraph{\acrshortpl{cpu}.}
\Acrlongpl{cpu} are highly flexible platforms that can, in principle, support low-bit-width quantization in software as well as unstructured sparsity.  
In practice, however, performance gains are most pronounced when using datatypes directly supported by the hardware, such as 8-bit integers, together with structured pruning that aligns with SIMD vector units and cache hierarchies.  
This makes \acrshortpl{cpu} particularly effective at exploiting structured sparsity, while unstructured pruning or very low-bit quantization typically require extreme compression ratios to compensate for the overhead of irregular memory access and instruction scheduling.  

\paragraph{\acrshortpl{gpu}.}
\Acrlongpl{gpu} benefit from both quantization and structured pruning.  
Tensor cores and \acrshort{simd} units efficiently accelerate dense matrix operations and support low-precision arithmetic (e.g., FP16, INT8, and in recent architectures even INT4).  
Structured pruning can also improve throughput, provided sparsity patterns align with the block-parallel execution paradigm, while unstructured pruning is largely ineffective due to irregular memory access.  
In practice, both structured pruning and quantization are employed on \acrshortpl{gpu}, with the relative advantage depending on the specific architecture, workload characteristics, and precision requirements.  

\paragraph{\acrshortpl{fpga}.}
\Acrlongpl{fpga} excel at extreme quantization, as arbitrary data formats and customized compute pipelines can be implemented directly.  
This makes them particularly suitable for binary or ternary neural networks.  
Pruning is also supported, but benefits are limited by on-chip memory and routing overhead.  
In practice, FPGAs are most effective when paired with aggressively quantized models that meet strict real-time constraints.  

\paragraph{Domain-Specific Accelerators.}
Dedicated accelerators such as Google’s \acrshort{tpu} are optimized for dense linear algebra, often implemented as systolic arrays.  
They achieve high utilization on structured workloads and natively support accelerator-specific low-precision datatypes such as 16-bit floating point or 8-bit integer, which enable speedups for accordingly quantized models by design.  
However, these platforms provide little flexibility for arbitrary precisions or fine-grained sparsity, so pruning is generally less effective unless explicitly supported by hardware extensions.  

\section{Evaluation on Embedded Hardware}
\label{sec:rennfes:experiments}

\begin{figure}
	\centering
    \begin{subfigure}{0.32\textwidth}
        \includegraphics[width=\textwidth]{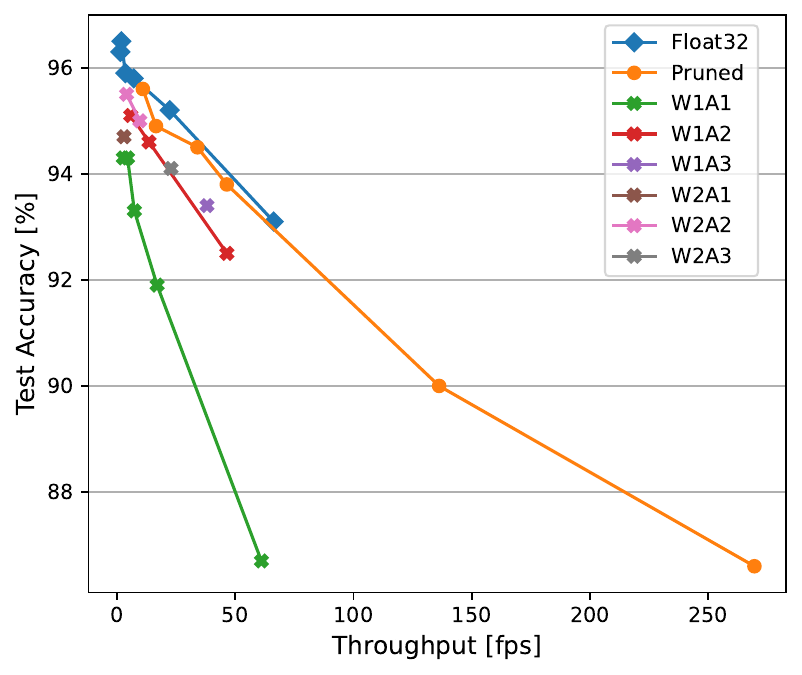}
        \caption{\acrshort{arm} \acrshort{cpu}}
        \label{fig:rennfes:tradeoff}
    \end{subfigure}
    \hfill
    \begin{subfigure}{0.32\textwidth}
        \includegraphics[width=\textwidth]{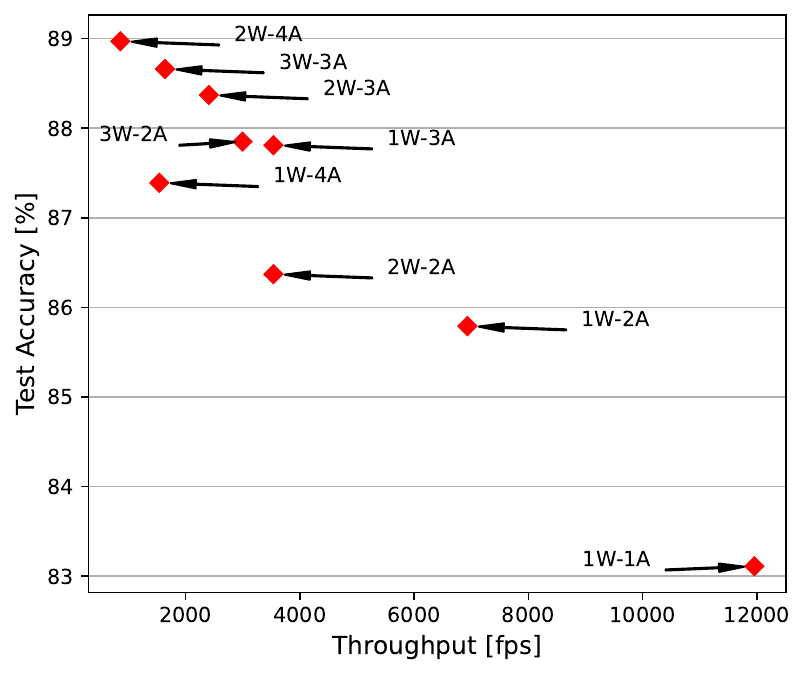}
        \caption{\acrshort{fpga}}
        \label{fig:rennfes:finn}
    \end{subfigure}
    \hfill
    \begin{subfigure}{0.32\textwidth}
        \includegraphics[width=\textwidth]{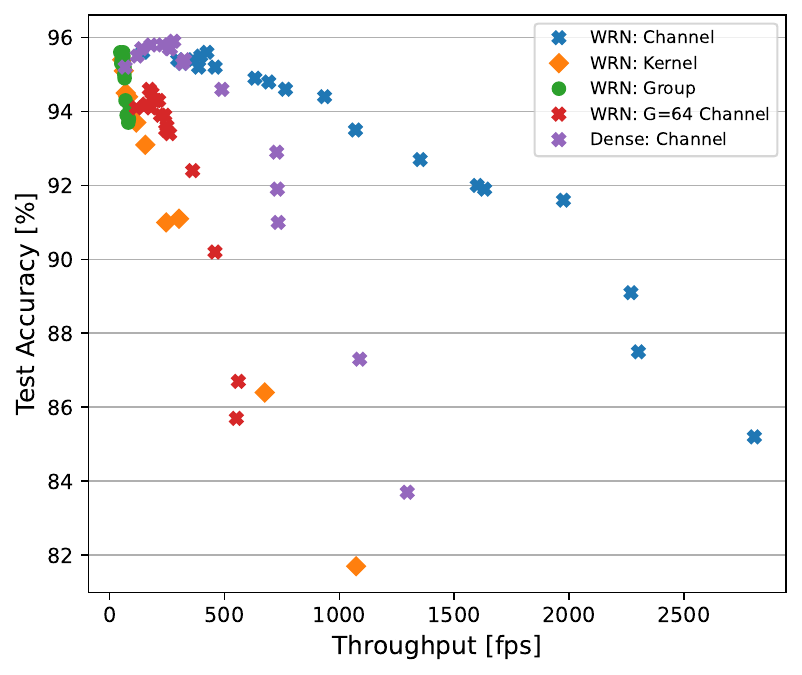}
        \caption{Embedded \acrshort{gpu}}
        \label{fig:rennfes:throughput_cifar}
    \end{subfigure}
    \caption{Throughput–accuracy trade-offs of compressed models on CIFAR-10 across embedded hardware platforms:  
	(a) quantized and pruned \acrshort{wrn}~\cite{WRN} models on an \acrshort{arm} \acrshort{cpu},  
	(b) quantized \acrshort{vgg}~\cite{Simonyan15} models on an \acrshort{fpga} data-flow architecture using FINN~\cite{Umuroglu2017}, and  
	(c) different pruning methods on an embedded \acrshort{gpu}.  
    \reprofrom{roth2024jmlr}}
    \label{fig:rennfes:tradeoffs}
\end{figure}
To complement the survey of compression techniques, we briefly review empirical results of executing compressed models on embedded processors.  
We compare quantization and pruning across representative embedded devices: an \acrshort{arm} \acrshort{cpu}, a reconfigurable \acrshort{fpga}, and an embedded \acrshort{gpu}.  
All devices operate in a comparable power envelope of about 5~W, which enables a fair comparison of efficiency–accuracy trade-offs.  

\paragraph{Embedded \acrshort{arm} \acrshort{cpu}}  
Figure~\ref{fig:rennfes:tradeoffs}~(a) shows throughput–accuracy trade-offs of pruned and quantized \acrshort{wrn}~\cite{WRN} models on an \acrshort{arm} Cortex-A53.  
Both quantization and pruning improve throughput over the baseline, but strong compression in either form reduces accuracy.  
Structured channel pruning achieves the highest throughput while maintaining competitive accuracy, reflecting that \acrshortpl{cpu} exploit sparsity more effectively than low-bit arithmetic.  

\paragraph{\acrshort{fpga}}  
On reconfigurable hardware, quantization is essential to fit models into limited on-chip resources.  
Figure~\ref{fig:rennfes:tradeoffs}~(b) shows quantized \acrshort{vgg}~\cite{Simonyan15} models on a XILINX Ultra96 \acrshort{fpga} using the FINN framework~\cite{Umuroglu2017}.  
Throughput rises sharply at very low bit-widths, but accuracy falls as precision decreases.  
The Pareto front is reached with extremely low-precision weights (1–2 bits) and moderately higher-precision activations (3–4 bits), confirming that activation quantization dominates predictive accuracy on \acrshortpl{fpga}.  

\paragraph{Embedded \acrshort{gpu}}  
Figure~\ref{fig:rennfes:tradeoffs}~(c) reports results for pruned \acrshort{wrn}~\cite{WRN} and DenseNet~\cite{Huang2017densenet} models on a Jetson Nano \acrshort{gpu}.  
All pruning methods improve throughput, but excessive compression reduces accuracy.  
Channel pruning strikes the best balance of accuracy and efficiency, while group and kernel pruning cut \acrshort{flops} without speeding up inference, underscoring that \acrshort{flops} are a poor runtime proxy on \acrshortpl{gpu}.  
Here, efficiency is more closely tied to reducing activation memory and aligning with the accelerator’s software stack.  

\begin{figure}
	\begin{center}
		\includegraphics[width=0.5\textwidth]{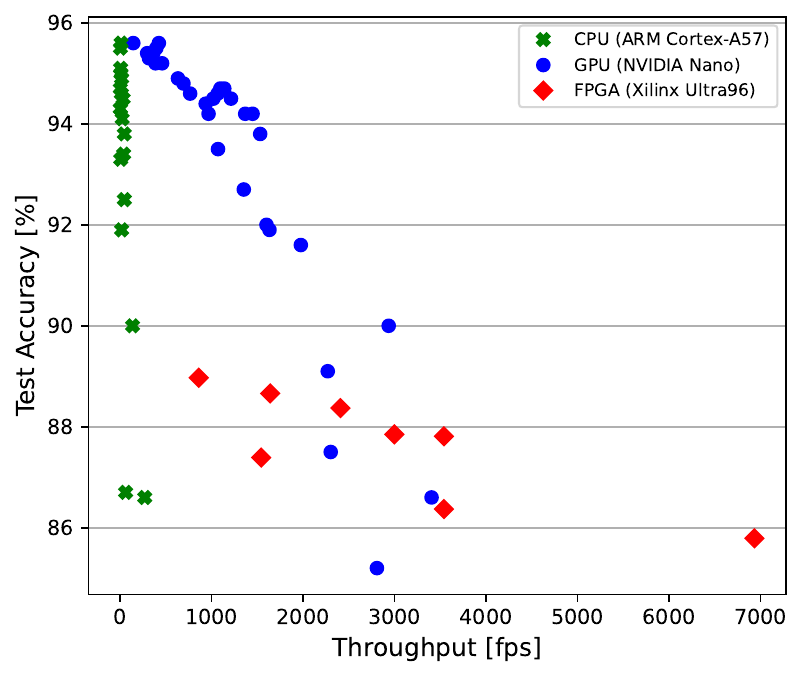}
		\caption{Throughput–accuracy trade-offs of compressed models on the CIFAR-10 task across different processor architectures (\acrshort{cpu}, \acrshort{fpga}, \acrshort{gpu}).  
		\acrshortpl{gpu} dominate the high-accuracy, high-throughput regime, being well suited for dense and batched matrix multiplications.  
		\acrshortpl{fpga} can deliver very high throughputs, but their limited on-chip resources often necessitate aggressive quantization, which reduces predictive accuracy.  
		\acrshortpl{cpu} offer high flexibility and support fine-grained sparsity, but in this comparison their general-purpose design yields lower throughput than the more parallel accelerator architectures when used together with structured compression.  
		\reprofrom{roth2024jmlr}}
		\label{fig:rennfes:throughput_final}
	\end{center}
\end{figure}

\paragraph{Cross-Platform Comparison}
Figure~\ref{fig:rennfes:throughput_final} summarizes throughput–accuracy trade-offs across \acrshort{cpu}, \acrshort{fpga}, and \acrshort{gpu} platforms.  
\acrshortpl{cpu} offer high flexibility and support fine-grained sparsity, but in this comparison their general-purpose design yields lower throughput than the more parallel accelerator architectures when used together with structured compression.  
\acrshortpl{gpu} dominate the high-accuracy, high-throughput regime, benefiting from massive parallelism and large memory.  
\acrshortpl{fpga} achieve outstanding throughput at low precision, but limited on-chip resources often necessitate aggressive quantization that reduces accuracy.  

These results highlight that the effectiveness of compression is highly hardware-dependent.  
While quantization is mandatory on \acrshortpl{fpga}, pruning aligns better with \acrshortpl{cpu}, and \acrshortpl{gpu} require compression strategies that optimize memory and software support rather than raw FLOPs.

\section*{Summary}
\label{sc:rennfes:summary}

This chapter introduced the foundations of resource-efficient inference of deep neural networks on embedded systems.  
We reviewed three main approaches: quantization, pruning, and \acrlong{nas}, and discussed how their benefits depend critically on the target hardware.  

From the survey and experiments, several lessons emerge:  
\begin{itemize}
\item \textbf{Quantization} reduces memory traffic and enables low-precision execution.  
    In image classification, accuracy is particularly sensitive to activation bit-width, while weight quantization is often more benign.  
    The actual benefits, however, depend strongly on memory hierarchies and data layouts.  
\item \textbf{Pruning} exploits over-parameterization to remove redundant weights or structures. Structured pruning yields practical speedups on dense hardware, whereas \acrshort{mac} reductions alone often fail to translate into latency gains.  
    \item \textbf{\Acrlong{nas}} provides a framework to automate design under accuracy--efficiency trade-offs, and recent work has extended these ideas to compression-aware settings.  
\end{itemize}

Hardware evaluations reinforced that compression is not universally effective: \acrshortpl{cpu} can leverage fine-grained sparsity, \acrshortpl{fpga} excel with low-precision quantization, and \acrshortpl{gpu} demand memory-efficient dense mappings.  
Crucially, FLOP counts and parameter numbers are poor predictors of real performance; hardware co-design and direct measurements are essential to identify viable efficiency gains.  

Together, these observations highlight both the promise and the complexity of resource-efficient inference.  
While individual techniques succeed under specific conditions, practical deployment requires balancing multiple strategies and aligning them with device characteristics.  
The next chapter introduces \textsc{Galen}~\cite{krieger2023galen}, a framework that addresses this challenge by automating the search over pruning and quantization parameters with reinforcement learning and hardware-in-the-loop feedback.  
Galen builds directly on the lessons of this chapter: rather than applying quantization and pruning in isolation, it integrates them into an automated framework that searches for compression strategies tailored to the constraints of a given hardware platform.  
\chapter{Galen: Automatic Model Compression}
\label{ch:galen}

\epigraph{Perfection is achieved not when there is nothing more to add, but when there is nothing left to take away.}{\textnormal{--- Antoine de Saint-Exupéry, \textit{Wind, Sand and Stars} (1939)}}

\noindent 
In practice, compression is often applied with \emph{global} settings—uniform bit widths and a single sparsity target across all layers—because selecting safe \emph{per-layer} quantization and pruning levels requires expertise in \acrshort{dnn} design and compression, as well as in the efficient implementation on the specific hardware architecture.
Per-layer compression, by contrast, allocates precision and sparsity at the layer granularity to exploit heterogeneity in hardware payoff.
Yet without automation and reliable hardware feedback, this flexibility remains largely unused in deployed systems.

Building on the previous chapter’s background on pruning and quantization, we now address the question of \emph{how to automate per-layer compression decisions for a concrete deployment device}.
Proxy objectives such as \acrshort{flops}, \acrshortpl{mac}, or \acrshort{bops} provide only weak guidance for end-to-end latency on real systems, where compiler optimizations, memory hierarchies, parallel execution, and vectorization dominate runtime.

\emph{Galen} targets this gap with a hardware-in-the-loop formulation that learns device-specific, layer-wise compression policies~\cite{krieger2023galen}.
The key idea is to align the objective with actual execution by coupling policy learning to two grounded signals: (i) \emph{measured} on-device latency obtained by compiling candidates with \acrshort{tvm}~\cite{chen2018tvm} and benchmarking on the target platform, and (ii) \emph{per-layer sensitivity} that quantifies how tolerant each layer is to pruning and quantization.
Optimizing against these signals enables the search to prefer policies that are robust in accuracy while improving real latency on the final hardware.

Conceptually, Galen extends the promise of combining compression methods—shown effective in \emph{Deep Compression}~\cite{han2016deepcompression}—to an automatic, hardware-aware setting.
In contrast to prior automatic approaches that optimize a single dimension under proxy cost models (e.g., pruning in \acrshort{amc}~\cite{he2018amc} or quantization in \acrshort{haq}~\cite{wang2019haq}), Galen \emph{jointly} learns per-layer policies guided by measured latency and sensitivity cues.
This design systematically exploits per-layer heterogeneity and closes the loop between compression decisions and measured on-device latency.

\section{Automatic Model Compression}

A large body of research has sought to automate model compression by framing it as an optimization problem. NetAdapt~\cite{yang2018netadapt} introduced a platform-aware greedy procedure that iteratively prunes channels and validates the effect on device latency, thereby integrating hardware measurements into the compression loop. Other methods replaced this heuristic search by more sophisticated optimization strategies, such as simulated annealing in AutoCompress~\cite{liu2020autocompress} or evolutionary algorithms in Automatic Structure Search~\cite{lin2020channelpruning}. While differing in their algorithms, these approaches are similar in that they largely optimize proxy metrics such as \acrshort{flops}, \acrshortpl{mac}, or parameter counts, which only approximate the behavior of the underlying hardware.

In parallel, mixed-precision quantization has been explored as a way to further reduce the cost of inference. Approaches in this line of work differ primarily in how they determine layer-wise bit widths. Some rely on sensitivity analysis to estimate the robustness of layers, with Hessian-aware methods such as HAWQ and HAWQ-V2 being prominent examples~\cite{dong2019hawq,dong2020hawqv2}. Others instead formulate policy generation as a learning problem, training reinforcement learning agents that predict quantization strategies, as in AutoQ and ReLeQ~\cite{lou2020autoq,elthakeb2020releq}. Despite the difference in methodology, these frameworks share a reliance on proxy estimates of latency—derived from look-up tables or analytic surrogates—rather than direct measurements of the target device.

The effectiveness of combining pruning and quantization was first demonstrated by \emph{Deep Compression}~\cite{han2016deepcompression}, which showed that high compression ratios can be achieved without compromising accuracy when multiple techniques are applied in sequence. Building on this insight, later work explored joint optimization formulations, for instance in CLIP-Q~\cite{tung2018clipq}, differentiable joint pruning–quantization objectives~\cite{wang2020djpq}, and joint architecture–compression searches such as APQ~\cite{wang2020apq}. These contributions underscore the value of multi-dimensional compression, yet they typically continue to optimize with respect to abstract metrics such as \acrshort{flops} or \acrfull{bops}~\cite{baskin2021uniq} rather than actual device behavior.

A related but distinct line of research lies in hardware-aware \acrshort{nas}, which aims to design new architectures under hardware constraints. Here, approaches diverge in the way device cost is incorporated. Some works continue to use simple analytic metrics such as \acrshort{flops} or parameter counts as a stand-in for latency. Others, such as FBNet, train cost models to predict latency from sampled measurements, thereby providing a differentiable surrogate objective~\cite{wu2019fbnet}. More recent frameworks, including MnasNet~\cite{tan2019mnasnet} and Once-for-All~\cite{cai2019once}, take a step further by relying on direct measurements on the target device. This progression from abstract metrics to cost models and finally to measured performance mirrors the developments observed in compression research, highlighting the importance of grounding optimization in real device characteristics.

Among these many directions, \acrshort{amc}~\cite{he2018amc} and \acrshort{haq}~\cite{wang2019haq} are most closely related to our work. Both adopt a reinforcement learning formulation to derive layer-wise compression policies, where a policy network sequentially decides pruning or quantization actions. While these frameworks demonstrated the feasibility of automated compression, they remain limited to a single dimension—pruning for \acrshort{amc} and quantization for \acrshort{haq}—and rely on proxy metrics to enforce efficiency, namely \acrshort{flops} or parameter counts in \acrshort{amc}, and look-up-table latency estimates in \acrshort{haq}. In contrast, Galen~\cite{krieger2023galen} jointly reasons about pruning and quantization, incorporates layer sensitivity into the optimization, and bases its decisions on on-the-fly hardware measurements, thereby closing the gap between proxy objectives and actual device performance.

\section{Galen Methodology}

The goal of \emph{Galen} is to automatically derive hardware-aware compression policies that balance model accuracy and inference latency. To this end, Galen frames compression as a reinforcement learning problem, where an agent incrementally decides how much to prune or quantize each layer of a network. After applying a complete policy, the resulting model is compiled and benchmarked on the target device, and the measured accuracy–latency trade-off serves as feedback for learning. This section describes the algorithmic concept, the formulation of pruning and quantization policies, the design of Galen’s agents, and the integration of sensitivity analysis and hardware latency measurements.

\begin{figure}
	\centering
	\includegraphics[width=0.49\textwidth]{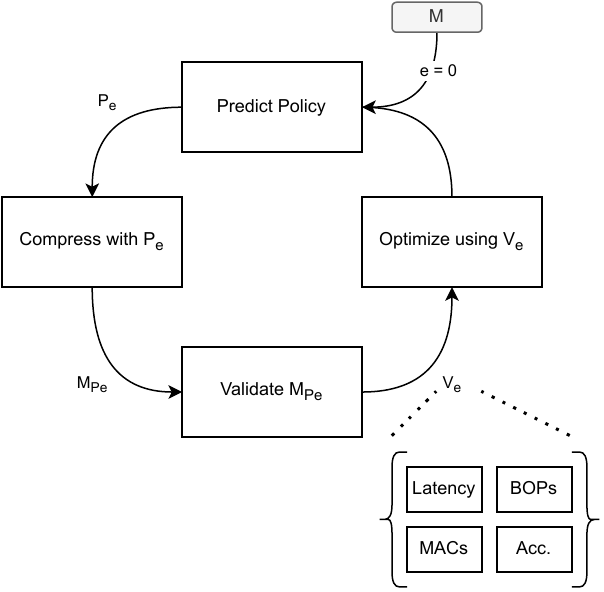}
	\caption{Episode overview in \emph{Galen}: (1) predict a compression policy; (2) apply it to obtain the compressed model; (3) evaluate on-device (compile, benchmark, and validate accuracy); (4) update the agent via the reward. \reprofrom{krieger2023galen}}
	\label{fig:galen:alg-schema:episode}       
\end{figure}

\begin{figure}
	\centering
	\includegraphics[width=0.7\textwidth]{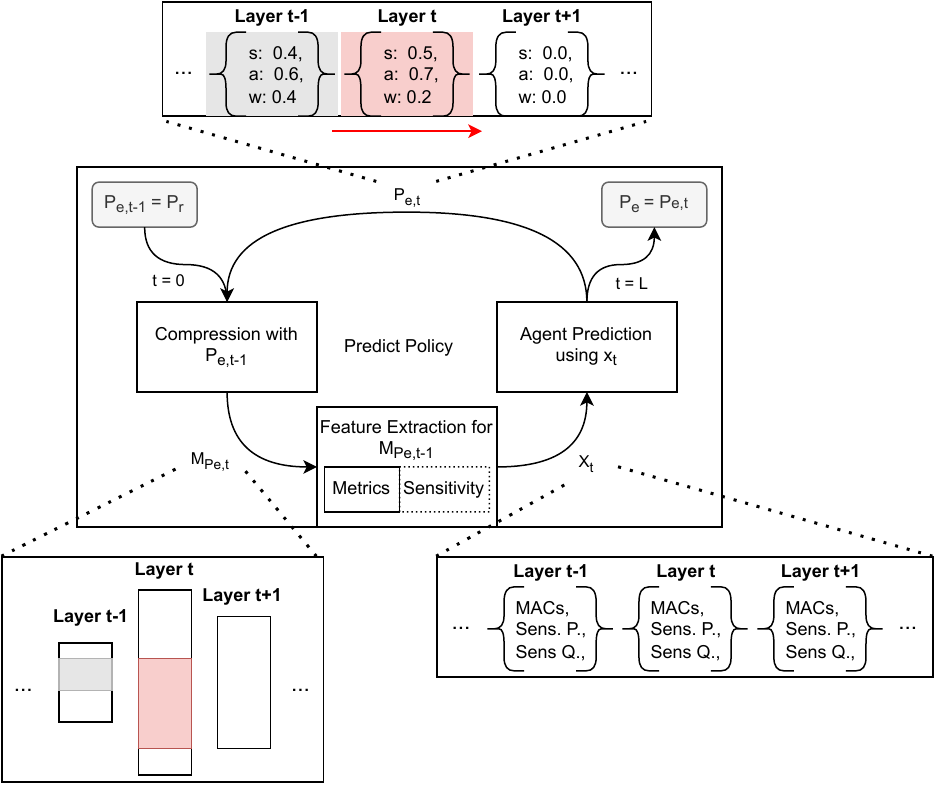}
\caption{Intra-episode policy prediction cycle: at each step the agent proposes pruning/quantization parameters for one layer and appends them to the policy before proceeding. \reprofrom{krieger2023galen}}
	\label{fig:galen:alg-schema:step} 
\end{figure}

\subsection{Algorithmic Concept}

At the center of Galen is the notion of a \emph{compression policy} $P$, which specifies how each layer of a model $M$ is transformed. Applying $P$ yields the compressed model $M^P$. A policy consists of normalized compression method parameters (CMPs), where each parameter lies in the interval $[0,1]$:

\begin{equation}
P \in \{r \in \mathbb{R}^K \mid r_i \in [0,1]\}^{L \times M},
\end{equation}

with $L$ the number of layers, $M$ the number of supported methods, and $K$ the dimensionality of parameters per method. Although pruning and quantization are discrete in nature (channels, datatypes, bit widths), CMPs are expressed as continuous variables to provide a uniform representation across layers. A mapping step later converts these normalized values into discrete hardware-specific configurations. This design reduces the complexity of the search space: the reinforcement agent always operates on normalized CMPs, without first having to learn layer-specific channel counts or quantization ranges.

The search problem is formulated as a constrained optimization task:

\begin{equation}
\hat{P} = \arg\max_P \; acc(M^P(\theta;x), y) \quad \text{s.t.} \quad cost(M^P) \leq c \cdot cost(M),
\end{equation}

where $acc(\cdot)$ is the accuracy of $M^P$ on data $(x,y)$, $\theta$ the model parameters, and $c$ the target cost ratio. 
Unlike prior works that approximate cost with \acrshort{flops}, parameter counts, or \acrshort{bops}~\cite{baskin2021uniq,size2020how,umuroglu2019bitserial}, \emph{Galen} instead uses measured inference latency on the target device as the actual optimization constraint~\cite{krieger2023galen}.  

\subsection{Algorithmic Schema}

As shown in Figures~\ref{fig:galen:alg-schema:episode} and~\ref{fig:galen:alg-schema:step}, an \emph{episode} corresponds to compressing a model once.  
At each step $t$, the agent observes the current layer features $s_t$ (layer type, index, number of parameters, sensitivity score) and predicts continuous actions $a_t$, which encode pruning and quantization policies.  
The sequence of decisions across all layers forms a complete compression policy $P_e$.  
The model $M^{P_e}$ is compiled with Apache \acrshort{tvm}~\cite{chen2018tvm}, executed on the target hardware, and evaluated for accuracy.  
Leveraging \acrshort{tvm} enables \emph{Galen} to target a wide range of \acrshort{cpu} and \acrshort{gpu} architectures.  
Retraining of compressed models for 1–5 epochs is included in each episode to recover accuracy.  
This step is particularly important for pruning, where performance initially drops substantially but typically recovers after only a few epochs.  
The measured latency and accuracy together define the reward $R(P_e)$ used to update the agent.  

\subsection{Compression Methods}

\paragraph{Structured pruning} Galen performs structured pruning by removing \emph{output channels} from convolutional or linear layers, following an $\ell_1$-norm criterion~\cite{li2017pruning,he2017channel}. Skip connections and other structural constraints are respected by the pruning procedure.

\paragraph{Quantization}
\emph{Galen} supports three quantization schemes: mixed precision with independent bit widths between 1 and 8 (\emph{MIX}), fixed-point 8-bit integer quantization (\emph{INT8}), and no quantization (\emph{FP32}). Mixed-precision support is hardware-dependent and not available for all hardware backends.

For each layer, the agent predicts continuous policy values $q_t^W, q_t^A \in [0,1]$ for weights and activations. These are first mapped to one of the three datatypes by a threshold rule:
\begin{equation}
d(q) =
\begin{cases}
\text{FP32}, & q \le \tau_{\text{fp}}, \\[4pt]
\text{INT8}, & \tau_{\text{fp}} < q \le \tau_{\text{int}}, \\[4pt]
\text{MIX},  & q > \tau_{\text{int}}.
\end{cases}
\end{equation}

If a layer is assigned FP32 or INT8, its tensors are kept in full precision or quantized to 8 bits, respectively. If the layer is assigned MIX, a bit-serial operator is used, and the agent’s policy additionally determines the bit width $b \in \{1,\dots,8\}$ for weights and activations. In this case, a real-valued input $r$ is mapped to its quantized form as
\begin{equation}
Q(r) = \max\!\left(-n, \; \min\!\left(n, \; \left\lfloor s \cdot r - z \right\rfloor \right)\right),
\end{equation}
with
\begin{equation}
\label{eq:galen:qparams}
n = 2^{b} - 1, \qquad
s = \frac{n}{x_{\max} - x_{\min}}, \qquad
z = \left\lfloor s \cdot x_{\min} \right\rfloor + 2^{b-1}.
\end{equation}
where $x_{\min}$ and $x_{\max}$ denote the calibrated tensor range. Quantization uses uniform asymmetric quantization with dynamic range calibration. During search, we employ fake quantization to simulate reduced precision in training.

\subsection{Sensitivity Analysis}

To guide compression, Galen augments the agent state with per-layer sensitivity scores. These scores quantify how strongly a layer reacts to compression. For each candidate layer $l$, only that layer is compressed at a chosen strength, while all others remain uncompressed. The output distribution of the modified network is then compared with the baseline network using the Kullback–Leibler divergence~\cite{kullback1951kl}, following the ZeroQ methodology~\cite{cai2020zeroq}:

\begin{equation}
S(l) = \frac{1}{N}\sum_{j=1}^{N} D_{\text{KL}}\!\big( f(x_j) \,\|\, f^{(l,q)}(x_j)\big),
\end{equation}

where $f$ is the original network, $f^{(l,q)}$ the network with only layer $l$ compressed at strength $q$, and $x_j$ calibration inputs. Low $S(l)$ indicates tolerance, high $S(l)$ indicates fragility. These scores are included in the state representation $s_t$, steering the agent toward hardware–robust trade-offs.

\subsection{Direct Hardware Latency Benchmarks}

Proxy cost models such as \acrshort{flops}, \acrshortpl{mac}, or \acrshort{bops} do not capture real device behavior~\cite{size2020how,umuroglu2019bitserial}. Galen therefore benchmarks each compressed model directly on the target hardware. Compiled with TVM~\cite{chen2018tvm}, each candidate $M^P$ is executed to measure $T^{M^P}$, which is used as part of the reward.

\subsection{Reward Function}

The reward function follows the absolute formulation of Bender et al.~\cite{bender2020tunas}, balancing accuracy and latency:

\begin{equation}
r(P) = acc(M^P) + \beta \cdot \left| \frac{T^{M^P}}{c \cdot T^M} - 1 \right|,
\end{equation}

where $acc(M^P)$ is the accuracy of the compressed model, $T^{M^P}$ and $T^M$ are the measured latencies of compressed and original model, $c$ is the target compression ratio, and the negativ cost exponent $\beta < 0$ penalizes exceeding the latency budget. Exponential reward functions~\cite{tan2019mnasnet} were also evaluated but were less stable in practice.

\subsection{Compression Agents}

All Galen agents build on the \acrfull{ddpg} algorithm~\cite{lillicrap2015ddpg}, which supports continuous actions. \acrshort{ddpg} combines an actor–critic architecture with experience replay. Actor outputs are perturbed by Ornstein–Uhlenbeck noise~\cite{uhlenbeck1930theory} for exploration. Both actor and critic are implemented as fully connected networks with two hidden layers (400 and 300 neurons, ReLU). Target networks use soft updates with $\tau=10^{-3}$, and the replay buffer stores $10^6$ transitions.

Three agent types were trained: (i) a pruning agent predicting $p_t$; (ii) a quantization agent predicting $(q_t^W, q_t^A)$; and (iii) a joint agent predicting both. While sharing the same backbone and training setup, the agents differ in state encoding and action dimensionality. 

\section{Experimental Evaluation and Discussion}

We evaluate Galen on ResNet-18~\cite{he2016deep} trained on CIFAR-10~\cite{krizhevsky2009learning}, deployed on a Raspberry~Pi~4B with an ARM Cortex-A72 \acrshort{cpu}. Models are compiled with Apache \acrshort{tvm}~\cite{chen2018tvm}, and candidate configurations are executed on the device to obtain measured inference latencies.
During the search, each candidate model is briefly retrained for 1-5 epochs before validation, while the final compressed models are retrained for 30 epochs.  

The quantization agent is trained for 310 episodes, while pruning and joint agents are trained for 410 episodes each, including 10 warm-up episodes. The reward function follows the absolute penalty form with a negativ cost exponent of $\beta = -3.0$, ensuring that policies converge close to the target compression budget.

\paragraph{General Performance} 
Table~\ref{tab:galen:agents} summarizes the quantitative results. At a moderate compression target ($c{=}0.3$), all agents achieve latency reduction close to the target and maintain accuracy near the baseline. At more aggressive compression ($c{=}0.2$), however, the quantization-only agent collapses to 45\% accuracy because it is forced into extremely low bit widths. The pruning and joint agents, in contrast, remain accurate. As expected, pruning minimizes \emph{\acrshortpl{mac}}, quantization reduces \emph{\acrshort{bops}} most efficiently, and the joint agent balances both—yielding the best accuracy–latency trade-off even though it is not maximal in either MAC or BOP reduction.

\begin{table}%
	\caption{Compressed model performance per agent with target compression ratio $c$. \reprofrom{krieger2023galen}}
	\centering
	\small	
	\begin{tabular}{l|c|l|r|r|r}
	\bfseries Method & \bfseries c & \multicolumn{1}{c|}{\bfseries \acrshortpl{mac}} & \multicolumn{1}{c|}{\bfseries \acrshort{bops}} &  \multicolumn{1}{c|}{\bfseries Latency} & \multicolumn{1}{c}{\bfseries Accuracy} \\ \hline
	Uncompressed 			& 							& $4.75 \cdot 10^{10}$ 	&  $4.86 \cdot 10^{13}$ 		&  $330$\,ms 		&  	$93.0$\,\% \\ 
	\hline
	Pruning Agent			& \multirow{3}{*}{0.3} & $1.42 \cdot 10^{10}$ 	&  $1.45 \cdot 10^{13}$ 		&  $98$\,ms 		& 	$93.0$\,\% \\
	Quantization A.		&							& $4.75 \cdot 10^{10}$ 	&  $8.23 \cdot 10^{11}$ 		&  $98$\,ms 		&  	$92.5$\,\% \\
	Joint Agent			&							& $4.35 \cdot 10^{10}$ 	&  $9.42 \cdot 10^{11}$ 		&  $99$\,ms 		& 	$93.2$\,\% \\
	\hline
	Pruning Agent			& \multirow{3}{*}{0.2}	& $9.24 \cdot 10^{9}$ 	&  $9.45 \cdot 10^{12}$ 		&  $66$\,ms 		& 	$92.4$\,\% \\
	Quantization A.		&							& $4.75 \cdot 10^{10}$ 	&  $4.01 \cdot 10^{11}$ 		&  $57$\,ms 		&  	$45.0$\,\% \\
	Joint Agent			&							& $2.82 \cdot 10^{10}$ 	&  $6.74 \cdot 10^{11}$ 		&  $64$\,ms 		& 	$92.8$\,\% \\
	\end{tabular}
	\label{tab:galen:agents}
\end{table}

\begin{figure}
    \centering
    \includegraphics[width=0.7\textwidth]{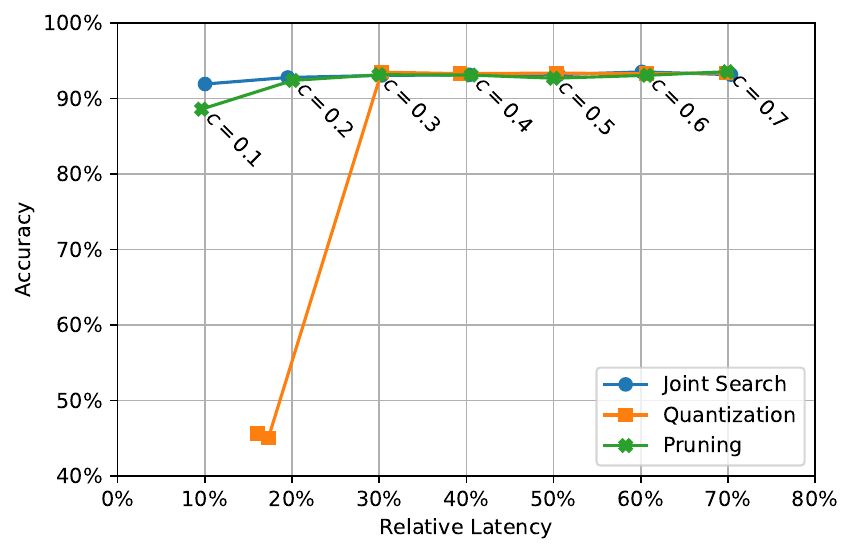}
\caption{Accuracy–latency trade-offs across target compression ratios $c$: all three agents perform well for moderate $c$; under aggressive targets the quantization agent suffers accuracy loss from very low bit widths, while pruning and the joint agent remain competitive, with the joint agent slightly superior at the extreme. \reprofrom{krieger2023galen}}
	\label{fig:galen:exp02:frontier}
\end{figure}

Figure~\ref{fig:galen:exp02:frontier} depicts the accuracy–latency trade-offs as the target compression ratio $c$ varies.
As expected, the pruning agent consistently achieves the largest reductions in \acrshortpl{mac} and only influences \acrshort{bops} indirectly via structural changes, whereas the quantization agent is most effective at reducing \acrshort{bops}.
The joint agent consistently lands closer to the Pareto front: it is \emph{not} the extreme winner on \acrshortpl{mac} or \acrshort{bops} alone, yet achieves the best latency–accuracy trade-off for a given target latency by effectively balancing pruning and quantization.  

\begin{figure}
	\centering
    \begin{subfigure}{0.7\textwidth}
        \centering
        \includegraphics[width=\textwidth]{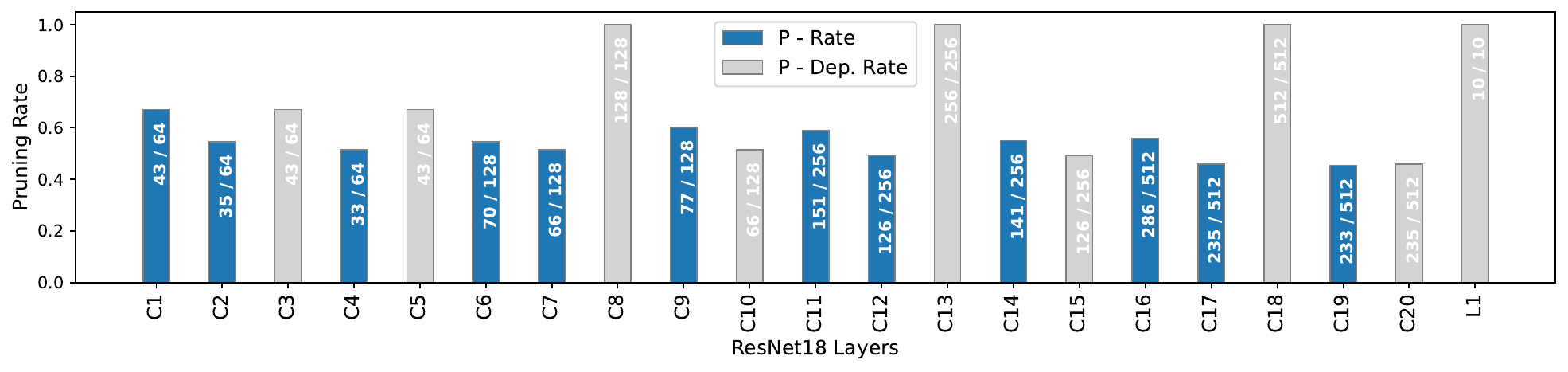}
        \caption{Pruning agent}
    \end{subfigure}

    \begin{subfigure}{0.7\textwidth}
        \centering
        \includegraphics[width=\textwidth]{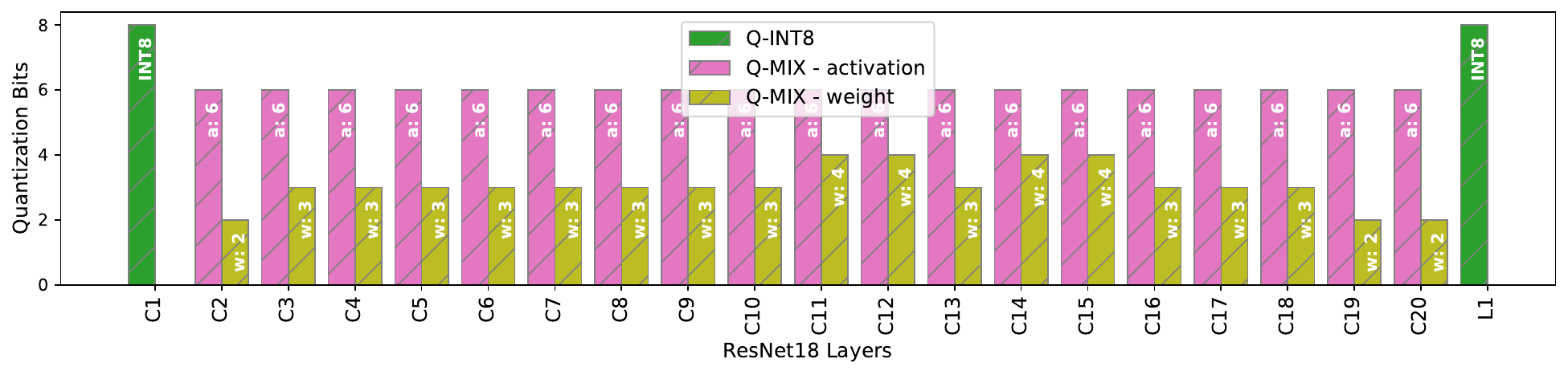}
        \caption{Quantization agent}
    \end{subfigure}

    \begin{subfigure}{0.7\textwidth}
        \centering
        \includegraphics[width=\textwidth]{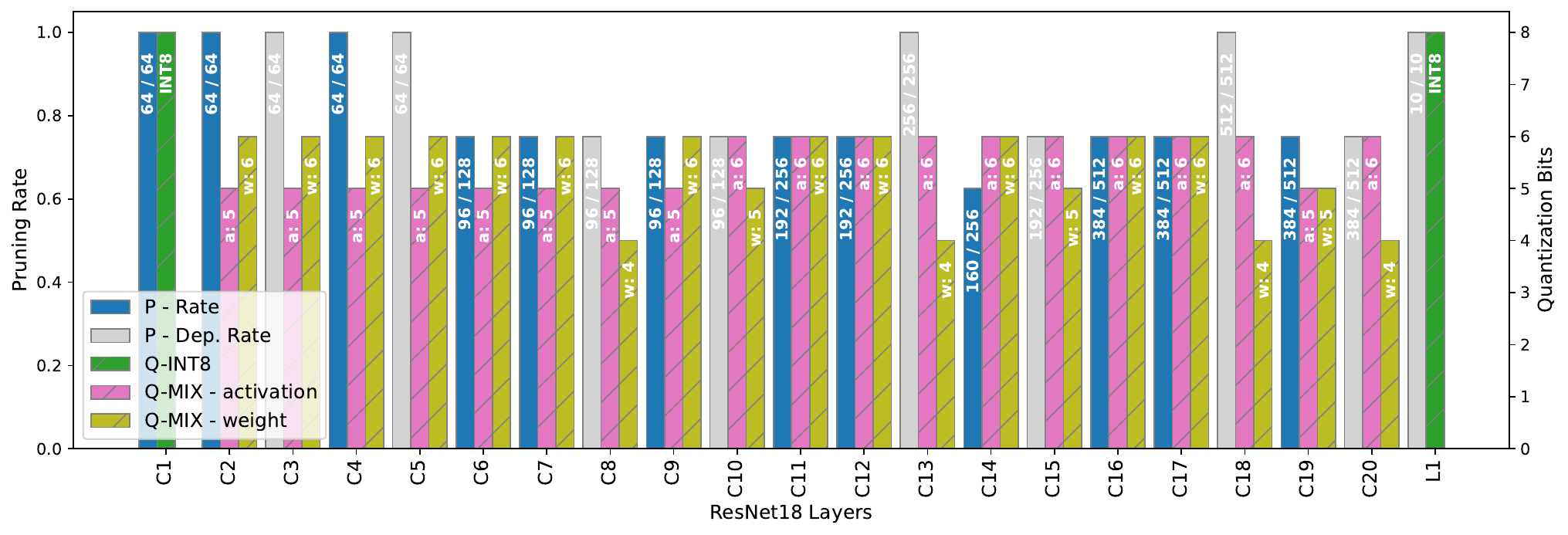}
        \caption{Joint agent}
    \end{subfigure}
\caption{Predicted per-layer policies at $c=0.3$ for pruning, quantization, and joint agents; bars show remaining channels (pruning) and activation/weight bit widths (quantization). \reprofrom{krieger2023galen}}
	\label{fig:galen:exp01}
\end{figure}

\paragraph{Policy Analysis}
Figure~\ref{fig:galen:exp01} illustrates the layer-wise policies for $c{=}0.3$. The pruning agent spreads sparsity across most layers, with a tendency to prune later layers more strongly. The quantization agent varies bit widths across layers, typically quantizing weights more aggressively than activations—an observation in line with expert knowledge for image classification models. As in common practice, the first and last layers are quantized less aggressively. The joint agent combines both methods in a balanced manner, moderating pruning and quantization simultaneously to preserve accuracy while meeting latency.

\begin{figure}
	\centering
    \begin{subfigure}{0.7\textwidth}
        \centering
        \includegraphics[width=\textwidth]{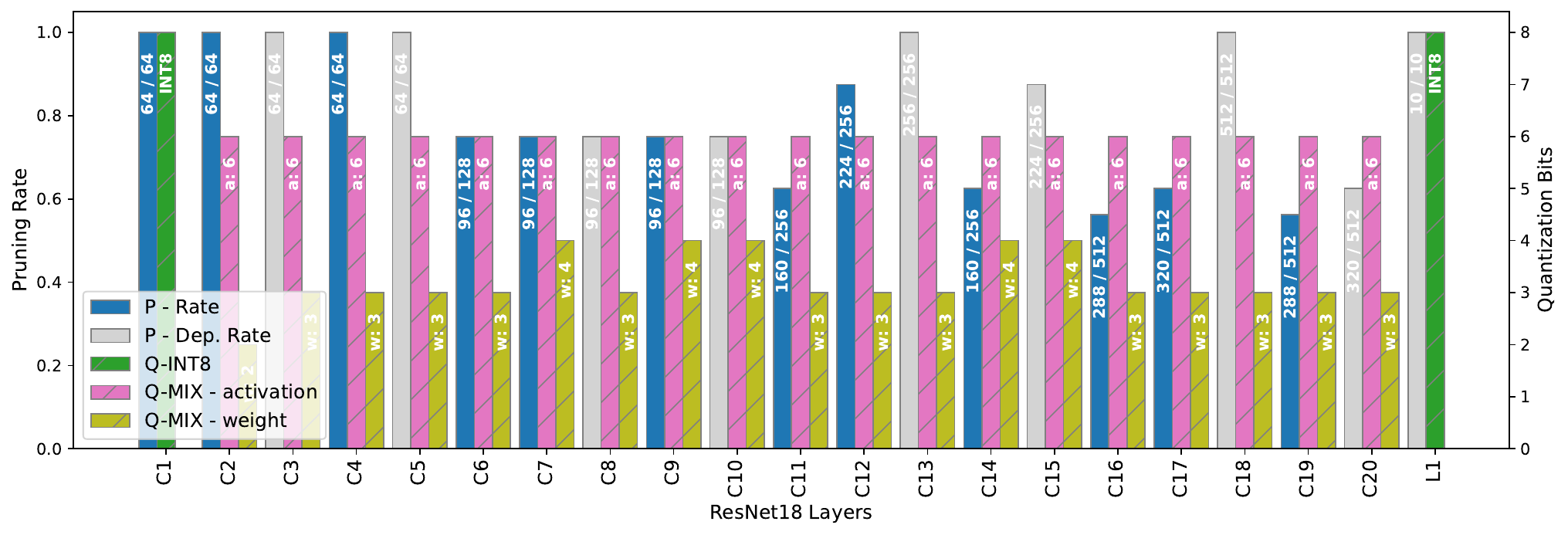}
        \caption{Pruning $\rightarrow$ quantization}
    \end{subfigure}

    \begin{subfigure}{0.7\textwidth}
        \centering
        \includegraphics[width=\textwidth]{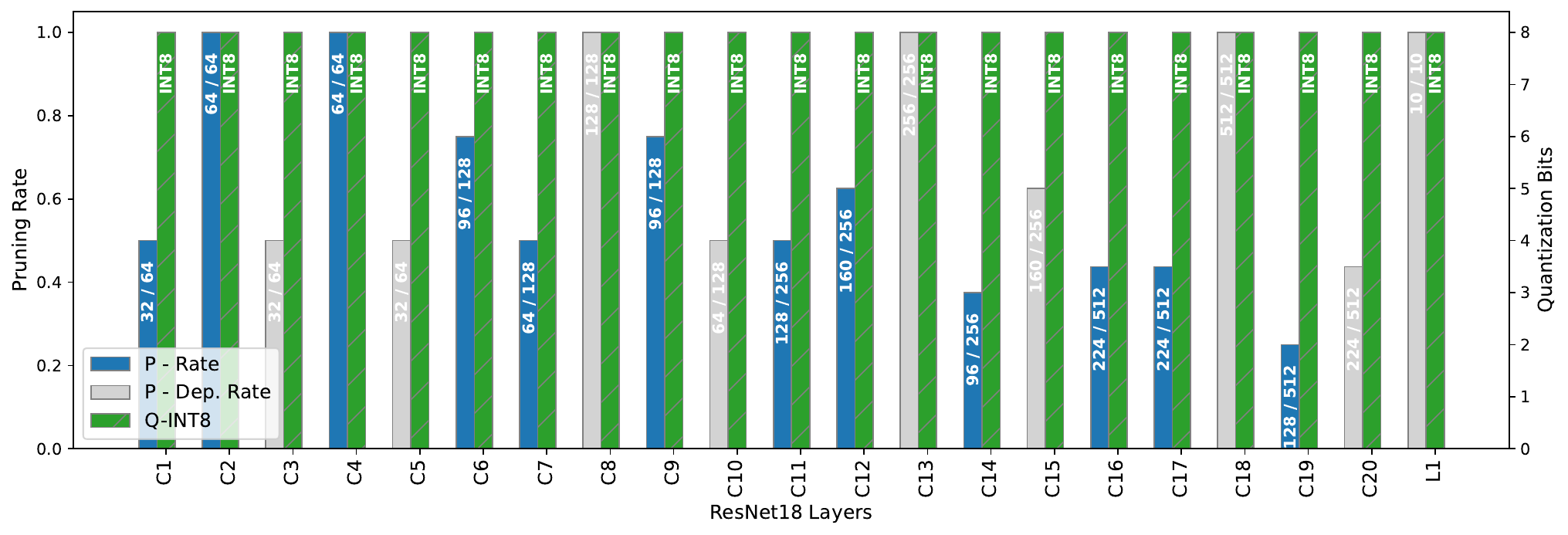}
        \caption{Quantization $\rightarrow$ pruning}
    \end{subfigure}

	\begin{subfigure}{0.7\textwidth}
        \centering
        \includegraphics[width=\textwidth]{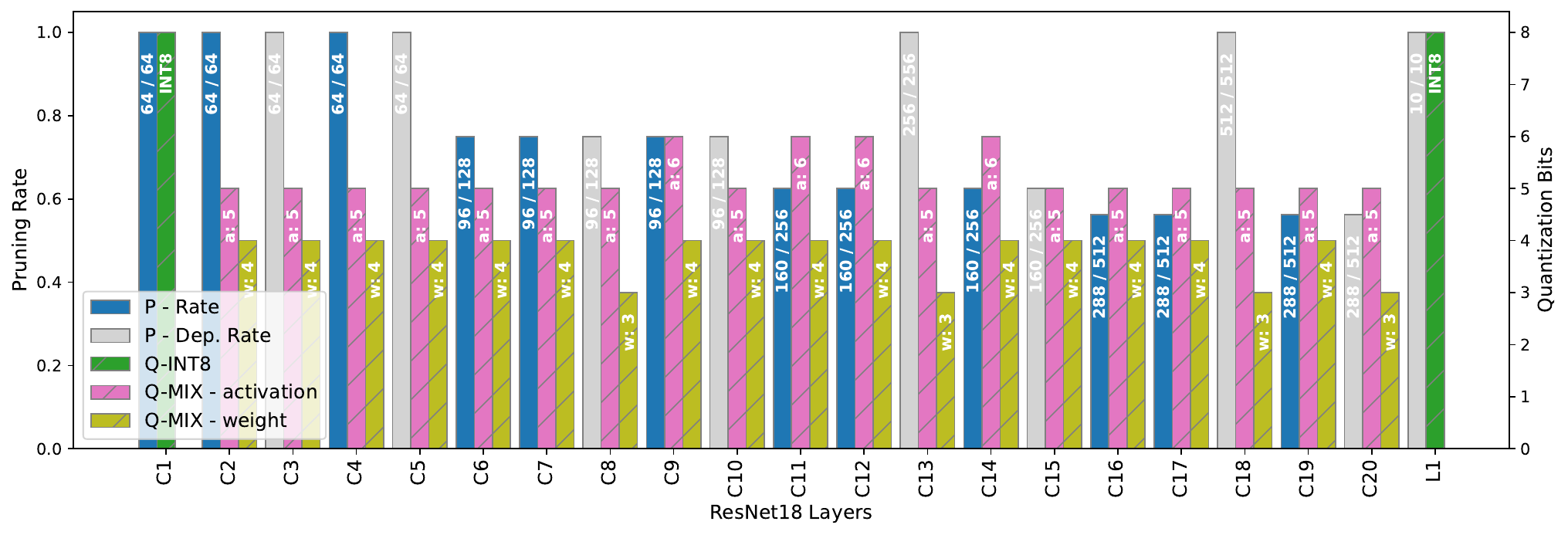}
        \caption{Joint search}
    \end{subfigure}
	\caption{Sequential vs.\ joint search at $c=0.2$: sequential pipelines overuse the second stage (harsher compression), whereas joint search balances pruning and quantization across layers. \reprofrom{krieger2023galen}}
	\label{fig:galen:exp03:policies}
\end{figure}

\paragraph{Sequential versus Joint Search}  
To test the importance of joint optimization, we also evaluate sequential application of pruning and quantization. As Figure~\ref{fig:galen:exp03:policies} shows, sequential pipelines tend to overuse the second stage, producing harsher compression and more accuracy loss. Joint search instead balances both simultaneously, yielding smoother policies and higher accuracy. This confirms that joint optimization is essential for Galen’s effectiveness.

\begin{figure}
    \centering
    \begin{subfigure}{0.3\textwidth}
        \centering
        \includegraphics[width=\textwidth]{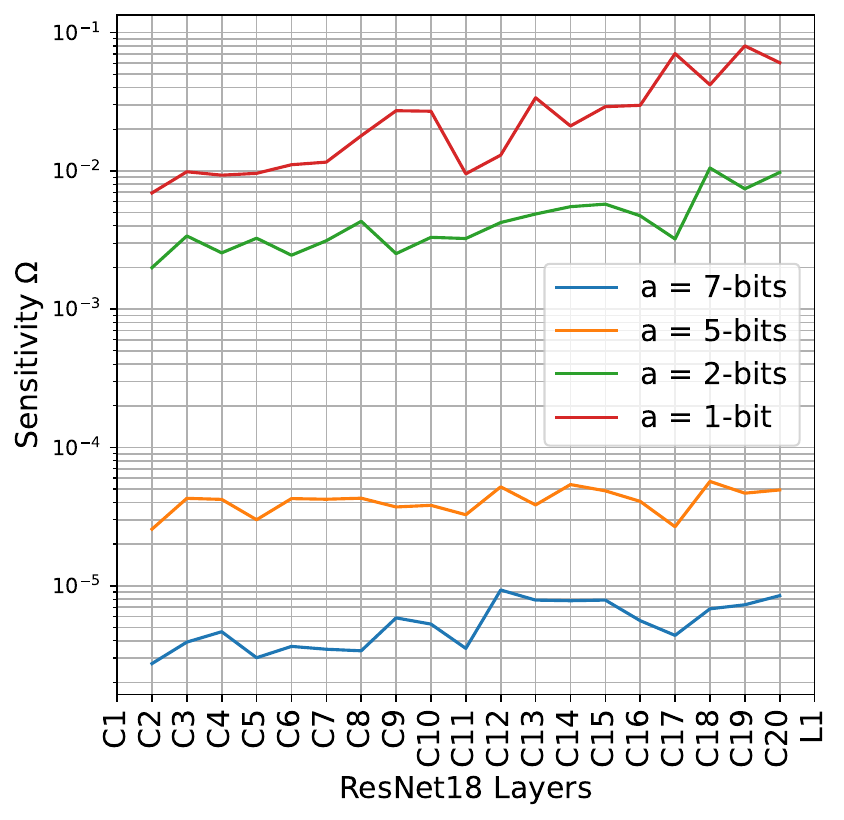}
        \caption{Activation sensitivity}
    \end{subfigure}
    \begin{subfigure}{0.3\textwidth}
        \centering
        \includegraphics[width=\textwidth]{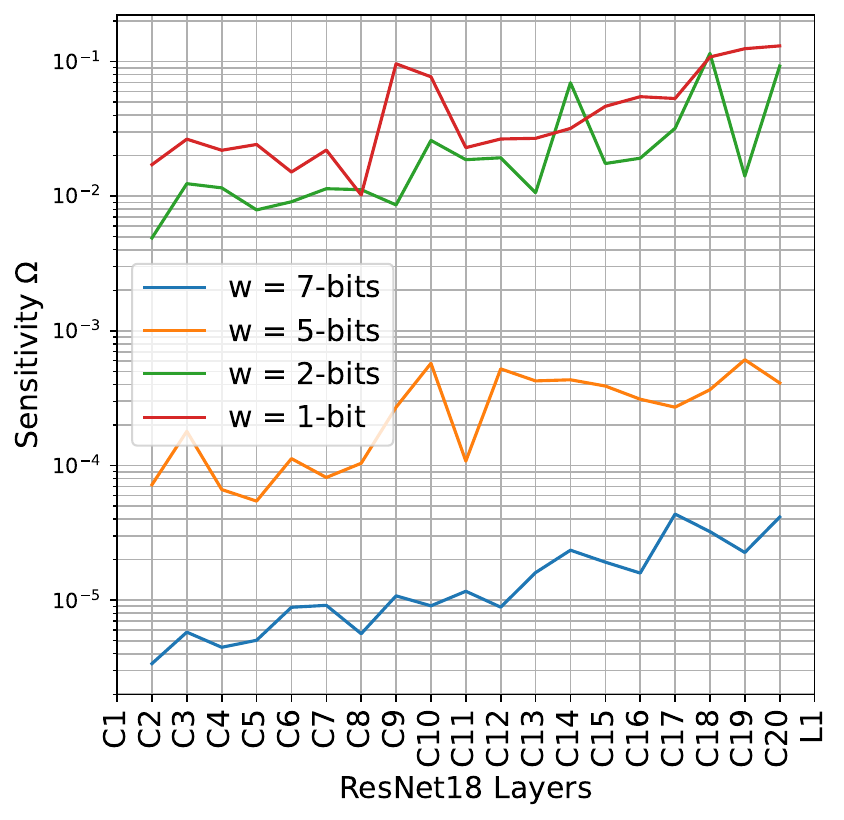}
        \caption{Weight sensitivity} 
    \end{subfigure}
	\begin{subfigure}{0.3\textwidth}
        \centering
        \includegraphics[width=\textwidth]{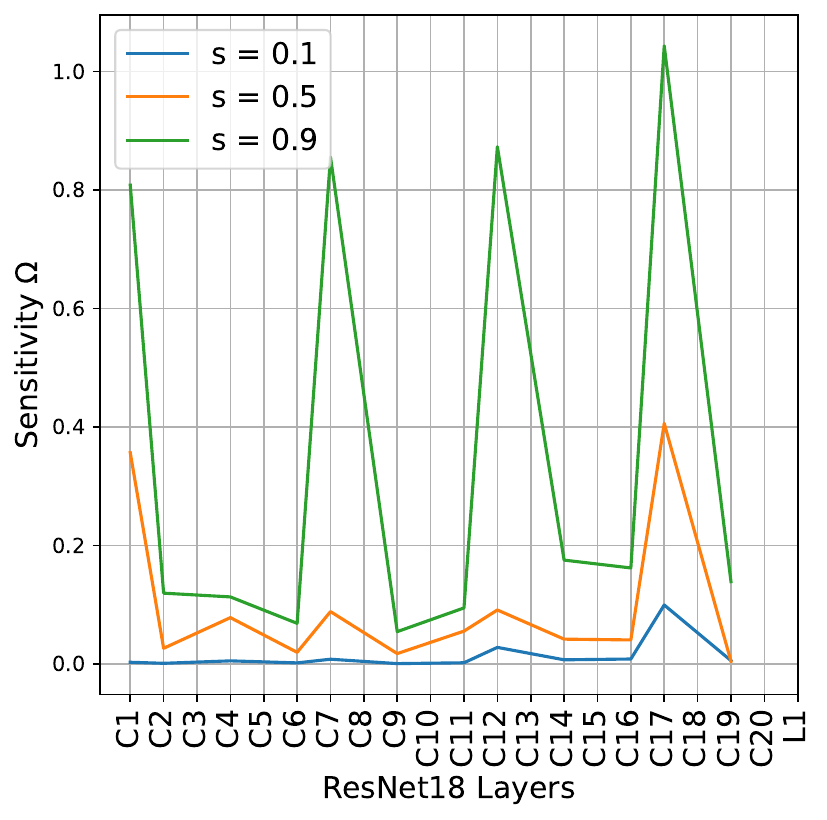}
        \caption{Pruning sensitivity} 
    \end{subfigure}
\caption{Per-layer sensitivity to activation quantization, weight quantization, and pruning. \reprofrom{krieger2023galen}}
	\label{fig:galen:eval:sens:aw}
\end{figure}

\begin{figure}
    \centering 
    \begin{subfigure}{0.7\textwidth}
        \centering
        \includegraphics[width=\textwidth]{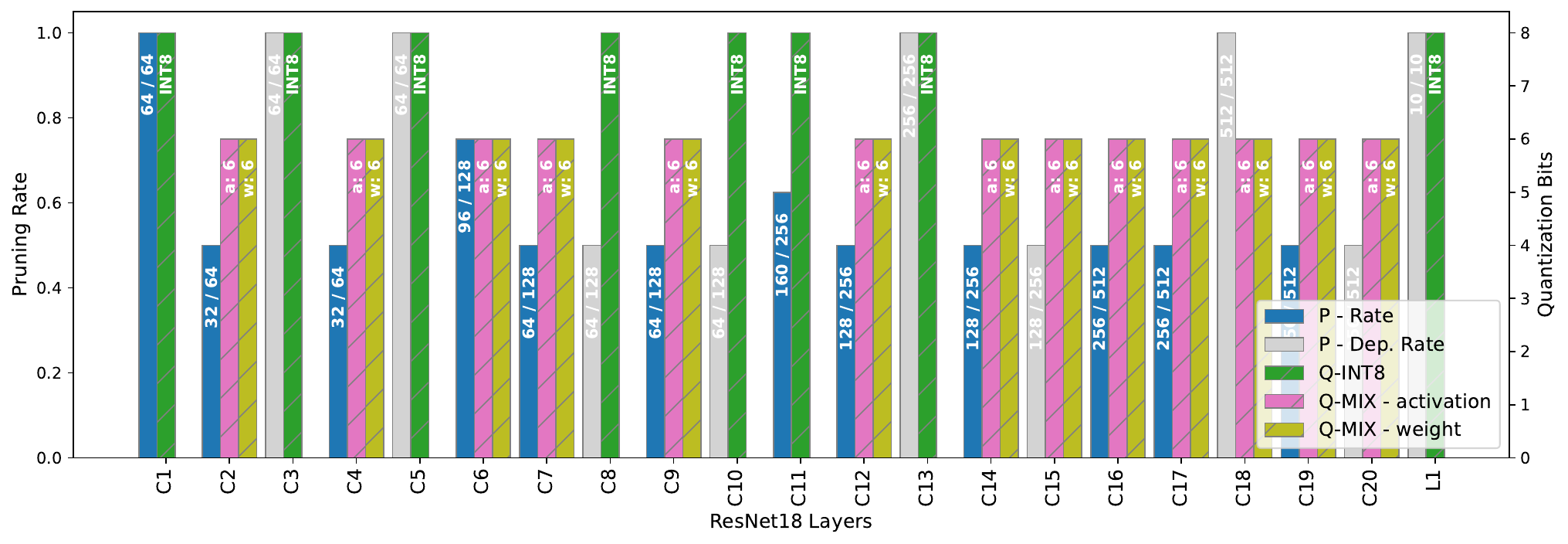}
        \caption{Without sensitivity}
    \end{subfigure}

    \begin{subfigure}{0.7\textwidth}
        \centering
        \includegraphics[width=\textwidth]{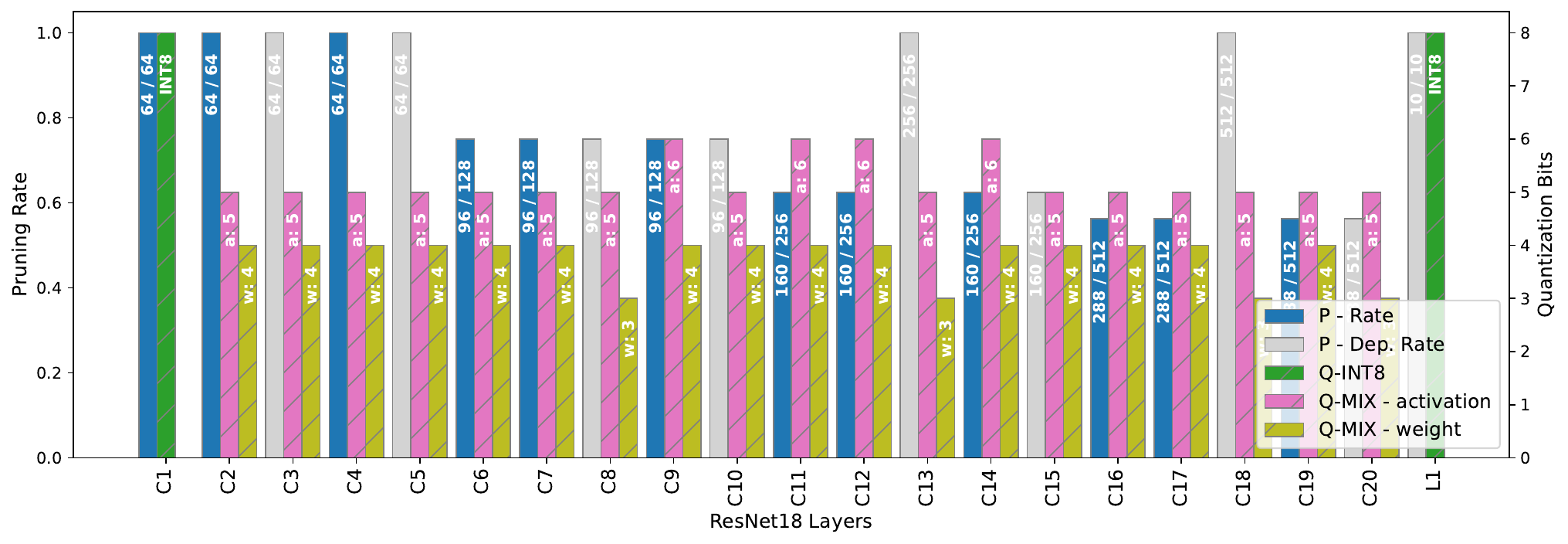}
        \caption{With sensitivity}
    \end{subfigure}
	\caption{Joint policies at $c=0.2$ without (a) and with (b) sensitivity features; sensitivity enables exploitation of per-layer heterogeneity. \reprofrom{krieger2023galen}}
    \label{fig:galen:sens}
\end{figure}

\paragraph{Role of Sensitivity Information}
Finally, we test whether Galen benefits from sensitivity features. Figure~\ref{fig:galen:eval:sens:aw} shows that sensitivity varies across layers: later layers are especially fragile to quantization, while pruning sensitivity exhibits distinct heterogeneity probably relating to the skip connections of residual architectures.
Table~\ref{tab:galen:exp:sens:overview} and Figure~\ref{fig:galen:sens} compare policies with and without sensitivity input. Without sensitivity, the agent produces almost uniform policies with nearly no heterogeneous structure. With sensitivity enabled, policies adapt to layer robustness, compressing tolerant layers more aggressively and protecting fragile ones. This demonstrates that sensitivity information is essential for Galen to exploit per-layer heterogeneity effectively.

\begin{table} %
\caption{Quantitative results of joint search with sensitivity enabled vs.\ disabled at $c=0.2$. \reprofrom{krieger2023galen}}
	\centering
	\small
	\begin{tabular}{l|r|r|r|r}
	\bfseries   & \multicolumn{1}{c|}{\bfseries \acrshortpl{mac}} & \multicolumn{1}{c|}{\bfseries \acrshort{bops}} &  \multicolumn{1}{c|}{\bfseries Rel. Lat.} & \bfseries Accuracy \\ \hline
	Reference 	& $4.75 \cdot 10^{10}$          & $4.86 \cdot 10^{13}$ 		& $100.0$\,\% 	& $93.04$\,\% \\
	Disabled 			& $1.68 \cdot 10^{10}$ 	& $8.10 \cdot 10^{11}$ 		& $20.0$\,\% 	& $92.66$\,\% \\
	Enabled 		    & $2.82 \cdot 10^{10}$  & $6.75 \cdot 10^{11}$ 		& $19.4$\,\% 	& $92.77$\,\% \\
	\end{tabular}
	\label{tab:galen:exp:sens:overview} 
\end{table}

\section*{Summary}

In this chapter, we introduced \emph{Galen}, a framework for automatic, hardware-aware model compression.  
Galen learns per-layer compression policies directly on the deployment device, aligning the optimization objective with real execution behavior and bridging the gap between algorithmic design and hardware realization.  

A central component of the framework is its \emph{hardware-in-the-loop} approach, in which candidate models are compiled and benchmarked on the target system.  
This ensures that optimization is guided by measured latency rather than by analytical proxies, capturing real compiler, memory, and parallelization effects.  
Through \emph{joint compression}, a single agent coordinates structural and precision reductions, yielding consistently favorable accuracy–latency trade-offs.  
Incorporating layer-wise \emph{sensitivity information} further allows the agent to adapt compression strength to architectural heterogeneity, avoiding uniform and suboptimal policies.  
Finally, Galen demonstrates \emph{expert-consistent behavior}: it rediscovers established heuristics such as conservative treatment of early and late layers, stronger pruning in deeper stages, and higher tolerance of weights to quantization.  
This alignment between learned and human-designed strategies highlights both the interpretability and the robustness of the method, underscoring Galen’s role as a step toward automated, trustworthy, and hardware-grounded model compression.

\chapter{Modeling Analog Hardware Accelerators}
\label{ch:analog:hardware_models}

\epigraph{Analog is more beautiful than digital, really,\\but we go for comfort.}{\textnormal{--- Anton Corbijn}}

\noindent
The previous chapter demonstrated how model compression can markedly reduce the computational footprint of neural networks by pruning redundant parameters and lowering numerical precision, while simultaneously accounting for hardware-specific constraints.  
Compression addresses one side of the efficiency spectrum: reducing the model’s size and computational demand to facilitate execution on digital processors.  

An orthogonal route to efficiency is to reduce the cost of computation itself by changing the underlying computing paradigm and performing inference directly in the physical domain, for example using analog or mixed-signal hardware accelerators.  
Rather than compressing the model, this approach makes the \emph{computations} themselves cheaper by exploiting physical quantities such as currents, voltages, or charges to represent and manipulate numerical values.  
Such devices promise orders-of-magnitude gains in energy efficiency and compute density~\cite{Murmann2021,schemmel2021accelerated}, yet they introduce new challenges arising from device mismatch, nonlinearities, saturation effects, and intrinsic noise~\cite{klein2021bss2whitebox,kuhn2023nonassociativity,Rekhi2019,Zhou2020}.  

In the following chapter, we therefore turn from digital compression to analog hardware modelling, building on our publications~\cite{klein2021bss2whitebox,kuhn2023nonassociativity} and recounting how progressively refined models helped us both accelerate training for the BrainScaleS-2 (\acrshort{bss2})~\cite{schemmel2021accelerated} system and uncover unexpected properties such as the non-associativity of analog dot products.

\section{Analog Computing}
Analog computing lacks the ``safety net'' of discretization and is therefore directly exposed to imperfections that can fundamentally alter the outcome of computations.  
Among the most critical are nonlinearities, saturation effects, dynamic phenomena such as leakage or crosstalk between circuit components, and various sources of noise, all of which can interact in complex ways and degrade accuracy~\cite{klein2021bss2whitebox,schemmel2021accelerated}.  
Analog accelerators are further affected by static device mismatch and variability across hardware instances due to manufacturing tolerances, as well as increased sensitivity to environmental conditions such as supply fluctuations and temperature~\cite{Murmann2021,Tsividis2018Spectrum}.  

Na\"ive deployment of digitally trained models onto such accelerators can severely degrade predictive quality, in some cases reducing performance to the level of random guessing and thereby destroying all practical utility \cite{klein2021bss2whitebox}. 

Nevertheless, artificial neural networks exhibit a remarkable tolerance to computational uncertainties when training explicitly accounts for them. 
This is well established in the context of digital quantization and pruning, both of which can be viewed as unsafe optimizations that introduce noise into the computation, yet typically retain accuracy after retraining \cite{Courbariaux2015binaryconnect,Liu2019}. 
Building on this principle, several methods have been proposed to enhance robustness for analog accelerators, for example through knowledge distillation \cite{Zhou2020}, explicit noise-injection during training \cite{Murray1994,qin2018noisycomputations}, or quantization-aware training \cite{Rekhi2019}. 
These results demonstrate that, given appropriate training strategies, neural networks can adapt to significant levels of hardware-induced imprecision. 

The most accurate way to expose a model to the full set of imperfections of a concrete analog chip instance is to include the hardware directly in the training loop. 
Such \acrfull{hil} training lets learning absorb device-specific variations and dynamic effects, and has been shown to be highly effective on the \acrshort{bss2} neuromorphic system \cite{cramer2020training,schemmel2021accelerated}. 

Yet \acrlong{hil} training suffers from practical limitations: dependence on physical device access, retraining overhead for each hardware instance, and limited throughput on some accelerators such as \acrshort{bss2}, which slows down training compared to the scalability of digital simulation on conventional compute clusters. 
This motivates software models of analog accelerators that can be inserted into the training loop~\cite{klein2021bss2whitebox}.

In this chapter we recount our path on \acrshort{bss2}: we first built a white-box model that encodes measured nonidealities via simple, parallelizable components and demonstrated substantial training speedups and accuracy gains over simplistic baselines \cite{klein2021bss2whitebox}. 
Persisting discrepancies led us to suspect temporal and ordering effects. 
We therefore tried an intentionally over-dimensioned transformer-set model to test whether sequence dependence mattered. 
That experiment revealed an unexpected and fundamental property: effective non-associativity of analog dot products on \acrshort{bss2}~\cite{kuhn2023nonassociativity}. 
Both models extract the hardware behavior from measured data, using these measurements to characterize and encode the specific inaccuracies of the concrete device instance \cite{klein2021bss2whitebox,kuhn2023nonassociativity}. 
Taken together, these models illustrate a central principle of our work: the value of hardware modelling lies not only in reducing the cost of training, but also in revealing which device characteristics fundamentally shape computation.

\subsection{Related Work}

The robustness of neural networks to imprecise computation has long been investigated in the context of quantization and pruning~\cite{Courbariaux2015binaryconnect,Liu2019}.  
Noise injection and knowledge distillation have been proposed to further enhance robustness against analog-style perturbations~\cite{Zhou2020,Rekhi2019,Murray1994,Torres2017}.  
For analog accelerators in particular, early modeling approaches often approximate imperfections as additive Gaussian noise or uniform quantization.  
While such abstractions facilitate \acrlong{hil} training, they overlook device heterogeneity and structured nonlinear effects.  

In summary, prior work has demonstrated the promise of analog accelerators \cite{Murmann2021,schemmel2021accelerated,Shen2017,Joshi2020,Lin2018} and has explored robustness through quantization, noise injection, and distillation \cite{Courbariaux2015binaryconnect,Zhou2020,Rekhi2019,Murray1994,Torres2017}.
Yet most models remain either too simplistic---ignoring heterogeneity and dynamic nonlinearities---or too domain-general to capture device-specific effects.
Our contribution is twofold: (i) a compact white-box model that leverages device-characterized data to encode static variations and stochastic noise, thereby bridging much of the gap between digital pretraining and hardware-in-the-loop retraining \cite{klein2021bss2whitebox}; and (ii) a transformer-set model that revealed effective non-associativity of dot products in analog accelerators \cite{kuhn2023nonassociativity}.
Together, these models highlight a guiding principle: software models of analog hardware are most valuable when they not only replicate hardware behavior but also \emph{reveal} which imperfections are consequential for training and deployment.  

\section{\acrlong{bss2}}

\begin{figure}
    \centering
	\resizebox{0.5\textwidth}{!}{%
\definecolor{blue}{HTML}{1f77b4}%
\definecolor{red}{HTML}{d62728}%
\definecolor{green}{HTML}{2ca02c}%
\definecolor{orange}{HTML}{ff710e}%
\definecolor{yellow}{HTML}{fee23e}%
\definecolor{cadc}{HTML}{1f77b4}%
\definecolor{input}{HTML}{ff7f0e}%
\definecolor{hidden}{HTML}{2ca02c}%
\definecolor{output}{HTML}{555555}%
\tikzset{block/.style={font={\rmfamily\footnotesize},align=center}}%
\tikzset{box/.style={draw=black!90}}%
\tikzset{block label/.style={fill=white,font={\rmfamily\footnotesize},inner sep=0.05cm}}%
\tikzset{%
	neuron/.style = {%
		draw=black,%
		circle,%
		inner sep=0pt,%
		minimum width=0.4cm%
	},%
	driver/.style = {%
		minimum height=0.45cm,%
		draw=black,%
		regular polygon,%
		regular polygon sides=3,%
		shape border rotate=-90,%
		inner sep=0pt%
	},%
}%
\begin{tikzpicture}[
	scale=0.8,
	>=stealth,
	transform shape,
	line width=1.0\pgflinewidth,
	anchor=center,
	spy using outlines=circle,
]
	\pgfdeclarelayer{background layer}
	\pgfsetlayers{background layer,main}
	\draw[use as bounding box,inner sep=0pt,draw=none] (0.0,0.0) rectangle ++(7.0,4.5);

	\begin{scope}[scale=1.15]
		\foreach \x in {0,1,...,6} {

			\node[neuron,output,thick] (nrn \x) at (0.8 + \x*0.5,0.35) {};
			\draw (nrn \x.north) ++ (0.0,0.01) -- ++(0.0,2.5);

		}

		\foreach \y in {0,1,...,4} {
			\node[driver,thick] (drv \y) at ($(nrn 0) + (-0.5,0.5 + \y*0.5)$) {};
			\draw (drv \y.+45) -- ($(drv \y.center) + (0.15, 0.125)$) -- ++(3.5,0.0);
			\draw (drv \y.-45) -- ($(drv \y.center) + (0.15,-0.125)$) -- ++(3.5,0.0);

			\foreach \x in {0,1,...,6} {

				\fill[hidden] (drv \y -| nrn \x) ++ (0.0, 0.125) circle (0.05cm);
				\draw[white] (drv \y -| nrn \x) ++ (0.0, 0.125) ++ (-0.045,0.0) -- ++(0.09,0.0);
				\draw[white] (drv \y -| nrn \x) ++ (0.0, 0.125) ++ (0.0,-0.045) -- ++(0.0,0.09);
				\fill[hidden] (drv \y -| nrn \x) ++ (0.0,-0.125) circle (0.05cm);
				\draw[white] (drv \y -| nrn \x) ++ (0.0,-0.125) ++ (-0.045,0.0) -- ++(0.09,0.0);
			}
		}

		\node[rectangle,thick,draw,cadc,inner sep=2pt,minimum width=3.5cm] (cadc) at ($(nrn 2) + (0.5,3.0)$) {\fontsize{6}{6}\selectfont Columnar ADC};
		\foreach \x in {0,1,...,6} {
			\draw[blue] (nrn \x.55) -- ++(55:0.12) coordinate (tmp) -- (tmp |- cadc.south);
		}

		\node[rectangle,thick,draw,inner sep=2pt,minimum width=3.5cm,above=0.07cm of cadc] (ppu) {\fontsize{6}{6}\selectfont SIMD Processor};

	\end{scope}

	\coordinate (sherlock) at (5.6,2.2);
	\spy[draw,height=1.2cm,width=1.2cm,magnification=2.5,connect spies] on (nrn 6 |- drv 0) in node at (sherlock);
	\node[align=center] at ($(sherlock) - (0.0,1.1)$) {\fontsize{7}{7}\selectfont signed\\[-0.3em] \fontsize{7}{7}\selectfont synapse};
\end{tikzpicture}
}
	\caption*{(a) Block diagram of the \acrshort{bss2} analog core, showing synapse drivers (triangles), synapses (small circles in the matrix), and neurons (large circles at the column ends). Reproduced with permission from \cite{cramer2020training}.}

    \vspace{1em}
    
    \includegraphics[width=0.99\textwidth]{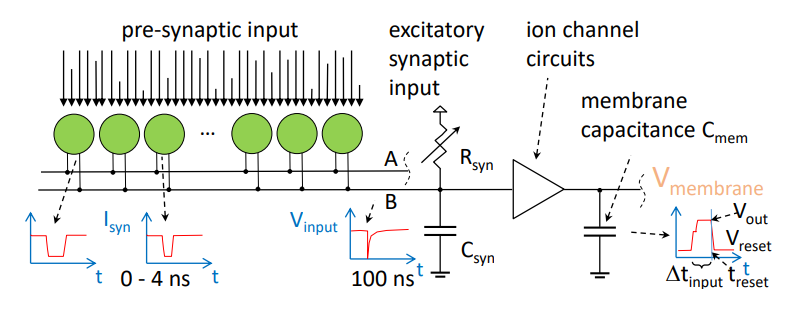}
	\caption*{(b) Detailed view of a single \acrshort{bss2} column in non-spiking mode, illustrating how synaptic pulses are integrated onto the neuron membrane. Reproduced with permission from \cite{schemmel2021accelerated}.}
        
    \caption{Architecture of the BSS-2 analog neuromorphic system.  
    Panel (a) shows the overall crossbar structure of the analog core, while panel (b) highlights the electrical details of a single column.}
    \label{fig:bss2-architecture}
\end{figure}

\acrshort{bss2} is a mixed-signal neuromorphic \acrshort{soc} manufactured in 65\,nm CMOS technology, designed to support both the accelerated emulation of spiking neural networks and the efficient execution of analog matrix--vector products within its analog core \cite{schemmel2021accelerated,cramer2020training}. 

As illustrated in Figure~\ref{fig:bss2-architecture}(a), the core implements a crossbar array comprising 512 columns—corresponding to neuron circuits—and 128 rows that represent the input activations.  
To support signed weights, each logical column is realized by a pair of physical columns (256 virtual neurons in total), and each column is equipped with a dedicated per-column \acrfull{adc} for digitized readout.  

During a matrix--vector multiplication, each input activation $a_i$ is encoded as a pulse length $\Delta t_i$ (5\,bit precision), while the synaptic weight $w_{i,j}$ is implemented as a current amplitude $I_{i,j}$. 
Each synapse thereby generates a current pulse with charge $Q_{i,j} = I_{i,j}\Delta t_i$, which is accumulated on the column capacitor of its target neuron via \acrfull{ota} circuits. 
The resulting membrane potential is then digitized by the column’s 8\,bit \acrshort{adc}. 
This integration of current pulses is illustrated in Figure~\ref{fig:bss2-architecture}(b).

This architecture, together with its characteristic nonidealities, makes \acrshort{bss2} a representative example for studying both the benefits and the limitations of analog matrix--multiply accelerators. 

\subsection*{Experimental Setup}
\label{sec:experimental-setup}

\paragraph{Dataset.}
All experiments in this chapter are based on the \acrfull{gsc} dataset~\cite{warden2018speechcommands}, a widely adopted benchmark for keyword spotting. 
It consists of one-second audio snippets of spoken keywords drawn from a vocabulary of 35 classes. 
We followed the standard split into training, validation, and test sets as provided in the dataset release. 

\paragraph{Preprocessing.}
Input waveforms were converted into log-Mel spectrogram features. 
This representation is well suited for keyword spotting and, importantly, produces non-negative values only. 
The latter property aligns with the constraints of the \acrshort{bss2} hardware, which supports only positive activations.

\begin{figure}[t]
  \centering
  \begin{tikzpicture}[
    node distance=5mm and 12mm,
    every node/.style={font=\sffamily},
    box/.style={draw, rounded corners=2pt, minimum width=42mm, minimum height=8mm, align=center},
    det/.style={box, fill=gray!15},        %
    act/.style={box, fill=blue!15},        %
    arrow/.style={-{Stealth[length=2.6mm]}, thick},
    dimlabel/.style={pos=0.55, above right=-6pt and -1pt, font=\scriptsize\sffamily}
  ]

  \node[det] (in) {Input};
  \node[act, below=of in] (fc1) {Dense (1024)};
  \node[det, below=of fc1] (relu1) {ReLU};
  \node[act, below=of relu1] (fc2) {Dense (128)};
  \node[det, below=of fc2] (relu2) {ReLU};
  \node[act, below=of relu2] (fc3) {Dense (12)};
  \node[det, below=of fc3] (out) {Output (12)};

  \draw[arrow] (in)  -- (fc1)  node[dimlabel] {$D_{\text{in}}$};
  \draw[arrow] (fc1) -- (relu1) node[dimlabel] {$1024$};
  \draw[arrow] (relu1) -- (fc2) node[dimlabel] {$1024$};
  \draw[arrow] (fc2) -- (relu2) node[dimlabel] {$128$};
  \draw[arrow] (relu2) -- (fc3) node[dimlabel] {$128$};
  \draw[arrow] (fc3) -- (out)   node[dimlabel] {$12$};

  \end{tikzpicture}

  \caption{Neural network architecture used for keyword spotting on the Google Speech Commands~\cite{warden2018speechcommands} dataset. 
  A 3-layer MLP with hidden sizes 1024 and 128 and a 12-way output; activation dimensions are annotated on the arrows. 
  The first (wide) layer is mapped to hardware in multiple sequential calls, while subsequent layers map directly. 
  Reproduced with permission from~\cite{klein2021bss2whitebox}.}
  \label{fig:network-architecture}
\end{figure}
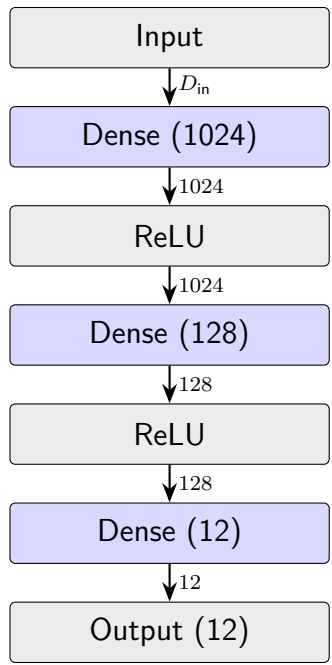

\paragraph{Model architecture.}
The same network architecture was used for all experiments with both the white-box and transformer-set models (Fig.~\ref{fig:network-architecture}). 
It is designed to balance task performance with hardware interpretability: large enough to solve the keyword spotting task, yet compact enough to make hardware effects transparent. 
The first layer is considerably wider than subsequent layers and therefore cannot be mapped onto the hardware in a single call; instead it is split across multiple sequential calls. 
All following layers fit directly on the hardware without partitioning. 
This design respects hardware constraints while keeping the analysis of imperfections tractable.

\section{White-Box Model of \acrshort{bss2}}
\label{sec:whitebox}

Hardware-in-the-loop retraining is the most accurate way to adapt neural networks to analog accelerators, since it directly exposes the model to the device’s imperfections. 
However, it is time-consuming and scales poorly when device throughput and access are limited. 
As an alternative, we investigated whether a compact and interpretable software model of hardware nonidealities can replace most \acrlong{hil} training epochs without compromising accuracy.  
In the following, we present the design of this white-box model and summarize its key findings; a more detailed treatment can be found in~\cite{klein2021bss2whitebox}.

\subsection{Model Description}
\begin{figure}%
	\centering
	\includegraphics[width=0.4\textwidth]{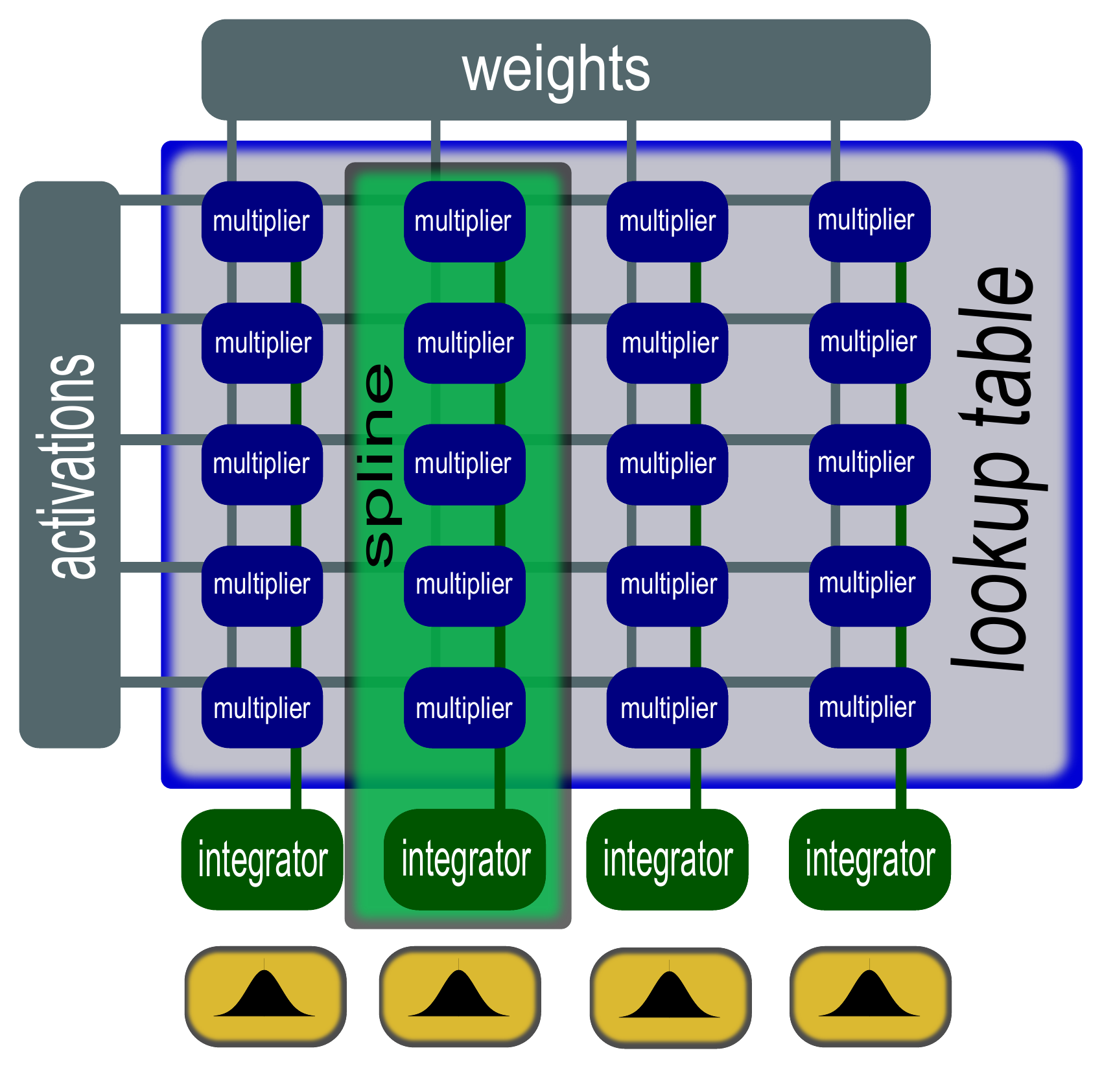}
	\caption{Schematic overview of the white-box representation: \emph{lookup table} for the static variations of the multiplier array,
	\emph{splines} for the column and integrator specific variations, and additive \emph{Gaussian noise} that models the electrical noise, among others. Reproduced with permission from~\cite{klein2021bss2whitebox}.}
	\label{fig:whitebox:schema}
\end{figure}

The white-box model explicitly encodes the main sources of imperfections observed on \acrshort{bss2}.
Specifically, it combines lookup tables for per-synapse nonlinearities, spline functions that capture column-specific saturation effects of the membrane integrators, and Gaussian noise terms modeling stochastic fluctuations. 
Figure~\ref{fig:whitebox:schema} illustrates this schematically, highlighting their components and their relation to the underlying hardware components.
Each component was parameterized from direct hardware measurements.
Together, these components provide a structured yet lightweight representation of the device’s non-ideal behavior.

\paragraph{\Acrlong{lut} for synaptic multipliers.}
To capture static local nonidealities at the level of individual synapses, the model uses \acrfullpl{lut} that encode their nonlinear characteristics. 
Each hardware synapse is intended to realize a multiplication between an activation $a_i$ (encoded as a pulse length $\Delta t_i$) and a weight $w_{i,j}$ (encoded as a current amplitude $I_{i,j}$). 
In practice, unavoidable manufacturing variations give rise to device mismatch, which in turn causes deviations from the ideal behavior, resulting in nonlinear and synapse-specific responses.
We characterized these effects by measuring row-wise transfer functions, activating one synapse at a time with controlled inputs so that column saturation did not interfere with the measurement. 
The resulting input–output curves were stored as \acrlongpl{lut} $e_{\text{lookup}}[r, c, a, w]$, which constitute the first stage of the white-box model and provide a calibrated mapping from operand values to synaptic contributions.

\paragraph{Column-wise splines for saturation effects.}
When multiple synapses in a column are active, their contributions interact nonlinearly due to line saturation and \acrshort{ota} limitations.
To model this, the summed \acrshort{lut} outputs of a column
are passed through a column-specific spline,
\begin{equation}
	\tilde{y}_c = f_c \left( \sum_{r} e_{\text{lookup}}[r, c, a_{rc}, w_{rc}] \right).
\end{equation}
These splines were fitted to measurements taken under full-range excitations (activating many synapses at once), capturing the column’s characteristic saturation curve.
The spline stage thus models column-specific interaction effects beyond isolated multipliers.

\begin{figure}[t]
    \centering
    \includegraphics[width=0.7\textwidth]{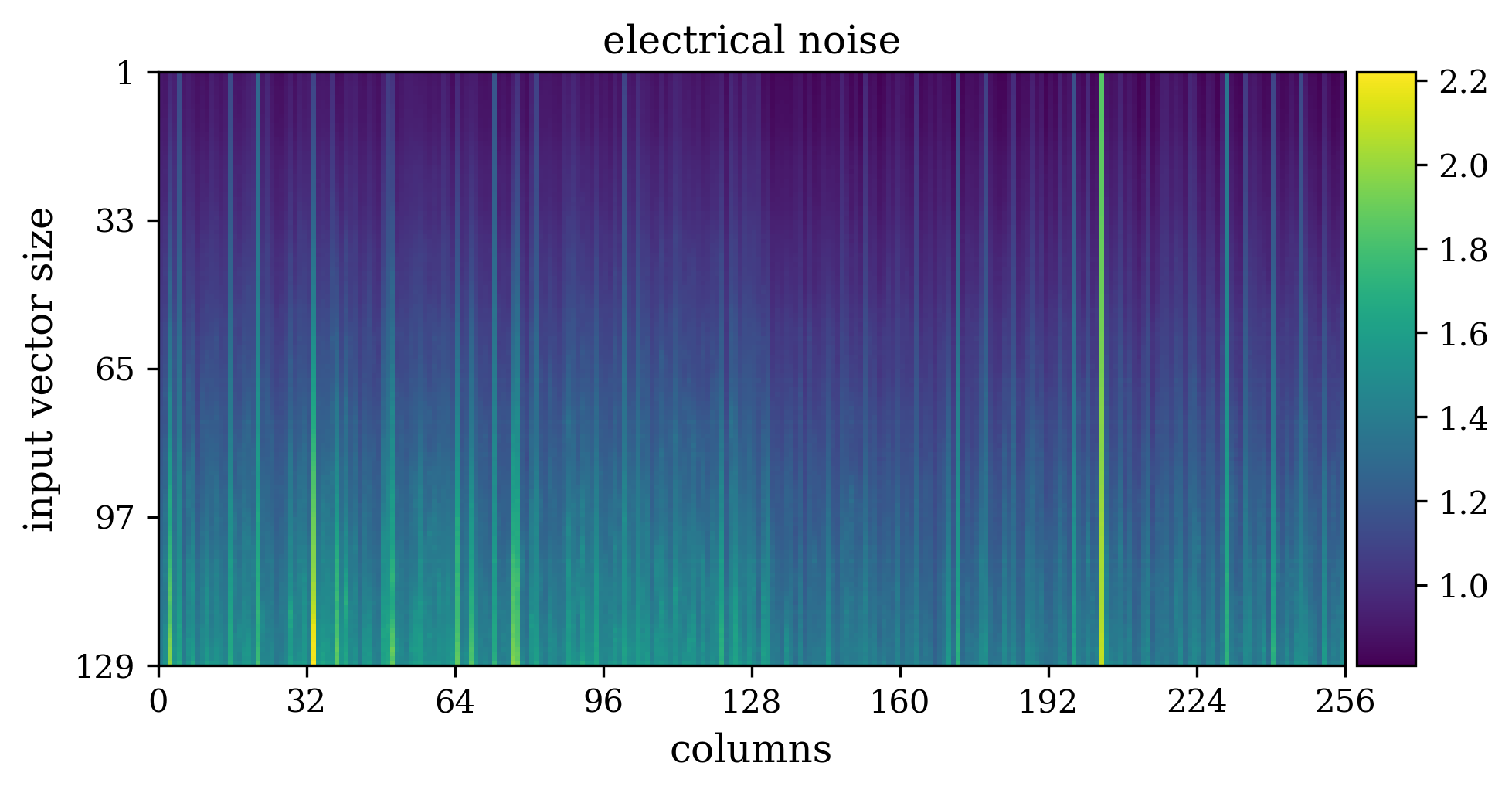}
    \caption{Electrical noise measured as the standard deviation of repeated calculations with identical inputs. 
    With larger input vectors the average noise increases (vertical gradient), since more noisy values are accumulated. 
    At the same time, variances between columns are significant, as shown by the vertical spread. 
    Reproduced with permission from~\cite{klein2021bss2whitebox}.}
    \label{fig:whitebox:noise_input_vector_size_column}
\end{figure}

\paragraph{Gaussian noise with per-neuron variance.}
Electrical noise further perturbs outputs and is not constant across neurons and varies with the input vector lengths.
We estimated noise variances by repeatedly measuring identical inputs and recording output variability.
This yielded a variance profile $\sigma^2(\text{col}, N)$ depending on the column index and the number of active inputs $N$, as illustrated in Figure~\ref{fig:whitebox:noise_input_vector_size_column}.
At runtime, we add this calibrated zero-mean Gaussian noise to each neuron output, reproducing the measured stochastic behavior,
\begin{equation}
	y_c = \tilde{y}_c + \mathcal{N}(0,\sigma^2_c).
\end{equation}

\paragraph{Incremental noise training.}
A further improvement was achieved through incremental noise training. 
Here, the Gaussian noise level was gradually increased from 0\% to 100\% of the measured variance over the course of training, following a curriculum strategy.  
This gradual exposure improved robustness by allowing the network to adapt progressively to hardware imperfections, and its effect is clearly visible in Table~\ref{tab:whitebox:pretraining}. 

\subsection{White-box Model Results}

\begin{figure}%
	\centering
	\includegraphics[width=\textwidth]{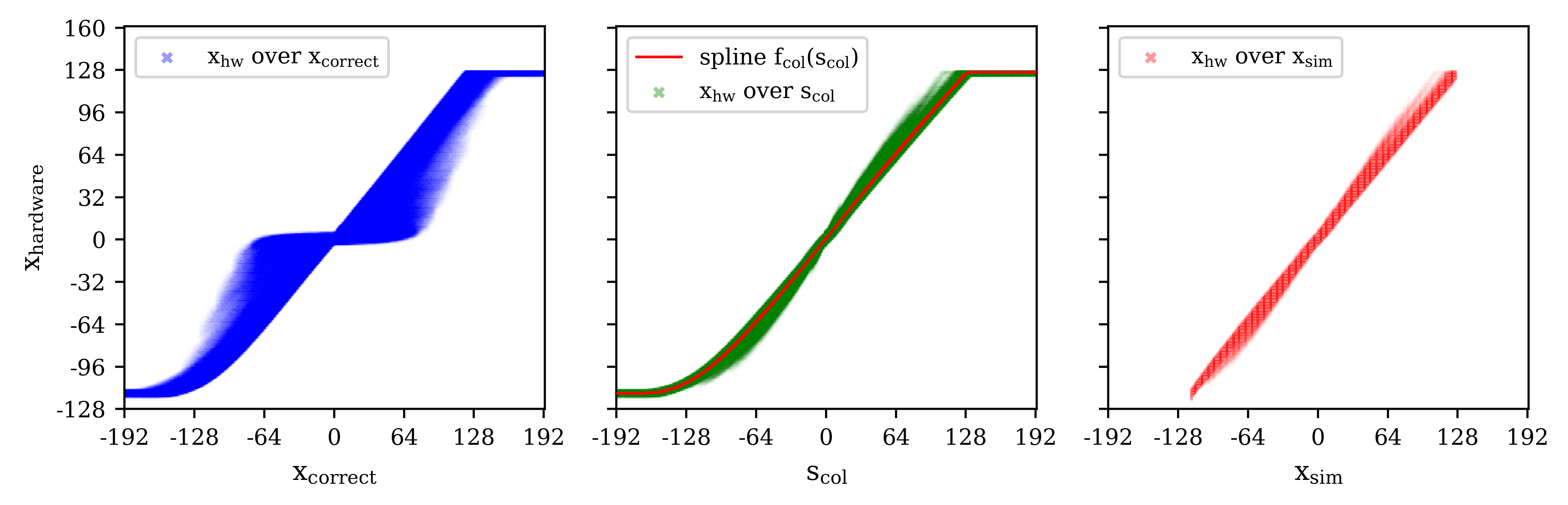}
	\caption{
	Mismatch between the ideal matrix multiplication result $x_{\text{correct}}$ and the measured hardware output $x_{\text{hardware}}$ for a random column (left). 
	A large fraction of this error can be compensated by applying the lookup table sums $s_{\text{col}}$ (middle) together with a spline interpolation that captures column-specific saturation effects (right). 
	An ideal representation would follow the identity relation, broadened only by electrical noise. 
	Reproduced with permission from~\cite{klein2021bss2whitebox}.
	}
	\label{fig:whitebox:spline_example}
\end{figure}

Figure~\ref{fig:whitebox:spline_example} illustrates the progressive effect of the white-box corrections.  
On the left, raw hardware results deviate strongly from the ground-truth matrix multiplication. 
Applying the \acrlong{lut} corrections (middle) compensates for much of this error, accounting for synapse-specific nonlinearities. 
Adding the spline corrections (right), which represent column-level saturation effects, brings the response close to the ideal identity relation. 
The resulting curve is very close to the ideal identity relation; the remaining deviations are attributable to residual effects not captured by the model, and ultimately to stochastic noise. 

\begin{figure}
    \centering
    \includegraphics[width=0.7\textwidth]{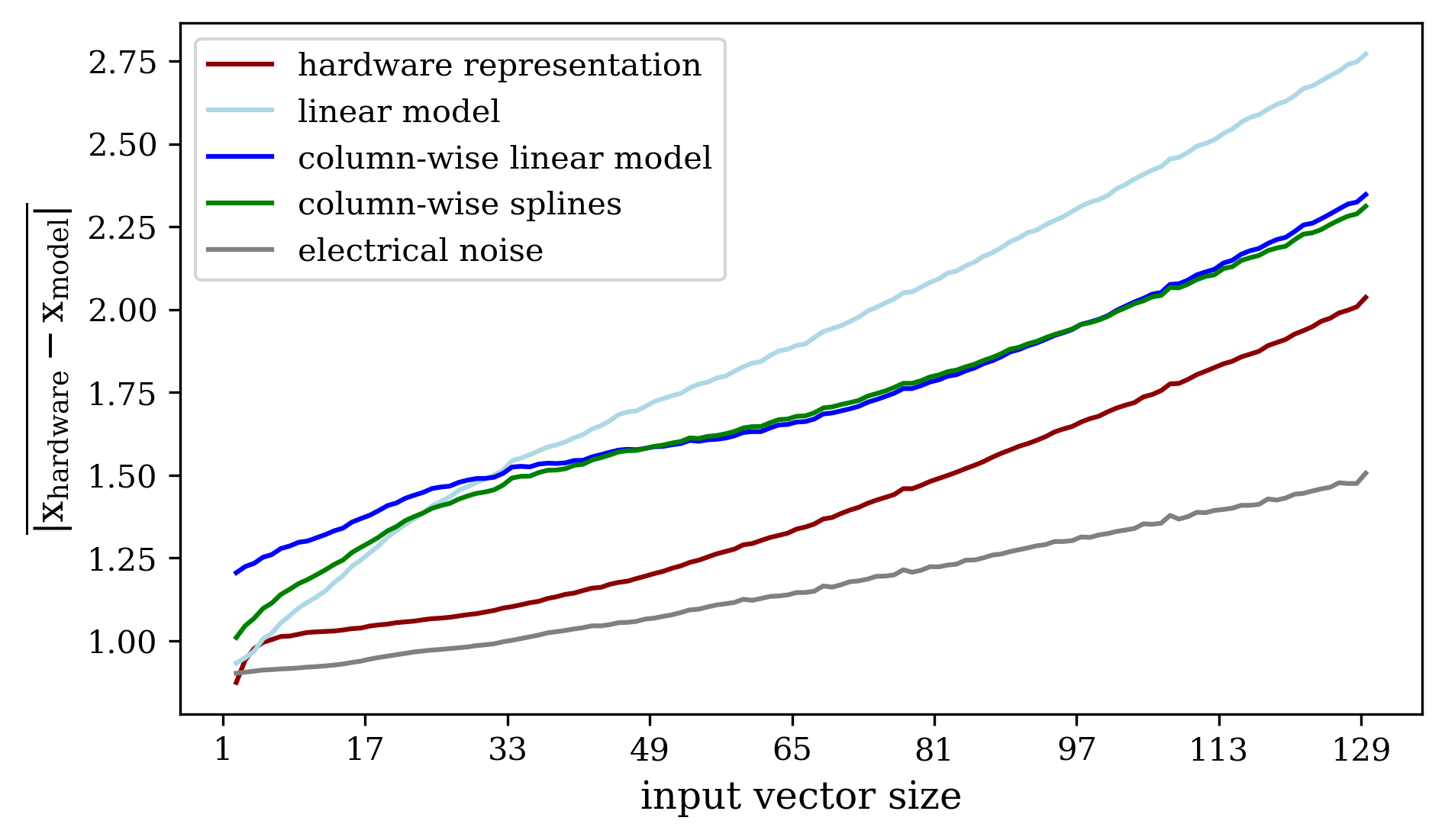}
    \caption{Model performance measured as the deviation from hardware results on random inputs. 
    An ideal model would align with the electrical noise (gray) observed on hardware. 
    The proposed hardware representation outperforms linear-regression and spline-based models. 
    Both noise and model imperfections increase with input vector size due to more involved components and saturation effects. 
    Reproduced with permission from~\cite{klein2021bss2whitebox}.}
    \label{fig:whitebox:models_over_input_size}
\end{figure}

To quantify how closely different representations approximate hardware behavior, we compared the error of several modeling approaches across input sizes (Fig.~\ref{fig:whitebox:models_over_input_size}). 
A simple linear regression model performs worst, as it cannot capture either local nonlinearities or saturation. 
Columnar approximations improve on this by accounting for average saturation behavior, but remain limited. 
Our proposed white-box model consistently performs best, closely tracking the hardware outputs across input sizes. 
Nonetheless, a residual gap to the hardware remains, indicating that not all effects are captured by the white-box model. 
At this stage we could only speculate that additional phenomena—such as temporal or ordering effects—might play a role. 
This hypothesis motivated the development of a more expressive transformer-set model (Sec.~\ref{sec:transformer-model}). 

As summarized in Table~\ref{tab:whitebox:pretraining}, these corrections translate into substantial accuracy gains in end-to-end training. 
The white-box model consistently outperformed the quantization baseline and narrowed the gap to full \acrlong{hil} training. 
When combined with a small number of final \acrlong{hil} epochs, it even surpassed prolonged pure-\acrlong{hil} training while reducing retraining time from more than 10\,h to under 2\,h. 
This demonstrates that device-characterized models can both accelerate training and improve accuracy compared to simplistic approximations.

\begin{table}
\centering
\scriptsize
\renewcommand{\arraystretch}{1.0}
\caption{Comparison of retraining strategies. \reprofrom{klein2021bss2whitebox}}
\label{tab:whitebox:pretraining}
\resizebox{\textwidth}{!}{%
\begin{tabular}{l|l|c|c|c|c}
\multicolumn{2}{c|}{\bfseries Method} & \multicolumn{2}{c|}{\bfseries Retraining} & \multicolumn{2}{c}{\bfseries Accuracy} \\ \hline
\multirow{2}{*}{\bfseries Name} & \multirow{2}{*}{\bfseries Description} & \multirow{2}{*}{\bfseries Epochs} & \bfseries Time & \multirow{2}{*}{\bfseries SW} & \multirow{2}{*}{\bfseries BSS-2} \\
 & & & \bfseries [min] & & \\ \hline
Plain & Full-precision baseline & 0 & 0.0 & 80.8\,\% & 12.3\,\% \\
Quantized & INT6 weights, UINT5 activations & 300 & 13.1 & 79.6\,\% & 25.8\,\% \\
Noise only & Baseline with Gaussian noise & 300 & 10.4 & 80.5\,\% & 18.4\,\% \\
Rep. no noise & Static variances, no noise & 300 & 83.4 & 76.5\,\% & 26.4\,\% \\
Rep. with noise & Hardware representation with noise & 300 & 83.4 & 73.5\,\% & 35.1\,\% \\
Rep. incr. noise & Hardware representation, incr. noise & 300 & 83.6 & 76.0\,\% & 41.0\,\% \\ \hline
HIL (300) & Full hardware-in-the-loop & 300 & 652.3 &  & 66.8\,\% \\
HIL (350) & Full hardware-in-the-loop & 350 & 769.5 &  & 66.7\,\% \\ 
Comb. quant. (5) & Quantization (300) + HIL (5) & 305 & $13.1 + 11.5$ &  & 62.1\,\% \\
Comb. quant. (50) & Quantization (300) + HIL (50) & 350 & $13.1 + 117.5$ &  & 67.3\,\% \\
Comb. rep. (1) & Representation (300) + HIL (1) & 301 & $83.6 + 2.2$ &  & 64.9\,\% \\
Comb. rep. (5) & Representation (300) + HIL (5) & 305 & $83.6 + 11.5$ &  & 67.4\,\% \\
Comb. rep. (10) & Representation (300) + HIL (10) & 310 & $83.6 + 23.5$ &  & 69.7\,\% \\
Comb. rep. (50) & Representation (300) + HIL (50) & 350 & $83.6 + 117.5$ &  & \textbf{70.1\,\%} \\
\end{tabular}%
}
\end{table}

The white-box model highlights that explicitly encoding static nonlinearities and noise structure is crucial for analog accelerator training.
Its main advantages are efficiency, parallelizability, and interpretability: each component maps directly to a physical imperfection.
However, residual discrepancies remained, especially under temporally clustered high inputs.
Because the LUT + spline + noise model is fundamentally static, it cannot capture ordering- or time-dependent effects.
This limitation motivated our next step: testing a more expressive transformer-set model, which ultimately revealed the phenomenon of effective non-associativity.

\section{Transformer-Set Model and Non-Associativity}
\label{sec:transformer-model}

\subsection{Hypothesis: Could Ordering Matter?}
Despite the progress of the white-box model, a residual gap to the hardware remained (Sec.~\ref{sec:whitebox}), which could not be explained by static nonlinearities, saturation, or stochastic noise alone. 
At this stage we speculated that the order in which synaptic pulses arrive might influence the computation. 
If true, this would violate the assumption of associativity: mathematically, addition is associative and commutative, $(x+y)+z=x+(y+z)$ and $x+y=y+x$, and therefore the dot product $\sum_i a_i w_i$ should be order-independent. 

However, we hypothesized that analog dynamics such as line saturation or leakage could lead to order dependence. 
To test this hypothesis, we designed a deliberately expressive model with sufficient capacity to learn ordering effects if they exist.

\subsection{Testing the Hypothesis with a Transformer Set}

\begin{figure}
    \centering
    \includegraphics[width=0.35\textwidth]{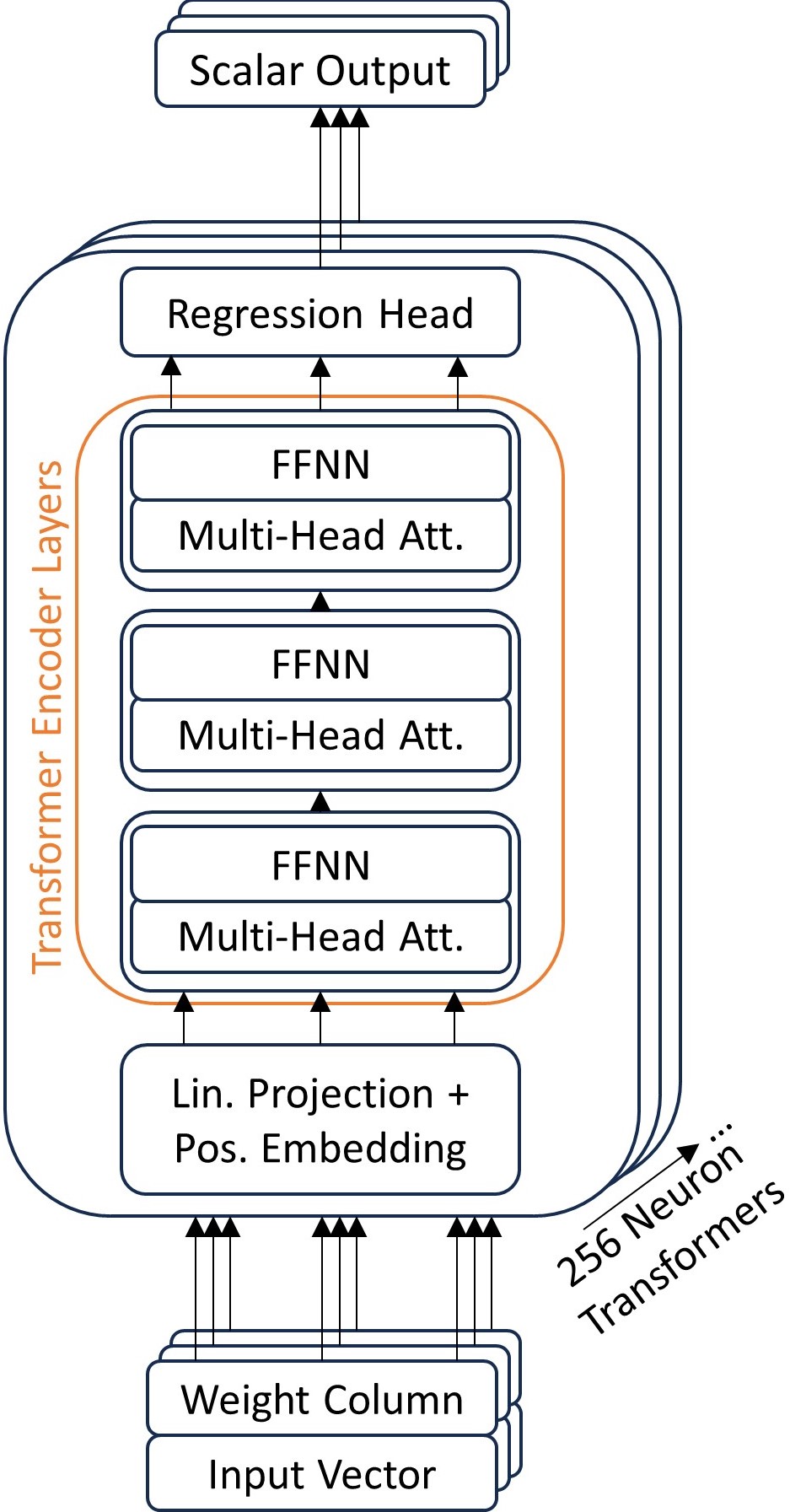}
    \caption{Transformer-set model: a deliberately expressive black-box model to probe for ordering effects. Reproduced with permission from~\cite{kuhn2023nonassociativity}.}
    \label{fig:non-associativity:transformer}
\end{figure}

We constructed a set of Transformers~\cite{Vaswani2017Attention}, one independent Transformer per column (256 in total), trained to map input sequences $(a_i, w_i)$ to the corresponding analog output (Fig.~\ref{fig:non-associativity:transformer}). 
Transformers were chosen because of their ability to capture sequence dependencies and thus provide an upper bound on what could be learned from ordering information. 
To isolate the role of order, we prepared two datasets: 
\begin{itemize}
    \item \textbf{Ordered:} operands presented in the true sequence as processed by the hardware. 
    \item \textbf{Non-ordered:} operands permuted randomly but consistently across all rows, thereby preserving the multiset of inputs but destroying ordering information.
\end{itemize}
Both datasets were based on the same hardware-characterized measurements and included per-column Gaussian noise consistent with Sec.~\ref{sec:whitebox}. 

\subsection{Findings: The Emergence of Non-Associativity}

\begin{figure}
    \centering
    \includegraphics[width=0.6\textwidth]{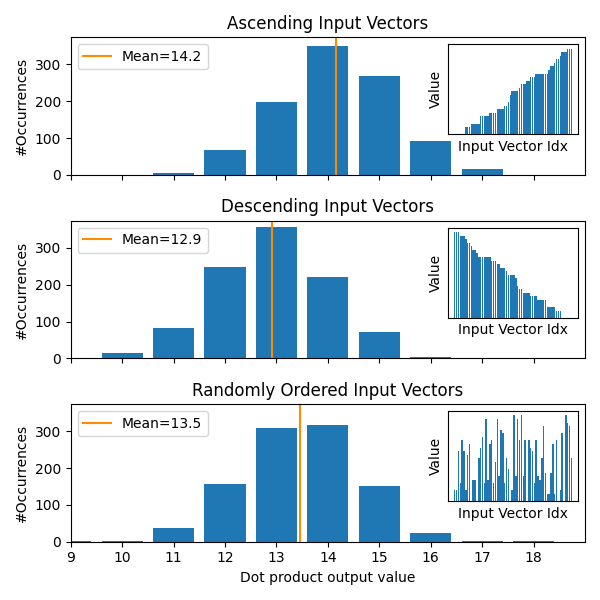}
    \caption{Example of non-associativity: identical operands produce different results depending on their order of arrival. Reproduced with permission from~\cite{kuhn2023nonassociativity}.}
    \label{fig:non-associativity:example}
\end{figure}

\begin{table}\renewcommand{\arraystretch}{1.0}
	\centering
	\small
	\caption{Comparison of hardware models. 
	MSE to BSS-2 measurements and Google Speech Commands test accuracy when retrained with each model and deployed on hardware. 
	The ordered transformer set performs clearly better, showing that operand order is a decisive factor. 
	Reproduced with permission from~\cite{kuhn2023nonassociativity}.}
	\begin{tabular}{c|  >{\centering}p{15mm}|| >{\centering}p{15mm} |  >{\centering\arraybackslash}p{15mm}}
	\multirow{2}*{\bfseries Method} & \bfseries MSE  & \multicolumn{2}{c}{\bfseries GSC Test Acc.}	\\ \cline{3-4}
	 &  \bfseries to BSS-2 & \bfseries SW & \bfseries BSS-2 \\ \hline 
	Ordered Transformer Set & 0.5 & 76.6\% & 53.4\% \\
	Non-ordered Transformer Set & 0.9 & 77.8\% & 44.5\% \\
	White-box Model & -  & 76.9\% & 41.2\% \\
	\end{tabular}
	\label{tab:results}
\end{table}

Figure~\ref{fig:non-associativity:example} illustrates an example of non-associativity: identical operands yield different results depending on their input order. 
Table~\ref{tab:results} quantifies this effect: \acrlong{gsc} models trained with the ordered transformer-set representation fit the hardware significantly better (lower MSE) and achieved higher test accuracy after deployment on \acrshort{bss2} than models trained with either the non-ordered transformer set or the white-box model. 
The improvement, amounting to gains of up to 9--12\% on keyword spotting, demonstrates that ordering is not a nuisance factor but a first-order effect of analog computation. 
Nonetheless, a smaller residual gap to pure \acrlong{hil} retraining remained, indicating that additional time-dependent effects beyond operand ordering are also present. 

Finally, we compared how the models behave when combined with a subsequent \acrlong{hil} retraining phase (Table~\ref{tab:hwLoopResults}). 
Here, the transformer-set and the white-box model reached comparable accuracy, both reducing the required number of \acrlong{hil} epochs by an order of magnitude compared to training without any hardware model. 
This highlights the complementary role of such models: they accelerate training and reveal which device effects matter, even if ultimate performance still requires direct hardware adaptation. 

\begin{table}\renewcommand{\arraystretch}{1.0}
	\centering
	\small
	\caption{Google Speech Commands test accuracy after hardware-in-the-loop retraining. 
	Using either the transformer set or the white-box model reduces the number of required \acrshort{hil} epochs dramatically compared to training without any model. 
	Reproduced with permission from~\cite{kuhn2023nonassociativity}.}
	\begin{tabular}{c|c|c|c}
	\multirow{2}*{\bfseries Method} &\bfseries HW Model &\bfseries BSS-2& \bfseries Accuracy on \\
	 & \bfseries Epochs & \bfseries Epochs & \bfseries BSS-2  \\ \hline 
	Without HW Model & - & 350 & 69.6\% \\
	Ordered Transformer Set & 300 & 50 &71.8\% \\
	Non-ordered Transformer Set & 300 & 50 &71.6\% \\ 
	White-box Model & 300 & 50 &71.9\% \\ 
	\end{tabular}
\label{tab:hwLoopResults}
\end{table}

\section*{Summary}
The two hardware models were developed for distinct purposes, yet together they have shaped how we train and reason about analog accelerators within this work.  
The white-box model was designed for interpretability and efficiency: it encodes static heterogeneity and stochastic noise in a compact form, runs efficiently on \acrshortpl{gpu}, and substantially reduces the need for \acrlong{hil} retraining. 
The transformer-set model, by contrast, was deliberately over-expressive and served as a diagnostic tool. 
It revealed the surprising phenomenon of non-associativity due to ordering-dependent saturation---a property invisible to static models but essential for understanding analog computation. 

From these insights a pragmatic recipe emerges: start with a conventionally trained digital model, apply quantization-aware training to obtain a suitable quantized baseline~\cite{Courbariaux2015binaryconnect,Liu2019}, continue with white-box training (including incremental noise curricula) to capture device-specific variations, and conclude with a short \acrlong{hil} fine-tuning. 
Where order effects are suspected to be significant, a targeted transformer probe can quantify their impact and guide both algorithmic and hardware-side mitigations. 

Looking ahead, more detailed circuit-level models that explicitly represent hardware components and integrate their dynamics over time may be required to fully capture complex effects beyond ordering. 
Such physically faithful models can provide valuable insights, but their computational cost makes them impractical for use in the training loop of large-scale networks. 
Instead, lightweight representations such as the white-box model strike a balance between fidelity and usability. 
More broadly, hardware models serve a dual purpose: they inform circuit design and calibration strategies on the hardware side, while guiding neural architecture and training design on the algorithmic side. 
The guiding principle that has emerged from our work is clear: models should not aim to perfectly emulate hardware, but to provide understanding and enable the training of networks that are robust to its peculiarities.

\chapter{Robustness Against Noisy Computations}
\label{ch:robustness}

\epigraph{What does not kill me makes me stronger.}{\textnormal{--- Friedrich Nietzsche, \textit{Twilight of the Idols} (1888)}}

\noindent
The preceding chapter addressed analog hardware accelerators from a device-centric perspective, focusing on how their nonidealities---such as nonlinearities, saturation, noise, and even ordering effects---can be captured in faithful models.  
These models serve as powerful tools for understanding the behavior of specific hardware instances and for enabling hardware-in-the-loop training.  
Yet, their strength lies precisely in their detail, and this level of fidelity also makes them less practical as a basis for generic robustness strategies on the algorithmic side.  

This chapter therefore shifts the focus from \emph{modeling hardware} to \emph{hardening neural networks against noisy computations}.  
Rather than reproducing the peculiarities of one device instance, we take a broader perspective: How can robustness be measured, understood, and systematically improved such that neural networks remain reliable when deployed on analog accelerators?  
The emphasis thus moves to the machine learning side of the co-design problem.  

To structure this discussion, we follow a trajectory across three works that form the backbone of this chapter.  
We begin with \emph{Walking Noise}~\cite{borras2024walkingnoiseECML}, which extends our earlier workshop paper~\cite{borras2022walkingnoiseAccML} and introduces a methodology to measure robustness at the granularity of individual layers, uncovering characteristic learning dynamics such as weight scaling and self-binarization.  
Building on these diagnostic insights, we then turn to \emph{hardening methods}~\cite{wang2025hardening}, where quantization-aware training and noisy training are systematically compared.  
This establishes noisy training as the baseline countermeasure, while clarifying the complementary role of quantization.  
Finally, we extend noisy training to \emph{\acrfull{vant}}~\cite{wang2025vant}, which addresses the fundamental limitation of static noise assumptions by introducing time-varying noise schedules, thereby improving robustness to practical effects such as drift and environmental variability.

Together, these contributions trace a coherent narrative: from \emph{measuring robustness}, to \emph{establishing effective countermeasures}, to \emph{sustaining robustness under dynamic and realistic conditions}.  
This progression highlights how robustness is not a singular property but a layered challenge that requires both diagnostic tools and adaptive training methods.  

This chapter builds on three of our publications---\emph{Walking Noise}~\cite{borras2024walkingnoiseECML}, \emph{Hardening}~\cite{wang2025hardening}, and \emph{\acrshort{vant}}~\cite{wang2025vant}---and focuses on their overarching narrative and distilled insights.  
Detailed methodologies, experimental setups, and quantitative results are provided in the respective original works.

\section{Robustness Against Hardware Noise}

The robustness of neural networks against noise has been studied from multiple perspectives, spanning generalization in machine learning, adversarial robustness, and hardware-induced uncertainty.  
A classical line of research established \emph{noise injection} as a form of regularization to improve generalization, with early works comparing it to weight decay and early stopping~\cite{grandvalet1997noise,jiang2009comparisonnoise}.  
With the advent of adversarial attacks, noise injection was also investigated as a defense strategy, often in combination with adversarial training~\cite{goodfellow2014explaining}.  
Variants include additive Gaussian perturbations to the inputs~\cite{bishop1995training}, ensembles with layer-wise noise injection~\cite{liu2018robustness}, and differentiating diffusion- versus jump-term randomness in continuous-time neural \acrlongpl{sde}~\cite{liu2020towardsrobust}.  

A second line of work focuses on \emph{noisy hardware}, where perturbations do not only affect the inputs but arise intrinsically from the computations themselves, e.g., during weight readout or multiply-accumulate operations.  
Training with additive Gaussian noise has been shown to mitigate such hardware-induced degradation~\cite{zhang2019precisionaware,camus2019trainingaware}, with further extensions such as Bayesian fine-tuning in \emph{BayesFT}~\cite{yang2020bayesianft}.  
In resistive memories, perturbation models have captured effects such as drift in \acrlong{rram} (\acrshort{rram}), and training with injected noise has been shown to mitigate performance degradation~\cite{du2020randomtelegraph}.  

To counteract accuracy degradation due to noisy computations, noisy training has also been successfully applied: while some works injected additive zero-mean Gaussian noise with varying variances during training~\cite{zhang2019precisionaware,camus2019trainingaware}, \emph{Noisy Machines}~\cite{Zhou2020} extended this approach with knowledge distillation from a digitally trained teacher to a noisy student.  
Beyond proposing distillation, it also provided insights into the higher sensitivity of deeper architectures compared to wider ones, explained through a mutual information analysis of information loss across layers.  
This line of work established noisy training, in combination with auxiliary mechanisms such as distillation, as a fundamental tool for enhancing robustness on noisy analog accelerators.  

Beyond Gaussian injection, \emph{perturbation-based methods} have been studied to improve both generalization and robustness.  
Perturbing weights can guide \acrshort{sgd} into flatter regions of the loss landscape, as formalized by \acrlong{sam} (\acrshort{sam})~\cite{foret2021sharpness}.  
By penalizing sharp minima, \acrshort{sam} improves generalization and offers a degree of robustness against input perturbations.  
Related studies investigated perturbations of both weights and inputs to strengthen adversarial robustness~\cite{gowal2020uncovering}.  

\emph{Quantization} provides a complementary perspective.  
Originally motivated by efficiency on digital accelerators, quantization implicitly injects structured noise by reducing numerical precision.  
Its effect on robustness has been studied extensively: some works report non-monotonic behavior of adversarial robustness across bit-widths~\cite{giacobbe2020quantization}, while others show that moderate quantization can increase resilience with little accuracy loss~\cite{duncan2018quantized}.  
Defensive quantization methods explicitly control the Lipschitz constant to prevent error amplification~\cite{lin2019defensive}, and further studies found that complex DNNs can absorb severe quantization through retraining, whereas smaller networks degrade more substantially~\cite{sung2015quantized}.  
From the analog accelerator perspective, quantization is often indispensable due to limited DAC/ADC precision, motivating the use of \acrlong{qat} to preserve robustness under such constraints~\cite{krishnamoorthi2018quantization}.  

In summary, prior research established noise injection as a versatile tool for generalization and robustness.  
For analog accelerators, noisy training and quantization emerged as common countermeasures.  
Building on these insights, the following sections first examine per-layer robustness through \emph{Walking Noise}, highlighting how neural networks adapt to tolerate noise when trained with noise injection.  
We then compare hardening strategies such as quantization and noisy training, quantifying their effectiveness for analog computations.  
Finally, motivated by observations on real analog hardware, noisy training is extended to account for dynamic variations in noise—such as those caused by temperature fluctuations—through \acrshort{vant}.

\section{Walking Noise}
\label{sc:analog:walkingnoise}

Robustness of neural networks is usually quantified at the network level, for example as overall accuracy degradation under noise.  
Such aggregate measures, however, conceal that sensitivity to perturbations is highly heterogeneous across layers: early layers may amplify noise, intermediate layers can compensate through redundancy, while later layers often remain fragile.  
For analog accelerators this distinction is particularly relevant, as noise may arise at different points in the computation chain and its impact depends on where in the architecture it occurs.  

\emph{Walking Noise}~\cite{borras2024walkingnoiseECML} addresses this by probing robustness at the level of individual layers.  
By injecting noise sequentially layer by layer, it enables the identification of weak spots and provides insights into how networks adapt to tolerate noise.  
This fine-grained view complements global robustness measures and serves as a diagnostic basis for developing hardening methods.  

\subsection{Methodology}

The key idea of \emph{Walking Noise}~\cite{borras2024walkingnoiseECML} is to inject noise selectively at one layer at a time, while keeping the rest of the network unaffected.  
By “walking” the noise across layers, robustness can be probed in a fine-grained manner.  
Three types of perturbations are considered: additive Gaussian noise, multiplicative Gaussian noise, and combinations thereof.  

Depending on the accelerator, noise may arise at different stages of a matrix multiplication: (1) during weight readout, (2) in the multiply–accumulate itself, or (3) when forwarding activations.  
We focus on the last case, as it captures accumulated effects from both weight and computation noise.  

\begin{figure}%
	\centering
	\includegraphics[width=0.75\columnwidth]{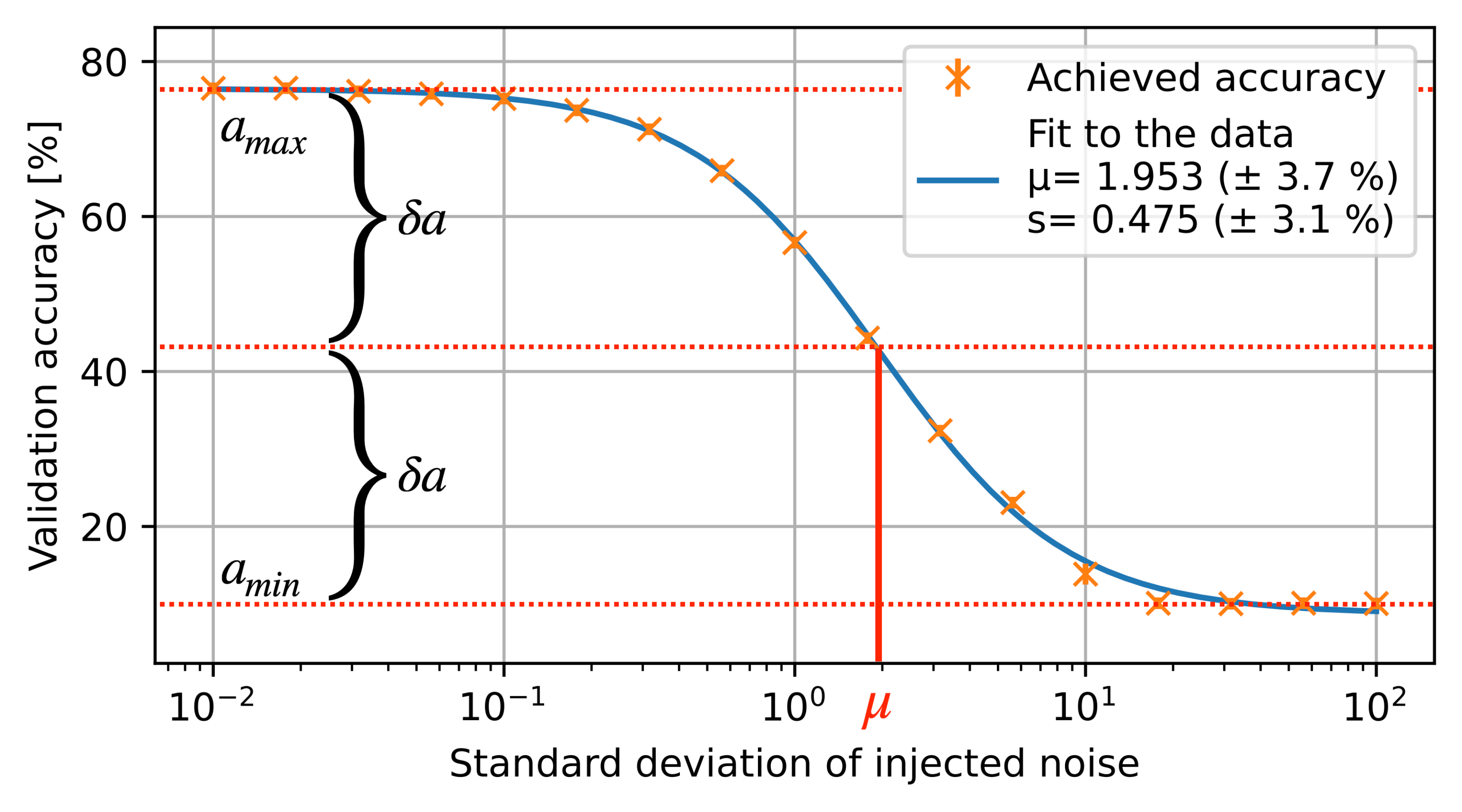}
	\caption{\emph{Midpoint noise level $\mu$} for the example of LeNet-5/CIFAR-10/\acrshort{bn} and globally injected additive noise. \reprofrom{borras2024walkingnoiseECML}}
	\label{fig:walkingnoise:method:acc_v_std_global_LeNet-BN-CIFAR10}
\end{figure}

To quantify robustness, the \emph{midpoint noise level}~$\mu$ is introduced.  
When increasing the globally injected noise, model accuracy typically follows a characteristic curve: it remains stable for low noise levels, then drops once a critical region is reached (Fig.~\ref{fig:walkingnoise:method:acc_v_std_global_LeNet-BN-CIFAR10}).  
This behavior is well captured by fitting a logistic function to the accuracy–noise relation,  

\begin{equation} \label{eq:logistic_CDF}
	F(\sigma; \mu, s, \delta a, a_{min}) = \frac{2}{1+e^{(\sigma-\mu)/s}} \cdot \delta a + a_{min}
\end{equation}
where $\delta a = (a_{\text{max}} - a_{\text{min}})/2$ denotes half of its maximum accuracy, $\sigma$ the varying injected noise level, $s$ the slope of the curve.
The parameter $\mu$ of this fit denotes the standard deviation at which accuracy falls to the midpoint between its clean performance and random guessing.  
It thus provides a natural and comparable measure of robustness: the higher $\mu$, the more noise the network or individual layer can tolerate before collapsing.  
Experiments cover standard convolutional networks on image classification benchmarks, trained both with and without noise injection, with details provided in~\cite{borras2024walkingnoiseECML}.

\subsection{Findings}

Applying the Walking Noise methodology reveals distinct robustness mechanisms depending on the type of perturbation and highlights strong heterogeneity across layers.  

\paragraph{Additive noise.}  
Under additive Gaussian perturbations, networks develop a compensatory mechanism by increasing the magnitude of their weights.  
This self-organized adaptation enhances the effective signal-to-noise ratio, allowing activations to remain separable despite the presence of noise.  
This behavior is clearly reflected in the characteristic accuracy–noise curves, which degrade smoothly with increasing perturbation strength before eventually collapsing to random guessing.  
The resulting robustness metric~$\mu$, derived from these curves, enables systematic comparison across datasets and architectures.  

\begin{figure} 
	\centering
	\subfloat[MLP on MNIST.\label{fig:walkingnoise:apdx:add:agg_MLP-MNIST}]{%
		\includegraphics[width=0.90\textwidth]{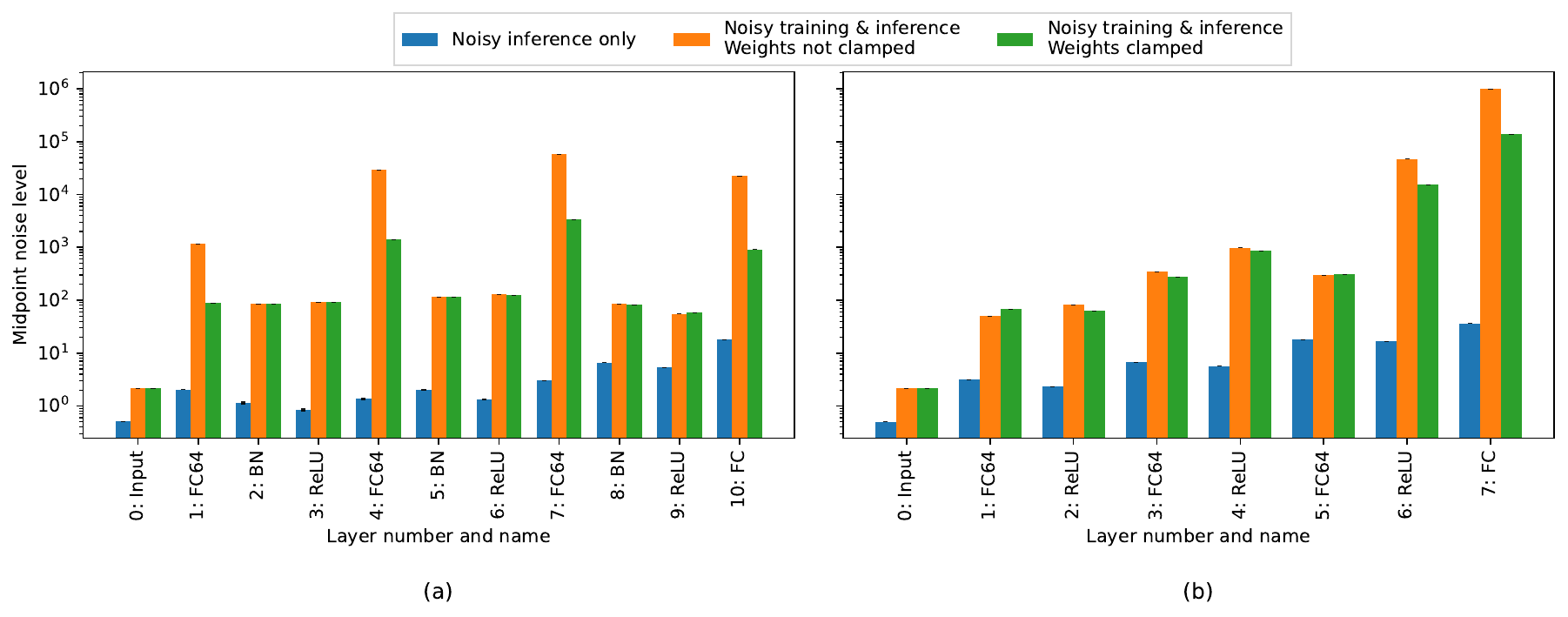}}
	\\
	\subfloat[LeNet-5 on MNIST.\label{fig:walkingnoise:apdx:add:agg_LeNet-MNIST}]{%
		\includegraphics[width=0.90\textwidth]{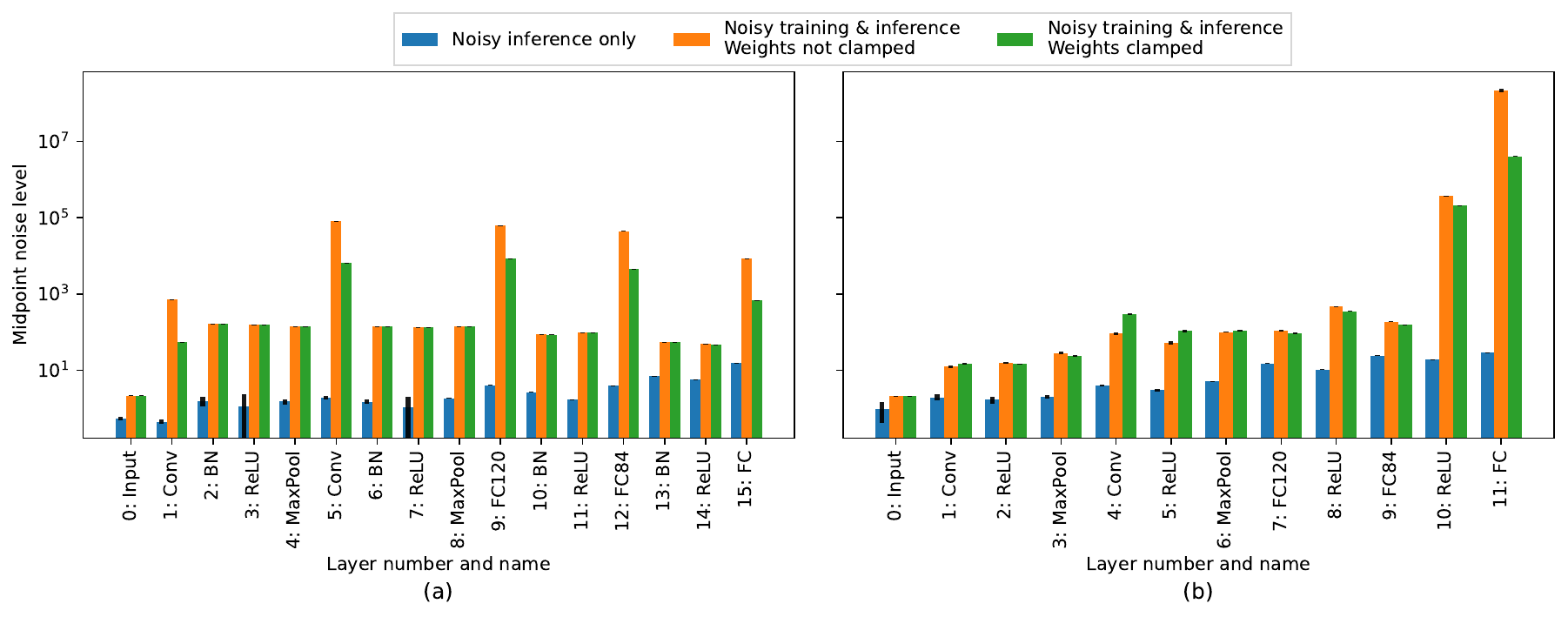}}
	\\
	\subfloat[MLP on CIFAR-10.\label{fig:walkingnoise:apdx:add:agg_MLP-CIFAR10}]{%
		\includegraphics[width=0.90\textwidth]{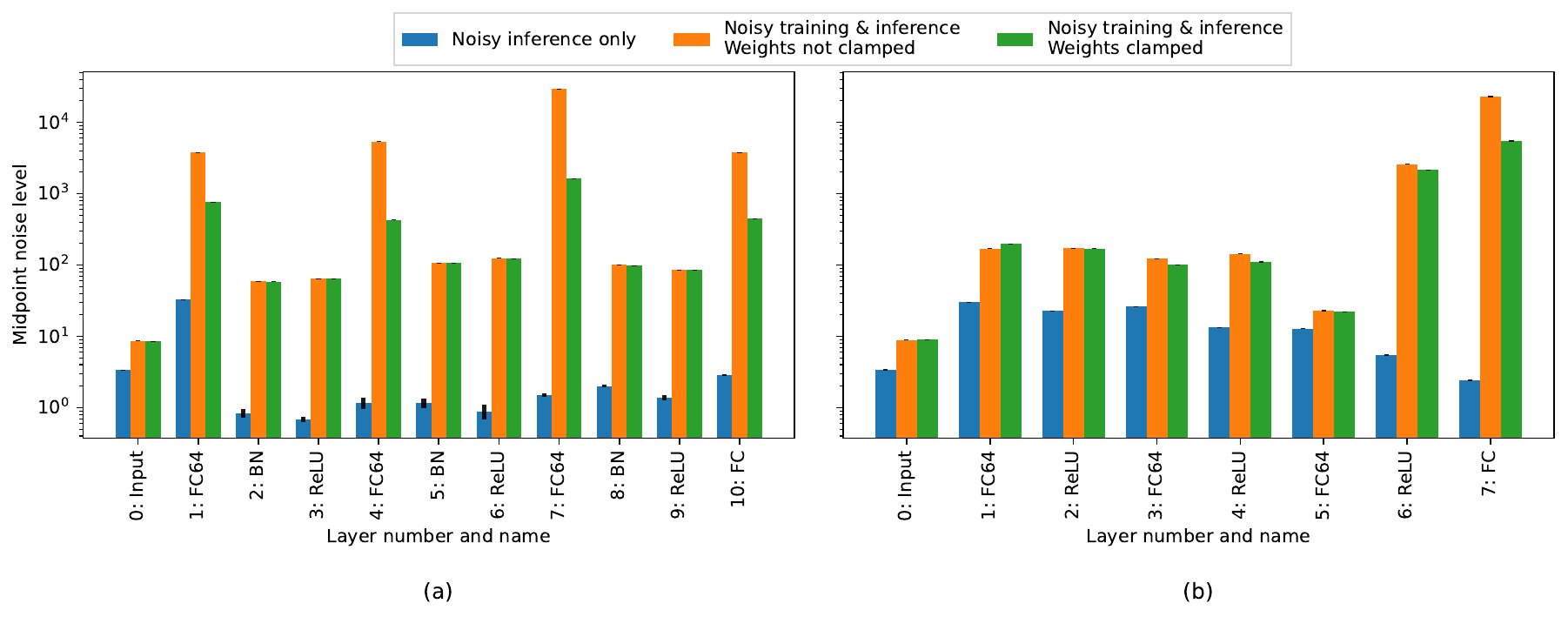}}
	\caption{Midpoint noise level for various model architectures and image classification datasets based on Walking Noise. The x axis shows layer number and name. Figures to the left are with \acrshort{bn}, Figures to the right are without \acrshort{bn}. \reprofrom{borras2024walkingnoiseECML}}
	\label{fig:walkingnoise:apdx:add:agg_ct}
\end{figure}

\emph{\Acrfull{bn}} interacts strongly with this mechanism.  
Since \acrshort{bn} normalizes activation scales after each layer, it counteracts weight growth and therefore suppresses the network’s ability to exploit weight magnitude as a compensation strategy.  
As a result, models with \acrshort{bn} often show lower robustness to additive noise than their non-\acrshort{bn} counterparts (Fig.~\ref{fig:walkingnoise:apdx:add:agg_ct}).  

\begin{figure}%
	\centering
	\includegraphics[width=0.7\textwidth]{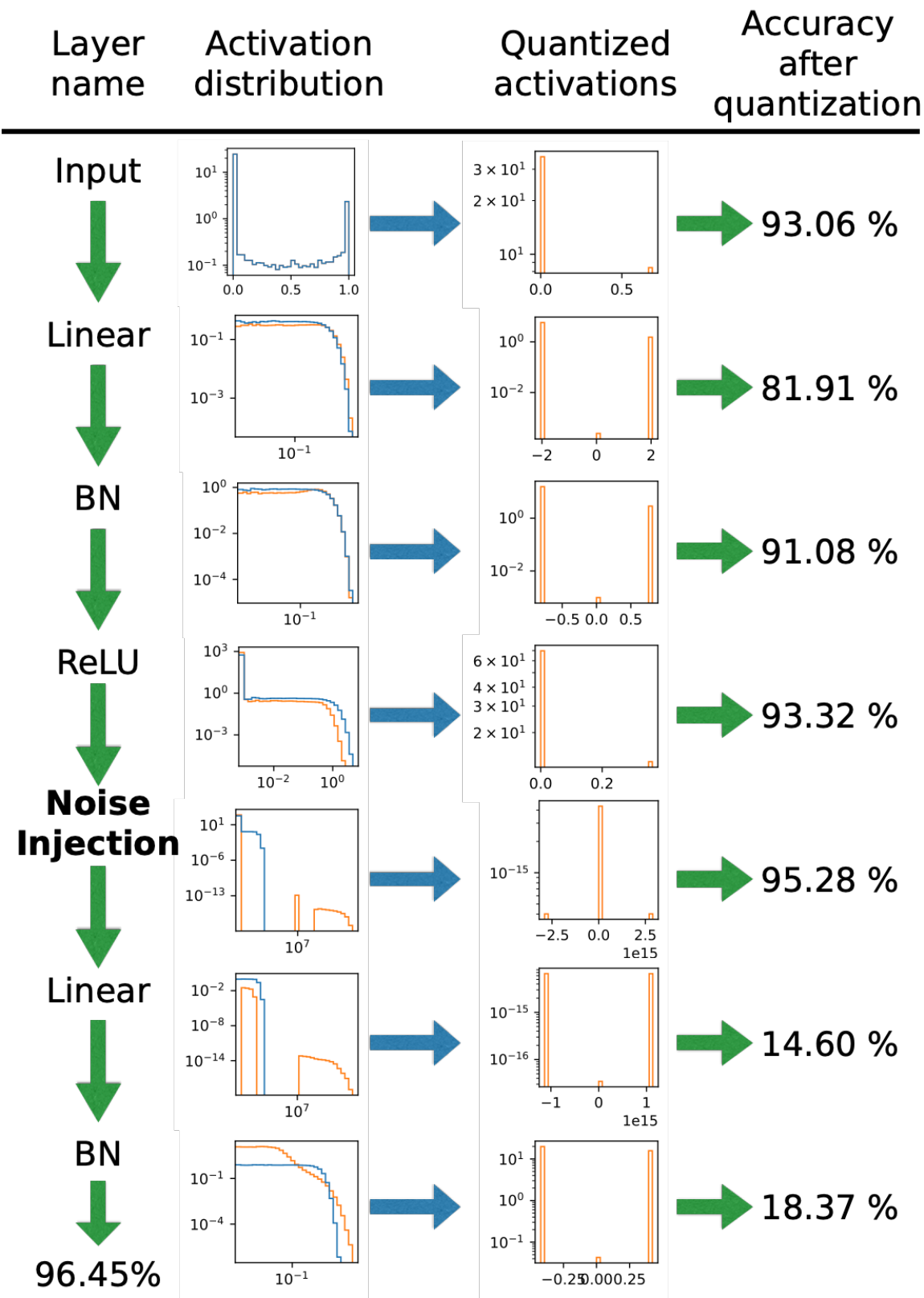}
	\vspace{-0.35cm}
\caption{Visualization of \emph{self-binarization} in an MLP trained on MNIST under multiplicative \emph{Walking Noise}.  
The histograms on the left show the distribution of activation values with noise (orange) and without noise (blue) after layer execution,  
using both logarithmic and linear x-axis scales to emphasize differences between the two networks.  
On the right, the distribution of activation values is shown together with the resulting accuracy when applying simple threshold-based quantization without retraining to the respective layer.  
\reprofrom{borras2024walkingnoiseECML}}
	\label{fig:walkingnoise:mul:self-bin}
	\vspace{-0.6cm}
\end{figure}

\paragraph{Multiplicative noise.}  
In contrast to the additive case, multiplicative noise revealed an unexpected and remarkably strong form of robustness.  
When injecting multiplicative Gaussian perturbations during inference, we observed that some layers maintained high accuracy even for extremely large noise levels (up to $\sigma=10^{10}$), without collapsing to random prediction accuracy.  
This behavior was particularly surprising, as no model could withstand such noise levels under additive injection or global perturbations.  

Closer inspection of the activation distributions showed that networks learn to \emph{self-binarize}: activations split into two distinct peaks, one near zero and one near the noise standard deviation (Fig.~\ref{fig:walkingnoise:mul:self-bin}).  
Information can then be encoded simply in the presence or absence of an activation peak, making the network insensitive to the precise values randomized within each cluster.  
The emergent binary representation explains the remarkable tolerance to multiplicative noise.  
This was evidenced by the fact that explicit threshold-based quantization without retraining caused only a marginal accuracy drop, demonstrating that the network had already learned an internal binary representation.

\begin{figure}%
	\centering
	\includegraphics[width=0.7\columnwidth]{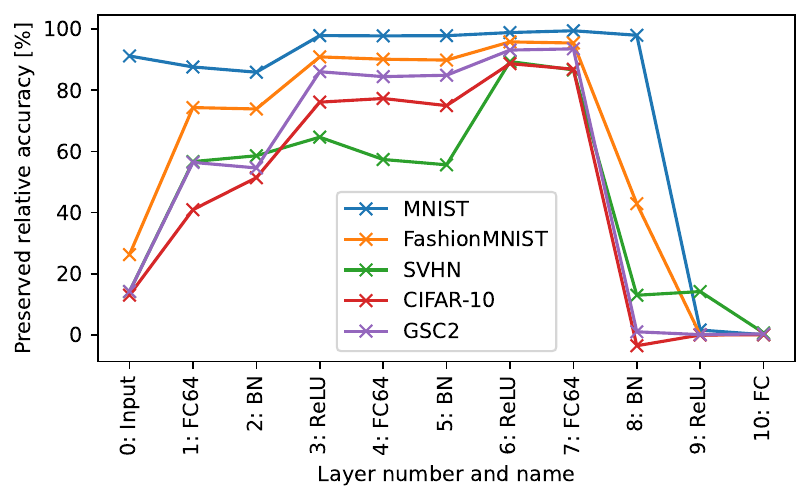}
	\vspace{-0.35cm}
	\caption{Accuracy preservation for MLP with \acrshort{bn} and various datasets, noise injected multiplicatively. Reproduced with permission from~\cite{borras2024walkingnoiseECML}.}
	\label{fig:walkingnoise:mul:MLP_BN_compare_datasets}
	\vspace{-0.4cm}
\end{figure}

Here, \emph{\acrlong{bn}} is essential.  
By stabilizing the scale of activations, \acrshort{bn} enables the formation and propagation of these bimodal distributions, and thus the self-binarization mechanism itself.  
Without \acrshort{bn}, networks largely lose this ability, and accuracy drops quickly when multiplicative noise is injected.  
The effect generalizes across architectures and datasets (Figs.~\ref{fig:walkingnoise:mul:MLP_BN_compare_datasets}), though sensitivity again varies by layer, with later layers eventually failing to decode the binarized representation.  

\paragraph{Mixed noise.}  
Since the self-learned robustness strategies for additive and multiplicative perturbations differ fundamentally, their interaction in a mixed setting is of particular interest.  
Real-world analog matrix–multiply accelerators are subject to multiple sources of noise, and depending on the hardware, additive or multiplicative components may dominate and occur earlier in the computational chain.  
We therefore evaluate both injection orders:
\begin{equation}
\begin{split}
\text{multiplicative-first: } & y(x) = \big(x \cdot \mathcal{N}(1,\sigma^{2}_\text{mul})\big) + \mathcal{N}(0,\sigma^{2}_\text{add}) \\
\text{additive-first: } & y(x) = \big(x + \mathcal{N}(0,\sigma^{2}_\text{add})\big) \cdot \mathcal{N}(1,\sigma^{2}_\text{mul}) . 
\end{split}
\end{equation}
Here, $x$ denotes the clean input, while $\mathcal{N}(\mu,\sigma^{2})$ represents Gaussian noise with mean~$\mu$ and variance~$\sigma^{2}$.  
The parameters $\sigma_\text{mul}$ and $\sigma_\text{add}$ control the standard deviations of the multiplicative and additive components, respectively.  

\begin{figure}
	\centering
	\begin{subfigure}{1.0\textwidth}
		\includegraphics[width=\textwidth]{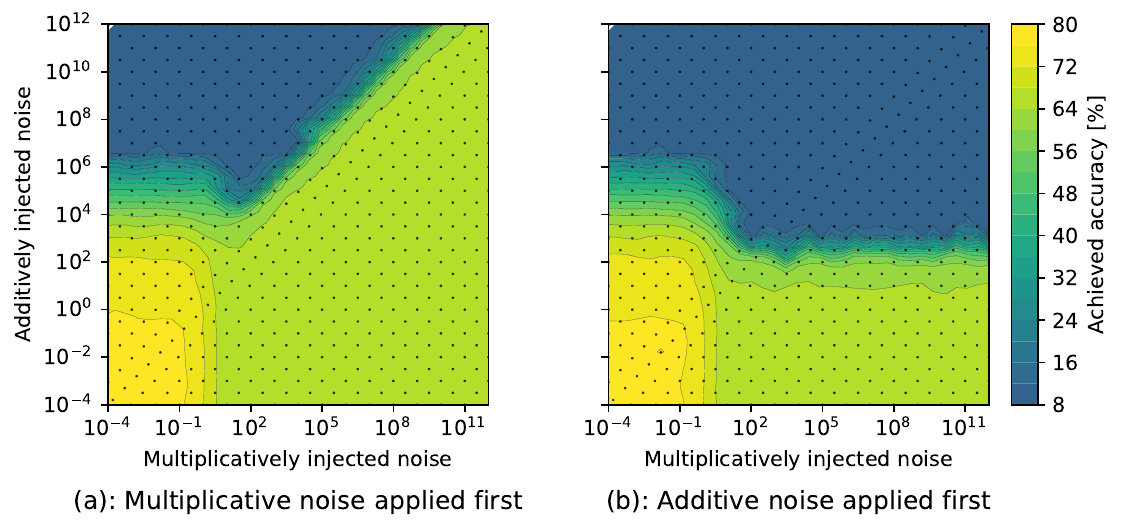}
	\end{subfigure}
	\caption{Accuracy for training with multiplicative and additive noise at layer 5 (Conv) of LeNet-5 trained on CIFAR-10. The black dots indicate points of measurement. Reproduced with permission from~\cite{borras2024walkingnoiseECML}.}
	\label{fig:walkingnoise:comb:contour_LeNet-BN-CIFAR10_both}
\end{figure}

Figure~\ref{fig:walkingnoise:comb:contour_LeNet-BN-CIFAR10_both} illustrates the resulting accuracy surfaces under both injection orders.  
A notable observation is that networks can still learn to self-binarize if multiplicative noise is applied first, thereby maintaining extreme robustness.  
This effect persists even when substantial additive noise is injected afterwards, in some cases up to one order of magnitude stronger.  
From a hardware perspective, this suggests a rather counterintuitive but powerful implication: in principle, even an accelerator dominated by strong additive noise could remain workable if its computation pipeline introduced multiplicative noise first.  
For practical accelerator design this means that hardware-in-the-loop trained networks may tolerate significant additive disturbances, provided these occur after multiplicative ones.  

\paragraph{Layer-specific sensitivity.}  
Walking Noise consistently demonstrates that robustness is highly heterogeneous across layers.  
By scanning noise layer by layer, weak spots of a network can be identified, often located in the first or last layers depending on the type of perturbation.  
This information can be exploited to guide targeted countermeasures.  
For example, selectively re-executing only the most sensitive layers yields significantly higher accuracy compared to uniformly repeating all layers (Table~\ref{tab:walkingnoise:repetitions}), thereby improving efficiency.  

\begin{table}[!t]
\centering
\scriptsize
\renewcommand{\arraystretch}{1.05}
\caption{Walking Noise guiding multi-execution to improve accuracy. \reprofrom{borras2024walkingnoiseECML}}
\label{tab:walkingnoise:repetitions}
\begin{tabular}{c|c|c|l|l|c|c}
 &  &  & \multicolumn{2}{c|}{\textbf{Executions per layer}} & \multicolumn{2}{c}{\textbf{Accuracy}} \\ \hline
\textbf{Dataset} & \textbf{Model} & \textbf{\acrshort{bn}} & \textbf{Uniform} & \textbf{Guided} & \textbf{Uniform} & \textbf{Guided} \\ \hline
MNIST & MLP & with & \{2,2,2,...\} & \{6,1,3,3,2,1,2,1,1,1,1\} & $71.4\pm0.7$\,\% & $\mathbf{80.6\pm0.5}$\,\% \\ \hline
MNIST & MLP & wo & \{2,2,2,...\} & \{8,1,2,1,1,1,1,1\} & $68.2\pm1.3$\,\% & $\mathbf{87.1\pm0.7}$\,\% \\ \hline
CIFAR-10 & MLP & with & \{2,2,2,...\} & \{1,1,3,4,2,2,3,2,1,2,1\} & $41.7\pm0.5$\,\% & $\mathbf{43.7\pm0.5}$\,\% \\ \hline
CIFAR-10 & MLP & wo & \{2,2,2,...\} & \{4,1,1,1,1,1,2,5\} & $38.9\pm0.6$\,\% & $\mathbf{43.4\pm0.7}$\,\% \\ \hline
CIFAR-10 & LeNet-5 & with & \{2,2,2,...\} & \{3,2,3,3,3,1,3,4,2,1,1,2,1,1,1,1\} & $58.5\pm1.4$\,\% & $\mathbf{61.1\pm1.0}$\,\% \\ \hline
CIFAR-10 & LeNet-5 & wo & \{2,2,2,...\} & \{4,3,4,3,1,3,1,1,1,1,1,1\} & $57.2\pm1.8$\,\% & $\mathbf{62.4\pm1.3}$\,\% \\ 
\end{tabular}
\end{table}

\paragraph{Summary.}  
Overall, Walking Noise uncovers characteristic mechanisms of robustness in neural networks.  
For additive perturbations, robustness is achieved by increasing weight magnitudes, a strategy that is suppressed when \emph{\acrlong{bn}} is applied, since \acrlong{bn} removes activation scale.  
For multiplicative perturbations, networks develop a self-binarization of activations, a mechanism that crucially depends on \acrlong{bn} to stabilize and propagate the bimodal distributions.  
Finally, under mixed noise conditions, the order of perturbation determines whether robustness can be preserved, with multiplicative-first injection enabling tolerance even in the presence of strong additive noise.  
Together, these findings establish robustness as a heterogeneous, layer-dependent property, and provide actionable insights for architecture design, compression strategies, and selective hardening.  

Having established how robustness emerges at the layer level, we next investigate systematic strategies to \emph{harden} neural networks against noise, comparing the effectiveness of quantization-aware training and noisy training.

\section{Hardening Methods}
\label{sc:analog:hardening}

The insights from Walking Noise highlight that robustness is not a uniform property, but instead emerges from heterogeneous layer-wise mechanisms.  
While this diagnostic perspective reveals how networks adapt to tolerate perturbations and helps to design more robust neural architectures, it leaves open the practical question of how to \emph{systematically improve robustness} during training.  

Two prominent methods are commonly considered.  
\acrfull{qat} has long been used to prepare models for low-precision deployment, and implicitly introduces structured perturbations that may also enhance robustness.  
In contrast, noisy training directly injects stochastic perturbations during training, explicitly mimicking the conditions of noisy inference.  
Both approaches are widely applied, yet their relative effectiveness for analog computations---where noise is unavoidable---had not been systematically compared.  

This motivates the following study: to evaluate \acrshort{qat} and noisy training side by side across different architectures, and quantify their robustness using the midpoint noise level~$\mu$.  

\subsection{Methodology}

To compare hardening strategies, we evaluate \acrshort{qat} and noisy training across several neural architectures, including LeNet-5, VGG-11, and ResNet-18, trained on the CIFAR-10 dataset.  
\acrshort{qat} is implemented using uniform quantization of weights and activations, with both constant and dynamic scaling factors considered.  
Constant scaling applies a fixed factor across activations, whereas dynamic scaling recomputes scaling factors adaptively during inference.  

Noisy training, in contrast, directly injects additive Gaussian noise into activations during training, with the same distribution applied during inference.  
This exposes the model to realistic perturbations and compels it to learn representations that remain robust under noisy computations.  

To quantify robustness, we use the \emph{midpoint noise level}~$\mu$, defined as the noise standard deviation at which accuracy falls to the midpoint between its clean baseline and random guessing (cf. Sec.~\ref{fig:walkingnoise:method:acc_v_std_global_LeNet-BN-CIFAR10}).  
This metric provides a stable measure of tolerance against noise and allows direct comparison across models and training strategies.  

The evaluation is performed by measuring accuracy under globally injected noise during inference, fitting the accuracy–noise curve, and extracting~$\mu$.  
This enables a systematic comparison of robustness for different bit-widths, scaling strategies, and training methods.  

\subsection{Findings}

The comparison between \acrshort{qat} and noisy training reveals several characteristic patterns.  

\begin{figure}
    \centering

	\begin{subfigure}{0.32\textwidth}
		\includegraphics[width=\textwidth]{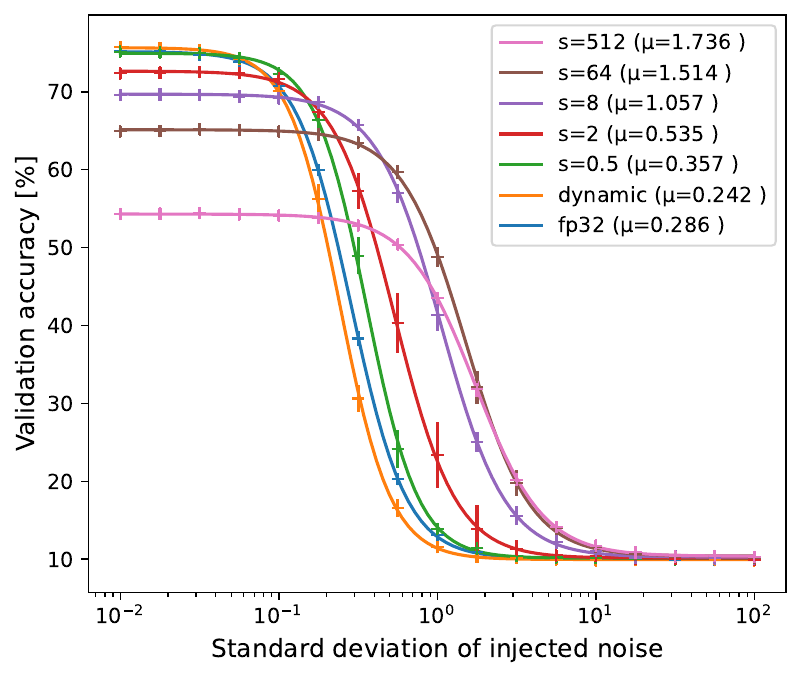}
		\caption{LeNet-5, 8-bit}
	\end{subfigure}
	\hfill
	\begin{subfigure}{0.32\textwidth}
		\includegraphics[width=\textwidth]{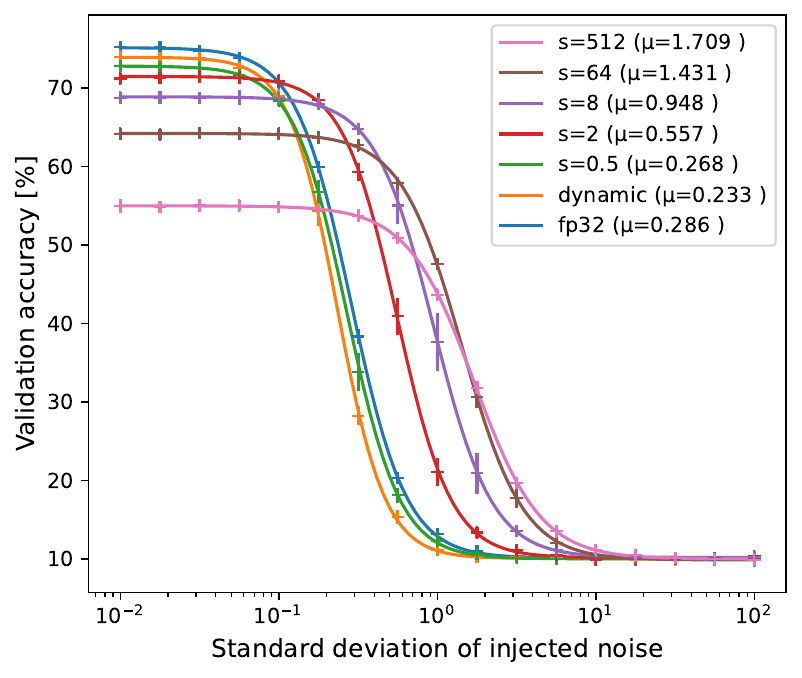}
		\caption{LeNet-5, 4-bit}
	\end{subfigure}
	\hfill
	\begin{subfigure}{0.32\textwidth}
		\includegraphics[width=\textwidth]{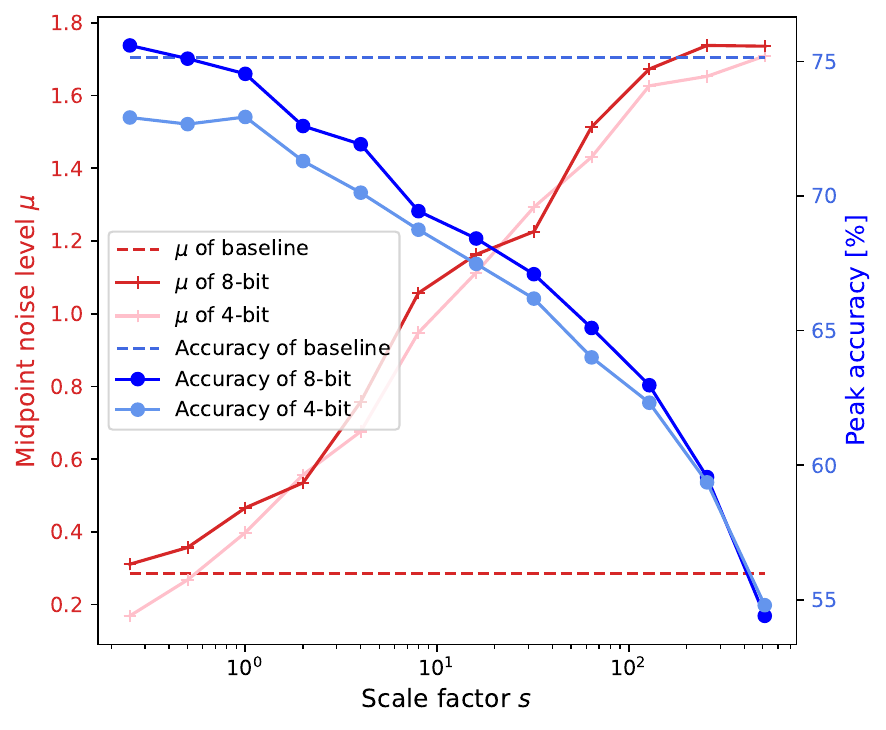}
		\caption{Accuracy--$\mu$ trade-off}
	\end{subfigure}

	\begin{subfigure}{0.32\textwidth}
		\includegraphics[width=\textwidth]{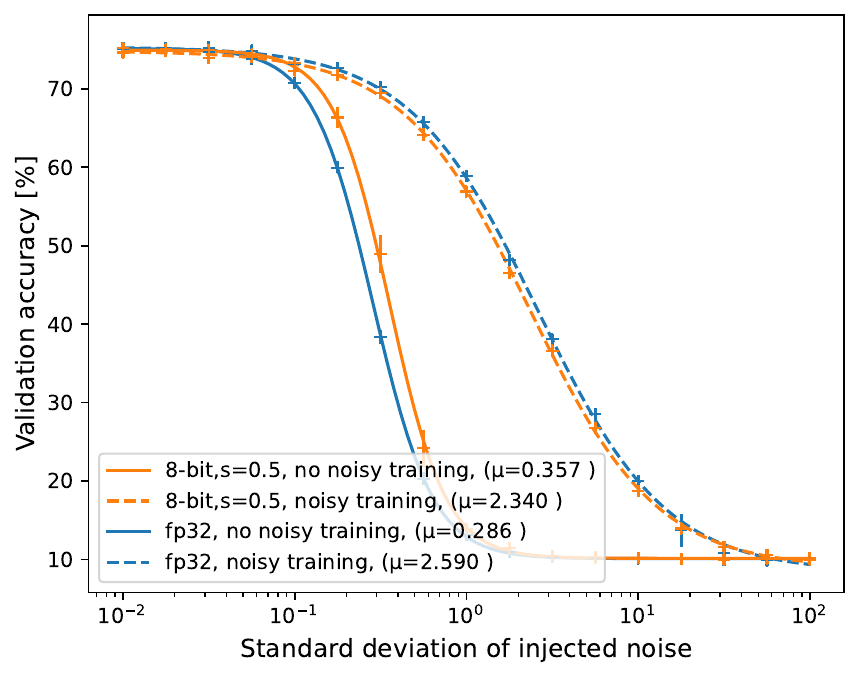}
		\caption{LeNet-5 + noisy training}
	\end{subfigure}
	\hfill
	\begin{subfigure}{0.32\textwidth}
		\includegraphics[width=\textwidth]{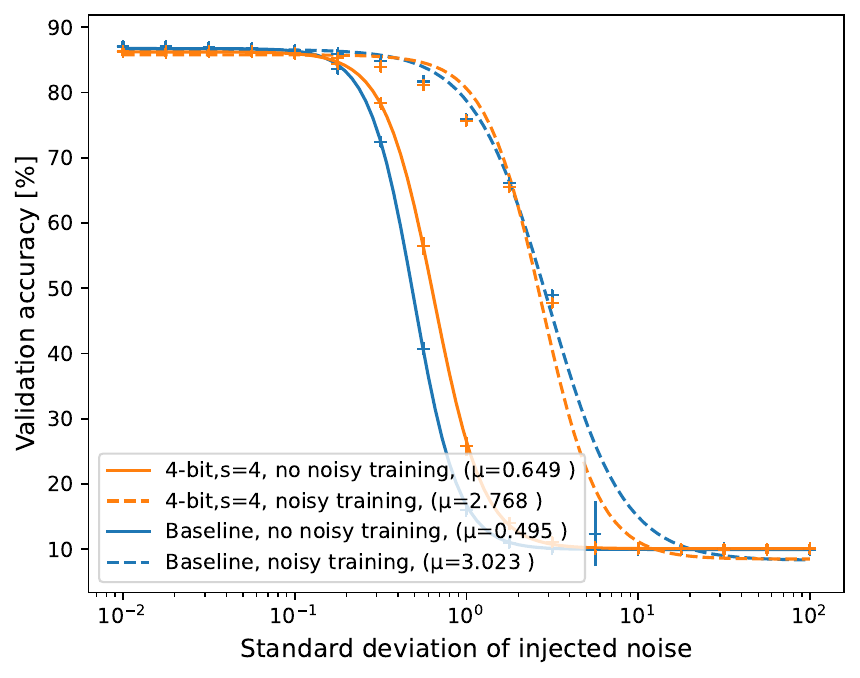}
		\caption{ResNet-18 + noisy train.}
	\end{subfigure}
	\hfill
	\begin{subfigure}{0.32\textwidth}
		\includegraphics[width=\textwidth]{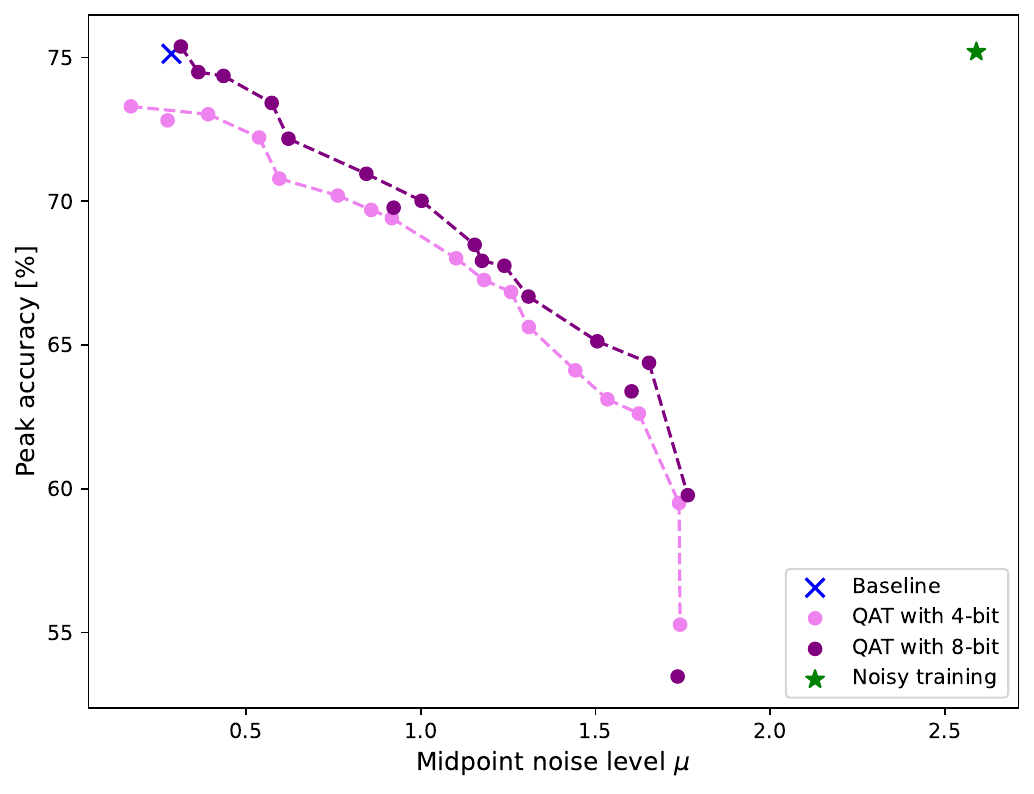}
		\caption{Pareto frontier}
	\end{subfigure}

 	\caption{Robustness of LeNet-5 and ResNet-18 on CIFAR-10 under quantization and noisy training.  
	Top row: quantization alone, showing the effect of bit width, scaling factors, and the trade-off between accuracy and midpoint noise level~$\mu$.  
	Bottom row: quantization combined with noisy training, demonstrating the substantial robustness gains across architectures and the resulting Pareto frontier.  
	The dramatic improvement in robustness of quantized models with noisy training is clearly visible.  
	All x-axes are logarithmic.  
	\reprofrom{wang2025hardening}}
    \label{fig:hardening:quant_noisy_combined}

\end{figure}

\paragraph{Quantization-aware training.}  
Quantization alone provides a modest degree of robustness, but the effect strongly depends on bit-width and the scaling strategy.  
For LeNet-5 on CIFAR-10, 8-bit quantization achieves higher robustness than 4-bit (Fig.~\ref{fig:hardening:quant_noisy_combined}), since larger dynamic ranges reduce clipping effects.  
Constant scaling factors can significantly increase the midpoint noise level~$\mu$, but at the cost of lower clean accuracy, leading to a trade-off between robustness and peak performance.  
Dynamic scaling preserves clean accuracy, but generally results in lower robustness.  
The Pareto trade-off between peak accuracy and robustness is highlighted in Fig.~\ref{fig:hardening:quant_noisy_combined}(c), and extends to deeper architectures such as VGG-11 and ResNet-18 (Table~\ref{tab:hardening:vgg_quant}).  
Overall, \acrshort{qat} improves tolerance to noise but does not fundamentally change the fragility of deeper models.  

\begin{table}
	\caption{Robustness of VGG-11 and ResNet-18 on CIFAR-10 under different bit widths and scaling factors using \acrshort{qat}.  
	Constant scaling yields higher robustness than dynamic scaling, and increasing the scaling factor improves tolerance up to a point where the network collapses completely.  
	\reprofrom{wang2025hardening}}
	\scriptsize
	\centering
	\begin{tabular}{c|cc|c|c} 
		\textbf{Model} & \textbf{Bitwidths} & \textbf{Scaling factors} & \textbf{Peak accuracy (\%)} & \textbf{Midpoint noise level $\mu$ }\\
		\hline
		\rule[-1ex]{0pt}{2.5ex} &fp32 & - & \textbf{87.7} & 0.154 ($\pm$0.5\%) \\
		\cline{2-5}
		\rule[-1ex]{0pt}{2.5ex} &8-bit & dynamic & 87.2 & 0.024 ($\pm$0.1\%) \\
		\rule[-1ex]{0pt}{2.5ex} & & 0.5 & 84.3 & 0.145 ($\pm$0.2\%)\\ 
		\rule[-1ex]{0pt}{2.5ex} & & 1 & 82.2 & 0.2 ($\pm$0.2\%)\\ 
		\rule[-1ex]{0pt}{2.5ex} & & 2 & 76.8 & 0.222 ($\pm$0.3\%)\\ 
		\rule[-1ex]{0pt}{2.5ex} \textbf{VGG-11}& & 3 & 10.0 & 0.013 ($\pm$0.0\%) \\ 
		\cline{2-5}
		\rule[-1ex]{0pt}{2.5ex} &4-bit  & dynamic & 86.5 & 0.031 ($\pm$0.1\%) \\ 
		\rule[-1ex]{0pt}{2.5ex} & & 0.5 & 84.5 & 0.12 ($\pm$0.2\%)\\ 
		\rule[-1ex]{0pt}{2.5ex} & &  1 & 82.6 & 0.177 ($\pm$0.2\%)\\ 
		\rule[-1ex]{0pt}{2.5ex} & &  2 & 78.3 & \textbf{0.23 ($\pm$0.3\%)}\\ 
		\rule[-1ex]{0pt}{2.5ex} & &  3 & 10 & 0.010 ($\pm$0.1\%) \\
		\hline
		\hline
		\rule[-1ex]{0pt}{2.5ex} &fp32 & - & 87.0 & 0.495 ($\pm$0.2\%) \\
		\cline{2-5}
		\rule[-1ex]{0pt}{2.5ex} &8-bit & dynamic & \textbf{87.3} & 0.452 ($\pm$0.3\%) \\
		\rule[-1ex]{0pt}{2.5ex} & & 0.5 & 87.1 & 0.526 ($\pm$0.3\%)\\ 
		\rule[-1ex]{0pt}{2.5ex} & & 1 & 87.0 & 0. 557 ($\pm$0.3\%)\\ 
		\rule[-1ex]{0pt}{2.5ex} & & 4 & 86.5 & 0.577 ($\pm$0.3\%)\\ 
		\rule[-1ex]{0pt}{2.5ex} & & 8 & 85.7 & 0.625 ($\pm$0.4\%) \\ 
		\rule[-1ex]{0pt}{2.5ex} \textbf{ResNet-18}& &  10 & 10 & 0.05 ($\pm$2.5\%) \\
		\cline{2-5}
		\rule[-1ex]{0pt}{2.5ex} &4-bit  & dynamic & 86.8 & 0.281 ($\pm$0.3\%) \\ 
		\rule[-1ex]{0pt}{2.5ex} & & 0.5 & 86.5 & 0.492 ($\pm$0.4\%)\\ 
		\rule[-1ex]{0pt}{2.5ex} & &  1 & 86.7 & 0.605 ($\pm$0.3\%)\\ 
		\rule[-1ex]{0pt}{2.5ex} & &  4 & 86.5 & 0.657 ($\pm$0.5\%)\\ 
		\rule[-1ex]{0pt}{2.5ex} & &  8 & 86.5 & \textbf{0.665 ($\pm$0.3\%)} \\
		\rule[-1ex]{0pt}{2.5ex} & &  10 & 10 & 0.005 ($\pm$1.3\%) \\
	\end{tabular}
	\label{tab:hardening:vgg_quant}
\end{table}

\paragraph{Noisy training.}  
In contrast, noisy training proves highly effective across all architectures.  
By exposing the network to Gaussian perturbations during training, the learned representations become resilient to noise injected during inference.  
As shown in Fig.~\ref{fig:hardening:quant_noisy_combined}, noisy training consistently raises robustness far beyond that of quantization, while maintaining competitive clean accuracy.  
Table~\ref{tab:hardening:flops-transposed} summarizes this effect across LeNet-5, VGG-11, and ResNet-18: the midpoint noise level~$\mu$ improves substantially when noisy training is applied, largely independent of model scale or parameter count.

\begin{table}
	\caption{Overall comparison of architectures on CIFAR-10.  
	ResNet-18 exhibits higher robustness than LeNet-5 and VGG-11, owing to its skip connections, which compensate for the increased depth that otherwise amplifies noise sensitivity.  
	Across all models, noisy training effectively hardens networks against noise.  
	\reprofrom{wang2025hardening}}
	\small
	\centering
	\begin{tabular}{ l|c|c|c } 
		\textbf{Property} & \textbf{LeNet-5} & \textbf{VGG-11} & \textbf{ResNet-18} \\
		\hline
		\textbf{FLOPS (M)} & 0.66 & 276.56 & 37.53 \\
		\textbf{Param (M)} & 0.06 & 132.86 & 11.69 \\
		\textbf{Noisy layers} & 12 & 27 & 33 \\
		\textbf{Peak accuracy (\%)} & 75 & 87.7 & 86.9 \\
		\textbf{Robustness $\mu$ w/o Noisy Training} & 0.286 & 0.154 & 0.494 \\
		\textbf{Robustness $\mu$ with Noisy Training} & 2.59 & 2.957 & 3.023 \\
	\end{tabular}
	\label{tab:hardening:flops-transposed}
\end{table}

\paragraph{Combining methods.}  
When quantization is combined with noisy training, robustness remains dominated by the effect of noisy training (Fig.~\ref{fig:hardening:quant_noisy_combined}).  
Quantization does not further increase robustness, but it does enable more compact models without losing the benefits of noise-aware training.  
This makes noisy training with quantization a practical approach for robust deployment on accelerators with limited precision.  

\paragraph{Summary.}  
In summary, \acrshort{qat} and noisy training both contribute to robustness, but in different ways.  
\acrshort{qat} shifts the trade-off between accuracy and tolerance by controlling scaling, whereas noisy training fundamentally reshapes network representations to withstand noisy inference.  
The combination yields compact and robust models, but noisy training emerges as the more decisive hardening method.

\section{Variance-aware Noisy Training}
\label{sc:analog:vant}

The previous section showed that noisy training is the most effective strategy to harden neural networks against noisy computations.  
However, it also assumes that the noise distribution observed during training is identical to the one encountered at inference time.  
This assumption rarely holds for analog accelerators: noise levels drift with temperature, voltage, device aging, and even between individual chips.  
As a result, models trained with noisy training at a fixed noise level often fail when the actual operating conditions deviate.  

To address this limitation, we introduce \acrfull{vant}.  
The key idea is to extend noisy training by injecting noise with varying strength, sampled from a distribution rather than fixed at a single level.  
In this way, models are explicitly exposed to the variability characteristic of real hardware, and can learn representations that remain robust even under fluctuating noise conditions.

\subsection{Methodology}

\begin{figure}
    \centering
    \begin{minipage}[t]{.495\textwidth}
        \centering
        \includegraphics[width=\linewidth]{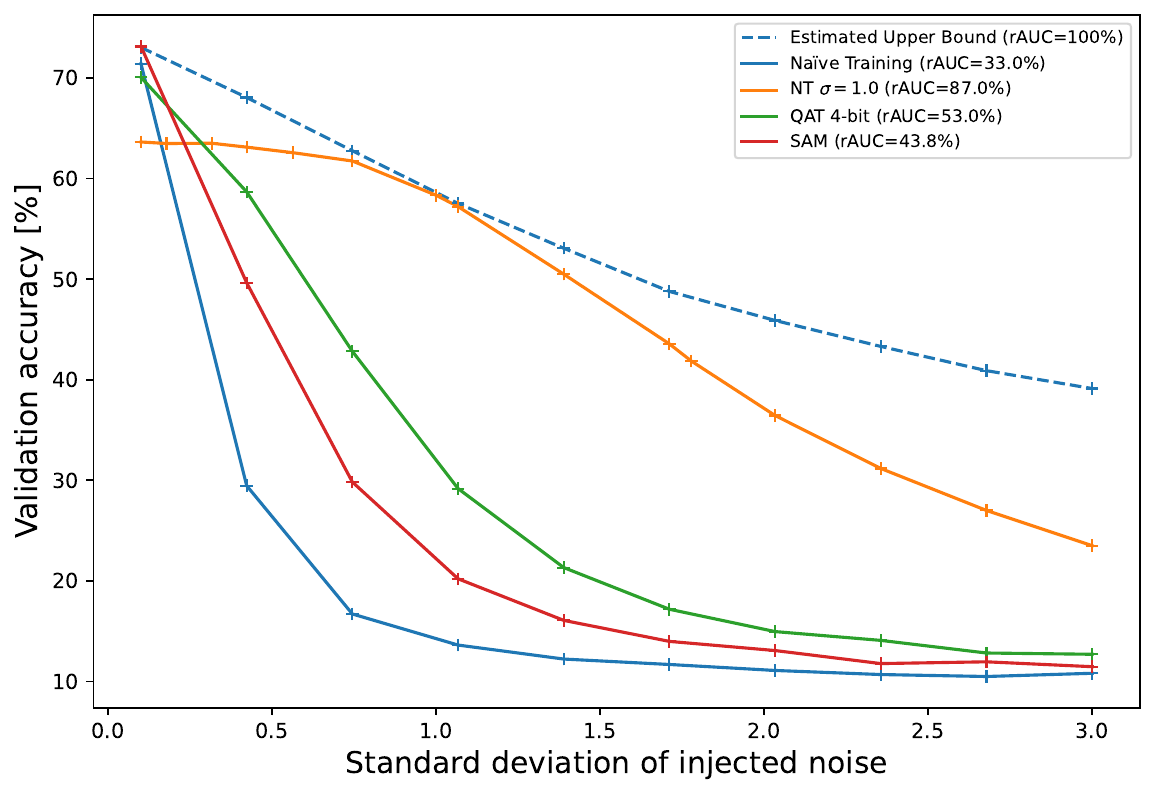}
        \subcaption{LeNet-5}\label{fig:vant:lenet-comp}
    \end{minipage}
    \begin{minipage}[t]{.495\textwidth}
        \centering
        \includegraphics[width=\linewidth]{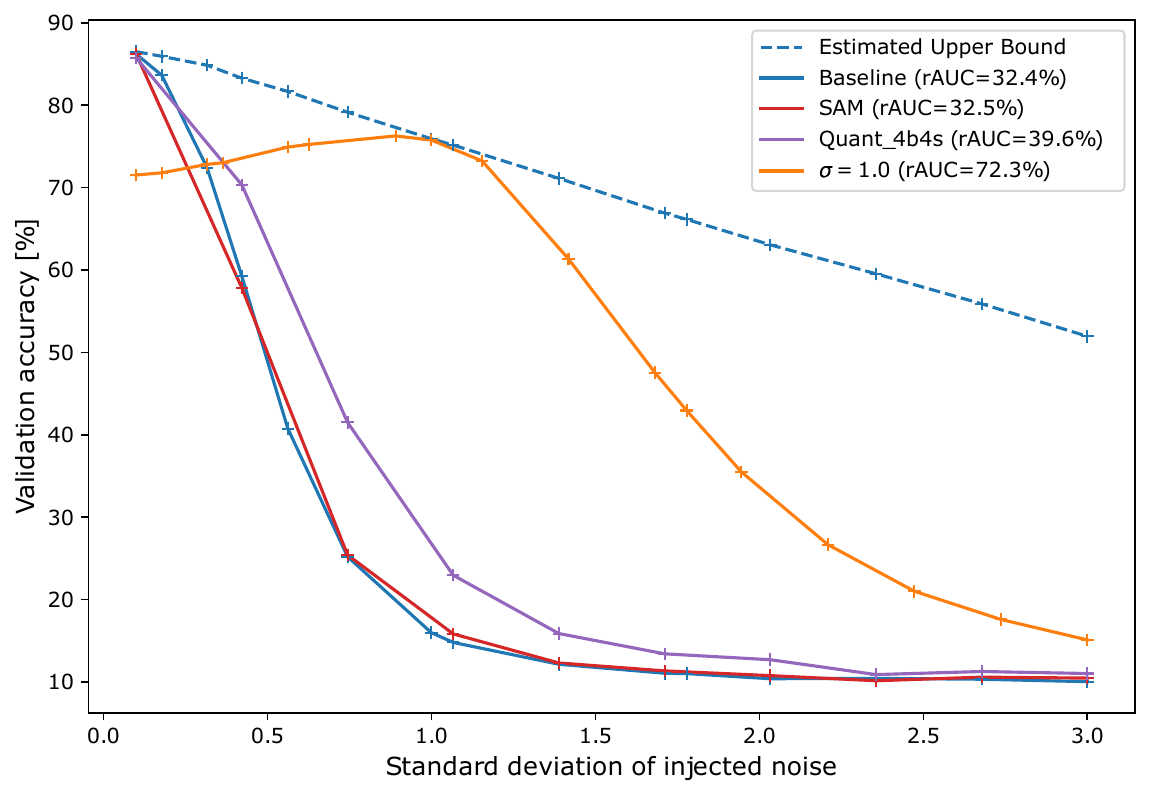}
        \subcaption{ResNet}\label{fig:vant:resnet-comp}
    \end{minipage}
    \hfill
    \begin{minipage}[t]{.495\textwidth}
        \centering
        \includegraphics[width=\linewidth]{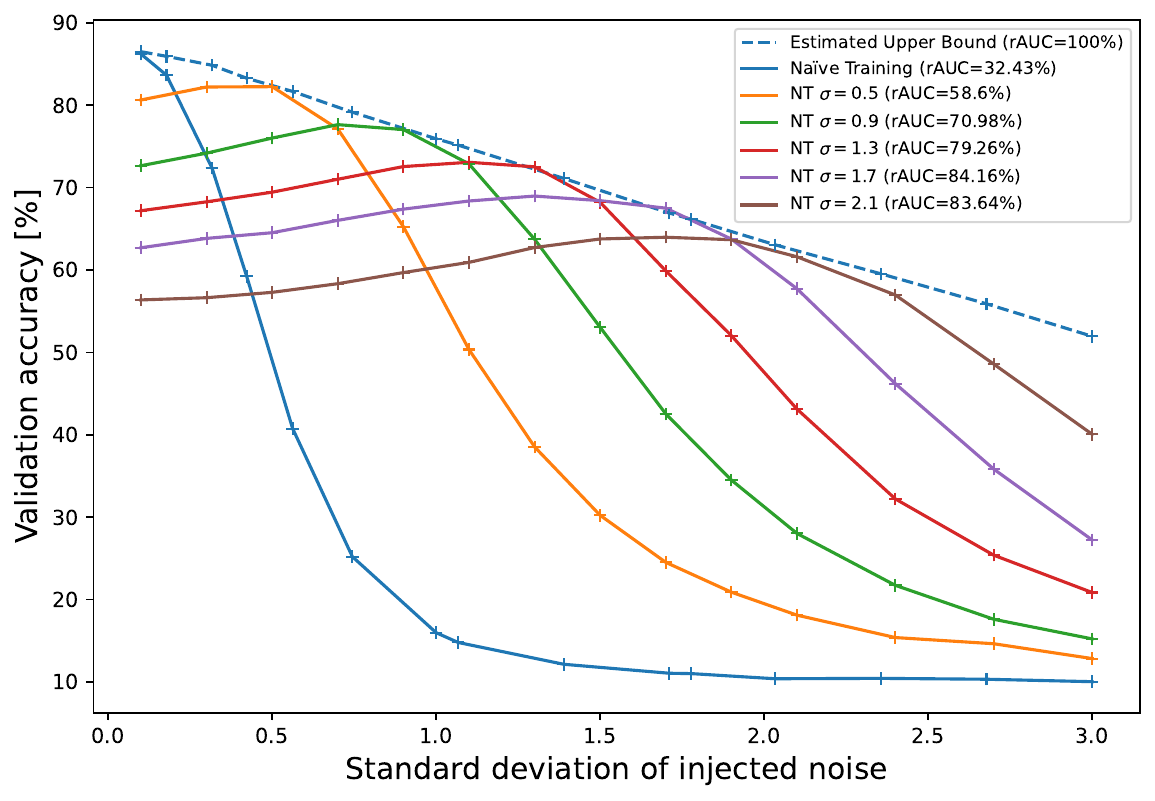}
		\subcaption{Noisy training with different training noise.}
		\label{fig:vant:resnet-basic}
    \end{minipage}
    \begin{minipage}[t]{.495\textwidth}
		\includegraphics[width=\textwidth]{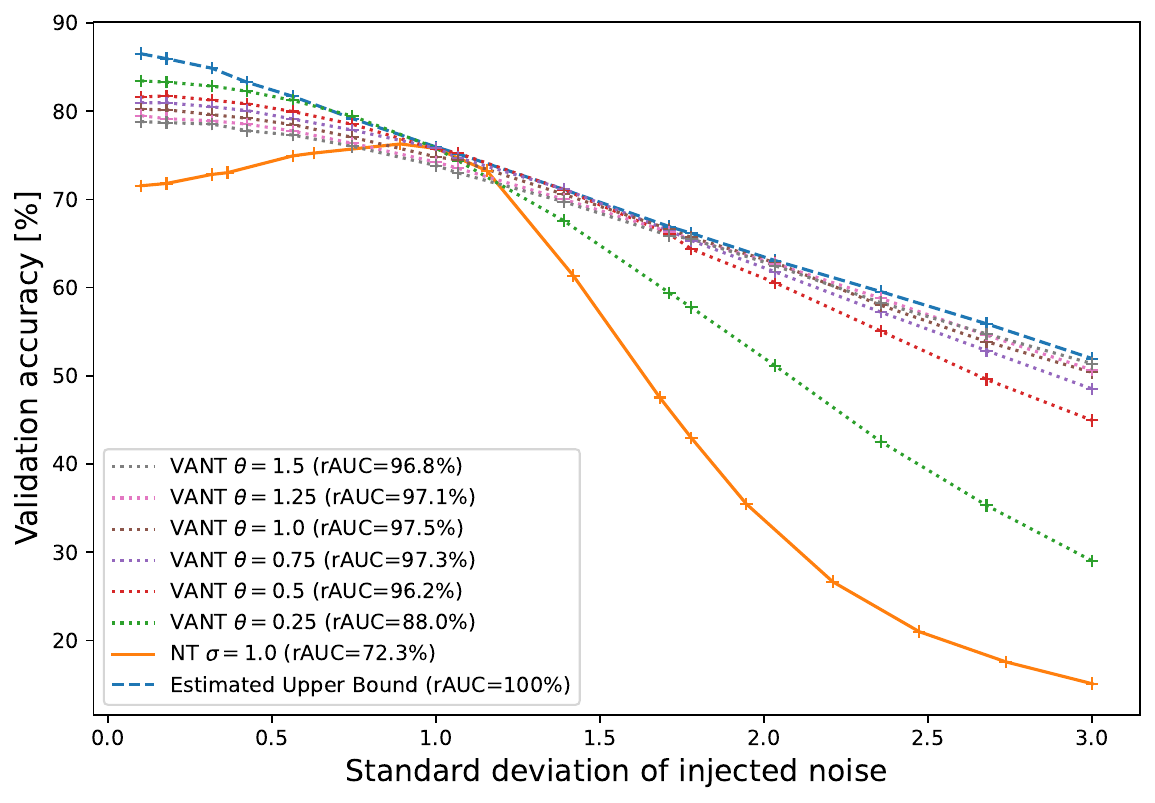}
        \subcaption{VANT with varying $\theta$ vs. noisy training}\label{fig:vant:cifar-10-theta-alpha-individual_a}
    \end{minipage}
	
	\caption{Comparison of hardening methods on CIFAR-10.  
	(\subref{fig:vant:lenet-comp}) LeNet-5 and (\subref{fig:vant:resnet-comp}) ResNet show that noisy training substantially outperforms quantization and \acrshort{sam}.  
	However, this robustness strongly depends on training and inference noise levels matching.  
	(\subref{fig:vant:resnet-basic}) ResNet under varying inference noise: each noisy training curve peaks where training and inference noise coincide, while the dashed line marks the upper bound with perfectly matched noise.  
    (\subref{fig:vant:cifar-10-theta-alpha-individual_a}) VANT with different variability levels $\theta$, demonstrating higher robustness across broader noise ranges compared to standard noisy training.  
	Adapted from~\cite{wang2025vant}.}
   \label{fig:vant:hardening-comparison}
\end{figure}

Previous results (Figs.~\ref{fig:vant:lenet-comp},~\ref{fig:vant:resnet-comp}) show that noisy training substantially outperforms both \acrshort{qat} and perturbation-based methods such as \acrlong{sam} (\acrshort{sam})~\cite{foret2021sharpness}, which improves generalization by guiding gradient descent towards flatter minima.  
They also reveal a critical limitation: robustness is only achieved when the noise level at inference closely matches the fixed noise level used during training.  
As soon as the two deviate, accuracy collapses rapidly (Fig.~\ref{fig:vant:resnet-basic}).  
This motivates \acrshort{vant}, which explicitly accounts for variability in noise strength over time and across devices.  

Standard noisy training injects Gaussian noise with a fixed standard deviation, thereby assuming that the accelerator’s noise is constant across devices and stable over time.  
In practice, however, analog accelerators exhibit fluctuations due to temperature, voltage, and device drift, such that fixed-noise training does not match inference conditions.  

To address this, \acrshort{vant} samples the injected noise level itself from a distribution:  
\begin{equation} \label{eq:vant}
    \begin{aligned}
        x &\sim \mathcal{N}\!\left(0, \sigma_\text{var}\right), \\
        \sigma_\text{var} &\sim \mathcal{N}\!\left(\alpha \cdot \sigma_\text{train}, \theta \right),
    \end{aligned}
\end{equation}
where $\sigma_\text{train}$ is the nominal noise level of the target hardware,  
$\alpha$ calibrates the sampled distribution to $\sigma_\text{train}$,  
and $\theta$ controls its variability.  
During training, $\sigma_\text{var}$ is sampled per input, and noise $x$ is injected additively into activations in the forward pass, exposing the network to a range of perturbation levels that mimic real hardware conditions.  

Robustness is quantified using the \acrfull{rauc}.  
Let $A(\sigma)$ denote the accuracy under noise level $\sigma$.  
We compute
\begin{equation}
    \acrshort{rauc} = \frac{\int A(\sigma)\, d\sigma}{\int A_\text{ideal}(\sigma)\, d\sigma},
\end{equation}
where the denominator represents the ideal case of perfectly matched training and inference noise.  
The \acrshort{rauc} is normalized between~0 and~1 and measures how closely a method approaches this upper bound of robustness.  

\subsection{Findings}

The comparison of hardening methods in Figs.~\ref{fig:vant:lenet-comp} and~\ref{fig:vant:resnet-comp} confirms that noisy training substantially outperforms both \acrshort{qat} and \acrshort{sam} in terms of robustness.  
However, Fig.~\ref{fig:vant:resnet-basic} illustrates a key limitation: each noisy training curve peaks precisely at the noise level used during training, and robustness collapses as soon as inference noise deviates.  
The dashed line shows the theoretical upper bound, corresponding to perfectly matched training and inference noise.  
This demonstrates that while noisy training is highly effective, it is also brittle under realistic conditions where noise is not static.  

\begin{figure}
	\centering
	\begin{subfigure}{0.49\textwidth}
		\includegraphics[width=\textwidth]{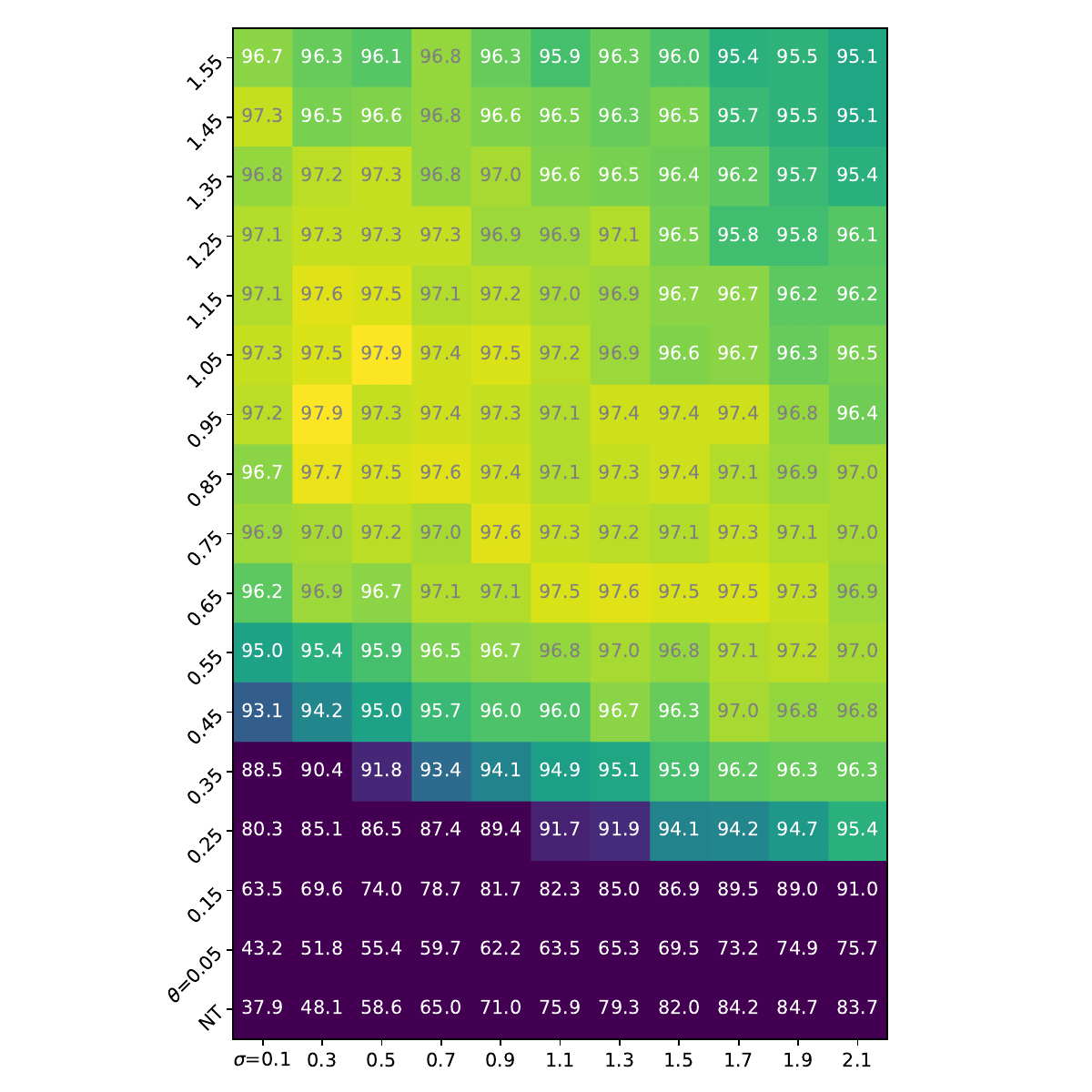}
		\caption{Robustness (rAUC \%)}
		\label{fig:vant:heatmap-resnet18-fix-alpha:a}
	\end{subfigure}
	\begin{subfigure}{0.49\textwidth}
		\includegraphics[width=\textwidth]{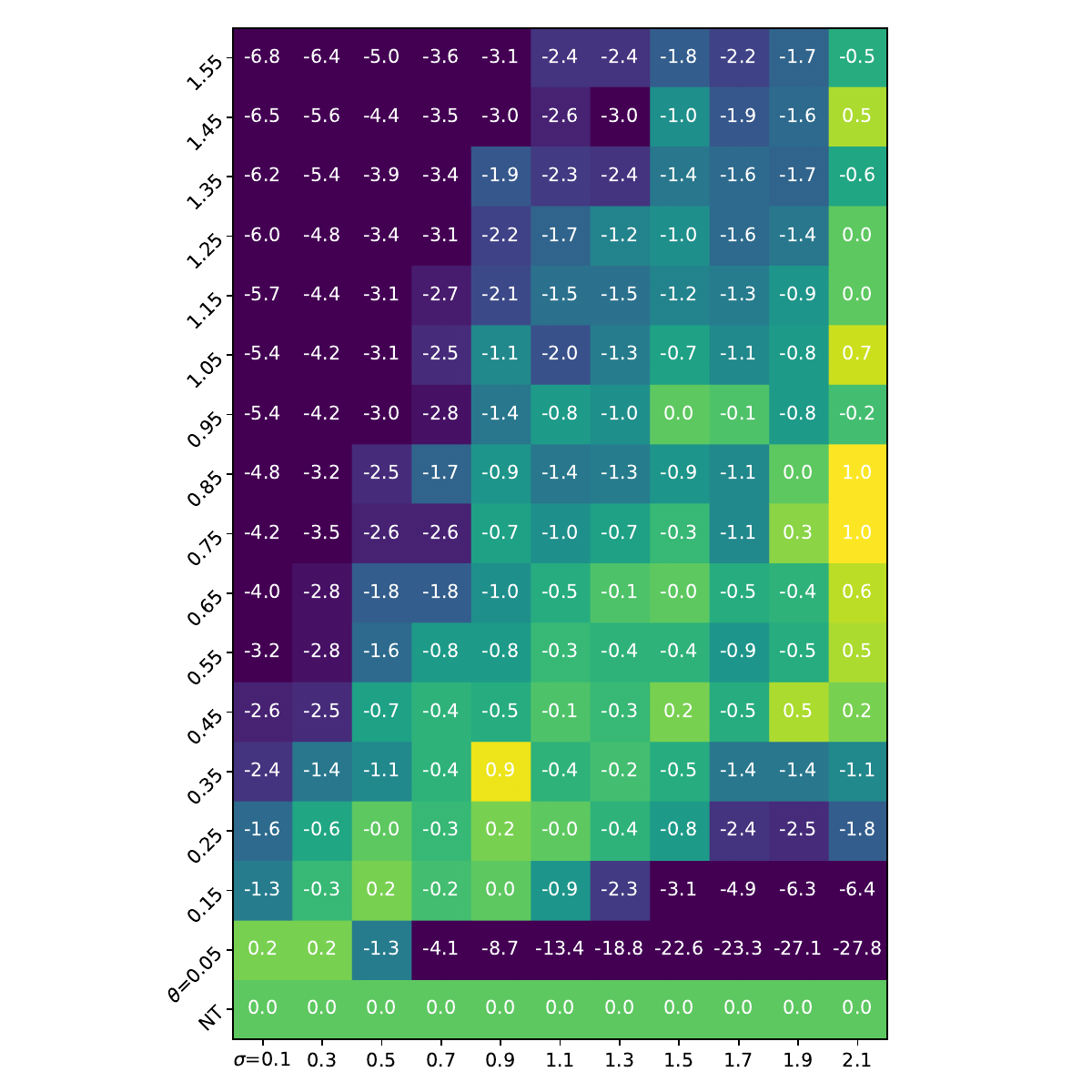}
		\caption{Preserved Accuracy}
		\label{fig:vant:heatmap-resnet18-fix-alpha:b}
	\end{subfigure}
	\caption{Heatmaps illustrating how the variability parameter $\theta$ influences robustness and accuracy of \acrshort{vant} on CIFAR-10 with ResNet-18.  
	(\subref{fig:vant:heatmap-resnet18-fix-alpha:a}) Robustness measured as rAUC increases with higher $\theta$.  
	(\subref{fig:vant:heatmap-resnet18-fix-alpha:b}) Preserved accuracy compared to standard noisy training decreases if variability becomes too large.  
	Together, these plots show how robustness and accuracy trade off under different noise environments and provide guidance for selecting suitable hyperparameters.  
	\reprofrom{wang2025vant}}
	\label{fig:vant:heatmap-resnet18-fix-alpha}
\end{figure}

\acrshort{vant} directly addresses this limitation.  
By sampling the training noise level from a distribution, the network is exposed to a range of perturbation strengths and learns representations that remain stable under variation.  
Figure~\ref{fig:vant:heatmap-resnet18-fix-alpha} illustrates how robustness depends on the choice of the variability parameter~$\theta$ and the baseline hardware noise level~$\sigma_\text{train}$.  
While larger values of $\theta$ generally increase robustness (rAUC), too much variability reduces preserved accuracy.  
An approximately linear relation $\theta \approx 0.4 \cdot \sigma_\text{train}$ was found to balance these effects, yielding robust yet accurate models.  
Figure~\ref{fig:vant:cifar-10-theta-alpha-individual_a} illustrates that while \acrshort{vant} remains sensitive to the variability parameter, it is significantly more robust across a broader range of noise levels than plain noisy training.  

\begin{figure}
	\centering
	\begin{subfigure}{0.49\textwidth}
		\includegraphics[width=\textwidth]{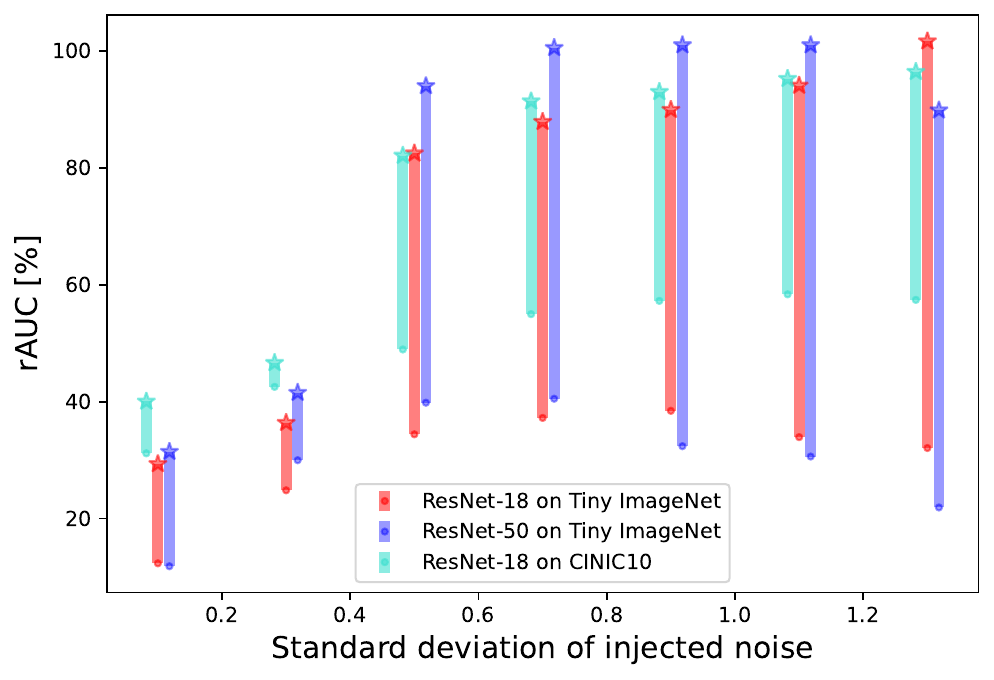}
		\caption{Robustness (rAUC)}
		\label{fig:vant:cinic-tinyImageNet_a}
	\end{subfigure}
	\begin{subfigure}{0.49\textwidth}
		\includegraphics[width=\textwidth]{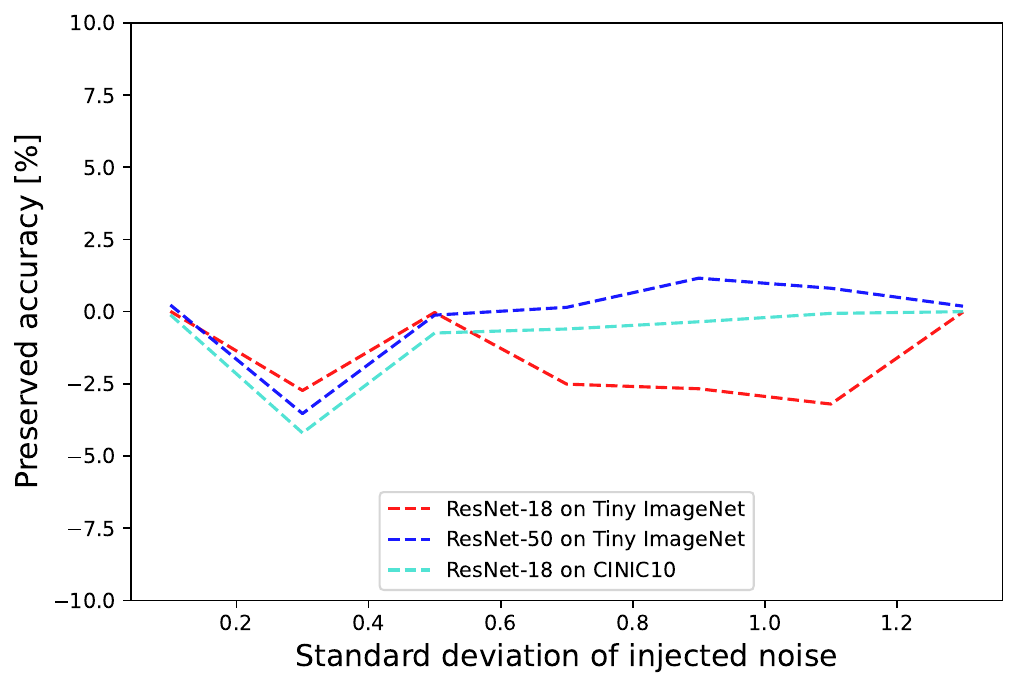}
		\caption{Preserved Accuracy}
		\label{fig:vant:cinic-tinyImageNet_b}
	\end{subfigure}
	\caption{Generalization of \acrshort{vant} to more complex datasets and deeper architectures.  
	(\subref{fig:vant:cinic-tinyImageNet_a}) Robustness (rAUC) improves consistently across CINIC-10 and Tiny ImageNet for both ResNet-18 and ResNet-50.  
	(\subref{fig:vant:cinic-tinyImageNet_b}) Preserved accuracy remains close to standard noisy training, with only minor deviations.  
	These results demonstrate that \acrshort{vant} reliably increases robustness across tasks and model scales, while incurring little accuracy cost.  
	\reprofrom{wang2025vant}}
	\label{fig:vant:cinic-tinyImageNet}
\end{figure}

The benefits of \acrshort{vant} generalize across datasets and architectures.  
As shown in Fig.~\ref{fig:vant:cinic-tinyImageNet}, it enhances robustness on more complex datasets such as CINIC-10 and Tiny ImageNet, and scales effectively to deeper architectures like ResNet-50.  
Crucially, \acrshort{vant} sustains high robustness even under strong and varying noise levels, where standard noisy training rapidly deteriorates.  

In conclusion, \acrshort{vant} overcomes the brittleness of standard noisy training by explicitly modeling noise variability.  
It provides stable robustness across datasets, architectures, and operating conditions, marking an important step toward sustainable deployment of analog accelerators.  
While already highly effective, further refinements may extend its applicability even more broadly.

\section*{Summary}

This chapter developed a progressive view on robustness in neural networks for analog accelerators, moving from diagnosis to hardening, and finally to methods that address real-world variability.  

First, \emph{Walking Noise} showed that robustness is not a uniform property but emerges through heterogeneous, layer-specific mechanisms such as weight scaling under additive perturbations and self-binarization under multiplicative noise.  
This diagnostic perspective highlighted the opportunity of targeted countermeasures rather than uniform approaches.  

Second, we compared systematic hardening strategies against noisy computations.  
While \acrshort{qat} can trade accuracy for tolerance, noisy training proved to be the most effective method for achieving robustness under static noise conditions, outperforming both \acrshort{qat} and perturbation-based baselines such as \acrshort{sam}.  

Finally, we introduced \acrshort{vant}, a training strategy that extends noisy training by explicitly modeling variability in noise levels.  
Instead of assuming a fixed noise distribution, it samples noise strengths during training, exposing the network to the fluctuations characteristic of analog hardware.  
This variance-aware formulation makes robustness an inherent property of the learned representations, rather than a brittle consequence of matched training and inference conditions.  

Taken together, these contributions trace a coherent trajectory: robustness can be understood at the layer level, strengthened with effective hardening, and sustained by accounting for the variability inherent to analog hardware.  
This progression lays the foundation for practical deployment of neural networks on analog accelerators, and points toward future refinements such as adapting methods to further device imperfections and extending beyond simple Gaussian noise to richer distributions and multiple injection points along the computational chain.  
At the same time, the results highlight that even simple noise models can be remarkably effective, suggesting that tailored yet computationally inexpensive hardening strategies remain a promising direction.

  \clearpage{}%
\clearpage{}%

\part{Accelerating Bayesian Neural Networks}
\label{pt:BNNs}

\chapter{Bayesian Neural Networks}
\label{ch:bnns}

\epigraph{Uncertainty is the only certainty there is, and knowing how to live with insecurity is the only security.}{\textnormal{--- John Allen Paulos}}

\noindent
Uncertainty lies at the heart of real-world machine learning.
Sensors produce noisy signals, environments evolve unpredictably, and training data can never fully capture the diversity of the world.  
When neural networks are deployed in safety-critical domains such as autonomous driving, medical diagnostics, or embedded control, it is therefore not sufficient to provide single deterministic outputs.
Instead, models must also provide a measure of confidence in their outputs, enabling downstream systems to react appropriately: they may defer to a human operator, trigger a safe fallback mechanism, or request additional information when predictions are unreliable.

Conventional deep neural networks, despite their outstanding predictive accuracy, are poorly equipped to serve this role.  
Outputs are typically interpreted through the softmax function, which is often miscalibrated and tends to produce overconfident predictions even on inputs far outside the training distribution~\cite{guo2017calibration}.  
This structural inability to quantify uncertainty has become a central obstacle to the trustworthy deployment of deep learning~\cite{gawlikowski2021survey}.  
A common failure mode is that networks issue highly confident but wrong predictions on out-of-distribution inputs.  
In such cases, models lack the ability to express \emph{I do not know}, a capability that is crucial for robust decision-making.

A principled approach to modeling uncertainty is provided by \acrfullpl{bnn}.  
By replacing deterministic weights with probability distributions, \acrshortpl{bnn} extend classical \acrlongpl{nn} into a probabilistic framework~\cite{mackay1992practical,neal1996bayesian}.  
Predictions thus become distributions rather than fixed values, capturing both data variability and model uncertainty.  
This perspective was consolidated by \citeauthor{jospin2022handsonBNN}~\cite{jospin2022handsonBNN}, who provide a comprehensive overview of Bayesian inference strategies and a unified workflow for \acrshortpl{bnn}.  

The theoretical foundation of \acrshortpl{bnn} is mathematically rigorous, but exact Bayesian inference over high-dimensional weight distributions is intractable.  
Approximate approaches such as sampling-based or variational inference methods require multiple posterior samples and forward passes, resulting in high runtime and memory demands~\cite{neal1996bayesian,murphy2023probabilisticML}.
Consequently, while \acrshortpl{bnn} provide the most principled approach to uncertainty estimation, they are also among the most resource-intensive models to train and deploy.

Within the structure of this work, this chapter marks the transition from deterministic computation to probabilistic machine learning.  
It introduces the foundations of uncertainty quantification, surveys the main Bayesian inference strategies for \acrshortpl{bnn}, and discusses their computational challenges.  
The overarching message is clear: reliable uncertainty estimation is not an auxiliary feature but a fundamental capability, and \acrshortpl{bnn} provide the mathematical basis upon which this capability can be built.  
Later chapters build on these foundations: first by deploying the \acrlong{pfp} on embedded systems to demonstrate how closed-form propagation of an extreme \acrshort{svi} approximation can replace costly sampling in \acrshort{bnn} inference, then by introducing ensemble-based methods as practical Bayesian approximations, and finally by exploring photonic accelerators where hardware noise itself becomes a controllable source of stochasticity.  
\section{Quantifying Uncertainty}

Uncertainty quantification provides the foundation for probabilistic modeling.  
It captures not only how confident a model is in its predictions, but also why predictions may be uncertain.  
In practice, two principal forms of uncertainty are distinguished~\cite{kiureghian2009aleatoricepisetmic,kendall2017uncertainties,jospin2022handsonBNN,murphy2023probabilisticML}.  

\paragraph{Aleatoric uncertainty.}  
Aleatoric uncertainty refers to the inherent randomness of the data-generating process.  
It arises from sources such as sensor noise, ambiguous labels, or intrinsic variability of physical systems.  
Since it is tied to the data itself, it cannot be reduced by collecting additional observations.  

\paragraph{Epistemic uncertainty.}  
Epistemic uncertainty corresponds to modeling error: it reflects incomplete knowledge about the data-generating process.  
This includes insufficient or unrepresentative training data, structural misspecification of the model class, or imperfections in the training procedure.  
Formally, it is represented by the posterior distribution $p(\theta \mid D)$, which quantifies the range of plausible parameter values given the data.  
Unlike aleatoric uncertainty, epistemic uncertainty can be reduced by collecting more representative data or by improving the model.  
It is particularly pronounced for inputs that lie outside the training distribution, making it crucial for out-of-distribution detection and active learning.  

\paragraph{Metrics.}  
In \acrshortpl{bnn} for regression tasks, aleatoric uncertainty is often modeled explicitly using an \emph{aleatoric head}.  
Following \citeauthor{kendall2017uncertainties}~\cite{kendall2017uncertainties}, each regression target is represented by two outputs in the final layer: one predicts the mean $\mu(x)$ and the other the variance $\sigma^2(x)$, which models input-dependent Gaussian noise.  
This heteroscedastic Gaussian formulation jointly learns predictive means and variances, enabling the model to capture heterogeneous noise levels across the input domain.  
For numerical stability, it is common to predict $\log \sigma^2(x)$ instead of $\sigma^2(x)$ directly.  
Such aleatoric heads are specific to regression, since classification outputs are categorical and their uncertainty is naturally expressed through the distribution over class probabilities rather than explicit variance terms.  

In classification, uncertainty is commonly quantified using information-theoretic metrics.  
Let $x$ denote an input with label $y \in \{1, \dots, C\}$, and let the predictive distribution be defined by marginalizing over the posterior,
\begin{equation}
    p(y \mid x, D) = \int p(y \mid x, \theta) \, p(\theta \mid D) \, d\theta.
\end{equation}
The total predictive uncertainty is measured by the Shannon entropy of this distribution~\cite{shannon1948entropy},
\begin{equation}
    H[y \mid x, D] = - \sum_{c=1}^{C} p(y=c \mid x, D) \, \log p(y=c \mid x, D).
\end{equation}
This quantity can be decomposed into contributions from aleatoric and epistemic sources~\cite{depeweg2018,huellermeier2021uncertainty}.  
The expected data uncertainty, also known as \acrfull{sme}, isolates the aleatoric part,
\begin{equation}
    \mathbb{E}_{p(\theta \mid D)} \Big[ H[y \mid x, \theta] \Big]
    = - \frac{1}{N} \sum_{n=1}^{N} \sum_{c=1}^{C} p(y=c \mid x, \theta_n) \, \log p(y=c \mid x, \theta_n),
\end{equation}
while the mutual information (MI) between predictions and parameters captures the epistemic part,
\begin{equation}
    I[y, \theta \mid x, D] = H[y \mid x, D] - \mathbb{E}_{p(\theta \mid D)} \Big[ H[y \mid x, \theta] \Big].
\end{equation}
It should be noted, however, that this decomposition is approximation-dependent and its exact interpretation remains debated in the community~\cite{huellermeier2021uncertainty,jimenez2025failepisetmic}.

Beyond accuracy and uncertainty decomposition, the \emph{calibration} of predictive probabilities is a key indicator of reliability.  
A well-calibrated model outputs confidence values that correspond to the true likelihood of correctness, ensuring trustworthy uncertainty estimates.  
To quantify calibration, two metrics are widely used.  

The \acrfull{nll} evaluates how well predicted probabilities align with observed labels,
\begin{equation}
    \text{NLL} = - \frac{1}{N} \sum_{n=1}^{N} \log p(y_n \mid x_n, D),
\end{equation}
with lower values indicating better calibration.  
The \acrfull{ece} measures the gap between predicted confidence and empirical accuracy by binning predictions~\cite{guo2017calibration},
\begin{equation}
    \text{ECE} = \sum_{m=1}^{M} \frac{|B_m|}{N} \; \big| \; \text{acc}(B_m) - \text{conf}(B_m) \; \big|,
\end{equation}
where $B_m$ is the set of samples in bin $m$, $\text{acc}(B_m)$ is their accuracy, and $\text{conf}(B_m)$ their average confidence.  

The \acrfull{auroc} is used to quantify separation between in-domain and \acrlong{ood} data.  
It integrates the \acrfull{tpr} over the \acrfull{fpr} across thresholds,  
\[
  \text{AUROC} = \int_0^1 \text{TPR}(\text{FPR}^{-1}(x)) \, dx.  
\]  
An \acrshort{auroc} of $1.0$ indicates perfect separation, while $0.5$ corresponds to random guessing~\cite{fawcett2006roc}.

Together, these metrics provide a principled way to assess both the decomposition of predictive uncertainty and the reliability of probability estimates.  
They also illustrate why probabilistic approaches are attractive: they offer a unified way to model uncertainty, distinguish its sources, and evaluate prediction quality.  
To realize these metrics in the context of neural networks, \acrshortpl{bnn} provide the natural mathematical framework, as they extend deterministic models into a probabilistic setting where both aleatoric and epistemic uncertainty can be quantified in a principled manner.  
\section{Bayesian Neural Networks: Foundations}

\Acrlongpl{bnn} extend classical neural networks by placing probability distributions over their parameters.  
Instead of learning a single point estimate $\theta^\ast$ for the weights, a \acrshort{bnn} treats weights as random variables and infers a posterior distribution given observed data $D$~\cite{neal1996bayesian}.  
By Bayes' rule~\cite{murphy2023probabilisticML}, the posterior is expressed as
\begin{equation}
    p(\theta \mid D) = \frac{p(D \mid \theta) \, p(\theta)}{p(D)},
\end{equation}
where $p(\theta)$ is a prior distribution, $p(D \mid \theta)$ the likelihood, and $p(D)$ the marginal likelihood or evidence.\footnote{The rule is named after Reverend Thomas Bayes, whose posthumous essay~\cite{bayes1763essay} first described the principle.}  
The prior reflects assumptions or domain knowledge about the weights, while the posterior captures updated beliefs after observing data~\cite{murphy2012probabilisticPerspective,jospin2022handsonBNN}.  
The evidence $p(D)$ serves as a normalizing constant but is intractable to compute for neural networks, since it requires integrating over all possible parameter configurations.  

Predictions for a new input $x$ are obtained by marginalizing over the posterior,
\begin{equation}
    p(y \mid x, D) = \int p(y \mid x, \theta) \, p(\theta \mid D) \, d\theta,
\end{equation}
which defines the posterior predictive distribution.  
This connects directly to the uncertainty measures introduced in the previous section: predictive entropy reflects total uncertainty, while its decomposition into aleatoric and epistemic parts depends on the likelihood and posterior, respectively.  

The Bayesian perspective contrasts with the frequentist view.  
In frequentist inference, parameters $\theta$ are fixed but unknown, and learning corresponds to estimating them via maximum likelihood or related criteria.  
Bayesian inference instead treats parameters as random variables with a prior distribution, and learning corresponds to updating this belief in light of data.  
This perspective yields predictive distributions that naturally quantify both aleatoric and epistemic uncertainty~\cite{murphy2023probabilisticML,jospin2022handsonBNN}.  

The posterior $p(\theta \mid D)$ involves high-dimensional integrals that lack closed form and are costly to approximate reliably, 
necessitating the use of approximate inference methods in practical \acrshortpl{bnn}.
Sampling-based approaches, such as Markov chain Monte Carlo, provide asymptotically exact posterior draws but scale poorly with model size.  
Optimization-based approaches, such as variational inference, trade statistical fidelity for computational scalability, typically relying on tractable Gaussian approximations.
More expressive variational families or advanced samplers can improve posterior fidelity but at substantially higher computational cost.

\begin{figure}[t]
    \centering
    \includegraphics[width=\linewidth]{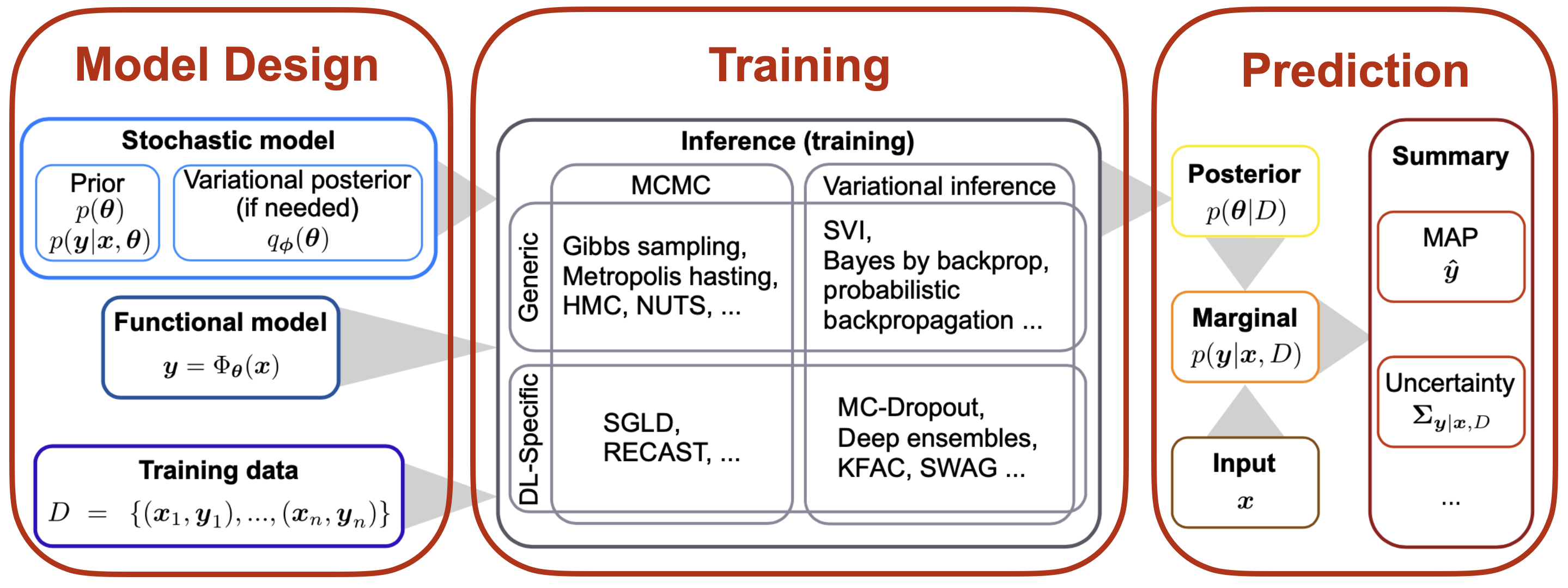}
	\caption[Workflow of a Bayesian neural network]{Workflow of a \acrshort{bnn}.  
	Design specifies both the architecture and prior distributions.  
	Training applies Bayesian inference to approximate the posterior. 
	Prediction marginalizes the posterior to form the predictive distribution, from which point estimates and uncertainty metrics can be derived.~\adjfrom{jospin2022handsonBNN}}
    \label{fig:bnn_workflow}
\end{figure}

The overall process of \acrshort{bnn} training and inference is illustrated in Figure~\ref{fig:bnn_workflow}, following the perspective of \citeauthor{jospin2022handsonBNN}~\cite{jospin2022handsonBNN}.
It consists of three stages:  
\begin{enumerate}[label=(\roman*), leftmargin=2em]
    \item \textbf{Design:} the neural architecture and prior distributions are specified.  
    \item \textbf{Training:} Bayesian inference techniques approximate the posterior distribution.  
    \item \textbf{Prediction:} the posterior is marginalized to form the posterior predictive distribution $p(y \mid x, D)$.  
\end{enumerate}
To obtain uncertainty estimates, multiple forward passes are performed by sampling from the posterior, yielding a distribution of predictions from which metrics such as variance, \acrshort{sme}, \acrshort{mi}, or predictive Shannon entropy can be derived.

\Acrshortpl{bnn} thus provide a principled probabilistic extension of neural networks by treating parameters as random variables.  
They yield posterior predictive distributions that naturally capture both aleatoric and epistemic uncertainty.  
While conceptually elegant, their exact inference is computationally infeasible for modern architectures, motivating the approximate methods discussed in the next section.  

\section{Bayesian Inference}
\label{sec:bnns:bi}

\subsection{\acrfull{mcmc}}
\Acrlong{mcmc} methods provide the classical foundation of Bayesian inference and remain the gold standard for drawing asymptotically exact samples from complex posteriors~\cite{brooks2011handbook}.  
They construct a Markov chain whose stationary distribution is the posterior $p(\theta \mid D)$ and generate samples $\{\theta_n\}_{n=1}^N$ from this chain, from which posterior expectations can be approximated by averaging.  
This makes \acrshort{mcmc} a conceptually simple yet broadly applicable tool, albeit one whose computational cost typically increases rapidly with dimensionality.

The most widely known instance is the Metropolis algorithm~\cite{metropolis1953equation} and its generalization by Hastings~\cite{hastings1970monte}.  
The key idea of Metropolis--Hastings is to construct a proposal distribution $q(\theta' \mid \theta)$ for candidate moves and to accept a proposal with probability
\begin{equation}
    \alpha(\theta, \theta') = \min\!\left(1, \; \frac{p(\theta' \mid D)\, q(\theta \mid \theta')}{p(\theta \mid D)\, q(\theta' \mid \theta)} \right).
\end{equation}
This accept--reject step, together with the Hastings correction for asymmetric proposals, enforces detailed balance and guarantees that the posterior $p(\theta \mid D)$ is the stationary distribution of the chain.  
The defining property of Metropolis--Hastings is that it only requires ratios of posterior densities, which ensures that the intractable evidence $p(D)$ cancels out.  
An acceptance rate strictly below one prevents the chain from collapsing onto the maximum a posteriori mode and ensures that the full posterior distribution, rather than only its peak, is represented.

Running \acrshort{mcmc} in practice requires careful handling of initialization and convergence.  
The initial part of each chain, known as the burn-in or warm-up phase, is discarded because early samples are still influenced by the arbitrary starting point rather than the stationary distribution~\cite{gelman2013bayesian}.  
During warm-up, proposal distributions or other tuning parameters are often adapted to improve efficiency, after which sampling proceeds with fixed settings to ensure validity.  
Convergence and sample quality are then assessed using standard diagnostics: autocorrelation functions and the \acrfull{ess} quantify the amount of independent information contained in a chain~\cite{geyer1992practical}, while the potential scale reduction statistic $\hat{R}$ compares within-chain to between-chain variance to detect lack of convergence~\cite{gelman1992inference,vehtari2021ranknormal}.  

Despite its generality, classical Metropolis--Hastings suffers from poor mixing in high-dimensional parameter spaces.  
For models such as \acrshortpl{bnn}, naive proposals lead to low acceptance rates and highly correlated samples, which results in prohibitively slow convergence.  
To address this limitation, gradient-informed methods exploit local geometry.  
\Acrfull{hmc}, originally introduced as Hybrid Monte Carlo~\cite{duane1987hybrid} and popularized in machine learning by \citeauthor{neal2011mcmc}~\cite{neal2011mcmc}, is inspired by classical physics and augments the parameter space $\theta$ with auxiliary momentum variables $r$.  
It defines a Hamiltonian consisting of potential and kinetic energy,
\begin{equation}
    \mathcal{H}(\theta,r) \;=\; U(\theta) + K(r) \;=\; - \log p(\theta \mid D) + \tfrac{1}{2} r^\top M^{-1} r \, ,
\end{equation}
where $U(\theta)$ is the potential energy given by the negative log posterior and $K(r)$ is the kinetic energy with mass matrix $M$.  
In this Bayesian interpretation, the posterior landscape acts as the potential energy surface, and by sampling random momenta $r$, the algorithm follows approximate Hamiltonian trajectories that traverse this landscape.  
Proposals are generated by simulating these dynamics with the leapfrog method using step size $\varepsilon$ and trajectory length $L$.  
Since integration is approximate, a Metropolis accept--reject step at the end of each trajectory corrects for numerical error and ensures that the posterior remains the stationary distribution.  
This accept--reject step is essential for validity and is what distinguishes \acrshort{hmc} from mere gradient-based optimization dynamics.  
The resulting proposals are long and directed, suppress random-walk behavior, and substantially improve mixing in continuous, high-dimensional spaces.  

The No-U-Turn Sampler (\acrshort{nuts}) addresses the need to set the trajectory length $L$ by dynamically expanding paths until a U-turn criterion is met and simultaneously adapting the step size $\varepsilon$ during warm-up using dual averaging~\cite{hoffman2014nuts}.  
This removes the need for manual tuning and makes \acrshort{nuts} the default choice in many modern probabilistic programming frameworks.  

\paragraph{Practical advice.}  
In the context of \acrshortpl{bnn}, manual tuning of \acrshort{hmc} parameters is particularly challenging.  
Selecting a suitable trajectory length $L$ and step size $\varepsilon$ is often infeasible in practice: inappropriate choices can either trap the chain in highly correlated states or cause excessive rejections, and the sensitivity grows with network size.  
\acrshort{nuts} has proven to be a real game changer in this setting.  
By adapting path lengths and step sizes automatically during warm-up, it makes \acrshort{hmc} one of the easiest inference methods to apply reliably, despite its high computational cost.  
In practice, we found that \acrshort{nuts} consistently yields stable results for \acrshortpl{bnn} without the extensive parameter tuning otherwise required.  

\paragraph{Limitations.}  
Nevertheless, \acrshort{mcmc} remains challenging for large, modern \acrlong{nn} architectures.  
Each proposal in \acrshort{mcmc} requires at least one evaluation of the model likelihood or its gradient with respect to all network parameters.  
For \acrshort{hmc}, this cost scales with the trajectory length $L$, since each trajectory requires $L$ gradient evaluations.  
\Acrshort{nuts} can be even more expensive, as it explores trajectories by recursively building a binary tree until a U-turn condition is met, leading to up to $2L-1$ gradient evaluations per iteration~\cite{hoffman2014nuts}.  
Multiple chains, warm-up, and the need for many effectively independent samples further amplify both compute and memory demands.  
As a result, \acrshort{mcmc} is mainly applicable to small-scale models, where it serves both as a practical tool and as a rigorous reference for evaluating approximate inference methods.  
Its prohibitive computational and memory cost in modern architectures contrasts with the more scalable approximations discussed next.

\subsection{Variational Inference (VI)}

While \acrshort{mcmc} provides asymptotically exact posterior samples, its computational demands scale poorly with model size.  
\Acrlong{vi} offers a scalable alternative by recasting Bayesian inference as an optimization problem.  
The idea is to introduce a family of tractable distributions $q_\phi(\theta)$, parameterized by variational parameters $\phi$, to approximate the intractable posterior $p(\theta \mid D)$~\cite{jordan1999vi,blei2017variational,goodfellow2016deep}.  
A seminal application to neural networks is \emph{Bayes by Backprop}~\cite{blundell2015}, which demonstrated variational learning of Gaussian weight distributions and laid the foundation for scalable approximate Bayesian deep learning.  
Instead of drawing samples directly from the posterior, the goal is to find the member of this family that is closest to the true posterior according to a divergence measure.

The most common choice is the \acrfull{kl}, which for two distributions $q(\theta)$ and $p(\theta)$ is defined as~\cite{kullback1951kl,goodfellow2016deep}
\begin{equation}
    \text{KL}(q(\theta) \;\|\; p(\theta)) \;=\; \mathbb{E}_{q(\theta)} \left[ \log \frac{q(\theta)}{p(\theta)} \right] 
    \;=\; \int q(\theta) \, \log \frac{q(\theta)}{p(\theta)} \, d\theta.
\end{equation}
Although asymmetric, the KL divergence is non-negative and equals zero if and only if $q(\theta) = p(\theta)$ almost everywhere.  

Directly minimizing $\text{KL}\!\left(q_\phi(\theta) \,\|\, p(\theta \mid D)\right)$ is infeasible, since the true posterior $p(\theta \mid D)$ contains the intractable evidence term $p(D)$.  
Instead, the problem is reformulated in terms of the \emph{\acrfull{elbo}}, which is a tractable lower bound on the marginal log-likelihood $\log p(D)$.  
Maximizing the \acrshort{elbo} is equivalent to minimizing the KL divergence, but avoids the need to compute $p(D)$ explicitly~\cite{jordan1999vi,blei2017variational,goodfellow2016deep}:  
\begin{equation}
    \mathcal{L}(\phi) \;=\; \mathbb{E}_{q_\phi(\theta)} \big[ \log p(D \mid \theta) \big] - \text{KL}\!\big(q_\phi(\theta) \,\|\, p(\theta)\big).
\end{equation}
The first term encourages data fit, while the second term regularizes the approximation towards the prior.  

Optimization of the \acrshort{elbo} can be carried out with stochastic gradient methods.  
In this setting, \acrfull{svi} employs mini-batches of data and stochastic gradient estimates to scale variational inference to large datasets~\cite{hoffman2013svi}.  
The reparameterization trick~\cite{kingma2013auto,rezende2014stochastic}, first applied to Bayesian neural networks in \emph{Bayes by Backprop}~\cite{blundell2015}, enables low-variance gradient estimates by rewriting stochastic sampling as a deterministic transformation of parameters and auxiliary noise.  
For instance, if $q_\phi(\theta) = \mathcal{N}(\theta \mid \mu, \sigma^2)$, then sampling can be expressed as
\begin{equation}
    \theta = \mu + \sigma \, \epsilon, \qquad \epsilon \sim \mathcal{N}(0,1),
\end{equation}
which disentangles the randomness $\epsilon$ from the variational parameters $(\mu, \sigma)$ and allows gradients to propagate through them efficiently.  
Intuitively, this transformation lets gradients bypass the stochastic sampling step and flow into the variational parameters, making optimization with backpropagation feasible.  

The expressiveness of the variational family $q_\phi(\theta)$ critically determines the quality of the approximation.  
Mean-field Gaussians, which assume independence across parameters, offer an attractive compromise: they are simple, scale efficiently to large models, and often yield sufficiently good uncertainty estimates in practice~\cite{blei2017variational,goodfellow2016deep}.  
Despite their tendency to underestimate posterior uncertainty, they remain widely used due to their efficiency and ease of implementation.  

More expressive families can be employed when higher fidelity is required.  
Matrix-variate Gaussians, for example, introduce correlations across rows and columns of weight matrices, which is particularly natural for convolutional filters where nearby weights tend to co-vary.  
Normalizing flows instead transform simple base distributions through sequences of invertible mappings, allowing the capture of multimodal or skewed posteriors~\cite{rezende2015flows,murphy2023probabilisticML}.  
These richer families yield more faithful approximations, but at the cost of additional parameters and computational overhead, since each transformation must be optimized alongside the base network.

\paragraph{Practical advice.}  
\acrshort{svi}-trained \acrshortpl{bnn} are highly sensitive to hyperparameter choices, and poor configurations often lead to degenerate solutions.  
If the posterior variance is initialized too high, the network quickly learns that predicting ``I do not know'' is a valid strategy and may never recover to produce meaningful outputs.  
Several strategies have proven effective in practice to mitigate this issue:

\begin{itemize}
    \item \textbf{Warm-starting the means.} Initializing the posterior means from a pretrained deterministic model provides a strong starting point for predictions.  
    \item \textbf{Underestimating the prior variance.} Explicitly initializing prior variances significantly smaller than the means biases the model toward confident predictions in early epochs, preventing collapse into trivial uncertainty.  
	\item \textbf{Balancing KL and data fit.}  
	In practice, the KL term in the \acrshort{elbo} must be down-weighted relative to the likelihood.  
	Too much weight causes posterior collapse into high uncertainty, while too little leads to overfitting and ignoring the prior.  
	Finding this balance is among the most sensitive hyperparameters in \acrshort{svi}.
    \item \textbf{KL annealing.} Replacing the fixed balance with a dynamic schedule, where the KL weight is gradually increased during training, has proven particularly effective. Linear annealing is a common choice. While this approach reduces sensitivity to initialization, the final balance between KL and likelihood terms still requires hyperparameter tuning.  
\end{itemize}

Together, these strategies stabilize optimization and substantially improve the reliability of \acrshort{svi}-trained \acrshortpl{bnn}, though at the cost of increased hyperparameter search.

Compared to sampling-based methods, \acrshort{vi} trades asymptotic exactness for computational scalability.  
It enables deterministic optimization procedures with fast convergence and amortized inference.
Its scalability on \acrshortpl{gpu} and compatibility with modern deep learning toolchains explain why \acrshort{svi} has become the standard variational approach for training large-scale \acrshortpl{bnn}.

\subsection*{Beyond Classical Bayesian Inference}
The discussion so far has focused on the two principal methods for Bayesian inference:  
\acrshort{mcmc}, with \acrshort{hmc} and \acrshort{nuts} as its most effective variants for high-dimensional \acrshortpl{bnn}, and \acrshort{vi}, with \acrshort{svi} as the key scalable instantiation.  
These approaches represent the classical Bayesian route to training \acrshortpl{bnn}: asymptotically exact sampling on the one hand, and optimization-based approximations on the other.  

Beyond these methods, more pragmatic approximations have become popular in deep learning.  
\Acrfull{mcdo}~\cite{gal2016mcdo} and \acrfullpl{de}~\cite{lakshminarayanan2017deepensembles} trade theoretical rigor for ease of use and empirical effectiveness.  
They will be revisited in Chapter~\ref{ch:ensembles}, which broadens the perspective to ensemble-based methods.  
For a comprehensive overview of approximate Bayesian inference in \acrshortpl{bnn}, we refer to \textcite{jospin2022handsonBNN}.

\section{Evaluation Datasets}
\label{sec:bnns:datasets}

Before evaluating the practical behavior of different inference methods, it is essential to clarify what constitutes a meaningful assessment of uncertainty estimation.  
In deterministic models, performance can often be summarized by a single accuracy or loss value, whereas for \acrshortpl{bnn}, evaluation must capture not only predictive quality but also how well epistemic and aleatoric uncertainty are represented.
This places specific requirements on the datasets used for analysis.  
Ideally, they should provide clear distinctions between in- and \acrlong{ood} samples, expose ground-truth aleatoric variability, and remain low-dimensional enough to allow direct visualization of predicted uncertainties.  
At the same time, suitable quantitative metrics are needed to compare uncertainty quality across methods in a consistent and interpretable manner.  

While low-dimensional datasets are invaluable for interpreting model behavior and visualizing uncertainty, they do not reflect the full complexity of realistic machine learning tasks.  
A comprehensive evaluation of \acrshortpl{bnn} therefore requires benchmarks that jointly address interpretability, scalability, and diversity in task type.  
To this end, we employ three complementary datasets that together span these requirements.  
The \emph{Noisy Sine} benchmark provides a one-dimensional regression task with explicit ground-truth uncertainty, enabling direct comparison between predicted and true aleatoric variance.  
The \emph{Two Half Moons} dataset introduces a low-dimensional classification problem that remains simple enough for visualization but exposes the challenge of epistemic uncertainty both near decision boundaries and with increasing distance from the training data.  
Finally, the \emph{Dirty-MNIST} benchmark extends the analysis to high-dimensional image classification, allowing us to assess whether findings from the simpler settings generalize to more realistic and complex data domains.

\subsection{Noisy Sine}
\label{sec:datasets:noisy-sine}

\begin{figure}[t]
  \centering
  \includegraphics[width=0.9\linewidth]{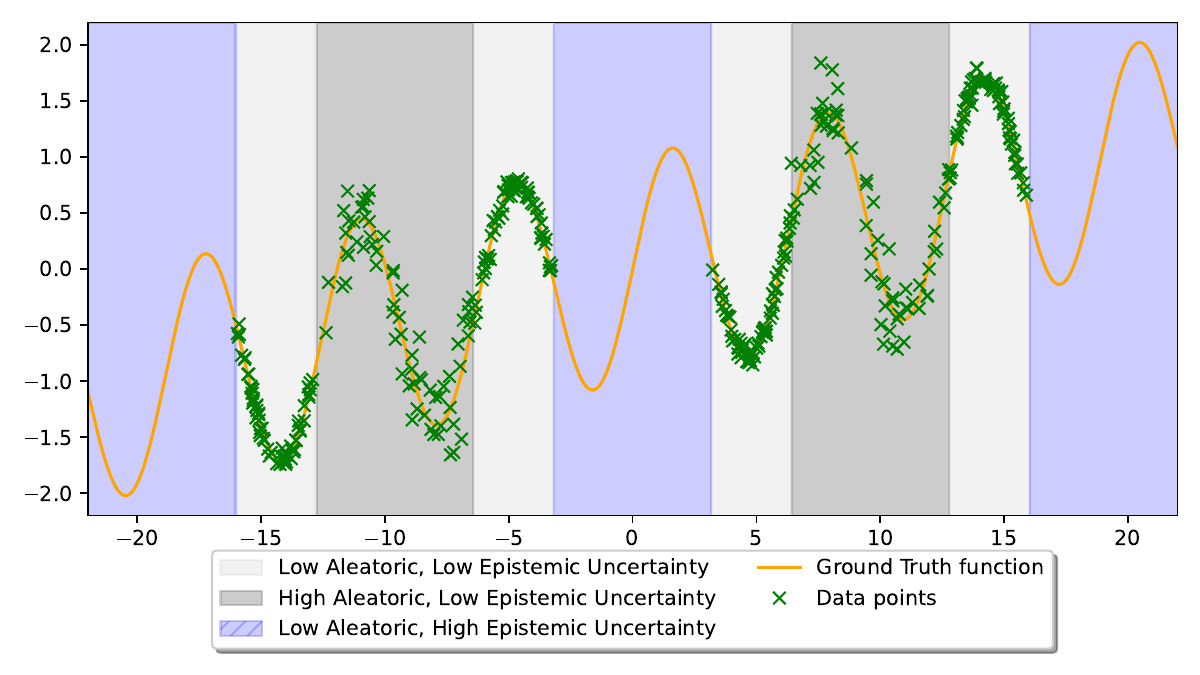}
  \caption[Noisy Sine: dataset regimes]{Schematic of the Noisy Sine dataset showing the generative function $f(x)$, 
  data-bearing regimes with different Gaussian noise levels (aleatoric), and data-free gaps that induce epistemic uncertainty. 
  Evaluation points are placed at regime midpoints and gap midpoints.~\reprofrom{simonides2025ma}}
  \label{fig:noisy_sine_schematic}
\end{figure}

\begin{figure}[t]
  \centering
  \includegraphics[width=0.9\linewidth]{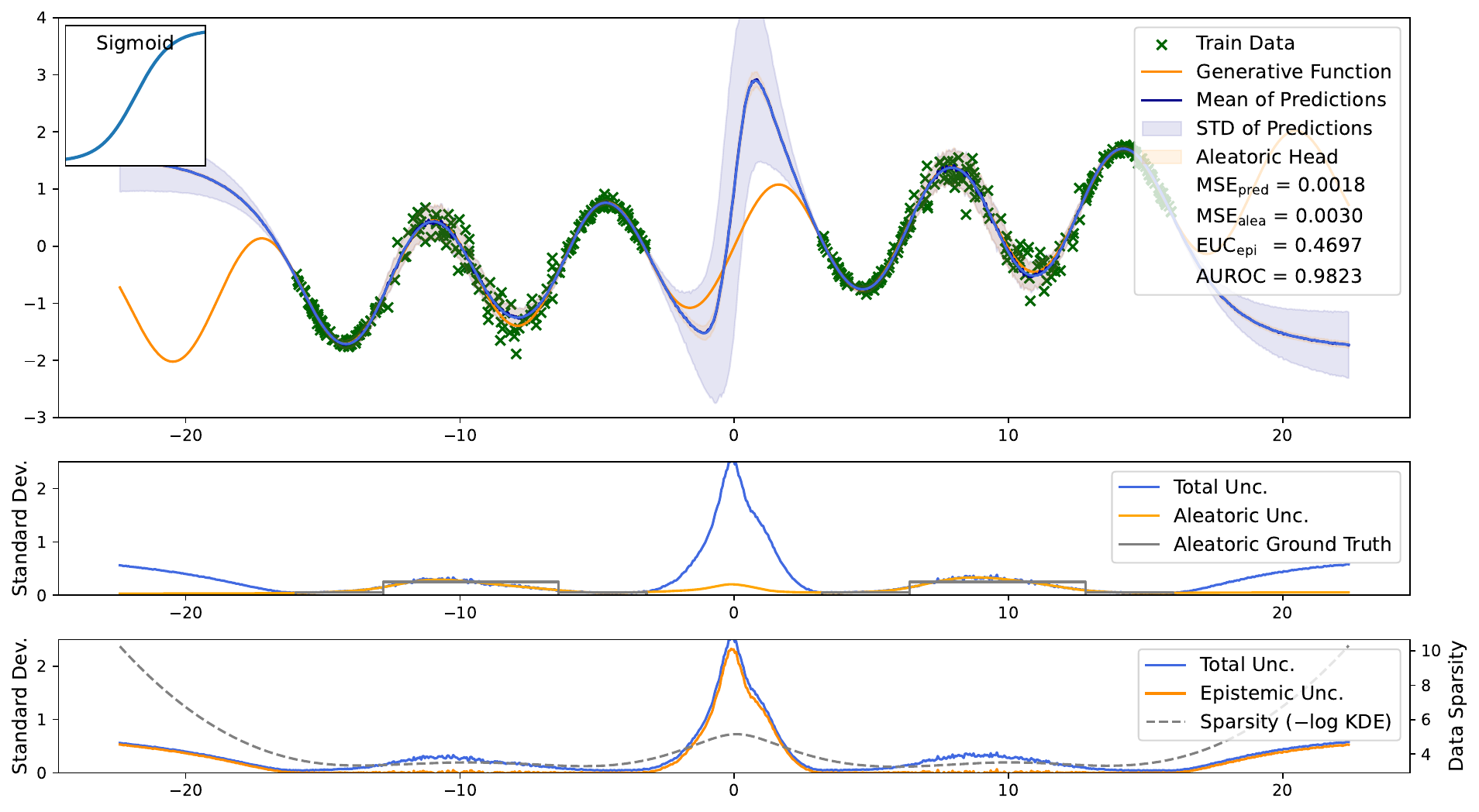}
	\caption[Noisy Sine: visualization layout]{Visualization convention for the Noisy Sine benchmark.  
	Each panel is divided into three rows:  
	\emph{Top:} predictive mean with a band for the \emph{total} predictive standard deviation, overlaid with training samples.  
	\emph{Middle:} \emph{aleatoric} uncertainty—predicted by the aleatoric head and compared against the ground-truth noise profile.  
	\emph{Bottom:} \emph{epistemic} uncertainty together with the negative log \acrshort{kde} of the training data (dashed line),  
	illustrating its correlation with data sparsity.}
  \label{fig:noisy_sine_convention}
\end{figure}

The \emph{Noisy Sine} benchmark is a one-dimensional regression task designed to probe both aleatoric and epistemic uncertainty in a controlled setting.  
The generative function is
\begin{equation}
    y = \sin(x) + 0.05 x + \epsilon, \qquad \epsilon \sim \mathcal{N}(0, \sigma^2(x)),
\end{equation}
where $\epsilon$ represents aleatoric uncertainty in the form of heteroscedastic Gaussian noise.  
The input domain is partitioned into multiple regimes, as illustrated in Figure~\ref{fig:noisy_sine_schematic}.  
Some regimes contain training points and use distinct Gaussian noise levels, inducing different levels of \emph{aleatoric} uncertainty.  
Other regimes contain no training points at all, which elicit high \emph{epistemic} uncertainty.  
To establish ground-truth uncertainty patterns, three regime types alternate along the input axis:  
(\emph{i}) dense regions with low Gaussian noise $\sigma(x){=}0.05$,  
(\emph{ii}) mid-density regions with half as many samples and higher noise $\sigma(x){=}0.25$, and  
(\emph{iii}) data-free gaps that express purely epistemic uncertainty.  
This structure creates alternating segments where uncertainty is dominated by aleatoric effects within the data regimes and by epistemic effects in the gaps and beyond the training range.  

Figure~\ref{fig:noisy_sine_convention} illustrates the visualization layout used throughout this work.  
Each plot consists of three rows: (\emph{top}) predictive mean with total uncertainty, (\emph{middle}) predicted aleatoric variance alongside the ground-truth noise profile, and (\emph{bottom}) epistemic uncertainty together with the reference sparsity of training data.

\paragraph{Prediction quality.}  
Since the ground-truth is known, predictive accuracy can be measured by the mean squared error (\acrshort{mse}) between the predicted mean $\mu(x)$ and the noiseless ground-truth target $f(x) = \sin(x) + 0.05 x$.  
Evaluation is restricted to data-bearing regimes, as extrapolation outside these regions is not meaningful for assessing fit.  

\paragraph{Aleatoric evaluation.}  
Since the ground-truth noise standard deviation $\sigma(x)$ is defined within each data regime, aleatoric uncertainty can be evaluated directly.  
We compute the \acrshort{mse} between the predicted \acrshort{std} $\hat\sigma(x)$ (from the aleatoric head) and the true injected \acrshort{std} $\sigma(x)$.  
Lower values indicate more accurate estimation of the heteroscedastic noise profile.

\paragraph{Epistemic evaluation.}  
Assessing epistemic uncertainty is more difficult because no explicit ground truth exists.  
We therefore adopt the \emph{\acrfull{euc}} proposed by \textcite{simonides2025ma}.  
The \acrshort{euc} is defined as the Pearson correlation between predicted epistemic uncertainty and the sparsity of training data, estimated by a \acrfull{kde}.  
Given training samples $\{x_i\}_{i=1}^N$, the \acrshort{kde} at a point $x$ with bandwidth $h$ is
\begin{equation}
    f(x) = \frac{1}{Nh} \sum_{i=1}^N K\!\left(\frac{x - x_i}{h}\right),
\end{equation}
where $K$ is typically chosen as a Gaussian kernel,
\begin{equation}
    K(x) = \frac{1}{\sqrt{2\pi}} \exp\!\left(-\tfrac{1}{2}x^2\right).
\end{equation}
Since well-covered regions correspond to high density and sparse regions to low density, epistemic uncertainty is expected to correlate with the \emph{negative log density}.  
The \acrshort{euc} therefore measures whether the "shape" of the predicted epistemic uncertainty aligns with data coverage.  

\begin{figure}[t]
    \centering
    \begin{subfigure}{0.49\linewidth}
        \includegraphics[width=\linewidth]{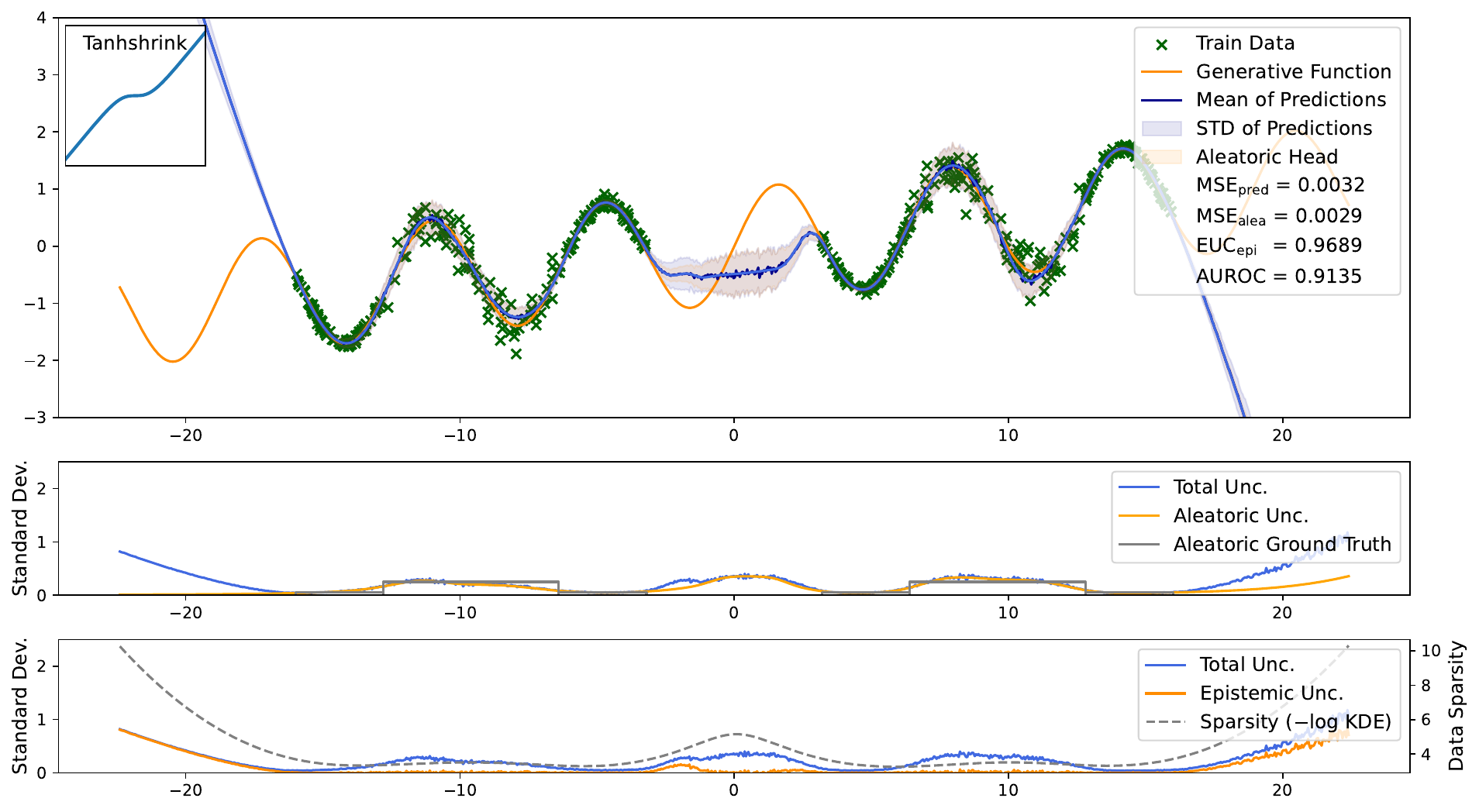}
		\caption{Tanhshrink (\acrshort{euc} $= 0.9689$)}
        \label{fig:noisy_sine_hardtanh}
    \end{subfigure}
    \hfill
    \begin{subfigure}{0.49\linewidth}
        \includegraphics[width=\linewidth]{noisysine_Sigmoid.pdf}
		\caption{Sigmoid (\acrshort{euc} $= 0.4697$)}
        \label{fig:noisy_sine_sigmoid}
    \end{subfigure}
	\caption[Noisy Sine: EUC limitation]{Illustration of the limitations of the \acrlong{euc}.  
	Panel (a) shows a model with Tanhshrink activation function, which achieved the highest \acrshort{euc} across experiments, yet clearly underestimates uncertainty in data-free regions.  
	Panel (b) shows a model with Sigmoid activation function, which yielded a lower \acrshort{euc} but produced more realistic epistemic uncertainty.  
	This highlights that \acrshort{euc} captures correlation with data sparsity but not absolute calibration, motivating the use of \acrshort{auroc} as a complementary metric.}
	\label{fig:noisy_sine_euc_limitation}
\end{figure}

While intuitive, \acrshort{euc} has an important limitation: it does not capture absolute calibration.  
It is possible for a model to systematically underestimate uncertainty and still achieve a high \acrshort{euc}, as long as the spatial correlation with data density is preserved.  
For instance, we observed cases such as a \acrshort{bnn} with a Tanhshrink activation where \acrshort{euc} reached the highest value among all experiments, yet visual inspection revealed clear underestimation of uncertainty (Figure~\ref{fig:noisy_sine_euc_limitation}).  

\paragraph{Practical alternative.}  
From a practitioner’s perspective, the more relevant question is whether uncertainty estimates can reliably flag predictions outside the training domain.  
We therefore complement \acrshort{euc} with an \acrfull{auroc}-based evaluation of \acrlong{ood} detection.  
By labeling samples inside data regimes as \acrfull{id} and those in gaps as \acrfull{ood}, we compute the \acrshort{auroc} to quantify how well epistemic uncertainty predictions separates the two cases.  
Unlike \acrshort{euc}, this metric evaluates the discriminative power of uncertainty for decision-making: a high \acrshort{auroc} indicates that the model provides a trustworthy signal of when its predictions should not be relied upon.

\subsection{Two Half Moons}
\label{sec:datasets:twohalfmoons}

Following the one-dimensional regression task, the \emph{Two Half Moons} dataset introduces a low-dimensional classification problem that remains simple enough for visualization while capturing key aspects of epistemic uncertainty.  
It is a two-dimensional binary classification benchmark with a non-linear decision boundary, frequently used to analyze how different methods capture uncertainty in classification settings.  
While aleatoric effects appear near the decision boundary, the different inference methods behave similarly in this region.  
Our evaluation therefore focuses on epistemic effects and explicit \acrshort{ood} detection relative to \acrshort{id} regions, where the methods diverge more clearly.

\begin{figure}[t]
    \centering
    \includegraphics[width=0.95\linewidth]{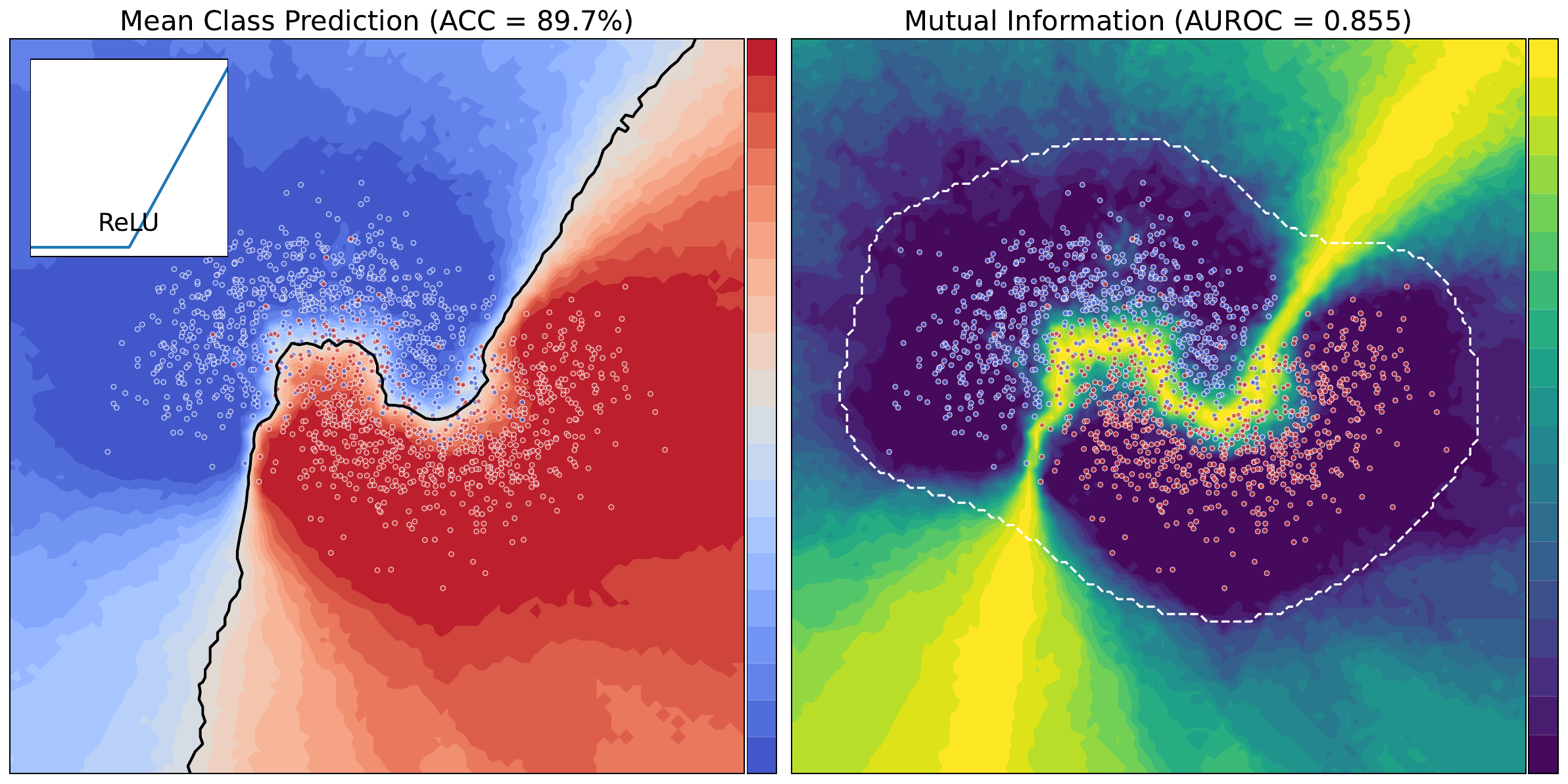}
    \caption[Two Half Moons: epistemic uncertainty and \acrshort{ood} regions]{Visualization of the \emph{Two Half Moons} dataset.  
    Training samples are shown as small circles: blue for one class and red for the other.  
    \emph{Left:} mean class predictions with the non-linear decision boundary overlaid.  
    \emph{Right:} epistemic uncertainty quantified by \acrlong{mi}, where higher values (see colorbar) indicate higher uncertainty.  
    The dashed contour marks the \acrshort{ood} region obtained via \acrshort{kde}-based density thresholding.  
    This setup enables evaluation of \acrshort{ood} detection quality via \acrshort{auroc}.}
    \label{fig:two_moons}
\end{figure}

\paragraph{\acrshort{id}/\acrshort{ood} labeling.}  
In contrast to the Noisy Sine benchmark, the \emph{Two Half Moons} dataset does not provide ground-truth \acrshort{id}/\acrshort{ood} regions.  
We therefore construct reference labels via \acrlong{kde}, similar to its use in the calculation of \acrshort{euc}, but here applied in the two-dimensional input space.  
Given training samples $\{x_i\}_{i=1}^N \subset \mathbb{R}^2$, the density at a test point $x$ is estimated as
\begin{equation}
    \hat{f}(x) \;=\; \frac{1}{N h^2} \sum_{i=1}^{N} K\!\left(\frac{x - x_i}{h}\right),
\end{equation}
with Gaussian kernel $K(u) = \tfrac{1}{2\pi}\exp\!\big(-\tfrac{1}{2}\|u\|^2\big)$ and bandwidth $h$.  
The bandwidth is determined automatically using Scott’s rule~\cite{scott1992multivariate}, a widely used heuristic that adapts to dataset size and dimensionality and provides stable density estimates without manual tuning.  
Binary \acrshort{id}/\acrshort{ood} labels are then obtained by thresholding this density at a chosen quantile: points above the threshold are treated as \acrshort{id}, while points below are labeled as \acrshort{ood}.  

We quantify \acrshort{ood} detection by ranking test points with predicted epistemic uncertainty via \acrlong{mi} and comparing against these \acrshort{id}/\acrshort{ood} labels using the \acrshort{auroc}.  
Figure~\ref{fig:two_moons} illustrates this setup: the left panel shows mean predictions and decision boundary, while the right panel shows epistemic uncertainty with the \acrshort{kde}-derived \acrshort{ood} contour.

\begin{figure}[t]
    \centering
    \includegraphics[width=0.95\linewidth]{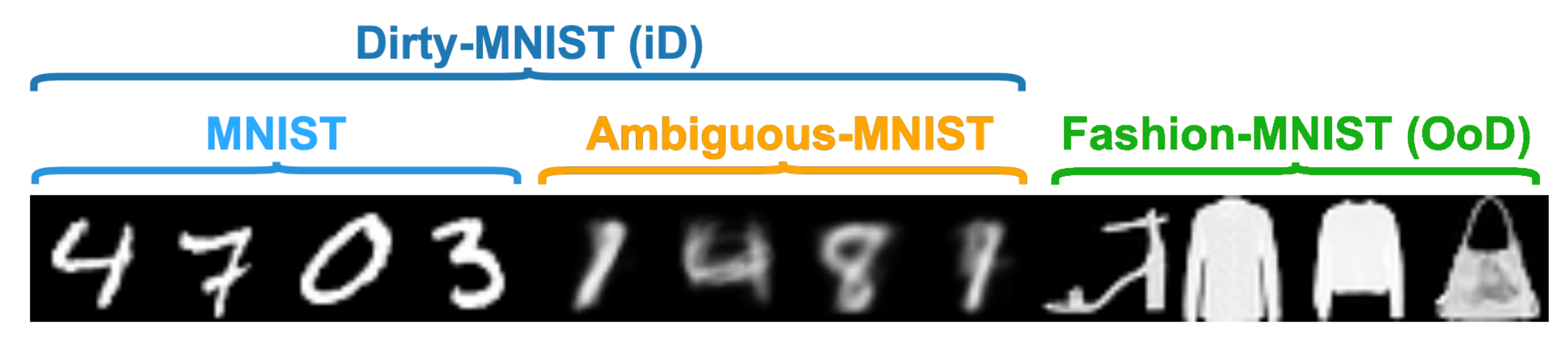}
    \caption[Dirty-MNIST benchmark]{Schematic illustration of the Dirty-MNIST benchmark.  
	From left to right: clean MNIST digits (\acrshort{id}), Ambiguous-MNIST samples generated by variational autoencoder interpolation (aleatoric uncertainty), and Fashion-MNIST images used as explicit \acrshort{ood} inputs (epistemic uncertainty).  
    Figure adapted from~\cite{Mukhoti2022dirtyMNIST}.}
    \label{fig:dirtymnist_schematic}
\end{figure}

\subsection{Dirty-MNIST}
\label{sec:datasets:dirtymnist}

The \emph{Dirty-MNIST} benchmark extends the evaluation to high-dimensional image classification, providing a more realistic testbed for uncertainty estimation.  
It builds upon the classical MNIST dataset~\cite{lecun1998mnist} and is specifically designed to probe both aleatoric and epistemic uncertainty in a controlled yet complex setting~\cite{Mukhoti2022dirtyMNIST}.  
It combines three complementary subsets, which we collectively refer to as \emph{Dirty-MNIST} in the following:

\begin{itemize}
    \item \textbf{MNIST:} clean in-distribution digit images, serving as the reference domain where predictive uncertainty should remain low.  

    \item \textbf{Ambiguous-MNIST:} digits generated by interpolating between latent codes of two MNIST digits using a variational autoencoder~\cite{pkingma2014auto-encoding}.  
    These samples are inherently ambiguous, making them a source of strong \emph{aleatoric} uncertainty.  

    \item \textbf{Fashion-MNIST:} an auxiliary dataset of clothing images~\cite{xiao2017fashionMnist}, unseen during training and used as explicit \acrshort{ood} data.  
    Predictions on these samples should exhibit high \emph{epistemic} uncertainty.  
\end{itemize}

Together, these subsets form a compact yet comprehensive benchmark.  
Following the setup of~\textcite{Mukhoti2022dirtyMNIST}, models are trained jointly on MNIST and Ambiguous-MNIST, treating both clean and ambiguous digits as \acrlong{id} data.  
MNIST provides the clean baseline, Ambiguous-MNIST emphasizes aleatoric effects, and Fashion-MNIST serves as the \acrlong{ood} probe for epistemic uncertainty.  
Figure~\ref{fig:dirtymnist_schematic} illustrates the three components side by side.  

\section{Empirical Insights into BNN Inference}
\label{sec:bnns:comparison}

The practical behavior of \acrshortpl{bnn} cannot be judged by theory alone.  
Even methods with solid theoretical foundations may succeed or fail depending on architectural choices and dataset characteristics.  
This sensitivity implies that a deterministic architecture cannot simply be made ``Bayesian'' by placing distributions on its weights; instead, small architectural adjustments must be considered to obtain reliable uncertainty estimates.  

In the following, we analyze this effect on the \emph{Noisy Sine} and \emph{Two Half Moons} benchmarks.  
Together, these results highlight systematic patterns in how activation functions influence uncertainty quality, independent of raw predictive performance.  

\subsection{Impact of Activation Functions}
\label{sec:bnns:experiments:activations}

One of the most striking observations across our experiments is that \acrshortpl{bnn} are highly sensitive to the choice of activation function.  
While predictive accuracy varies little across activation functions, uncertainty estimates differ substantially.  

\paragraph{Experimental setup.}  
All activation function experiments were conducted with an \acrshort{mlp} of two hidden layers and 50 neurons each.  
We use \acrshort{hmc} with \acrshort{nuts}, an asymptotically exact sampler whose adaptive warm-up reduces manual tuning and which serves as the de facto reference baseline for \acrshort{bnn} inference quality.
On the Noisy Sine dataset, we drew 1000 posterior samples after 1000 warm-up iterations, while for the Two Half Moons dataset we used 2000 samples with 2000 warm-up iterations to account for its higher complexity.  

\begin{figure}
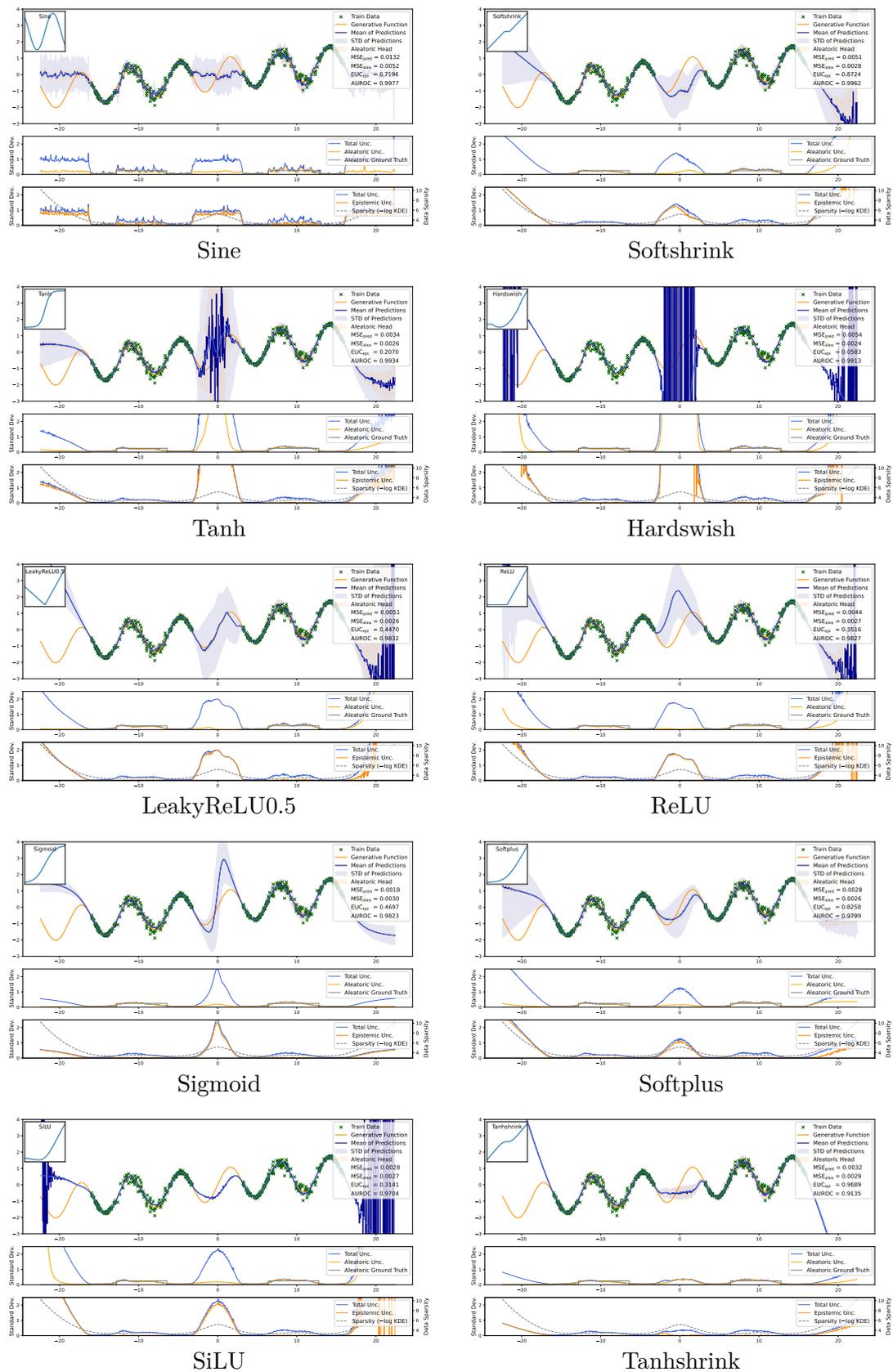
%
	\centering

	\def\ImgList{
	noisysine_Sine.pdf,
	noisysine_Softshrink.pdf,
	noisysine_Tanh.pdf,
	noisysine_Hardswish.pdf,
	noisysine_LeakyReLU0.5.pdf,
	noisysine_ReLU.pdf,
	noisysine_Sigmoid.pdf,
	noisysine_Softplus.pdf,
	noisysine_SiLU.pdf,
	noisysine_Tanhshrink.pdf
	}

	\newcommand{\ActPanel}[1]{%
	\StrBetween{#1}{noisysine_}{.pdf}[\ActName]%
	\begin{subfigure}[t]{0.48\linewidth}
		\centering
		\includegraphics[width=\linewidth]{#1}
		\vspace{-7mm}%
		\caption*{\small \ActName}
	\end{subfigure}%
	}

	\foreach \img [count=\i] in \ImgList {%
	\ActPanel{\img}%
	\ifnum\i=0\relax\else%
		\ifnum\numexpr\i-\numexpr2*\intcalcDiv{\i}{2}\relax=0\relax
		\par\vspace{2mm}%
		\else
		\hfill
		\fi
	\fi
	}

	\caption[Noisy Sine activation comparison]{Comparison of activation functions on the Noisy Sine task, using \acrshort{mcmc}-based \acrshortpl{bnn}.  
	Uncertainty behavior differs markedly in \acrshort{ood} regions despite similar predictive fits, ranging from oscillatory and unstable to smooth but overconfident.  
	Panels are ordered by decreasing \acrshort{auroc}.}
	\label{fig:bnns:activations:noisysine}
\end{figure}

\begin{table}
	\center
	\small
	\caption[Activation functions on Noisy Sine]{Activation functions on the Noisy Sine dataset, using \acrshort{mcmc}-based \acrshortpl{bnn}.  
    Metrics cover predictive accuracy, aleatoric variance estimation, and \acrshort{ood} detection quality.}
	\label{tab:bnns:activations:noisysine}
	\begin{tabular}{l|c|c|c|c}
        Activation  & MSE (Pred.) $\downarrow$ & MSE (Aleatoric) $\downarrow$ & \acrshort{euc} $\uparrow$ & \acrshort{auroc} $\uparrow$ \\
		\hline
		Cosine 			 & 0.0137 & 0.0057 & 0.6988 & \textbf{0.9985} \\
		Sine 			 & 0.0132 & 0.0052 & 0.7196 & 0.9977 \\
		Softshrink 		 & 0.0051 & 0.0028 & 0.8724 & 0.9962 \\
		Hardtanh 		 & 0.0066 & 0.0025 & 0.0978 & 0.9948 \\
		PReLU 			 & 0.0049 & 0.0028 & 0.4673 & 0.9935 \\
		Tanh 			 & 0.0034 & 0.0026 & 0.2070 & 0.9934 \\
		Hardswish 		 & 0.0054 & 0.0024 & 0.0583 & 0.9913 \\
		Hardsigmoid 	 & 0.0037 & 0.0035 & 0.7305 & 0.9906 \\
		Softsign 		 & 0.0042 & 0.0029 & 0.4661 & 0.9900 \\
		LeakyReLU 		 & 0.0050 & 0.0026 & 0.4801 & 0.9894 \\
		Mish 			 & 0.0050 & 0.0029 & 0.4195 & 0.9886 \\
		SELU 			 & 0.0053 & \textbf{0.0024} & 0.6794 & 0.9868 \\
		GELU 			 & 0.0029 & 0.0027 & 0.0443 & 0.9854 \\
		LogSigmoid 		 & 0.0024 & 0.0031 & 0.9264 & 0.9840 \\
		LeakyReLU0.5	 & 0.0051 & 0.0026 & 0.4470 & 0.9832 \\
		ReLU 			 & 0.0044 & 0.0027 & 0.3516 & 0.9827 \\
		Sigmoid 		 & \textbf{0.0018} & 0.0030 & 0.4697 & 0.9823 \\
		Softplus 		 & 0.0028 & 0.0026 & 0.8258 & 0.9799 \\
		ELU 			 & 0.0040 & 0.0028 & 0.3459 & 0.9761 \\
		CELU 			 & 0.0040 & 0.0028 & 0.3459 & 0.9761 \\
		SiLU 			 & 0.0028 & 0.0027 & 0.3141 & 0.9704 \\
		RReLU 			 & 0.0351 & 0.0205 & 0.9284 & 0.9497 \\
		LogSoftmax 		 & 1.2008 & 0.5332 & 0.4433 & 0.9212 \\
		Tanhshrink 		 & 0.0032 & 0.0029 & \textbf{0.9689} & 0.9135 \\
		Softmin 		 & 0.2612 & 0.1641 & 0.1442 & 0.7065 \\
		Softmax 		 & 0.3499 & 0.2152 & 0.1170 & 0.6440 \\
		Hardshrink 		 & 0.2690 & 0.3039 & 0.0339 & 0.4760 \\
	\end{tabular}
\end{table}

\paragraph{Noisy Sine.}
On the one-dimensional Noisy Sine task, \acrshortpl{bnn} with different activation functions generally fit the data regimes well and capture the injected heteroscedastic noise through the aleatoric head.
Overall, however, the data fit is strong, as reflected in low \acrshort{mse} for most functions, with only a few clear outliers in both predictive means and aleatoric variance (Table~\ref{tab:bnns:activations:noisysine}).  

The key differences arise in the epistemic component.  
Some activation functions, such as Sigmoid, produce smooth and moderate uncertainty growth with distance from the training data, whereas others, like Hardswish, exhibit erratic predictions and markedly inflated uncertainty bands.

Periodic activation functions (Sine, Cosine) form a special case: they yield consistently high uncertainties already a short distance away from the training data and therefore achieve the best \acrshort{auroc} values for \acrshort{ood} detection.  
This behavior can be attributed to their oscillatory structure, which tends to amplify epistemic uncertainty in regions beyond the training data.  
This benefit, however, comes at a slight cost to predictive \acrshort{mse}, reflecting the influence of their high-frequency components.  
The visualizations in Figure~\ref{fig:bnns:activations:noisysine} also highlight the limitations of scalar metrics: for example, the Tanhshrink activation attains the highest \acrshort{euc}, yet clearly underestimates uncertainty in data-free regions.  
Nearly all activation functions ultimately succeed in distinguishing \acrshort{id} from \acrshort{ood} regions, but they differ markedly in how they express epistemic uncertainty once outside the data regimes.

\begin{figure}
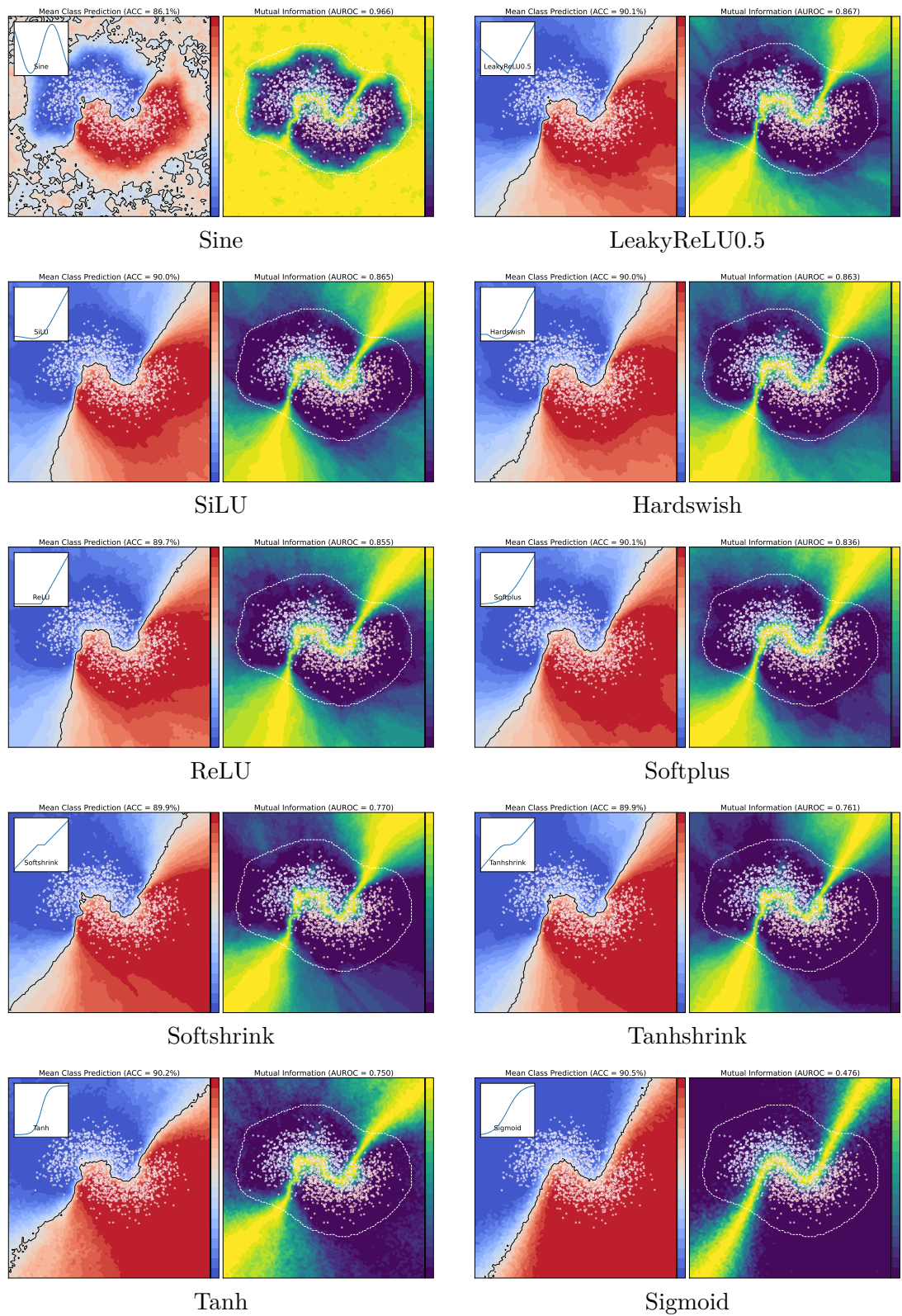
%
    \centering

    \def\ImgList{
        moons_Sine.pdf,
        moons_LeakyReLU0.5.pdf,
        moons_SiLU.pdf,
        moons_Hardswish.pdf,
        moons_ReLU.pdf,
        moons_Softplus.pdf,
        moons_Softshrink.pdf,
        moons_Tanhshrink.pdf,
        moons_Tanh.pdf,
        moons_Sigmoid.pdf
    }

    \newcommand{\ActPanel}[1]{%
        \StrBetween{#1}{moons_}{.pdf}[\ActName]%
        \begin{subfigure}[t]{0.48\linewidth}
            \centering
            \includegraphics[width=\linewidth]{#1}
			\vspace{-6mm}%
            \caption*{\small \ActName}
        \end{subfigure}%
    }

    \foreach \img [count=\i] in \ImgList {%
        \ActPanel{\img}%
        \ifnum\i=0\relax\else%
            \ifnum\numexpr\i-\numexpr2*\intcalcDiv{\i}{2}\relax=0\relax
                \par\vspace{2mm}%
            \else
                \hfill
            \fi
        \fi
    }

	\caption[Two Half Moons activation comparison]{Representative uncertainty visualizations for different activation functions on the Two Half Moons dataset, using \acrshort{hmc}-\acrshort{nuts}-based \acrshortpl{bnn}.  
    Each panel shows mean predictions (left) with decision boundary and mutual information (right) as a proxy for epistemic uncertainty.  
	While all predictions are uncertain at the decision boundary, the uncertainty increase with distance to the training data varies.
	Plots are ordered by decreasing \acrshort{auroc}.}
    \label{fig:bnns:activations:moons}
\end{figure}

\paragraph{Two Half Moons.}  
In the two-dimensional classification task, \acrshortpl{bnn} with different activation functions consistently capture the nonlinear decision boundary.  
However, they systematically misattribute the resulting ambiguity to epistemic rather than aleatoric uncertainty, which in theory should dominate in overlapping class regions.  
The crucial differences emerge in the outer \acrshort{ood} regions.  
Here, performance diverges sharply: periodic activation functions such as Sine and Cosine achieve excellent \acrshort{auroc} scores, displaying a sharp rise in uncertainty immediately beyond the data boundary, while some standard activation functions like Sigmoid remain overconfident even far outside the training support (Table~\ref{tab:mcmc_svi:moons}, Figure~\ref{fig:bnns:activations:moons}).  
Across activation functions, we observe a characteristic ordering of \acrshort{ood} behavior:
\begin{enumerate}[label=(\roman*)]
	\item \textbf{Periodic} activation functions (Sine, Cosine) yield the strongest \acrshort{ood} discrimination, characterized by a sharp rise in uncertainty immediately outside the data regions.
    \item \textbf{Smooth, positively sloped} activation functions on both sides of the origin (e.g., LeakyReLU0.5, SiLU, Hardswish) follow, typically increasing uncertainty with distance at a moderate rate.
    \item \textbf{One-sided or flatter} activation functions (e.g., \acrshort{relu}, Softplus) tend to delay uncertainty growth, remaining confident deeper into \acrshort{ood} regions.
    \item \textbf{Functions with shrinking response near the origin} (e.g., Softshrink, Tanhshrink) extend this behavior further.
    \item \textbf{Saturating sigmoidal} functions (e.g., Sigmoid, Tanh) are the most persistently overconfident.
\end{enumerate}
Importantly, this ordering is \emph{dataset-specific}: while it holds for Two Half Moons, it does not replicate on the Noisy Sine benchmark.  
This underlines that the relevance of a particular activation trait must be assessed in the context of the task rather than assumed to generalize universally.  

A recent study by \textcite{tempczyk2022simpletrick} highlighted the strong sensitivity of \acrshortpl{bnn} to activation choice in the context of mean-field \acrlong{vi}.  
They argued that \acrshort{relu} activation functions induce highly non-Gaussian posterior landscapes, which are poorly captured by Gaussian variational families, and showed that replacing \acrshort{relu} with an optimized LeakyReLU substantially improves uncertainty calibration without compromising accuracy.  
Although our experiments rely on \acrshort{hmc} with \acrshort{nuts} rather than \acrlong{vi}, and are not limited to mean-field Gaussian distributions, the broader message carries over: activation functions exert a pronounced influence on uncertainty quality.
To connect to this line of work, we evaluated their proposed best-performing variant---LeakyReLU with slope $0.5$ in the negative region, denoted \emph{LeakyReLU0.5} in our experiments.  
On the Noisy Sine task, its performance was nearly indistinguishable from standard \acrshort{relu} and outperformed by the default LeakyReLU, indicating little advantage in this setting.  
In contrast, on the Two Half Moons dataset, LeakyReLU0.5 clearly outperformed both baselines and ranked among the strongest activation functions overall, surpassed only by the periodic functions.  
These results reinforce the view that carefully designed activation functions can mitigate uncertainty failures, while also confirming that their benefits remain highly task-dependent.

\paragraph{Insights and Takeaway.}  
Across both benchmarks, several clear lessons emerge.  
First, predictive quality alone is not a reliable guide for architecture design: although most activation functions perform similarly within the data regime, their ability to capture uncertainty differs substantially.
Second, it seems there is no universally best activation function.  
Functions that work well in one setting may fail in another, as exemplified by Sigmoid, which produces smooth and realistic uncertainty on Noisy Sine but remains overconfident in Two Moons.  
Third, certain patterns emerge: for example periodic activation functions often excel at \acrshort{ood} detection in low-dimensional tasks.
However, these tendencies are not universal and do not necessarily transfer between datasets.

The overarching message is that activation function choice is a critical design aspect in \acrshortpl{bnn}.  
Unlike in deterministic models, where activation functions can often be swapped without major impact, uncertainty-aware models are highly sensitive to this decision.  
Small-scale architecture searches are therefore essential to identify suitable functions for a given dataset and model.  
Periodic activation functions appear particularly promising for robust \acrshort{ood} detection, but no universal best option exists, and careful tuning remains indispensable.

Looking ahead, more adaptive strategies may further improve activation design for \acrshortpl{bnn}.  
In preliminary work, we explored \acrfullpl{kan}~\cite{liu2025kan} to learn task-specific activation functions directly from data.  
These methods showed encouraging potential to discover novel activation functions that outperform the standard set in terms of uncertainty quality for \acrshortpl{bnn}.  
Although this line of research is not yet mature enough to be included in the present work, it suggests that the future of \acrshort{bnn} design may involve \emph{learning} activation functions tailored to uncertainty quantification rather than relying on fixed, hand-crafted functions.

\begin{figure}[t]
    \centering
    \includegraphics[width=0.9\linewidth]{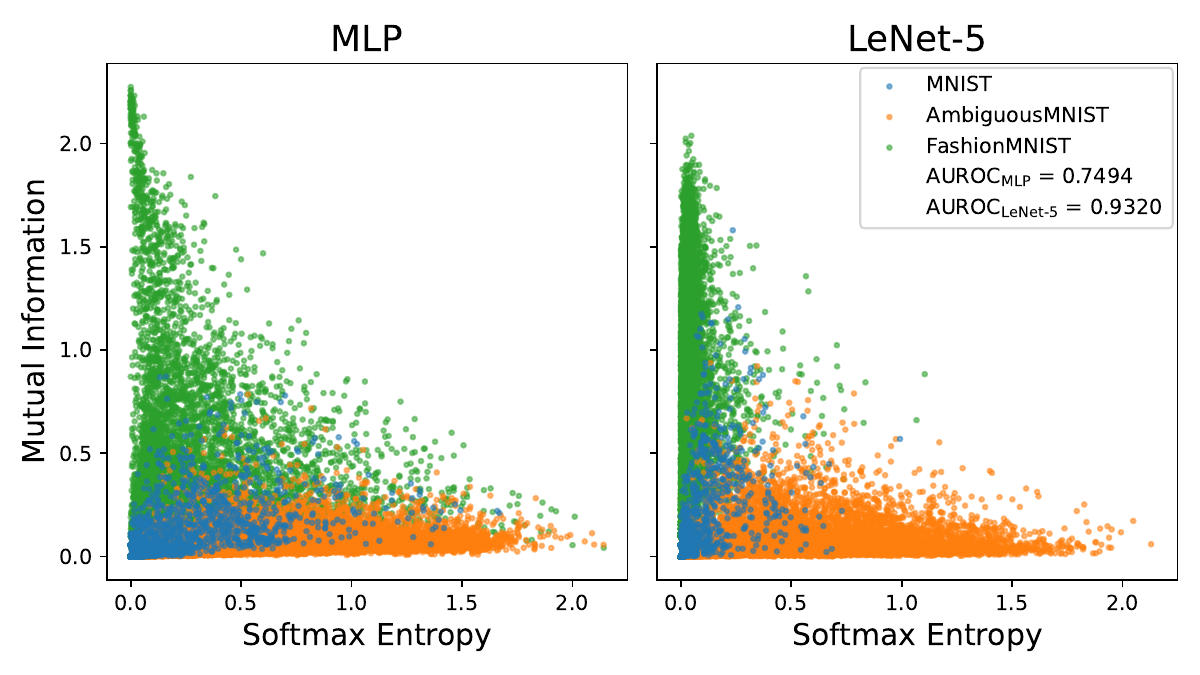}
    \caption[Architectural effects on Dirty-MNIST]{Comparison of two \acrshort{svi}-trained \acrshortpl{bnn} on the Dirty-MNIST benchmark:
    a one-hidden-layer \acrshort{mlp} (100 units) and a LeNet-5 convolutional network.
	Uncertainty separation is illustrated in the space of \acrlong{sme} (aleatoric uncertainty) and \acrlong{mi} (epistemic uncertainty).
	The LeNet-5 achieves stronger \acrshort{ood} separation and better uncertainty calibration across MNIST (\acrshort{id}), Ambiguous-MNIST (aleatoric), and Fashion-MNIST (\acrshort{ood}) subsets, highlighting the impact of architectural expressiveness.}
    \label{fig:bnns:architectures:dirtymnist}
\end{figure}

\paragraph{Other Architectural Effects.}

Beyond activation functions, other architectural choices also have a pronounced impact on uncertainty estimation in \acrshortpl{bnn}.  
Even within the same inference method, network depth and width strongly influence how uncertainty propagates.  
For \acrshort{mlp}-based models, shallow or narrow networks tend to be overconfident, whereas overly large ones often become unstable and degrade in uncertainty quality.  
This reflects a practical trade-off: increasing capacity improves expressiveness but complicates training.
Effective \acrshort{bnn} design therefore requires balancing model capacity against the fidelity of uncertainty estimation.

A similar pattern appears across architectural families.  
Figure~\ref{fig:bnns:architectures:dirtymnist} compares a fully connected \acrshort{mlp} (one hidden layer, 100 neurons) and a convolutional LeNet-5 on the Dirty-MNIST benchmark, both trained with mean-field \acrshort{svi} for 2000 epochs using \acrshort{kl} annealing.  
Although predictive accuracy on MNIST is comparable (96.6\% for the \acrshort{mlp}, 97.9\% for LeNet-5), the LeNet-5 achieves much stronger \acrshort{ood} detection (AUROC~0.93 vs.\ 0.75) and clearer uncertainty separation across the MNIST, Ambiguous-MNIST, and Fashion-MNIST subsets.  
This highlights that greater architectural expressiveness—here through convolutional feature extraction—can substantially improve uncertainty estimation under identical inference settings.  
Hence, model architecture is a decisive factor in the calibration quality of \acrshortpl{bnn}.

\begin{figure}
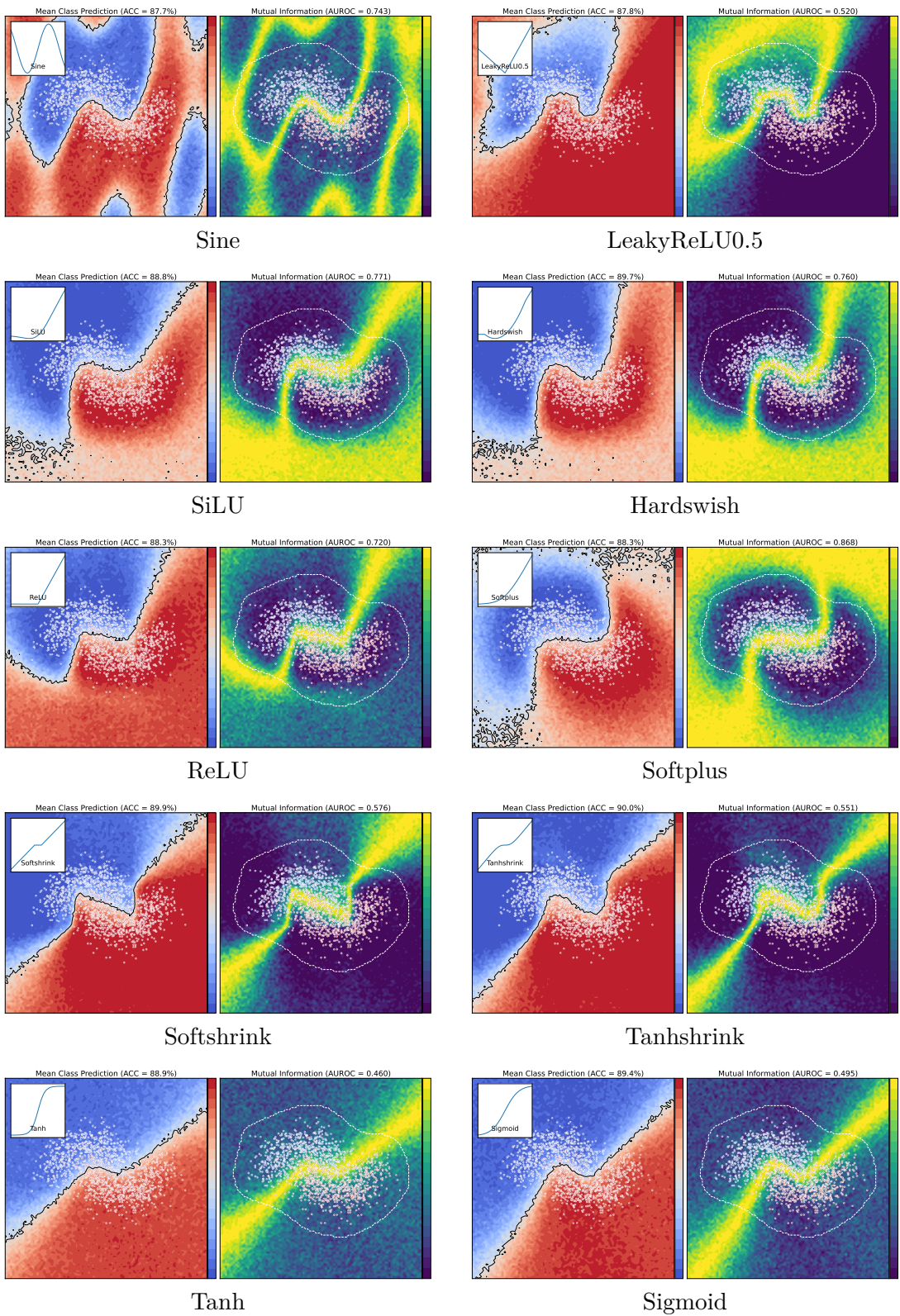
%
    \centering

    \def\ImgList{
        moonssvi_Sine.pdf,
        moonssvi_LeakyReLU0.5.pdf,
        moonssvi_SiLU.pdf,
        moonssvi_Hardswish.pdf,
        moonssvi_ReLU.pdf,
        moonssvi_Softplus.pdf,
        moonssvi_Softshrink.pdf,
        moonssvi_Tanhshrink.pdf,
        moonssvi_Tanh.pdf,
        moonssvi_Sigmoid.pdf
    }

    \newcommand{\ActPanel}[1]{%
		\StrBehind{#1}{svi_}[\ActName]%
		\StrBefore{\ActName}{.pdf}[\ActName]%
        \begin{subfigure}[t]{0.48\linewidth}
            \centering
            \includegraphics[width=\linewidth]{#1}
			\vspace{-6mm}%
            \caption*{\small \ActName}
        \end{subfigure}%
    }

    \foreach \img [count=\i] in \ImgList {%
        \ActPanel{\img}%
        \ifnum\i=0\relax\else%
            \ifnum\numexpr\i-\numexpr2*\intcalcDiv{\i}{2}\relax=0\relax
                \par\vspace{2mm}%
            \else
                \hfill
            \fi
        \fi
    }

	\caption[Two Half Moons SVI]{\acrshort{svi}-\acrshortpl{bnn} results on the Two Half Moons dataset with various activation functions. Trained with linear \acrshort{kl}--annealing for 500 epochs after deterministic pretraining. }
    \label{fig:bnns:svi:moons}
\end{figure}

\subsection{Comparison of MCMC and SVI}
To compare the practical behavior of different Bayesian inference methods, we evaluate sampling-based inference using \acrshort{hmc} with \acrshort{nuts}---referred to as \acrshort{mcmc} in the following---against \acrshort{svi} on the \emph{Two Half Moons} dataset.  
This setup enables a direct comparison between the posterior sampling approach and its variational alternative.

Comparison of Figure~\ref{fig:bnns:svi:moons} with Figure~\ref{fig:bnns:activations:moons} illustrates qualitative differences in the resulting uncertainty maps, while Table~\ref{tab:mcmc_svi:moons} summarizes their quantitative performance.  
As expected, \acrshort{mcmc} generally achieves higher \acrshort{auroc} values for \acrlong{ood} detection and produces more stable uncertainty patterns across activation functions.  
Only in two cases does \acrshort{svi} reach superior \acrshort{auroc}, and in merely three cases it slightly exceeds the predictive accuracy of \acrshort{mcmc}.  
This confirms the expected advantage of sampling-based inference in both robustness and calibration.

\begin{table}[t]
	\centering
	\small
	\caption{Impact of activation functions on \acrshortpl{bnn} trained with \acrshort{svi} and \acrshort{mcmc} on the Two Half Moons dataset. 
	Accuracy (\%) and AUROC are reported, with the better value per activation highlighted in bold.}
	\label{tab:mcmc_svi:moons}
	\begin{tabular}{l|c|c|c|c}
	\textbf{Activation} & \multicolumn{2}{c|}{\textbf{Accuracy [\%] $\uparrow$}} & \multicolumn{2}{c}{\textbf{AUROC $\uparrow$}} \\
	\hline
	& \textbf{SVI} & \textbf{MCMC} & \textbf{SVI} & \textbf{MCMC} \\
	\hline
	Sine           & \textbf{87.73} & 86.13 & 0.7427 & \textbf{0.9663} \\
	LeakyReLU0.5   & 87.80 & \textbf{90.07} & 0.5198 & \textbf{0.8666} \\
	SiLU           & 88.80 & \textbf{90.00} & 0.7706 & \textbf{0.8646} \\
	Hardswish      & 89.67 & \textbf{90.00} & 0.7602 & \textbf{0.8631} \\
	ReLU           & 88.27 & \textbf{89.67} & 0.7205 & \textbf{0.8551} \\
	Softplus       & 88.27 & \textbf{90.07} & \textbf{0.8680} & 0.8361 \\
	Softshrink     & \textbf{89.93} & 89.87 & 0.5763 & \textbf{0.7698} \\
	Tanhshrink     & \textbf{90.00} & 89.93 & 0.5514 & \textbf{0.7606} \\
	Tanh           & 88.93 & \textbf{90.20} & 0.4605 & \textbf{0.7498} \\
	Sigmoid        & 89.40 & \textbf{90.47} & \textbf{0.4953} & 0.4757 \\
	\end{tabular}
\end{table}

Beyond average performance, the visual comparison reveals qualitative distinctions that highlight the greater sensitivity of \acrshort{svi} models to architectural choices—particularly the activation function.  
While the hyperparameter sensitivity of \acrshort{svi} is well known in the community, its dependence on activation functions has received far less attention.  
As discussed earlier in Section~\ref{sec:bnns:experiments:activations} and shown by \textcite{tempczyk2022simpletrick}, activation choice can substantially affect uncertainty calibration in \acrshortpl{bnn}.  
Their study, however, was limited to \acrshort{relu} variants within mean-field \acrshort{svi} \acrshortpl{bnn}, whereas our experiments demonstrate that this sensitivity generalizes across a broader spectrum of activation functions and inference methods.  
The proposed \emph{LeakyReLU0.5}~\cite{tempczyk2022simpletrick} variant exhibited qualitatively distinct behavior but did not outperform the standard \acrshort{relu}, underscoring that such effects are strongly task-dependent.  

More broadly, our findings confirm that this phenomenon extends well beyond the \acrshort{relu} family.  
Across activation functions, \acrshort{svi} outcomes vary considerably more than under \acrshort{mcmc}, often leading to inconsistent or unstable uncertainty patterns.  
Similar activation shapes tend to induce similar artifacts: for instance, \emph{LeakyReLU0.5} produces a pronounced asymmetry, leaving one region of the input space with persistently low uncertainty, while periodic activation functions such as \emph{Sine} retain large-scale oscillatory patterns but lack the sharp \acrshort{id}/\acrshort{ood} transition observed in the \acrshort{mcmc} baseline.  
Interestingly, \emph{Softplus} emerges as the best-performing activation under \acrshort{svi}, achieving \acrshort{auroc} values on par with the strongest \acrshort{mcmc} models.  
This highlights a central observation: the optimal activation function for a given task and architecture is not invariant to the inference method.  
In practice, the most effective activation functions for \acrshort{svi} may differ from those for \acrshort{mcmc}, complicating \acrshort{bnn} research and the transfer of architectures between inference paradigms.  

From a practitioner's perspective, these results have two main implications.  
First, \acrshort{svi} can, in principle, match the uncertainty quality of \acrshort{mcmc}---but only under carefully tuned configurations.  
In this study, the Softplus--\acrshort{svi} combination reached comparable \acrshort{ood} detection quality, whereas most other activation functions underperformed the sampling-based models.  
Second, this tuning sensitivity emphasizes that \acrshort{svi}-based \acrshortpl{bnn} require substantially more effort in hyperparameter optimization and architecture adaptation to achieve reliable results.  
The stochastic optimization process and mean-field assumptions of \acrshort{svi} appear to amplify such sensitivities, in contrast to the more stable but computationally demanding \acrshort{mcmc} approach.

\begin{figure}
    \centering
    \includegraphics[width=0.9\linewidth]{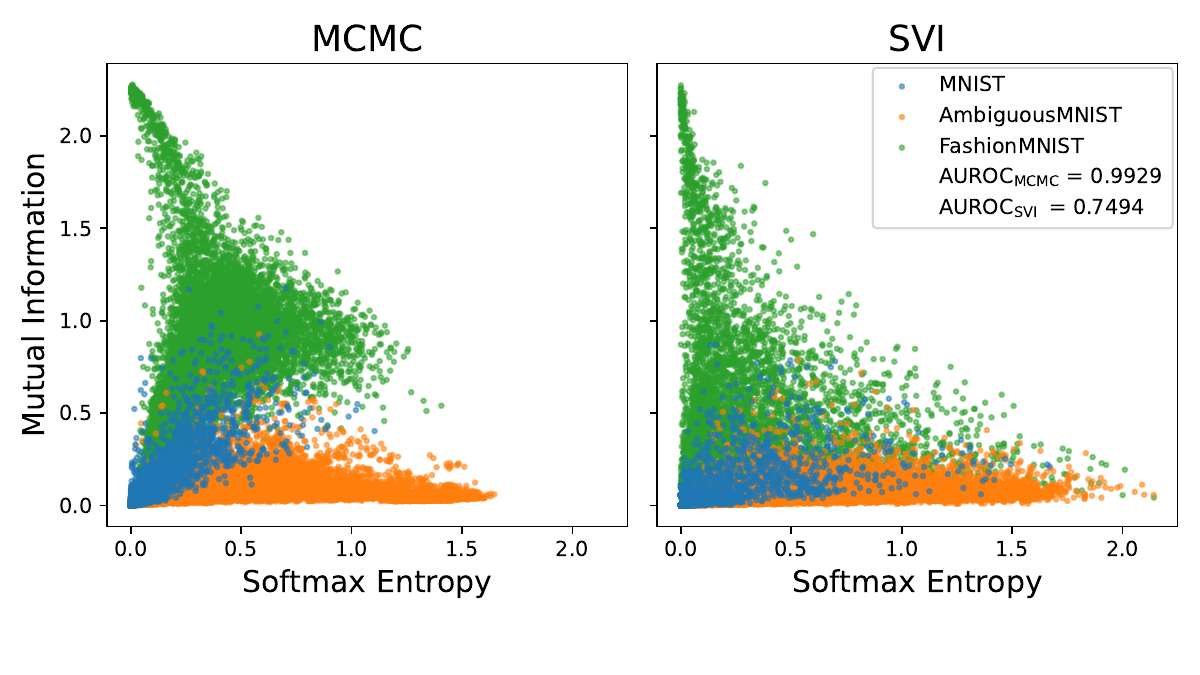}
	\caption[Dirty-MNIST comparison of MCMC and SVI]{Comparison of \acrshort{mcmc}- and \acrshort{svi}-trained \acrshortpl{bnn} on the Dirty-MNIST dataset using a \acrshort{mlp} (one hidden layer, 100 neurons).  
	Both models assign high \acrlong{sme} to Fashion-MNIST samples—an incorrect attribution, since these should be characterized by high \acrlong{mi}.
	The \acrshort{mcmc} model however, also attributes high \acrlong{mi}, correctly marking them as \acrshort{ood}, whereas \acrshort{svi} does in many cases not.
	Despite similar accuracy (95.8\% \acrshort{mcmc} vs.\ 96.6\% \acrshort{svi}), \acrshort{mcmc} delivers far superior \acrshort{ood} detection (\acrshort{auroc} \acrshort{mcmc}\,=\,0.99 vs.\ 0.75 for \acrshort{svi}).}
    \label{fig:bnns:mcmc_svi:dirtymnist}
\end{figure}

For the higher-dimensional Dirty-MNIST benchmark~\cite{Mukhoti2022dirtyMNIST}, this trade-off becomes particularly evident.  
Figure~\ref{fig:bnns:mcmc_svi:dirtymnist} compares predictions from \acrshort{mcmc}- and \acrshort{svi}-trained \acrshortpl{bnn} using a compact \acrshort{mlp} with one hidden layer of 100 neurons.  
While both models achieve comparable accuracy, they struggle to disentangle epistemic from aleatoric uncertainty.  
Both assign high \acrlong{sme} to Fashion-MNIST samples, but only the \acrshort{mcmc} model simultaneously reports high \acrlong{mi}, correctly identifying these samples as \acrshort{ood}.  
The \acrshort{svi} model, in contrast, misattributes many of them as aleatoric, resulting in weaker \acrshort{ood} discrimination.  

A complementary experiment, shown in Figure~\ref{fig:bnns:architectures:dirtymnist}, demonstrates that increased architectural expressiveness can mitigate these effects.  
Replacing the simple \acrshort{mlp} with a convolutional LeNet-5---trained via \acrshort{svi}---leads to markedly improved uncertainty separation and \acrshort{ood} detection performance.  
Despite relying on approximate inference, the LeNet-5 achieves near-perfect separation between MNIST, Ambiguous-MNIST, and Fashion-MNIST subsets while maintaining high predictive accuracy.  
However, this architecture already lies in a regime where training with \acrshort{mcmc} becomes prohibitively time-consuming.  

Together, these results highlight a clear trade-off: \acrshort{mcmc} remains the most reliable approach for uncertainty calibration but is computationally infeasible for more complex architectures, whereas \acrshort{svi} offers superior scalability and can leverage expressive models to compensate for its approximate nature.

Taken together, the three benchmarks reveal that the practical success of mean-field \acrshort{svi} for \acrshortpl{bnn} is highly dataset and architecture dependent.  
On simple, low-dimensional tasks such as \emph{Noisy Sine} and \emph{Two Half Moons}, the method often struggles to reproduce the well-calibrated uncertainty structure obtained with sampling-based inference, showing strong sensitivity to both architectural and hyperparameter choices.  
In contrast, on higher-dimensional datasets like \emph{Dirty-MNIST}, where full \acrshort{mcmc} becomes computationally infeasible, carefully tuned \acrshort{svi} models can achieve robust uncertainty separation and scale efficiently to larger networks.  
Overall, these results indicate that mean-field \acrshort{svi} remains a practical and scalable approximation, but its reliability---and the degree of tuning required---depend critically on the dataset characteristics and model architecture.

\subsection{Computational Scaling of Bayesian Inference}
\label{sec:bnns:scaling}

\begin{figure}
  \centering
  \begin{subfigure}[b]{0.378\textwidth}
    \includegraphics[width=\linewidth]{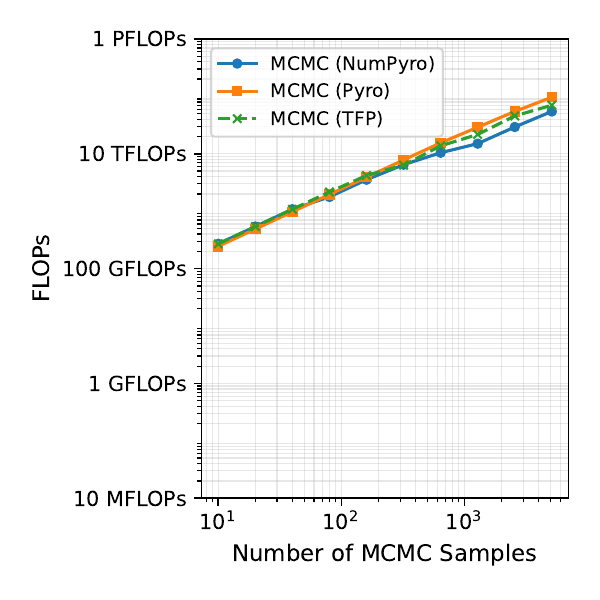}
    \caption{MCMC Samples}
    \label{fig:sampling}
  \end{subfigure}
  \hfill
  \begin{subfigure}[b]{0.28\textwidth}
    \includegraphics[width=\linewidth]{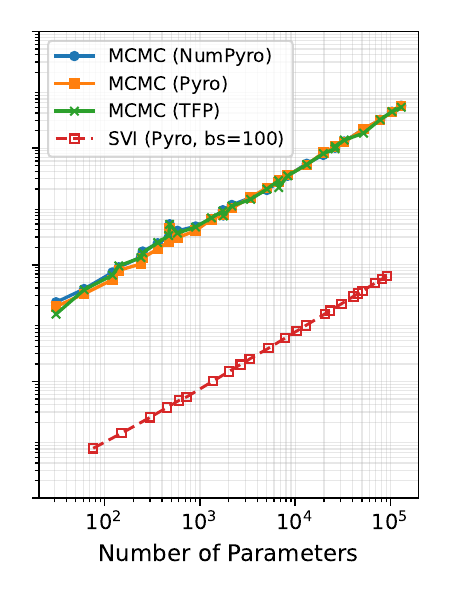}
    \caption{Model Size}
    \label{fig:total}
  \end{subfigure}
  \hfill
  \begin{subfigure}[b]{0.28\textwidth}
    \includegraphics[width=\linewidth]{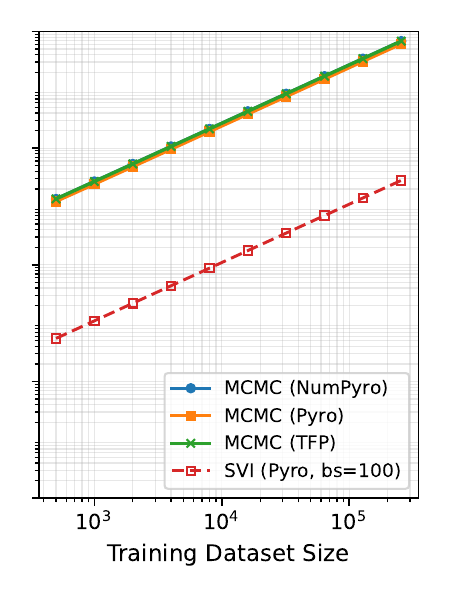}
    \caption{Dataset Size}
    \label{fig:dataset}
  \end{subfigure}

	\caption[FLOP scaling of Bayesian inference methods]{Scaling of \acrshort{flops} for \acrshort{mcmc} and \acrshort{svi} training of \acrshortpl{bnn}.  
	FLOP counts increase linearly with both model and dataset size across all probabilistic programming frameworks (Pyro, NumPyro, TensorFlow Probability).  
	While the overall trends are similar, \acrshort{svi} requires roughly two orders of magnitude fewer FLOPs, highlighting its superior computational efficiency.}
  \label{fig:bnns:scaling:flops}
\end{figure}

While previous sections have qualitatively noted that \acrshort{mcmc} tends to be computationally more demanding than variational approaches, we now quantify this difference empirically.  
Specifically, we evaluate the computational scaling of \acrshort{hmc}--\acrshort{nuts} and mean-field \acrshort{svi} across varying model and dataset sizes, as well as across three probabilistic programming frameworks: \emph{Pyro}~\cite{bingham2019pyro}, \emph{NumPyro}~\cite{Phan2019NumPyro}, and \emph{TensorFlow Probability}~\cite{Dillon2017TFP}.  
The goal of these experiments is not to optimize predictive performance, but to characterize how computational cost grows with problem complexity.  
Here, we specifically measure the computational cost of \emph{Bayesian inference} for \emph{training} \acrshortpl{bnn}---that is, approximating the posterior distribution over network weights---rather than the subsequent \emph{predictive inference} phase used for evaluating new inputs.  
This distinction, illustrated in Figure~\ref{fig:bnn_workflow}, is essential for understanding the trade-off between training and inference costs for \acrshortpl{bnn}:  
Approaches such as \acrshort{hmc} are dominated by expensive posterior sampling during training, whereas other techniques may instead shift the computational burden toward a larger number of required samples---and thus forward passes---at inference time.

The scalability experiments in this subsection were conducted as part of a master’s thesis project.\footnote{At Heidelberg University and carried out by Jonathan Bernhard.}
All experiments were executed on an AMD EPYC~7302P 16-core processor, with hardware-level metrics collected using the \texttt{likwid} performance monitoring framework~\cite{treibig2010likwid}.  
Unless otherwise stated, the evaluated networks were \acrshortpl{mlp} with three hidden layers of 50 neurons each.  
Model size was systematically varied by adjusting both the network depth (from one to five hidden layers) and layer width (from 25 to 200 neurons), while dataset size was scaled on the \emph{Noisy Sine} benchmark by increasing the number of training samples.  
For \acrshort{mcmc}, we employed \acrshort{hmc} with the \acrshort{nuts} sampler and collected 50 posterior samples following 100 warm-up iterations.  
For \acrshort{svi}, models were trained for 150 epochs to ensure comparable computational budgets across settings.  
While these configurations are insufficient to achieve fully converged predictive quality, they provide a consistent and representative basis for measuring computational cost and scaling trends.  
All Pyro experiments were executed with \acrshort{jit} compilation enabled to mitigate Python interpreter overhead.

\begin{figure}
  \centering
 	\begin{subfigure}[b]{0.49\textwidth}
    \includegraphics[width=\linewidth]{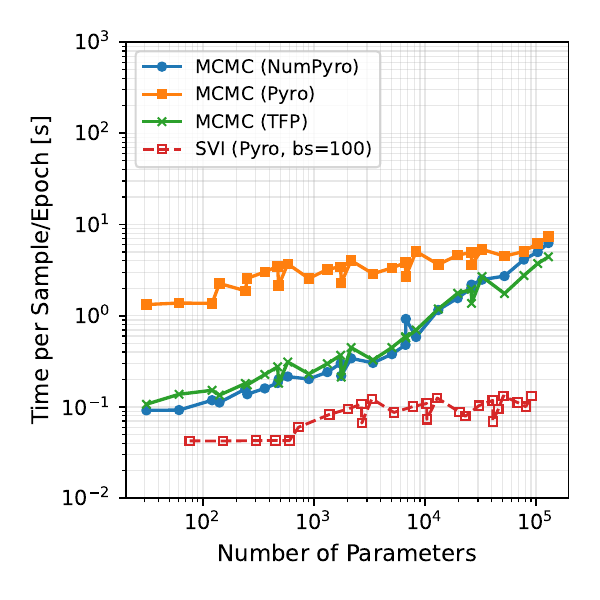}
    \caption{Model Size}
    \label{fig:total}
  \end{subfigure}
  \hfill
  \begin{subfigure}[b]{0.49\textwidth}
    \includegraphics[width=\linewidth]{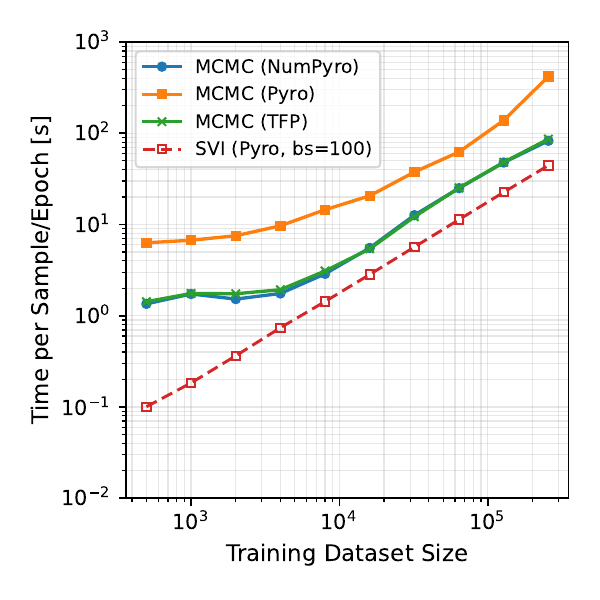}
    \caption{Dataset Size}
    \label{fig:dataset}
  \end{subfigure}

\caption[Runtime scaling of MCMC and SVI]{Average latency per sample (for \acrshort{mcmc}) and per epoch (for \acrshort{svi}) across model and dataset size.  
While both methods scale with model size, \acrshort{svi} remains significantly faster, whereas \acrshort{mcmc} incurs higher latency due to adaptive sampling steps.}
  \label{fig:bnns:scaling:latency}
\end{figure}

Across frameworks, the FLOP count increases approximately linearly with both model and dataset size (Figure~\ref{fig:bnns:scaling:flops}).  
The differences between frameworks are negligible, indicating that they perform essentially the same amount of arithmetic work for comparable models.  
Overall, \acrshort{svi} requires about two orders of magnitude fewer operations than \acrshort{mcmc}, consistent with the additional gradient-based sampling steps in \acrshort{hmc}.  

Figure~\ref{fig:bnns:scaling:latency} illustrates the measured latency per iteration—defined as time per \acrshort{mcmc} sample or per \acrshort{svi} training epoch—as a function of model and dataset size.  
With increasing model size, the computational gap between the two inference schemes widens markedly: while \acrshort{svi} exhibits a gradual, near-linear growth in runtime, \acrshort{mcmc} slows down disproportionately as the number of parameters increases.  
This trend is most evident for the NumPyro and TensorFlow Probability implementations, where time per sample rises steeply with model complexity.  
For Pyro, the effect is somewhat masked by interpreter overhead, which dominates total latency at smaller scales.  
Overall, the results highlight that \acrshort{mcmc} scales considerably worse with parameter count than \acrshort{svi}, reinforcing the computational advantage of optimization-based variational inference.

We attribute the poorer scaling of \acrshort{mcmc}---more precisely, \acrshort{hmc} with the \acrshort{nuts} sampler---relative to \acrshort{svi} mainly to its reliance on repeated gradient evaluations for each posterior sample.  
Each trajectory in \acrshort{hmc} involves simulating the system dynamics over many leapfrog steps, effectively multiplying the number of full forward and backward passes through the network.  
As model dimensionality increases, the posterior landscape becomes increasingly complex, demanding longer trajectories and smaller integration step sizes to maintain numerical stability.  
In contrast, \acrshort{svi} performs amortized inference by directly optimizing a variational objective, requiring only a single gradient update per mini-batch and scaling with only a modest increase in cost as model size grows.
Empirically, this effect appears as a significantly steeper growth in computation time for larger architectures under \acrshort{mcmc}.

\begin{figure}
  \centering
 	\begin{subfigure}[b]{0.49\textwidth}
    \includegraphics[width=\linewidth]{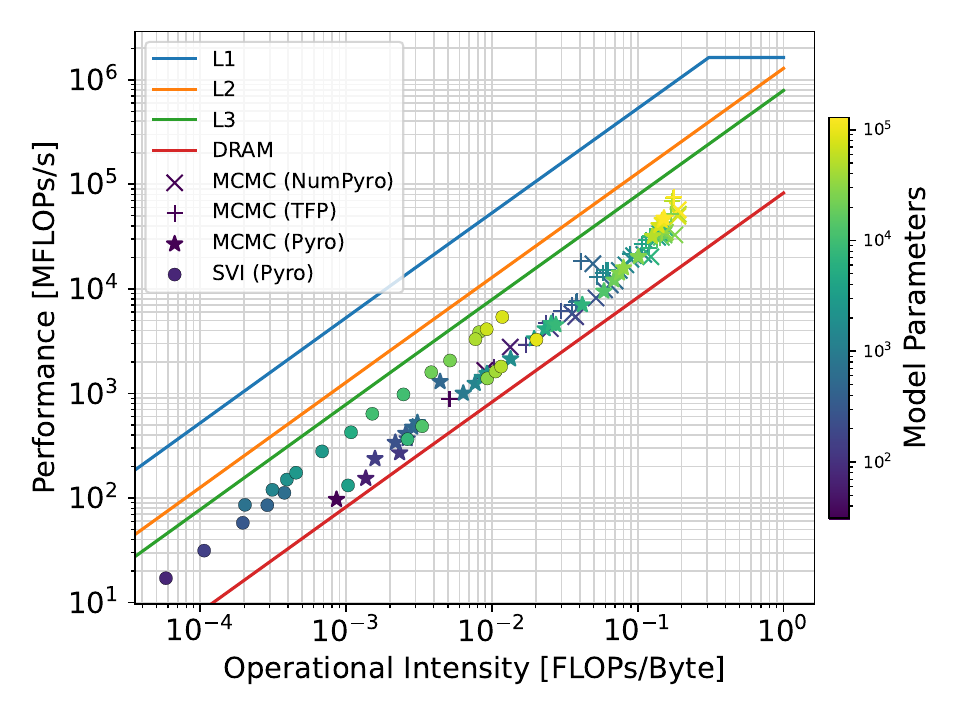}
    \caption{Model Size}
    \label{fig:total}
  \end{subfigure}
  \hfill
  \begin{subfigure}[b]{0.49\textwidth}
    \includegraphics[width=\linewidth]{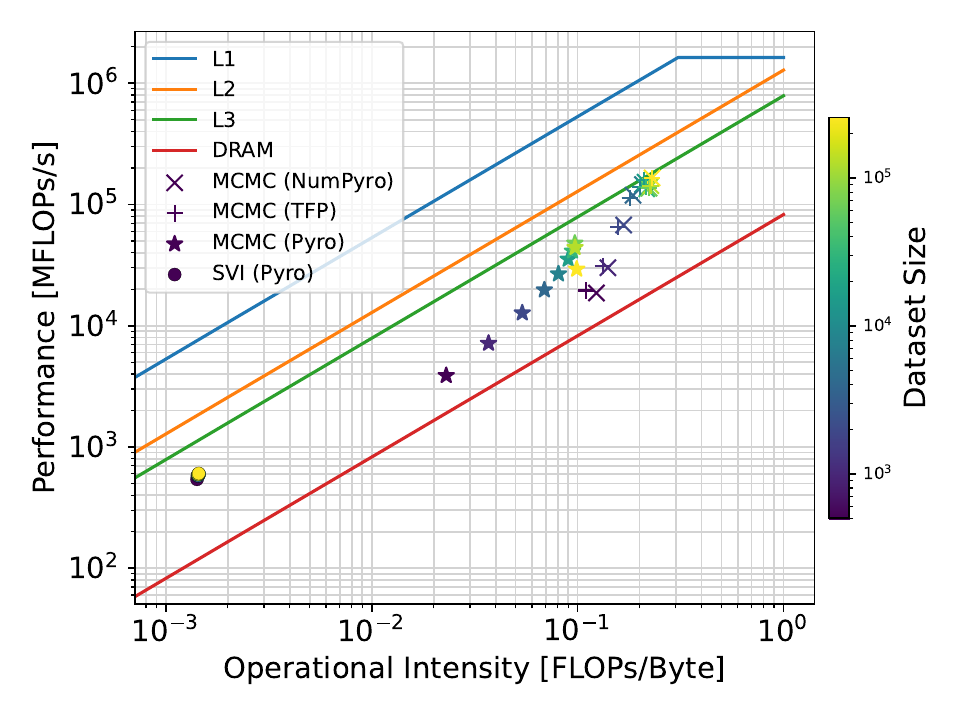}
    \caption{Dataset Size}
    \label{fig:dataset}
  \end{subfigure}

	\caption[Roofline analysis of MCMC training]{Roofline analysis of \acrshort{mcmc} and \acrshort{svi} training for \acrshortpl{bnn}.  
	All configurations operate in the memory-bound regime between DRAM and L3-cache bandwidth, with larger models and datasets increasing operational intensity and improving hardware utilization.}
  \label{fig:bnns:scaling:roofline}
\end{figure}

A more detailed view of computational efficiency is provided by the roofline analysis in Figure~\ref{fig:bnns:scaling:roofline}.  
Across all configurations, \acrshort{mcmc} sampling operates in a regime of low operational intensity, confirming that Bayesian sampling on modern \acrshortpl{cpu} is predominantly memory-bound rather than compute-bound.  
Most measurements fall between the effective DRAM and L3-cache bandwidths, indicating that limited data reuse and frequent memory transfers dominate runtime.  
Increasing either model or dataset size raises the operational intensity, as larger workloads improve cache locality and arithmetic-to-memory ratios.  
In particular, scaling the dataset shifts the computation closer to the L3-bandwidth limit, suggesting more efficient processor utilization for larger data regimes.  
From an efficiency standpoint, both larger models and datasets make better use of the available hardware, though at the cost of substantially higher absolute compute demand.  

For \acrshort{svi}, the picture differs due to mini-batch training: all dataset-size configurations cluster around a similar operational intensity, as memory access patterns are dominated by fixed-size batches rather than the total dataset.  
When scaling model size, operational intensity increases proportionally with the number of parameters, yet remains below that of \acrshort{mcmc}, reflecting the lower arithmetic density of its gradient-based optimization.  
Among the evaluated frameworks, NumPyro and TensorFlow Probability achieve the highest effective hardware utilization, while Pyro remains most constrained by interpreter and runtime overhead.

In summary, the scaling analysis highlights a clear computational trade-off.  
\acrshort{svi} consistently achieves lower runtime and scales more efficiently with both model and dataset size, whereas \acrshort{mcmc} with \acrshort{nuts} incurs substantially higher cost but remains more robust in terms of convergence and calibration.  
All implementations operate in a memory-bound regime, indicating that future efficiency gains will primarily depend on improving data movement and cache utilization rather than on increasing raw compute throughput.

\section*{Summary}

This chapter provided an empirical examination of Bayesian neural networks, analyzing how inference methods, architectural choices, and implementation factors shape uncertainty quality and computational efficiency.  
Across experiments, activation functions emerged as a major determinant of epistemic behavior: while predictive accuracy remained similar, uncertainty calibration varied drastically.  
Comparisons of \acrshort{mcmc} and \acrshort{svi} showed that sampling-based inference yields the most reliable uncertainty estimates, whereas mean-field \acrshort{svi} is more sensitive to architecture and activation choice yet scales to larger models.  
Architectural expressiveness also plays a critical role, exemplified by the transition from a simple \acrshort{mlp} to a slightly larger \acrshort{cnn}, markedly improved uncertainty separation.  
Finally, computational scaling analyses revealed that \acrshort{svi} is orders of magnitude faster and scales better with model size than \acrshort{mcmc}, though both are predominantly memory-bound.  

Overall, reliable \acrshort{bnn} inference depends as much on the network architecture and activation functions as on the chosen Bayesian inference method.
\acrshort{mcmc} remains the reference standard for small, precision-critical models, while \acrshort{svi} enables scalable Bayesian deep learning when coupled with well-chosen architectures.  

While this chapter established the principles of designing and training \acrlongpl{bnn}, deploying them on embedded hardware raises new challenges, where limited computation and memory dominate design trade-offs.  
The next chapters therefore move from training to deployment, exploring \emph{efficient \acrshort{bnn} inference} methods that sustain reliable uncertainty estimation under the tight constraints of real-world systems.

\chapter{Compiling Probabilistic Forward Pass BNNs for Embedded Systems}
\label{ch:pfp}

\epigraph{In theory, there is no difference between theory and practice, while, in practice, there is.}{\textnormal{--- Benjamin Brewster}}

\noindent
As discussed in the previous chapter, \acrshortpl{bnn} provide a principled way to quantify uncertainty, distinguishing between aleatoric and epistemic sources.  
However, nearly all established Bayesian inference methods, such as \acrlong{mcmc}, \acrlong{svi}, and even modern approximation techniques like Monte Carlo Dropout~\cite{gal2016mcdo} or Deep Ensembles~\cite{lakshminarayanan2017deepensembles}, remain computationally costly.  
In practice, reliable uncertainty estimation typically requires drawing many posterior samples and executing multiple forward passes per input, which prevents deployment on embedded and resource-constrained platforms.  
This chapter is based on our work \emph{“Accelerated Execution of Bayesian Neural Networks using a Single Probabilistic Forward Pass and Code Generation”}, currently under review at the \emph{ACM Transactions on Architecture and Code Optimization}~\cite{klein2025pfp}.

This chapter focuses on the \acrfull{pfp}~\cite{roth2016pfp,roth2021phd}, which can be regarded as an extreme form of \acrshort{svi}.  
While \acrshort{svi} models weights as Gaussian-distributed parameters under a mean-field assumption, \acrshort{pfp} extends this approximation to activations as well.  
This assumption enables distributions to be propagated in closed form, allowing both predictions and uncertainties to be computed within a single forward pass.  
By eliminating repeated stochastic evaluations, \acrshort{pfp} directly addresses the central computational bottleneck of \acrshortpl{bnn}.  
Although the Gaussian restriction reduces the ability to capture non-Gaussian activation distributions, it provides a tractable and efficient approximation particularly suited for inference on resource-constrained systems.  

So far, \acrshort{pfp} has remained largely theoretical, as its Gaussian-propagating operators are absent from standard \acrshort{ml} frameworks.  
Without this support, no practical path to deployment was available.  
In this chapter, we close this gap by presenting the first end-to-end realization of \acrshort{pfp}.  
We demonstrate how the deep learning compiler \acrshort{tvm}~\cite{chen2018tvm}, can be extended with custom operators to support \acrshort{pfp}.  
These compilers provide a generic implementation pathway transferable across hardware backends while maintaining seamless integration with common \acrshort{ml} workflows.  
As a result, models trained with \acrshort{svi} can be exported, optimized, and deployed with minimal effort—enabling a capability that was previously out of reach.  

The chapter begins by establishing a training pipeline based on \acrshort{svi} and shows that \acrshort{pfp} achieves comparable uncertainty estimation and out-of-domain detection to sampling-based baselines on the Dirty-MNIST dataset~\cite{Mukhoti2022dirtyMNIST}.  
We then extend \acrshort{tvm} with a library of custom operators for \acrshortpl{mlp} and \acrshortpl{cnn}, and apply both manual and automatic optimizations to improve the efficiency of computationally demanding operators.  
Finally, we benchmark the resulting implementation on embedded ARM processors, demonstrating speedups of up to four orders of magnitude compared to \acrshort{svi}-based baselines.  

Altogether, this chapter demonstrates how deep learning compilers, operator optimizations, and algorithmic approximations together make \acrshort{bnn} inference—and thus uncertainty quantification—feasible even on resource-constrained embedded systems.  
\section{Efficient Inference of \acrshortpl{bnn}}
\label{sec:pfp:rw}

Achieving efficiency is a central challenge for both standard neural networks and \acrshortpl{bnn}.  
This section reviews related efforts, beginning with compiler and compression techniques for deterministic models on embedded hardware, and then focusing on methods that aim to make \acrshortpl{bnn} feasible under similar constraints.  

\paragraph{Resource-Efficient Deployment on Mobile Devices.}  
Deploying \acrlong{ml} models on mobile hardware requires strict control of computational and memory budgets.  
To this end, \emph{deep learning compilers} have emerged as a key abstraction layer that bridges high-level \acrlong{ml} frameworks and heterogeneous hardware backends.  
They automate optimizations such as operator fusion, memory scheduling, and parallelization, and increasingly support automated tuning of implementation schedules.  
Prominent examples include \acrfull{tvm}~\cite{chen2018tvm} and \acrfull{mlir}~\cite{Lattner2021mlir}, which provide extensible infrastructures for graph- and operator-level optimizations across \acrshortpl{cpu}, \acrshortpl{gpu}, and reconfigurable hardware such as \acrshortpl{fpga}.  
\acrshort{tvm} in particular combines these optimizations with learning-based auto-tuning, enabling efficient code generation across diverse backends.  
We build on this abstraction in the present chapter by extending \acrshort{tvm} with custom operators to support the non-standard operators of the \acrlong{pfp}.  

Model compression techniques such as pruning and quantization reduce the computational and memory footprint of neural networks, enabling deployment on embedded hardware~\cite{jacob2018quantization,han2016deepcompression,schindler2018}.  
More advanced approaches use reinforcement learning or \acrlong{nas} to automatically adapt sparsity and precision across layers~\cite{heAMCAutoMLModel2019,krieger2023galen}.  
An introduction to compression methods and their hardware implications has already been provided in Chapter~\ref{ch:rennfes}, while Galen (Chapter~\ref{ch:galen}) extends these ideas with automatic compression guided by sensitivity analysis and hardware measurements.  
For a broader survey of efficient \acrfull{nn} inference, we refer to Roth~\cite{roth2024jmlr}.  

From a Bayesian perspective, \citeauthor{louizos2017} proposed compression techniques that embed priors into pruning and use posterior uncertainty to assign precision~\cite{louizos2017}.  
These methods, however, are designed for deterministic networks and do not address efficiency in \acrshortpl{bnn}.  

\paragraph{Bayesian Neural Networks on Resource-Constrained Devices.}  
Lightweight \acrshort{bnn} approximations such as \acrfull{mcdo}~\cite{gal2016mcdo} and \acrlongpl{de}~\cite{lakshminarayanan2017deepensembles} are widely used in practice.  
They reduce training cost compared to full Bayesian inference, but sacrifice theoretical rigor and, at inference time, still require multiple forward passes.  
Although significantly more efficient than sampling-based \acrshortpl{bnn}, \acrshort{svi} remains too costly for embedded systems, where latency and energy budgets are highly constrained.

Efforts that more directly target \acrshortpl{bnn} on constrained hardware remain limited.  
\citeauthor{banerjee2019}, for example, introduced AcMC2, a compiler that maps probabilistic models to optimized \acrshort{mcmc}-based accelerators, implemented either as \acrshortpl{asic} or on \acrshortpl{fpga}~\cite{banerjee2019}.  
However, the focus is on general probabilistic models rather than \acrshortpl{bnn}.  
Other dedicated accelerators, including ShiftBNN~\cite{shiftbnn}, $B^2N^2$~\cite{awano2023b2n2}, and VIBNN~\cite{cai2018}, use \acrshortpl{fpga} or \acrshortpl{asic} together with large-scale Gaussian random number generators to accelerate \acrshortpl{bnn}.  

While methods like \acrshort{mcdo} and ensembles provide practical workarounds, their reliance on repeated forward passes limits applicability on embedded devices.  
No existing work demonstrates efficient deployment of \acrshort{svi}-based \acrshortpl{bnn} on such platforms.  
\section{Probabilistic Forward Pass}
\label{sec:pfp:concept}

\begin{figure}  
    \centering
    \includegraphics[width=0.4\textwidth]{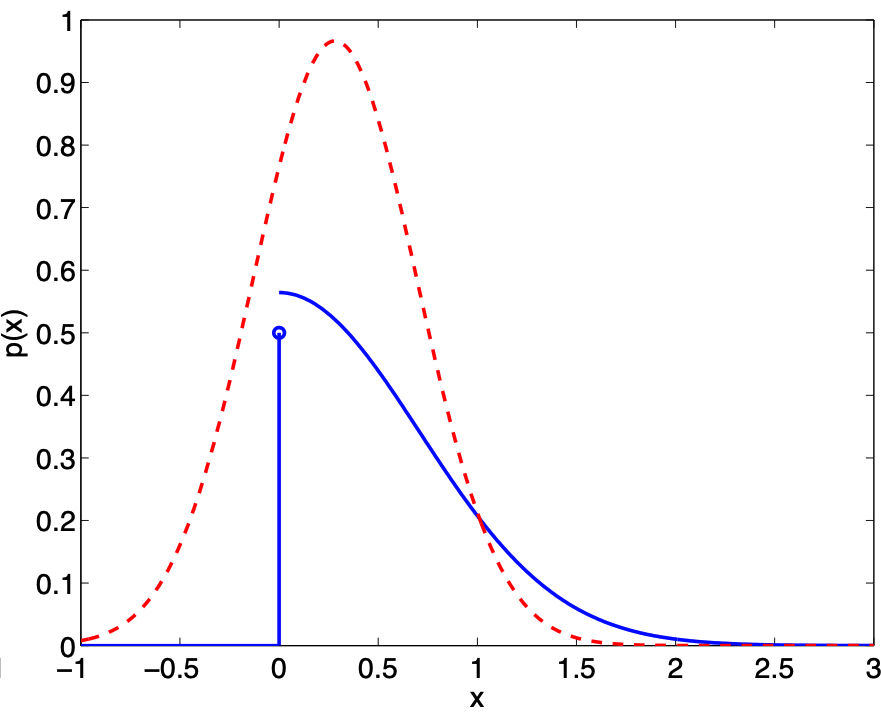}
	\caption{Moment matching in \acrshort{pfp}, shown for the \acrshort{relu} activation.  
	A Gaussian input distribution truncated by the non-linearity (solid) is approximated by a Gaussian (dashed) with matching first and second moment.}
   	\label{fig:pfp:relu_moment_matching}
\end{figure}

The concept of the \acrlong{pfp} was first introduced by Roth~\cite{roth2016pfp,roth2021phd} as an extreme approximation of \acrshort{svi}-based \acrshortpl{bnn}.  
By extending the Gaussian assumption from weights to activations, Roth showed that distributions can be propagated in closed form throughout the network, replacing repeated sampling with deterministic transformations.  
A key idea is \emph{moment matching}: whenever a non-Gaussian distribution arises within the network, it is mapped back to a Gaussian by matching the first two moments.  
This ensures that propagation remains tractable while still capturing predictive uncertainty.  
Figure~\ref{fig:pfp:relu_moment_matching} illustrates this principle for the widely used \acrshort{relu} activation, where a Gaussian activation is truncated by the non-linearity and then transformed back into Gaussian form.  

Consider a fully connected layer with Gaussian inputs characterized by mean and variance $(\mu_{\text{in}}, \sigma_{\text{in}}^2)$ and Gaussian weights with parameters $(\mu_w, \sigma_w^2)$.  
The output distribution is obtained by propagating expectations and variances through the linear transformation.  
The mean is computed by propagating expectations, while the variance reflects uncertainty from both weights and inputs.  
\acrshort{pfp} assumes independence between activations---a mean-field approximation that makes variance propagation tractable.  
Equations~\ref{eq:pfp_dense_scalar:mean} and~\ref{eq:pfp_dense_scalar:variance} show the scalar form for the mean and variance of neuron $i$ in layer $l$:  

\begin{align}
  \mu_{a_i^l} &= \sum_{j=1}^{d_{l-1}} \mu_{w^l_{ij}} \cdot \mu_{x_j^{l-1}} \label{eq:pfp_dense_scalar:mean} \\
	\sigma_{a_i^l}^2 &= \sum_{j=1}^{d_{l-1}} \sigma_{w_{ij}^l}^2 \cdot \mathbb{E} \left[ \left(x^{l-1}_j \right)^2 \right] + \mu_{w_{ij}^l}^2 \cdot \left( \mathbb{E} \left[ \left(x^{l-1}_j \right)^2 \right] - \mu_{x_j^{l-1}}^2 \right). \label{eq:pfp_dense_scalar:variance}
\end{align}

Here the \emph{second raw moment} $\mathbb{E}(x^2)=\mu^2 + \sigma^2$ is applied.  
The variable $d_{l-1}$ denotes the width of the previous layer, i.e., the input dimension to layer $l$.  
This can also be written in vectorized form using mean and variance directly:  

\begin{align}
  \bm{\sigma_a^2} &= \bm{\sigma_w}^2 \cdot \mathbb{E}[\bm{x}^2] + \bm{\mu_w}^2 \cdot \big( \mathbb{E}[\bm{x}^2] - \bm{\mu_x}^2 \big) \label{eq:pfp_dense_vectorized:mE} \\
                  &= \bm{\sigma_w}^2 \cdot \bm{\mu_x}^2 + \bm{\mu_w}^2 \cdot \bm{\sigma_x}^2 + \bm{\sigma_w}^2 \cdot \bm{\sigma_x}^2 \label{eq:pfp_dense_vectorized:mv}
\end{align}

Later implementations support both formulations and select the computationally cheaper one.  
To avoid costly conversions, the outputs of one layer and inputs of the next are kept in a consistent form, either as mean–variance pairs or as mean–second-raw-moment pairs.

A key challenge arises in non-linear activation functions, where Gaussianity is not preserved.  
For the widely used \acrshort{relu}, \acrshort{pfp} therefore applies the moment-matching procedure already introduced above (cf. Figure~\ref{fig:pfp:relu_moment_matching}).  
The truncated Gaussian produced by the non-linearity is projected back into Gaussian form by matching its first and second moments, keeping propagation in closed form.  
The resulting expressions for mean and second raw moment are:  

\begin{align}
  \mu_{x_i^l} &=  \mathbb{E}[x_i^l] = \frac{\mu_{a_i^l}}{2} \left( 1 + \text{erf} \left( \frac{\mu_{a_i^l}}{\sqrt{2 \sigma_{a_i^l}^2}} \right) \right) + \sqrt{\frac{\sigma_{a_i^l}^2}{2 \pi}} \exp \left( - \frac{\mu_{a_i^l}^2}{2 \sigma_{a_i^l}^2} \right) \\
  \mathbb{E} \left[ (x_i^l)^2 \right] &= \frac{\sigma_{a_i^l}^2 + \mu_{a_i^l}^2}{2} \left(  1 + \text{erf} \left( \frac{\mu_{a_i^l}}{\sqrt{2 \sigma_{a_i^l}^2}} \right) \right) + \mu_{a_i^l} \sqrt{\frac{\sigma_{a_i^l}^2}{2 \pi}} \exp \left( - \frac{\mu_{a_i^l}^2}{2 \sigma_{a_i^l}^2} \right). 
  \label{eq:pfp:relu}
\end{align}

Here $\text{erf}(u) = \frac{2}{\sqrt{\pi}}\int_0^u e^{-z^2} dz$ denotes the error function~\cite{roth2021phd}.  
Propagating distributions through successive layers in this way yields the final predictive distribution at the network output.  
During inference, predictions require only a single pass, directly producing expected outputs and uncertainties without ensembles or repeated runs.  
By reformulating the computation to operate in closed form on distributions, \acrshort{pfp} provides a scalable and efficient pathway to deploy \acrshortpl{bnn} in practice.  

\begin{figure}
    \centering
	\begin{subfigure}[t]{0.55\textwidth}
    	\includegraphics[width=\textwidth]{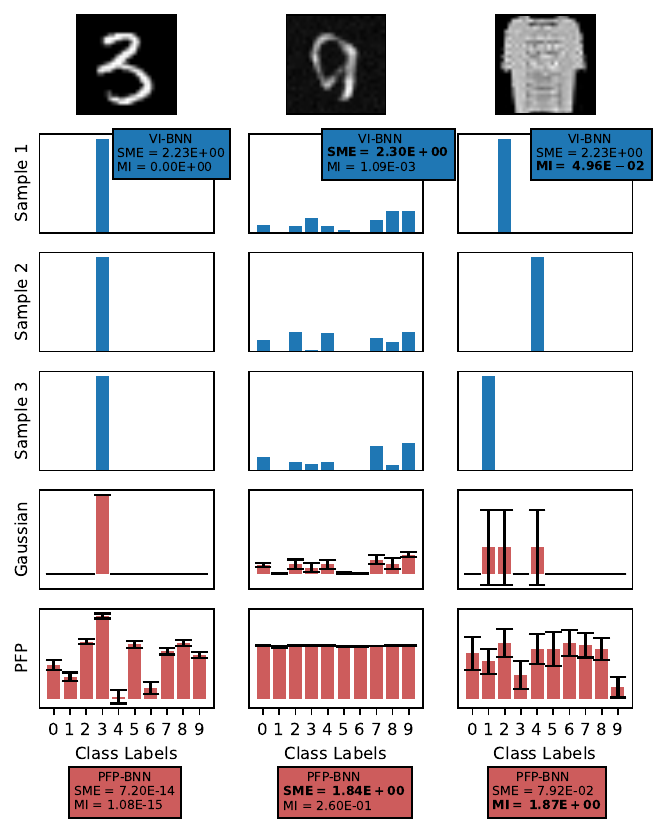}    
		\caption{Illustration of BNN Predictions}
    	\label{fig:pfp:example_samples}
	\end{subfigure}
	\hspace{1em}
	\begin{subfigure}[t]{0.4\textwidth}
    	\includegraphics[width=\textwidth]{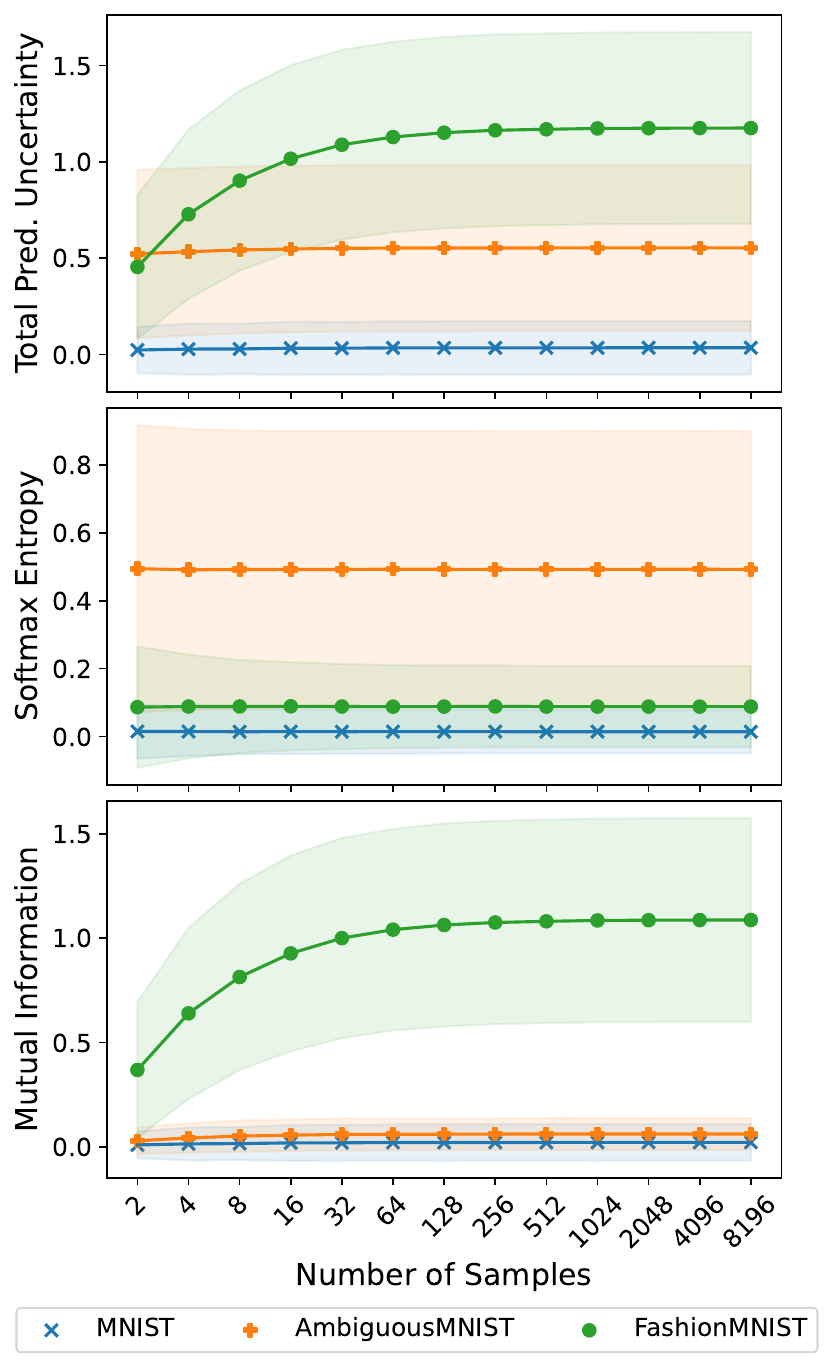}
		\caption{Effect of Number of Samples in SVI-BNNs}
    	\label{fig:pfp:number_of_samples_vi}
	\end{subfigure}
	\caption{Comparison of uncertainty estimation in \acrshort{svi}-based \acrshortpl{bnn} and \acrshort{pfp}.  
	(\subref{fig:pfp:example_samples}) Exemplary predictions on MNIST~\cite{lecun1998mnist}, Ambiguous-MNIST~\cite{Mukhoti2022dirtyMNIST}, and Fashion-MNIST~\cite{xiao2017fashionMnist} as \acrshort{ood} sample.  
	For each dataset, predictions are shown for an \acrshort{svi}-based \acrshort{bnn} (blue samples), its Gaussian approximation, and \acrshort{pfp}.  
	Variability across class probabilities captures aleatoric uncertainty (\acrshort{sme}), while disagreement between samples indicates epistemic uncertainty (\acrshort{mi}).  
	(\subref{fig:pfp:number_of_samples_vi}) Influence of the number of \acrshort{svi} predictive samples on uncertainty metrics, showing that reliable \acrlong{mi} estimates require many samples, whereas \acrshort{pfp} obtains them in a single pass.}
    \label{fig:pfp:example_samples_and_number_of_samples_vi}
\end{figure}

To illustrate the effect of predictive sampling and the advantages of \acrshort{pfp},  
Figure~\ref{fig:pfp:example_samples_and_number_of_samples_vi} compares uncertainty estimation in an \acrshort{svi}-based \acrshort{bnn} and in \acrshort{pfp}.  
Panel~\subref{fig:pfp:example_samples} shows predictions on three example images from MNIST~\cite{lecun1998mnist}, Ambiguous-MNIST~\cite{Mukhoti2022dirtyMNIST}, and Fashion-MNIST~\cite{xiao2017fashionMnist}, the latter serving as an \acrlong{ood} example.  

Recall that \acrfull{sme} quantifies \emph{aleatoric uncertainty} as the average class entropy across predictive samples.  
It reflects variability already present within individual predictions, but is aggregated by averaging over all samples.  
In contrast, \acrfull{mi} measures \emph{epistemic uncertainty} as the disagreement between predictive samples,  
and therefore requires variability across samples to become visible.  
Panel~\subref{fig:pfp:number_of_samples_vi} illustrates this for \acrshort{svi}:  
while \acrshort{sme} stabilizes quickly, reliable \acrshort{mi} estimates—critical for \acrshort{ood} detection—converge only after many samples,  
making \acrshort{svi} computationally demanding.  

\acrshort{pfp}, by contrast, propagates means and variances jointly in closed form.  
This eliminates the need for repeated sampling and produces both uncertainty measures in a single forward pass.  
While this comes at the cost of a less sharp separation between aleatoric and epistemic components,  
it drastically reduces computational overhead compared to sampling-based inference.  

\subsection{Conceptual Limitations}
\label{subsec:limitations}

The efficiency of \acrshort{pfp} comes at the price of assuming Gaussian-distributed logits.  
This structural simplification ensures closed-form propagation, but it limits the ability to represent non-Gaussian predictive distributions.  

As a result, total predictive uncertainty---quantified by Shannon Entropy---is largely preserved,  
but the decomposition into \acrshort{sme} (aleatoric) and \acrshort{mi} (epistemic) can become biased.  
Table~\ref{tab:pfp:limitations} illustrates this effect using artificial regimes with low, high-aleatoric, and high-epistemic uncertainty.  
While the Gaussian approximation faithfully reproduces total uncertainty in all cases,  
errors arise under high epistemic uncertainty, where \acrlong{mi} is underestimated by about $44\,\%$.  

In practice, this means that \acrshort{pfp} preserves overall uncertainty levels,  
but may compromise the disentanglement of aleatoric and epistemic components whenever predictive distributions deviate substantially from Gaussianity.  

\begin{table}
	\centering
	\small
	\caption{Influence of Gaussian approximation on uncertainty metrics.  
	While total uncertainty remains consistent, the separation of aleatoric and epistemic components may degrade under strong epistemic uncertainty.}
	\begin{tabular}{c|c|c|c|c|c|c}
	\textbf{Unc. Regime} & \multicolumn{2}{c|}{\textbf{Total}} & \multicolumn{2}{c|}{\textbf{Softmax Entropy}} & \multicolumn{2}{c}{\textbf{Mutual Information}} \\
	\hline
	& \textbf{True} & \textbf{Gauss} & \textbf{True} & \textbf{Gauss} & \textbf{True} & \textbf{Gauss} \\
	\hline
	Low            & 1.5008 & 1.5007 & 1.5007 & 1.5007 & 0.0     & 0.0     \\ 	
	Aleatoric high & 1.6094 & 1.6094 & 1.6094 & 1.6094 & 0.0     & 0.0     \\ 	
	Epistemic high & 1.6094 & 1.6094 & 1.5002 & 1.5484 & \textbf{0.1092}  & \textbf{0.0610} \\ 	
	\end{tabular}
	\label{tab:pfp:limitations}
\end{table}

\section{PFP Training and Uncertainty Estimation}
\label{sec:pfp:training}

A central advantage of the \acrlong{pfp} is its compatibility with pretrained \acrshort{svi} models.  
It benefits from the relatively fast training of \acrshort{svi} while leveraging established tools for constructing \acrshort{svi}-based \acrshortpl{bnn}.  
Probabilistic programming languages such as Pyro~\cite{bingham2019pyro}, Stan,\footnote{\url{https://mc-stan.org/}} and TensorFlow Probability\footnote{\url{https://github.com/tensorflow/probability}} provide flexible frameworks for designing, training, and evaluating probabilistic models.  
Here, \acrshortpl{bnn} are trained with Pyro \acrshort{svi} and then exported for use with \acrshort{pfp}.  

Two neural architectures are considered in the experiments.  
First, a simple \acrshort{mlp} with a single hidden layer of 100 neurons.  
Second, the LeNet-5~\cite{lecun1998mnist} architecture.  
In both cases, all weights are treated probabilistically with Gaussian priors.  
The mean-field assumption~\cite{farquhar2020meanfield} is applied to simplify training by neglecting correlations between Gaussian weight distributions.  

Training with \acrshort{svi} resembles standard neural network optimization but introduces additional loss terms and hyperparameter sensitivities.  
Due to the multi-objective nature of the \acrlong{elbo}, training is slower and requires careful initialization.  
In our setup, \acrshort{svi}-\acrshortpl{bnn} are trained for $1000$ epochs using the Adam optimizer~\cite{kingma2014adam} with a fixed learning rate of $0.001$.  
Variational posterior weights are initialized with $\mu = 0.08$ and $\sigma = 0.0001$, and a mini-batch size of $100$ is used for training.  

Balancing the expected log-likelihood with the \acrshort{kl} divergence term in the \acrshort{elbo} is non-trivial.  
Instead of a fixed scaling factor, we employ \emph{KL annealing}~\cite{abrol2014deterministic,Zhang2019advancesVI}.  
Here, the KL term weight $A(e)$ is gradually increased from $0$ to $\alpha_{\text{max}} = 0.25$ over the training epochs $e$,  
\begin{equation}  
	\text{ELBO}(q,e) = \mathbb{E}[\log p(\mathcal{D}|w)] - A(e) \cdot \text{KL}(q(w)||p(w|\mathcal{D})).  
\end{equation}  
This strategy improves robustness to initialization and avoids the need for non-probabilistic pretraining.  

The trained means and variances of the weights can be directly consumed by \acrshort{pfp}.  
A conversion from logarithmic to normal parameterization is performed, followed by an uncertainty calibration step.  
This calibration applies a global scaling to the variances and is referred to as the \emph{calibration factor}.  

We hypothesize that the need for calibration arises because moment-based propagation only tracks the first two moments of the distributions and ignores higher-order terms.  
At every non-linear transformation, the resulting non-Gaussian distribution is projected back to Gaussian form, which introduces systematic mismatches compared to full sampling.  
We assume that the main contribution to this miscalibration stems from such moment-matching errors, where higher-order moments beyond the mean and variance are neglected.  
In addition, approximations in the original derivation of \acrshort{pfp}, such as the use of first-order Taylor expansions for nonlinearities~\cite{roth2016pfp}, may further contribute to the discrepancy.  
These effects accumulate across network components, and the calibration factor provides a pragmatic global correction that aligns the propagated variances with the uncertainties observed under sampling-based inference.

To evaluate the ability of \acrshortpl{bnn} to capture both aleatoric and epistemic uncertainty, we use the Dirty-MNIST dataset introduced by \textcite{Mukhoti2022dirtyMNIST} as presented in Chapter~\ref{sec:bnns:datasets}.  

\subsection{Qualitative Performance}
\label{sec:pfp:evaluation}

\begin{figure}  
    \centering  
    \includegraphics[width=0.95\textwidth]{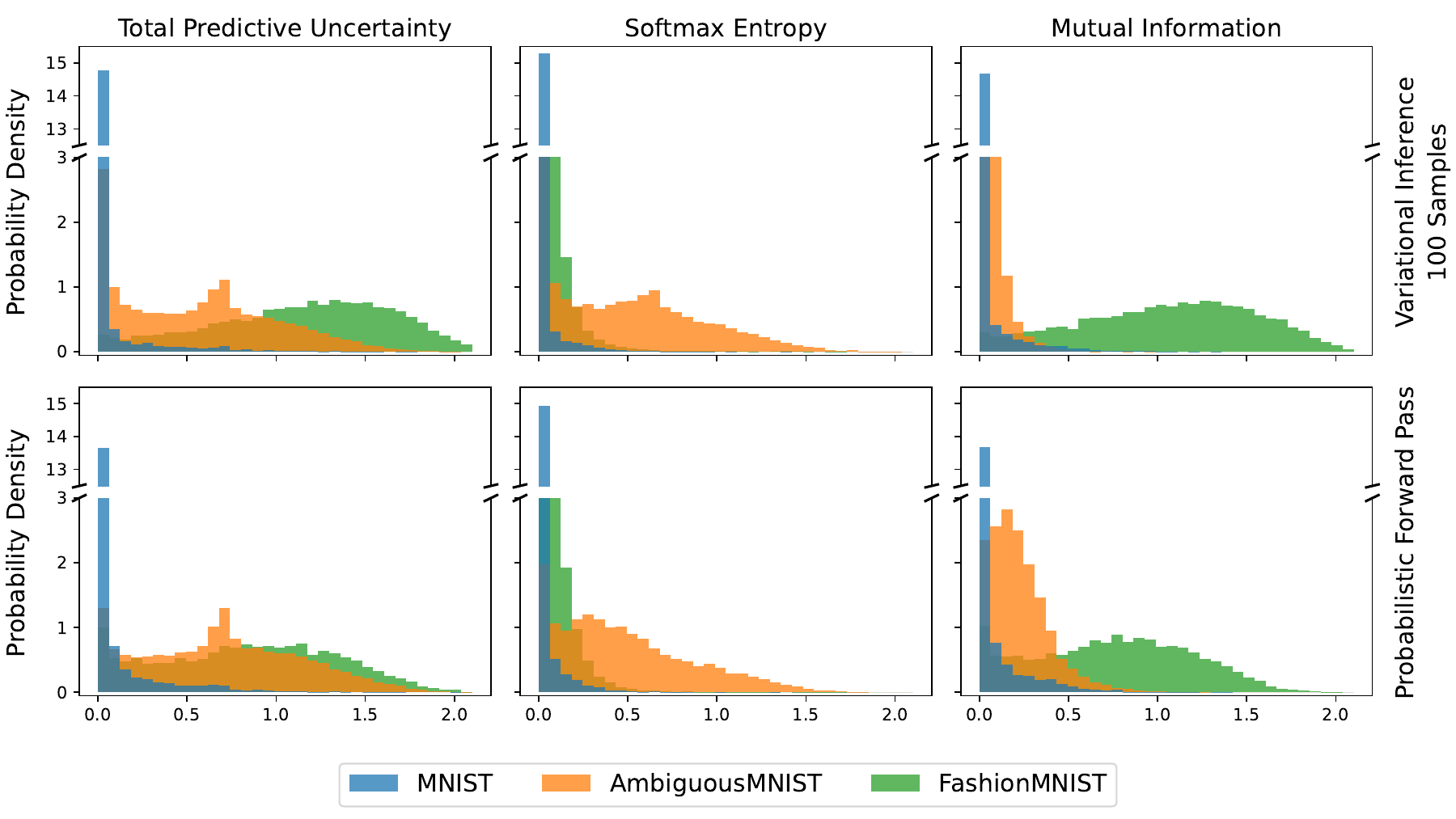}  
	\caption{Comparison of uncertainty predictions obtained with \acrshort{svi} and \acrshort{pfp}.  
	On MNIST~\cite{lecun1998mnist}, both approaches yield low uncertainty, as expected for in-domain data.  
	Ambiguous-MNIST~\cite{Mukhoti2022dirtyMNIST} exhibits higher aleatoric uncertainty, quantified by \acrlong{sme}.  
	Fashion-MNIST~\cite{xiao2017fashionMnist}, used here as an \acrlong{ood} dataset, shows increased epistemic uncertainty, reflected in higher \acrlong{mi}.  
	Overall, both methods correctly assign most samples to their expected domains.}
    \label{fig:pfp:hist_vi_pfp_TuSeMi}  
\end{figure}  

To assess the approximation quality of \acrshort{pfp}, we compare its predictions to sampling-based \acrshort{svi} across the Dirty-MNIST dataset.  
Figure~\ref{fig:pfp:example_samples} shows example predictions, while Figure~\ref{fig:pfp:hist_vi_pfp_TuSeMi} reports the aggregated metrics Shannon Entropy, \acrlong{sme}, and \acrlong{mi}.  
As illustrated in Figure~\ref{fig:pfp:number_of_samples_vi}, the number of samples strongly influences Shannon Entropy and, consequently, \acrlong{mi}.  
Unlike \acrshort{svi}, \acrshort{pfp} does not provide an explicit sampling dimension.  
To ensure comparability, we introduce an artificial sampling procedure based on the \acrshort{pfp}-predicted logit means $\mu_{\text{PFP}}$ and variances $\sigma^2_{\text{PFP}}$.  
Synthetic samples $l_{\text{PFP}}$ are generated as  
\begin{equation}  
    l_{\text{PFP}} \sim \mathcal{N}(\mu_{\text{PFP}}, \sigma^2_{\text{PFP}}).  
\end{equation}  
This \emph{logit sampling} is computationally lightweight, avoiding repeated forward passes.  
It enables the calculation of uncertainty metrics used in sampling-based methods.  
In highly constrained applications, the raw \acrshort{pfp}-predicted variances can be directly employed for decision making.  

Figure~\ref{fig:pfp:hist_vi_pfp_TuSeMi} shows that both \acrshort{svi} and \acrshort{pfp} report elevated uncertainty for Ambiguous-MNIST and Fashion-MNIST, compared to MNIST.  
Both methods capture the expected patterns: increased \acrlong{sme} for ambiguous inputs and increased \acrlong{mi} for \acrshort{ood} inputs.  
Overall, the predictions are consistent with theoretical expectations.  

\begin{figure}
    \centering  
    \includegraphics[width=\textwidth]{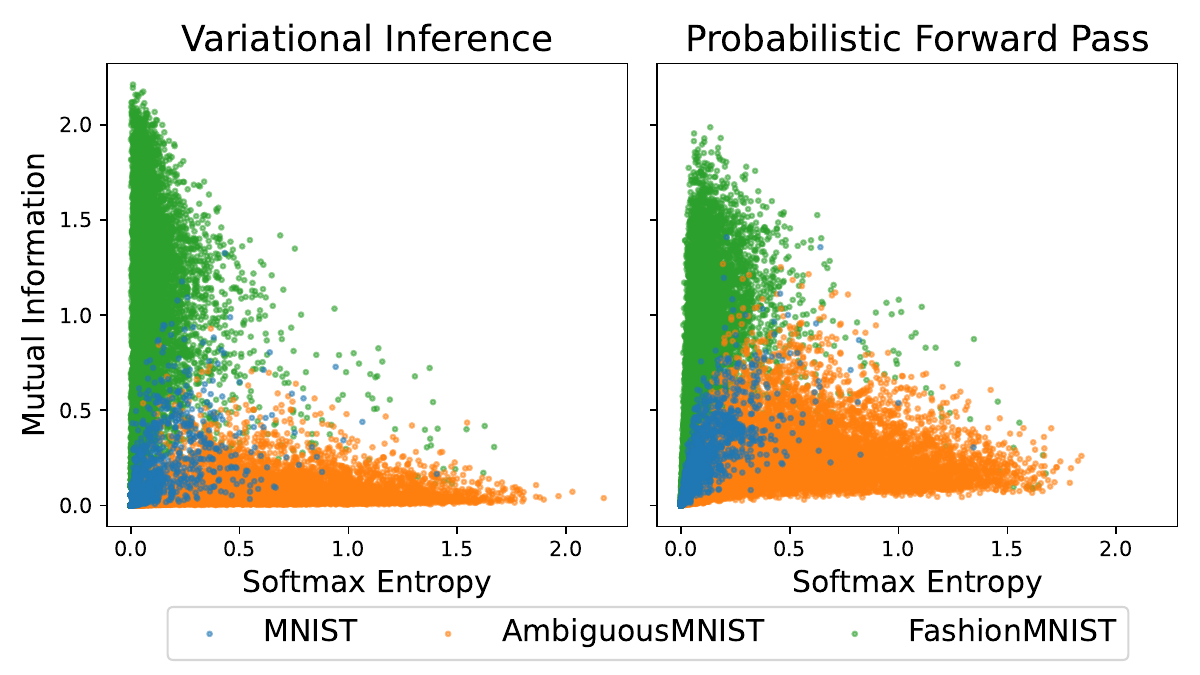}  
	\caption{Comparison of aleatoric and epistemic uncertainty estimates obtained with \acrshort{svi} and \acrshort{pfp} for a LeNet-5 \acrshort{bnn}.  
	The scatter plot displays \acrlong{sme} (aleatoric uncertainty) versus \acrlong{mi} (epistemic uncertainty).  
	While \acrshort{svi} provides a clearer disentanglement of the two uncertainty types, \acrshort{pfp} has some overlap, but still achieves a practically useful degree of separation.}
    \label{fig:pfp:scatter_vi_pfp_SeMi}  
\end{figure}

A more detailed analysis is presented in Figure~\ref{fig:pfp:scatter_vi_pfp_SeMi}, which plots \acrlong{sme} against \acrlong{mi} across all samples.  
Here, \acrshort{svi} achieves cleaner separation of aleatoric and epistemic uncertainty.  
Nevertheless, \acrshort{pfp} provides a practically sufficient distinction in most cases.
In rare edge cases both measures take high values, reducing separability.

\begin{table}
    \centering
	\caption{Comparison of \acrshort{svi} and \acrshort{pfp}-based \acrshortpl{bnn} on Dirty-MNIST~\cite{Mukhoti2022dirtyMNIST}.}
    \label{tab:auroc}
    \small
    \begin{tabular}{l|l|c|c}
        \textbf{Metric} & \textbf{Method} & \textbf{MLP} & \textbf{LeNet-5} \\ \hline
        \textbf{Calibration Factor} 
            & \acrshort{pfp} & 0.3 & 0.4 \\ \hline
        \multirow{2}{*}{\textbf{Accuracy}} 
            & \acrshort{svi} & 96.3\% & 98.7\% \\
            & \acrshort{pfp} & 96.3\% & 98.9\% \\ \hline
        \multirow{2}{*}{\textbf{AUROC}} 
            & \acrshort{svi} & 0.812 & 0.986 \\
            & \acrshort{pfp} & 0.858 & 0.966 \\
    \end{tabular}
\end{table}

Table~\ref{tab:auroc} compares the two methods on Dirty-MNIST.  
Both achieve similar predictive accuracy and \acrshort{auroc}, with \acrshort{pfp} requiring only a single forward pass.  
The influence of network architecture is visible, as the convolutional model achieves better accuracy and separation than the \acrshort{mlp}.  
In summary, \acrshort{pfp} delivers uncertainty estimates close to \acrshort{svi} while offering significantly improved efficiency.  

\section{\acrshort{pfp} Operator Library with \acrshort{tvm}}
\label{sec:operator_library}

\acrshort{tvm} provides multiple internal languages and intermediate representations, including TensorIR~\cite{feng2023tensorir}, \acrfull{te}, TVMScript, and Relax~\cite{Lai2023relax}.  
These abstractions are essential for implementing custom operators.  
\acrlongpl{te} define computation rules in a compact way, while TensorIR organizes computations as modular \emph{blocks}, enabling fine-grained scheduling and optimizations.  
Relax~\cite{Lai2023relax}, the successor of Relay~\cite{Roesch2018relay}, serves as a high-level intermediate representation, supporting dynamic shapes, control flow, and integration with TensorIR.\footnote{\url{https://tvm.apache.org/docs/deep_dive/relax/learning.html}}  
TVMScript, a Python-based frontend, allows direct definition and modification of TensorIR and Relax programs.  

To implement a custom operator, developers specify the computation in TE and generate IRModules via the BlockBuilder API.  
BlockBuilder creates primitive functions from TE expressions, which are then connected through Relax and optimized with TensorIR scheduling.\footnote{\url{https://mlc.ai/docs/get_started/tutorials/quick_start.html}}  
This workflow enables efficient execution of specialized operators across diverse hardware platforms while minimizing implementation complexity.  

\subsubsection*{Operating on Tuples}  
Neural network operators frequently take multiple input tensors, but operators producing multiple outputs are relatively uncommon.  
\acrshort{pfp} introduces a particular requirement, as both mean and variance must be propagated through the network.  
\acrshort{tvm} follows the principle of \emph{one operator = one compute rule}, meaning each operator executes a single sequential computation without divergence.  
As a result, \acrshort{pfp} operations may be split into separate operators, e.g., one for the mean and one for the variance.  
However, this design increases interconnection overhead, complicates network construction, and prevents reuse of shared sub-terms between the mean and variance paths.  
A joint formulation, in which both quantities are computed together, allows reuse of intermediate results and avoids redundant computations.  
Figure~\ref{fig:pfp:lib:operator_implementations} shows that such joint operators consistently outperform separate implementations by improving data reuse and memory locality.  

\subsubsection*{Variance and Second Raw Moment}  
\begin{figure}
    \centering
    \includegraphics[width=0.7\textwidth]{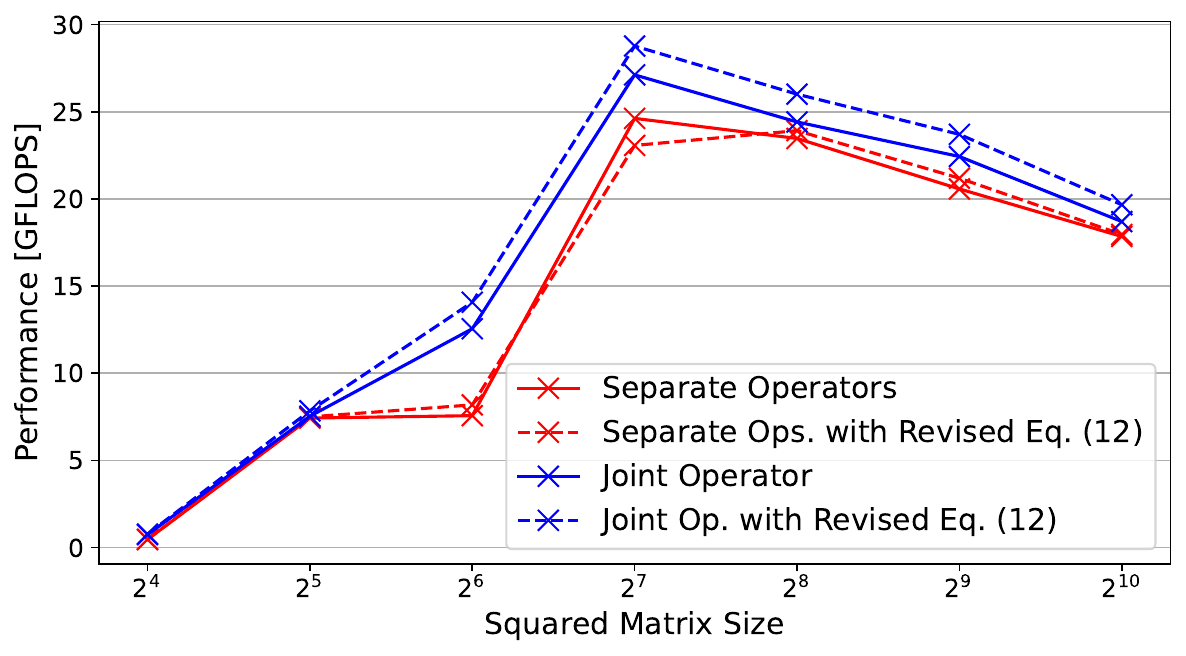}    
	\caption{Comparison of operator implementations on an ARM Cortex-A72.  
	The results demonstrate the performance gains from reformulating Equation~\ref{eq:pfp_dense_scalar:variance} into Equation~\ref{eq:pfp_dense_scalar:E:variance},  
	and from employing joint operators that combine mean and variance paths instead of implementing them separately.}  
    \label{fig:pfp:lib:operator_implementations}
\end{figure}

The original formulation of \acrshort{pfp} operators is based on mean and variance inputs and outputs.  
However, reformulating Equation~\ref{eq:pfp_dense_scalar:variance} in terms of second raw moments improves reuse and reduces conversions.  
In this form, the variance of activations in a dense layer becomes  
\begin{equation}
	\sigma_{a_i^l}^2 = \sum_{j=1}^{d_{l-1}} \mathbb{E}\!\left[\left(w_{ij}^l\right)^2\right] \cdot \mathbb{E}\!\left[\left(x_{j}^{l-1}\right)^2\right] - \left(\mu_{w_{ij}^l} \cdot \mu_{x_j^{l-1}}\right)^2,  
	\label{eq:pfp_dense_scalar:E:variance}
\end{equation}  
where the pre-computed second raw moments of the weights and activations can be directly reused.  
This eliminates conversions from activation outputs, which already produce second raw moments by design.  
The resulting tuple-based operator is more cache-efficient and reduces overall runtime.  
Figure~\ref{fig:pfp:lib:operator_implementations} illustrates the performance benefits of the second raw moment formulation and joint operators.  

When consecutive layers differ in their representation of variances and second raw moments, conversion is straightforward using $\mathbb{E}(x^2) = \mu^2 + \sigma^2$.  
However, performing repeated conversions across layers is wasteful.  
To address this, the operator implementation provides a configurable conversion function, while ensuring consistency between layers remains the responsibility of the model designer.  
Weights must also follow this convention, being stored either as means and variances (see Equation~\ref{eq:pfp_dense_vectorized:mv}) or as means and second raw moments (see Equation~\ref{eq:pfp_dense_scalar:variance}).  

By default, compute layers such as dense and convolutional layers expect second raw moments as inputs and produce variances as outputs.  
Conversely, activation functions take variances as inputs and produce second raw moments.  
This convention ensures compatibility between compute and activation functions operators.  
Additional layers, such as Max Pooling, which consume and produce variances, require explicit conversion at their interfaces.  

A special case occurs in the first network layer, where no input variance is available.  
In this case, the forward propagation simplifies to  
\begin{align}  
	\mu_{a_i^l} &= \sum_{j=1}^{d_{l-1}} \mu_{w^l_{ij}} \cdot \mu_{x_j^{l-1}}, \\
	\quad
	\sigma_{a_i^l}^2 &= \sum_{j=1}^{d_{l-1}} \sigma_{w_{ij}^l}^2 \cdot \mu_{x_j^{l-1}}^2.  
	\label{eq:pfp_dense_scalar:det_input}
\end{align}  
Here, weight variances are required explicitly.  
For subsequent layers, storing weights as second raw moments avoids additional conversions.  
Furthermore, compute layers support three bias configurations: no bias, deterministic bias, and probabilistic bias with variances.  

In summary, integrating custom \acrshort{pfp} operators into \acrshort{tvm} requires specific design considerations.  
Our analysis shows that combining joint operator implementations with second raw moment formulations yields the best efficiency across tested architectures.  

\section{Optimizing for Performance}
\label{sec:tuning}

The \acrshort{pfp} implementation introduced in the previous Section already enables functional \acrshortpl{bnn} with uncertainty estimation at a fraction of the cost of sampling-based methods.  
Nevertheless, further performance can be gained through implementation-level and hardware-aware optimizations.  
We first profile operator costs to identify performance bottlenecks.  
Based on these insights, we apply manual optimizations tailored to dense and pooling operators.  
Subsequently, we evaluate the effectiveness of automatic tuning frameworks provided by the \acrshort{tvm} compiler.  
Finally, we benchmark the tuned implementations across different \acrshortpl{cpu} and compare them against \acrshort{svi}-based \acrshortpl{bnn}.  

\subsection{Profiling Operators}

Optimizing for performance requires identifying operators that dominate execution time.  
\acrshort{tvm} provides three execution modes for compiled binaries: standard execution, benchmarking with averaging for high precision, and profiling, which reports latency on a per-operator basis.  
These profiling capabilities enable both quantitative evaluation of optimization strategies and visualization of operator cost distribution.  

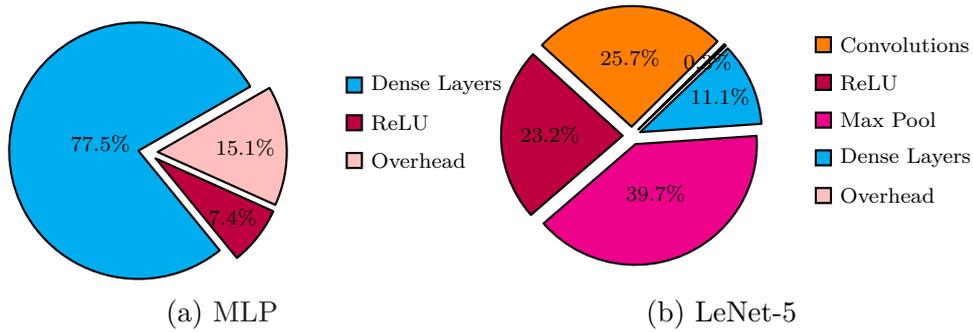
\begin{figure}
    \centering
	\begin{subfigure}[b]{0.45\textwidth}
		\adjustbox{valign=t}{
			\begin{tikzpicture}[scale=0.85]
				\scriptsize
				\pie[sum=100, text=legend, radius=2, explode=0.15, rotate=30, color={cyan, purple, pink}]
				{77.5/Dense Layers, 7.4/ReLU, 15.1/Overhead}
			\end{tikzpicture}
		}
		\caption{\acrshort{mlp}}
    	\label{fig:tune:profiling_pie:mlp}
    \end{subfigure}
	\begin{subfigure}[b]{0.45\textwidth}
		\adjustbox{valign=t}{
			\begin{tikzpicture}[scale=0.8]
				\scriptsize
				\pie[sum=100, text=legend, radius=2, explode=0.15, rotate=45, color={orange, purple, magenta, cyan, pink}]
				{25.7/Convolutions, 23.2/ReLU, 39.7/Max Pool, 11.1/Dense Layers, 0.3/Overhead}
			\end{tikzpicture}
		}
		\caption{LeNet-5}
    	\label{fig:tune:profiling_pie:lenet}
    \end{subfigure}
	\caption{Distribution of execution time across operator types for \acrshort{pfp}-based \acrshortpl{bnn}, measured on a Cortex-A72 with a mini-batch size of 10.  
	For the \acrshort{mlp}, dense layers dominate runtime.  
	For LeNet-5, latency is more evenly distributed, and operators considered simple in deterministic settings, such as \acrshort{relu} and Max Pooling, contribute significant overhead}  
    \label{fig:tune:profiling_pie:bs10}
\end{figure}

Figure~\ref{fig:tune:profiling_pie:bs10} shows the runtime share of different operator types.  
For the \acrshort{mlp}, dense layers are the dominant cost.  
For LeNet-5, latency is more balanced, with \acrshort{relu} and Max Pool operators accounting for a substantial portion of total runtime.  
This highlights that seemingly simple operators can become computationally expensive when propagating distributions rather than scalars.  

\subsection{Manual Optimizations}

The profiling results indicate that dense layers are the primary bottleneck in the \acrshort{mlp}.  
We therefore target the dense operator with optimization techniques commonly used for matrix-matrix multiplication.  
These include tiling, loop reordering, loop unrolling, vectorization, and parallelization.  

\begin{table}%
	\centering
	\small
	\caption{Evaluation of manual optimization techniques for the \acrshort{pfp} dense operator.  
	Measurements were performed on a Cortex-A72 using a 3-layer \acrshort{mlp} with mini-batch size 10.}
	\begin{threeparttable}
	\begin{tabular}{c|c|r|r|r}
		\multicolumn{2}{c|}{\textbf{Optimizations}}														& \multicolumn{2}{c|}{\textbf{Latency}}						& \textbf{Speedup}					\\ \hline
		\textbf{Name}													& \textbf{Other Opt.}			& \textbf{without Opt.}	& \textbf{with Opt.}				& 									\\ \hline 
		Baseline (no tuning)											& OFF							& 3.760\,ms 			& - 								& - 								\\ %
		Baseline (min. tuning) 											& OFF							& 3.681\,ms				& -									& - 								\\ %
		\hline
		Tiling\tnote{1}													& OFF 							& 3.672\,ms				& 0.747\,ms							& 4.91$\times$							\\ %
		Loop Reordering													& OFF 							& 3.681\,ms				& 1.940\,ms							& 1.90$\times$							\\ %
		Vectorization													& OFF 							& 3.681\,ms				& 8.837\,ms							& 0.42$\times$						\\ %
		Parallelization													& OFF 							& 3.681\,ms				& 0.729\,ms							& 5.05$\times$							\\ %
		Loop Unrolling													& OFF 							& 3.681\,ms				& 1.967\,ms							& 1.87$\times$							\\%
		\hline
		Tiling\tnote{1}													& ON 							& 0.754\,ms				& 4.237\,ms							& 0.18$\times$							\\ %
		Loop Reordering													& ON\tnote{2}  					& 0.750\,ms				& 0.743\,ms							& 1.01$\times$							\\ %
		Vectorization													& ON\tnote{2}  					& 0.759\,ms				& 0.743\,ms							& 1.02$\times$							\\ %
		Parallelization													& ON\tnote{2}  					& 1.953\,ms				& 0.743\,ms							& 2.63$\times$							\\ %
		Loop Unrolling													& ON\tnote{2}  					& 3.042\,ms				& 0.743\,ms							& 4.09$\times$							\\ %
		\hline
		All Optimizations 												& ON\tnote{2}  					& 3.760\,ms				& 0.743\,ms 						& 5.06$\times$						\\ %
	\end{tabular}
	\begin{tablenotes}
		\item[1] Without stochastic tuning.
		\item[2] All optimizations in use except tiling.
	\end{tablenotes}
	\end{threeparttable}
	\label{tab:dense_optimizations:bs10}
\end{table}

Table~\ref{tab:dense_optimizations:bs10} shows the effect of individual and combined optimizations.  
Some techniques, such as vectorization, degrade performance when used in isolation, as they require loop structures to be reordered first.  
Loop unrolling and parallelization are the most effective optimizations for the \acrshort{pfp} dense operator.  
When combined, they yield a speedup of more than $5\times$.  

Tiling requires separate evaluation.  
Applied independently with hand-tuned tile sizes, it yields strong performance gains.  
However, tiling is the only optimization that does not support stochastic tuning.  
Since the other optimizations benefit considerably from stochastic tuning, enabling tiling disables this option.  
Consequently, applying all optimizations including tiling but without stochastic tuning performs worse than tiling alone.  
The best results are obtained by combining all other optimizations with stochastic tuning while excluding tiling.  

\paragraph{Max Pool Operator}  
For LeNet-5, the generic Max Pooling operator introduced by~\cite{roth2021phd}, implemented as a reduction, proved inefficient.  
To address this, we implemented a specialized vectorized variant with fixed kernel size $k=2$.  
As shown in Table~\ref{tab:tune:maxpool:bs10}, automatic schedules fail to improve this operator and in some cases even degrade performance.  
Consequently, the custom Max Pool implementation is excluded from automatic tuning.  

\begin{table} %
	\centering
	\small
	\caption{Evaluation of Max Pool implementations for LeNet-5 on a Cortex-A72 with mini-batch size 10.  
	The specialized vectorized operator outperforms the generic reduction-based implementation, while automatic schedules fail to improve performance.}
	\begin{tabular}{c|c|c|r|r}
		\textbf{Arch.}  						& \textbf{Implementation}						& \textbf{Auto-tuning}					& \multicolumn{2}{c}{\textbf{Latency}}								 	\\ \hline 
												&												&										& \textbf{Max Pools}				& \textbf{Entire \acrshort{nn}}		 	\\ \hline
		LeNet-5 								& Generic Max Pool								& No									& 12.09\,ms							&	29.13\,ms					 	\\ %
		LeNet-5									& Generic Max Pool								& All operators							&  5.04\,ms							&	10.74\,ms					 	\\ %
		LeNet-5									& Generic Max Pool								& All except Max Pool					& 11.92\,ms							&	17.82\,ms					 	\\ %
		\hline
		LeNet-5 								& Vect. Max Pool 								& No									&  3.54\,ms							&	21.10\,ms					 	\\ %
		LeNet-5 								& Vect. Max Pool 								& All operators							& 27.28\,ms							&	33.42\,ms					 	\\ %
		LeNet-5 								& Vect. Max Pool 								& All except Max Pool					&  3.69\,ms							&	 9.79\,ms					 	\\ %
	\end{tabular}
	\label{tab:tune:maxpool:bs10}
\end{table}

\subsection{Automatic Optimizations}

Beyond manual schedules, \acrshort{tvm} provides advanced automatic tuning frameworks~\cite{chen2018tvm,Zheng2020ansor,Wu2023autotuning,Shao2022tvmMetaScheduler}.  
The \emph{Meta Scheduler}~\cite{Shao2022tvmMetaScheduler} automatically explores large optimization spaces by generating and benchmarking candidate schedules.  
This approach is slower than expert-crafted tuning but typically achieves comparable performance and requires no manual effort.  
Applying Meta Scheduler to the \acrshort{pfp} dense operator achieves latencies nearly identical to hand-tuned schedules ($0.742$\,ms versus $0.743$\,ms).  
We therefore rely on it in subsequent experiments.  

Table~\ref{tab:tune:profiling:bs10} summarizes profiling results for the \acrshort{mlp} and LeNet-5 before and after tuning.  
Dense and convolution layers benefit substantially from optimization, delivering the largest performance improvements.  

\begin{table} %
	\centering
	\small
	\caption{Profiling of \acrshort{pfp}-based neural architectures on a Cortex-A72 with mini-batch size 10.  
	The largest performance gains after tuning are achieved in dense and convolution layers.}
	\begin{threeparttable}
	\begin{tabular}{c|c|r|r|r|r|r}
		\textbf{Arch.}		& \textbf{Layer}		   	& \multicolumn{2}{c|}{\textbf{Baseline}} 	& \multicolumn{2}{c|}{\textbf{Tuned Impl.}} 						& \textbf{Speedup}			\\ \hline 
									&						   	& \textbf{Latency}					& \textbf{Fraction} 			& \textbf{Latency}			& \textbf{Fraction} 	&							\\ \hline 
		\multirow{5}*{MLP}			& Dense 1				   	&  2.931\,ms						&  62.8\,\%						&  0.642\,ms				&  33.7\,\%				&  4.6$\times$		\\ %
									& Dense 2				   	&  0.570\,ms						&  12.2\,\%						&  0.125\,ms				&   6.6\,\%				&  4.6$\times$		\\ %
									& ReLU\tnote{1}			   	&  0.195\,ms						&   4.2\,\%						&  0.185\,ms				&   9.7\,\%				&  1.1$\times$		\\ %
									& Dense 3				   	&  0.063\,ms						&   1.3\,\%						&  0.036\,ms				&   1.9\,\%				&  1.8$\times$		\\ %
									& Sum 						&  3.846\,ms						&  82.4\,\%						&  1.078\,ms				&  56.5\,\%				&  3.6$\times$		\\ %
									& Entire Model				&  4.668\,ms						& 100.0\,\%						&  1.908\,ms				& 100.0\,\%				&  2.4$\times$		\\ %
		\hline                                                                                                                                                                                
		\multirow{13}*{LeNet-5}	 	& Conv2d 2				   	&  7.207\,ms					 	&  30.4\,\%						&  1.509\,ms				&  12.4\,\% 			&  4.8$\times$		\\ %
								 	& ReLU 1				   	&  4.133\,ms						&  17.4\,\%						&  2.153\,ms				&  17.7\,\% 			&  1.9$\times$		\\ %
								 	& Dense 1				   	&  2.791\,ms						&  11.8\,\%						&  0.516\,ms				&   4.2\,\% 			&  5.4$\times$		\\ %
									& Max Pool 1\tnote{1,2}	   	&  2.780\,ms						&  11.7\,\%						&  2.807\,ms				&  23.1\,\% 			&  1.0$\times$		\\ %
								 	& ReLU 2				   	&  1.541\,ms						&   6.5\,\%						&  0.980\,ms				&   8.1\,\% 			&  1.6$\times$		\\ %
								 	& Conv2d 1				   	&  0.949\,ms						&   4.0\,\%						&  0.552\,ms				&   4.5\,\% 			&  1.7$\times$		\\ %
									& Max Pool 2\tnote{2}	   	&  0.854\,ms						&   3.6\,\%						&  0.902\,ms				&   7.4\,\% 			&  0.9$\times$		\\ %
								 	& Dense 2				   	&  0.589\,ms						&   2.5\,\%						&  0.111\,ms				&   0.9\,\% 			&  5.3$\times$		\\ %
								 	& ReLU 3				   	&  0.168\,ms						&   0.7\,\%						&  0.071\,ms				&   0.6\,\% 			&  2.4$\times$		\\ %
								 	& ReLU 4				   	&  0.107\,ms						&   0.5\,\%						&  0.051\,ms				&   0.4\,\% 			&  2.1$\times$		\\ %
								 	& Dense 3				   	&  0.057\,ms						&   0.2\,\%						&  0.027\,ms				&   0.2\,\% 			&  2.1$\times$		\\ %
									& Sum 		 				& 21.751\,ms						&  91.8\,\%						& 10.226\,ms				&  84.1\,\% 			&  2.1$\times$		\\ %
									& Entire Model	   			& 23.698\,ms						& 100.0\,\%						& 12.166\,ms				& 100.0\,\% 			&  1.9$\times$		\\ %

	\end{tabular}
	\begin{tablenotes}
		\item[1] Layers present multiple times in network
		\item[2] Layers excluded from tuning
	\end{tablenotes}
	\end{threeparttable}
	\label{tab:tune:profiling:bs10}
\end{table}

\subsection{Evaluation and Performance Gain}
\begin{table}
	\centering
	\tiny
	\caption{Latency and speedup comparison of deterministic, \acrshort{svi}, and \acrshort{pfp}-based networks on three different embedded ARM processors, using vectorized Max Pool.  
	While \acrshort{pfp} incurs some overhead compared to deterministic inference, it achieves orders-of-magnitude speedups over sampling-based \acrshort{svi}, making uncertainty-aware inference practical on embedded devices.}
	\begin{tabular}{c|c|c|r|r|r|r|r|r}
		\textbf{Arch.}   	& \textbf{\acrshort{bs}} 	& \textbf{Processor} 				& \multicolumn{2}{c|}{\textbf{Deterministic NN}}			& \multicolumn{1}{c|}{\textbf{SVI}}	& \multicolumn{2}{c|}{\textbf{PFP}}				& \textbf{Speedup} 	\\ \hline
		  						&							& 									& \textbf{not tuned} 	& \textbf{tuned} 					&  						& \textbf{not tuned}	& \textbf{tuned} 							\\ \hline							%
		\multirow{3}*{MLP}		&	\multirow{3}*{10}		&	Cortex-A53			 			&  14.02\,ms			&  0.933\,ms						& 		-				&  15.26\,ms			&   4.989\,ms			&	  				\\ %
								&							&	Cortex-A72			 			&   4.59\,ms			&  0.186\,ms						&  734.74\,ms			&   3.75\,ms			&   0.742\,ms			&	 990.2$\times$	\\ %
								&							&	Cortex-A76			 			&   1.64\,ms			&  0.071\,ms						&  307.52\,ms			&   1.89\,ms			&   0.341\,ms			&	 901.8$\times$	\\ %
		\hline                                                                                                                                                                                                              		            
		\multirow{3}*{MLP}		&	\multirow{3}*{100}	   	&	Cortex-A53			 			& 137.80\,ms			&  6.565\,ms						& 		-				& 147.61\,ms			&  15.358\,ms			&	      			\\ %
								&						   	&	Cortex-A72			 			&  45.81\,ms			&  1.134\,ms						&  775.32\,ms			&  36.33\,ms			&   5.182\,ms			&	 149.6$\times$	\\ %
								&						   	&	Cortex-A76			 			&  16.30\,ms			&  0.230\,ms						&  306.89\,ms			&  18.60\,ms			&   1.200\,ms			&	 255.7$\times$	\\ %
		\hline                   				                                                                                                                                                                            		            
		\multirow{3}*{LeNet-5}	&	\multirow{3}*{10}		&	Cortex-A53			 			&  21.14\,ms			&  4.726\,ms						& 		-				&  76.09\,ms			&  35.159\,ms			&	      			\\ %
								&							&	Cortex-A72			 			&   6.89\,ms			&  0.754\,ms						& 1196.42\,ms			&  21.23\,ms			&  10.022\,ms			&	 119.4$\times$	\\ %
								&							&	Cortex-A76			 			&   3.16\,ms			&  0.347\,ms						&  801.40\,ms			&   9.63\,ms			&   3.897\,ms			&	 205.6$\times$	\\ %
		\hline                   				                                                                                                                                                                            		            
		\multirow{3}*{LeNet-5}	&	\multirow{3}*{100}		&	Cortex-A53			 			& 209.28\,ms			& 41.697\,ms						& 		-				& 801.33\,ms			& 383.680\,ms			&	      			\\ %
								&							&	Cortex-A72			 			&  70.08\,ms			&  9.524\,ms						& 2708.16\,ms			& 240.94\,ms			& 116.330\,ms			&	  23.3$\times$	\\ %
								&							&	Cortex-A76			 			&  31.51\,ms			&  3.131\,ms						& 2488.73\,ms			& 119.76\,ms			&  45.039\,ms			&	  55.3$\times$	\\ %
	\end{tabular}
	\label{tab:performance:bs10}
\end{table}

\begin{figure}
    \centering
	\begin{subfigure}{0.4\textwidth}
    	\centering
    	\includegraphics[width=\textwidth]{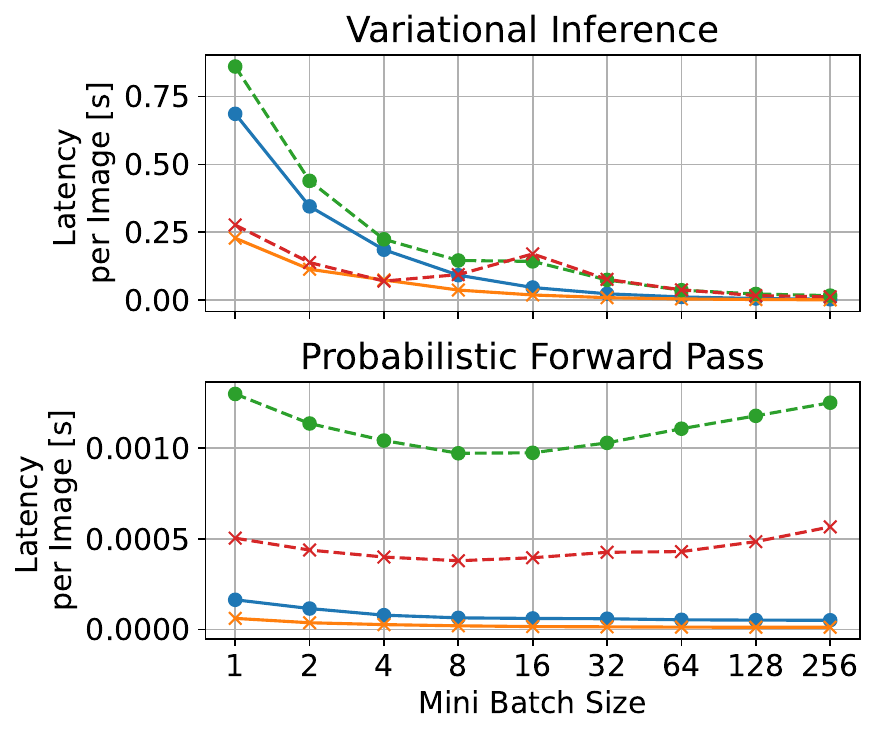}    
		\caption{Latency per image}
    	\label{fig:pfp:tune:speedup_and_latency_batchsize:latency}
	\end{subfigure}
	\begin{subfigure}{0.58\textwidth}
    	\centering
    	\includegraphics[width=\textwidth]{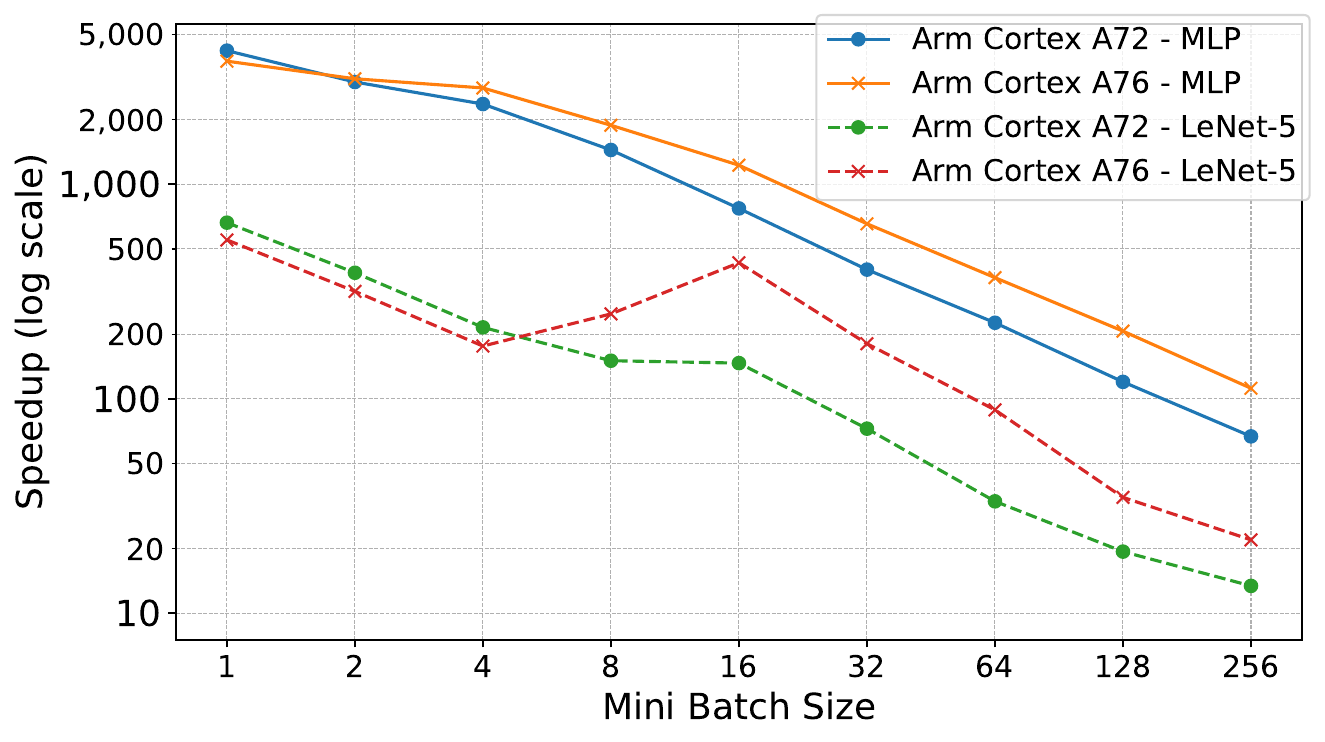}    
		\caption{Relative speedup of \acrshort{pfp} over \acrshort{svi}}
    	\label{fig:pfp:tune:speedup_and_latency_batchsize:speedup}
	\end{subfigure}
\caption{Latency and speedup in relation to mini-batch size.  
\acrshort{svi}-based \acrshortpl{bnn}, evaluated with 30 samples, show high per-image latency and poor scalability at small batch sizes.  
In contrast, \acrshort{pfp} maintains stable latency across all sizes due to targeted tuning.  
As a result, speedups range from tens to over four thousand, with the strongest gains appearing for batch size~1, which is a common case for real-time embedded applications.}
    \label{fig:pfp:tune:speedup_and_latency_batchsize}
\end{figure}

Table~\ref{tab:performance:bs10} and Figure~\ref{fig:pfp:tune:speedup_and_latency_batchsize} highlight the central result of this chapter:  
\acrshort{pfp}, combined with compiler-based optimizations, reduces inference cost by several orders of magnitude compared to sampling-based \acrshort{svi}.  
Across ARM processors, speedups reach on average $574\times$ for the \acrshort{mlp} and $101\times$ for LeNet-5, despite \acrshort{pfp} requiring more complex operators and doubling both parameters and activations relative to deterministic inference.  

The importance of these gains becomes most evident under conditions typical for embedded systems, where real-time operation often requires small mini-batch sizes.  
In this regime, \acrshort{svi}-based \acrshortpl{bnn}, implemented in Pyro with 30 samples---already a minimal configuration for meaningful uncertainty estimation (see also Figure~\ref{fig:pfp:number_of_samples_vi})---still incur substantial latency.  
Figure~\ref{fig:pfp:tune:speedup_and_latency_batchsize:latency} shows that their runtime scales poorly as batch sizes decrease.  
By contrast, \acrshort{pfp} benefits directly from compiler support: it is recompiled and optimized for each mini-batch size individually, allowing it to maintain stable latency across configurations.  
Only minor fluctuations remain, which can be attributed to cache effects or \acrshort{simd} alignment.  
The resulting speedups, shown in Figure~\ref{fig:pfp:tune:speedup_and_latency_batchsize:speedup}, range from $13\times$ to $112\times$ at batch size 256, and increase dramatically to between $550\times$ and $4200\times$ for batch size 1---a common real-time scenario.  

Altogether, these experiments demonstrate that the combination of probabilistic approximation and compiler-based code generation makes efficient, uncertainty-aware inference feasible on resource-constrained embedded \acrshortpl{cpu}---a capability that was previously out of reach for \acrshort{svi}-based \acrshortpl{bnn}.

\section*{Summary}
\label{sec:conclusion}

Although \acrlongpl{bnn} offer a principled framework for uncertainty estimation, their deployment on embedded systems remains constrained by the computational demands of sampling-based inference.  
This chapter demonstrated how combining algorithmic approximation with compiler-level optimization can overcome these limitations.  
By replacing repeated stochastic evaluations with a single closed-form forward pass, the \acrlong{pfp} enables efficient execution of \acrshort{svi}-trained networks on embedded ARM \acrshortpl{cpu}.  
While the Gaussian assumption limits expressiveness, it yields substantial speedups whenever the approximation remains adequate.  

To realize this approach in practice, we extended the deep learning compiler \acrshort{tvm} with specialized probabilistic operators and applied both targeted scheduling and automatic tuning.  
The resulting implementation achieved speedups of up to $4200\times$ on ARM processors while preserving accuracy and reliable uncertainty estimation under tight resource constraints.  
Together, these contributions establish a practical pathway from Bayesian approximation to deployment, making probabilistic inference feasible on embedded hardware.  

While \acrshort{pfp} provides an analytic route to efficient uncertainty estimation, the next chapter explores a complementary strategy based on ensemble methods.  
These approaches trade analytical simplicity for representational flexibility, capturing multiple modes of the predictive distribution with high parallel efficiency.

\chapter{Ensemble Methods for Practical Bayesian Neural Networks}
\label{ch:ensembles}

\epigraph{The best way to have a good idea is to have a lot of ideas.}{\textnormal{--- Linus Pauling}}

\noindent
Classical \acrlong{bnn} inference techniques such as \acrshort{mcmc} or \acrshort{svi} provide high quality posterior approximations and are mathematically grounded, but their computational cost scales poorly with model size \cite{murphy2023probabilisticML}.  
As shown in Chapter~\ref{ch:bnns}, the high dimensionality of modern neural networks renders exact or near-exact Bayesian inference methods impractical for most real-world tasks.  
This computational bottleneck is especially pronounced in resource-constrained environments such as embedded devices, where memory and energy budgets are tightly limited \cite{roth2024jmlr}.  

A practical alternative is to approximate the posterior through ensembles of predictors rather than through explicit weight distributions.  
In this paradigm, multiple predictors are trained or sampled, and their collective variability is used as a proxy for epistemic uncertainty.  
Aleatoric uncertainty remains captured within each individual model through its likelihood formulation, while epistemic uncertainty is quantified by the disagreement across ensemble members \cite{jospin2022handsonBNN}.  
This principle underlies a family of approaches that we refer to as \emph{ensemble-style Bayesian approximations}.  

Ensemble-style methods are attractive for two reasons.  
First, they are fully compatible with standard deep learning frameworks, requiring little to no modification of the training pipeline.  
Second, they substantially reduce the computational overhead compared to MCMC or SVI, while often providing competitive uncertainty estimates.  
They therefore represent a pragmatic approach inspired by theoretically rigorous but practically intractable Bayesian inference for large-scale neural architectures.  

In this chapter, we focus on three representative methods: \acrfull{mcdo}, \acrfullpl{de}, and \acrfullpl{rlle}.  
\acrlong{mcdo} and \acrlongpl{de} are well-established in the literature and serve as state-of-the-art reference points for scalable uncertainty estimation.  
By contrast, \acrshortpl{rlle} are a more recent development \cite{steger2024rlle}, to which we have contributed, and we therefore provide a more detailed account of its formulation while referring to the original publication for further technical details.  
Together, these methods illustrate how ensemble-style approximations can address the computational bottlenecks of \acrlongpl{bnn}.  
\section{Ensemble Methods}

\subsection{\acrfull{mcdo}}

Among the earliest and most widely used ensemble-style approximations is \acrlong{mcdo}~\cite{gal2016mcdo}, which has become a standard baseline in uncertainty-aware deep learning due to its simplicity and minimal additional training overhead.  
It builds on the well-known regularization technique of dropout, where units are randomly set to zero during training to prevent overfitting, and extends it by applying dropout also at inference time.  
\textcite{gal2016mcdo,gal2016bcnn} showed that dropout can also be interpreted as approximate variational inference, providing a Bayesian view of neural networks without modifying the underlying training procedure.  
Equivalently, each stochastic forward pass can be seen as evaluating one member of an implicit ensemble of subnetworks, where randomness in the dropout masks induces predictor diversity~\cite{jospin2022handsonBNN}.  
By performing multiple such passes, one obtains a set of predictions that approximate samples from the posterior predictive distribution.  

A key strength of \acrshort{mcdo} is its practicality.  
It requires no changes to the training pipeline and incurs only minimal additional cost compared to a deterministic neural network.
The method can even be retrofitted to pretrained models simply by enabling dropout at inference.  
Uncertainty estimates derived from the variability of the stochastic predictions have proven useful in applications such as out-of-distribution detection and active learning~\cite{gal2017deepBAL}.  

Despite these advantages, \acrshort{mcdo} has important limitations.  
The variational family defined by dropout masks is relatively crude, which often leads to an underestimation of epistemic uncertainty and a weaker separation between epistemic and aleatoric components compared to more principled approaches such as \acrshort{svi} or \acrshort{mcmc}.  
Its uncertainty estimates are further sensitive to the chosen dropout rate and network architecture.  
Moreover, while training remains inexpensive, inference typically requires a large number of stochastic forward passes to obtain sufficiently diverse samples.  
In practice, this number often exceeds that required by more sophisticated inference schemes, which, although more costly during training, represent the posterior diversity more efficiently and achieve comparable uncertainty quality with fewer samples.

In summary, \acrshort{mcdo} provides a practical and lightweight approach to probabilistic deep learning, but its posterior approximation remains coarse, and its inference-time overhead can still be significant.

\subsection{\acrfullpl{de}}

\acrlongpl{de}~\cite{lakshminarayanan2017deepensembles} represent another canonical baseline, widely regarded as the empirical state of the art for uncertainty estimation in deep learning, owing to their strong predictive performance and ease of use.  
The approach is conceptually simple: multiple networks are trained independently with different random initializations, and their predictions are aggregated to approximate Bayesian model averaging.  
\textcite{lakshminarayanan2017deepensembles} formalized this idea under the name \emph{\acrlongpl{de}}, showing that independently trained models with different random initializations can approximate sampling from diverse regions of the weight space.  
Aggregating their predictions yields an implicit approximation to Bayesian model averaging.

Given an ensemble of $M$ networks with parameters $\{\theta_m\}_{m=1}^M$, the predictive distribution for a new input $x$ is estimated as
\begin{equation}
    p(y \mid x, D) \;\approx\; \frac{1}{M} \sum_{m=1}^M p(y \mid x, \theta_m).
\end{equation}
The resulting set of ensemble predictions constitutes approximate posterior samples, which can be analyzed with the same information-theoretic metrics introduced earlier to quantify total, epistemic, and aleatoric uncertainty.

\acrlongpl{de} have several appealing properties.  
They can be trained with standard pipelines and do not require specialized loss functions or hyperparameter tuning beyond what is used for a single deterministic network.  
In practice, they often achieve strong accuracy and well-calibrated uncertainty estimates across a wide range of tasks~\cite{lakshminarayanan2017deepensembles,ovadia2019can,fort2019necessary}.  

These benefits come at the cost of increased computation and storage.  
Training scales linearly with the number of ensemble members $M$, making ensembles $M$ times more expensive than training a single deterministic model or using \acrshort{mcdo}.  
Nevertheless, this is typically still cheaper than \acrshort{mcmc} or \acrshort{svi}, where gradient-based sampling or optimization must be repeated for many posterior draws.  
At inference, ensembles require multiple forward passes, placing them on par with other approximate Bayesian methods in predictive cost.  
For prediction, ensembles also require storing $M$ complete sets of network parameters, which scales memory linearly with ensemble size.  
This overhead is similar to \acrshort{mcmc}, where multiple parameter samples from the chain must be maintained, but contrasts with \acrshort{mcdo} and Gaussian \acrshort{svi}, which only need a single set of parameters (plus variance terms in the case of \acrshort{svi}).  

While ensembles lack theoretical guarantees on covering the posterior, particularly in the high-dimensional parameter spaces of \acrshortpl{bnn}, their uninformed diversity often works surprisingly well in practice.  
Deep Ensembles are widely regarded as the empirical state of the art for large models where \acrshort{mcmc} and \acrshort{svi} are not feasible.  
In fact, they can outperform variational methods when simple approximations such as mean-field Gaussians fail to capture multimodal posteriors, whereas independently trained ensemble members may collectively represent multiple modes~\cite{Wilson2020ProbPersp,izmailov2021whatBNNposteriors}.  

Overall, \acrlongpl{de} provide robust uncertainty estimates and remain the empirical state of the art in large-scale settings, albeit at significant training and storage cost.  
This motivates lighter alternatives that can retain ensemble diversity at a fraction of the overhead, such as \acrlongpl{rlle}, which we discuss next.  

\subsection{\acrfullpl{rlle}}

\begin{figure}[t]
    \centering
    \includegraphics[width=0.60\textwidth]{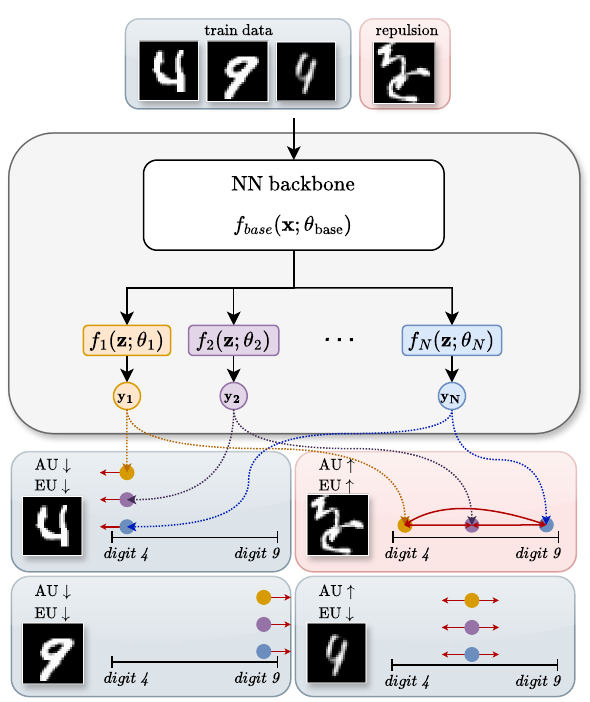}
	\caption{Illustration of a \acrlong{rlle} with $N$ particles, where each output head corresponds to one particle (colored dots).  
	Repulsion is enforced in function space using unlabeled samples from a different distribution, serving as \emph{repulsion samples}.  
	\acrlong{eu} (\acrshort{eu}) is low when particles agree and increases with their spread,  
	whereas \acrlong{au} (\acrshort{au}) arises from inherent label ambiguity, leading to predictions centered in uncertain probability regions. \reprofrom{steger2024rlle}}
   \label{fig:rlle:method}
\end{figure}

\paragraph{From \acrlongpl{de} to \acrlongpl{rde}.}  
While \acrlongpl{de} provide strong empirical uncertainty estimates, their diversity is incidental, arising only from random initialization, stochastic optimization, or random shuffling of the training inputs.  
D’Angelo and Fortuin formalized this diversity by introducing \emph{\acrlongpl{rde}} (\acrshortpl{rde})~\cite{angelo2021repulsiveensembles}, which explicitly encourage ensemble members to remain diverse during training.  
Their method augments the training loss with a repulsion term between predictive functions, grounded in \acrlong{povi} (\acrshort{povi})~\cite{liu2016stein,wang2019function,liu2017stein,liu2019understanding,chen2018unified}.  
This ensures that ensemble members (the particles) do not collapse onto similar solutions, but instead spread out to approximate multiple modes of the posterior.  

The key insight of \acrshortpl{rde} is that \emph{diversity in weight space is not sufficient}, since different parameter configurations can realize nearly identical functions.  
\acrshortpl{rde} therefore emphasize \emph{function-space diversity}, applying the repulsion term directly on the predictive distributions of each network.  
Empirical results demonstrated that such functional repulsion improves the estimation of epistemic uncertainty and out-of-distribution detection compared to standard \acrshortpl{de}~\cite{angelo2021repulsiveensembles}.  
However, \acrshortpl{rde} inherit the computational burden of classical ensembles: each network must still be trained and stored independently, making them costly for large-scale or resource-constrained applications.  

\paragraph{\acrlongpl{rlle}.}  
To address these computational limitations, Steger et al.\ proposed \emph{\acrlongpl{rlle}} (\acrshortpl{rlle})~\cite{steger2024rlle}.  
Building directly on the principle of function-space repulsion from \acrshortpl{rde}, \acrshortpl{rlle} restrict diversity to the final prediction layer.  
Instead of training multiple independent networks, a single shared backbone is equipped with several output heads, each representing one ensemble member.  
Function-space repulsion is then applied only to these heads, ensuring predictive diversity at minimal additional cost.  
This architectural simplification drastically reduces training and storage overhead and integrates seamlessly with pretrained backbones, while retaining the theoretical motivation of \acrshortpl{rde}.  

\paragraph{Architecture and objective.}  
Let $f_{\mathrm{base}}(x;\theta_{\mathrm{base}})$ denote a shared feature extractor and $\{f^{(i)}_{\mathrm{head}}(\cdot;\theta^{(i)}_{\mathrm{head}})\}_{i=1}^n$ denote $n$ last-layer heads.  
Each particle function is $f^{(i)}(x)=f^{(i)}_{\mathrm{head}}(f_{\mathrm{base}}(x))$.  
Training minimizes the standard predictive loss plus a repulsive term that penalizes functional similarity across heads on a set of \emph{repulsion samples}, thereby approximating function-space \acrshort{povi} with attraction–repulsion dynamics~\cite{angelo2021repulsiveensembles,steger2024rlle}.  
This multi-head design avoids duplicating the backbone and reduces both training and memory cost compared to \acrshortpl{de}~\cite{lakshminarayanan2017deepensembles,ovadia2019can}.  
Figure~\ref{fig:rlle:method} illustrates the concept. 

\paragraph{Attraction–repulsion field in function space.}
Following \acrshort{povi}, the update field for the $i$-th particle (head) acts on the last-layer parameters $\theta^{(i)}_{l}$ and combines a data-driven attraction term with a kernel-based repulsion in function space~\cite{steger2024rlle,angelo2021repulsiveensembles,liu2016stein}.
We write
\begin{equation}
v\!\left(\theta_{l}^{(i)}\right)
\;=\;
\underbrace{\nabla_{\theta_{l}^{(i)}} \log p\!\left(\theta_{l}^{(i)} \mid \mathcal{D}\right)}_{\text{attraction}}
\;-\;\gamma_{\mathrm{repulsion}}
\underbrace{\frac{\sum_{j=1}^{n} \nabla_{f\!\left(\theta_{l}^{(i)}\right)}
\,k\!\left(f\!\left(\theta_{l}^{(i)}\right),\, f\!\left(\theta_{l}^{(j)}\right)\right)}
{\sum_{j=1}^{n} k\!\left(f\!\left(\theta_{l}^{(i)}\right),\, f\!\left(\theta_{l}^{(j)}\right)\right)}}_{\text{repulsion in function space}},
\label{eq:rlle-repulsion}
\end{equation}
where $f(\theta_{l}^{(i)})$ denotes the predictive function of head $i$ evaluated on a batch of repulsion samples, $\gamma_{\mathrm{repulsion}}$ is the controllable repulsion strength, and $k$ is a positive-definite kernel that measures functional similarity.
A common choice is the RBF kernel applied to vector predictions on repulsion samples $\chi$,
\begin{equation}
k\!\left(f^{(i)}(\chi),\,f^{(j)}(\chi)\right)
\;=\;
	\exp\!\left(\frac{-\big\| f^{(i)}(\chi)-f^{(j)}(\chi)\big\|_{p}}{{\nu}}\right),
\end{equation}
which yields a normalized repulsion term and directly encourages predictive diversity at the heads, while the backbone remains shared~\cite{steger2024rlle}.

\paragraph{Choice of repulsion samples.}  
A critical design choice is where to enforce functional repulsion.  
Applying it directly on the training data artificially inflates epistemic uncertainty in regions where the model should be confident and thus harms accuracy.  
Instead, \acrshortpl{rlle} evaluate the repulsion term on unlabeled \acrshort{ood} data or on label-destroying augmentations, which probe regions that genuinely reveal epistemic uncertainty and thereby improve calibration and \acrshort{ood} detection.  
In high-dimensional settings, such auxiliary \acrshort{ood} or augmented samples serve as an effective approximation to full function-space coverage~\cite{wang2019function,steger2024rlle}.  

\paragraph{Compatibility with pretrained backbones.}  
Because diversity is confined to the last layer, \acrshortpl{rlle} can be naturally applied to pretrained networks: freeze (or lightly fine-tune) the backbone, replace the classifier with $n$ heads, and train with the function-space repulsion objective.  
This decouples representation learning from uncertainty-aware fine-tuning and is particularly effective when the backbone is regularized to avoid feature collapse, e.g., via spectral normalization or related constraints~\cite{miyato2018spectral,van2021feature}.  

\paragraph{Context within related work.}
Repulsive ensembles can be instantiated in different spaces—weights, features, input gradients, or functions.  
\acrshortpl{rlle} adopt the function-space variant while avoiding the cost of fully independent networks~\cite{angelo2021repulsiveensembles}.  
Compared to last-layer Bayesian baselines such as LL-Laplace or partially stochastic last layers, \acrshortpl{rlle} employ explicit functional repulsion among multiple heads rather than a single probabilistic classifier~\cite{daxberger2021laplace,steger2024rlle}.  

\paragraph{Limitations.}  
Despite their advantages, \acrshortpl{rlle} are not without caveats.  
Performance hinges on the calibration of repulsion strength and the quality and coverage of repulsion samples; poor choices can degrade accuracy or fail to elicit meaningful diversity~\cite{steger2024rlle}.  
Furthermore, because diversity is restricted to the last layer, expressiveness is more constrained than in fully independent \acrshortpl{de}, especially when the shared backbone provides limited feature diversity.  
Nevertheless, the combination of function-space grounding, empirical competitiveness, and low computational footprint makes \acrshortpl{rlle} compelling for scalable and embedded uncertainty estimation~\cite{steger2024rlle}.  
\section{Comparative Evaluation}

We first analyze the behavior of \acrshort{mcdo}, \acrshortpl{de}, and \acrshortpl{rlle} on the Noisy Sine regression task, as studied in Simonides’ master’s thesis~\cite{simonides2025ma}.  
This controlled benchmark is particularly suited to assess the decomposition of epistemic and aleatoric uncertainty, since the ground-truth aleatoric variance is known and out-of-distribution regions are clearly identifiable.  

Overall, all three methods exhibit a marked sensitivity to the choice of activation function.  
\acrshort{mcdo} performs worst: it fails to raise epistemic uncertainty in out-of-distribution regions, except for a single peak with the \texttt{SiLU} activation, and even then only in the central region rather than across the full support.  
\acrshortpl{de} are considerably more robust, working reliably with several activations, and require only modest ensemble sizes ($M\!\approx\!10$) to stabilize their predictions.  
\acrshortpl{rlle} outperform \acrshort{mcdo} but are more fragile than \acrshortpl{de}: they require selecting a suitable activation (\texttt{SeLU} being most effective) and carefully calibrating their two hyperparameters, variance regularization and repulsion strength.  
This flexibility allows balancing aleatoric and epistemic contributions, but it also demands more tuning effort.  

\begin{figure}[t]
    \centering

    \begin{subfigure}[t]{0.49\textwidth}
        \centering
        \includegraphics[width=\textwidth]{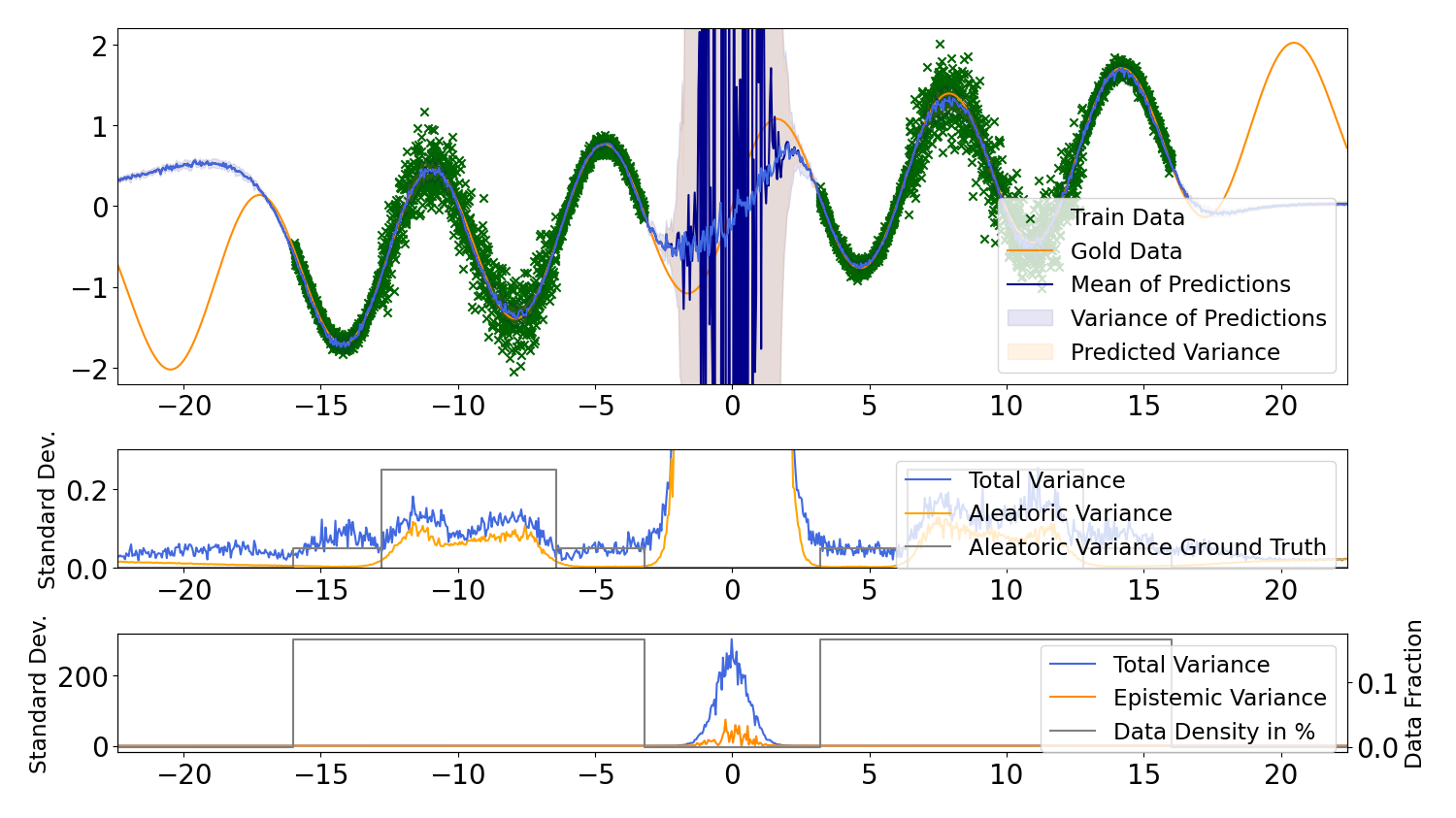}
		\caption{\acrshort{mcdo}}
        \label{fig:noisy-sine:mcdo}
    \end{subfigure}
    \hfill
    \begin{subfigure}[t]{0.49\textwidth}
        \centering
        \includegraphics[width=\textwidth]{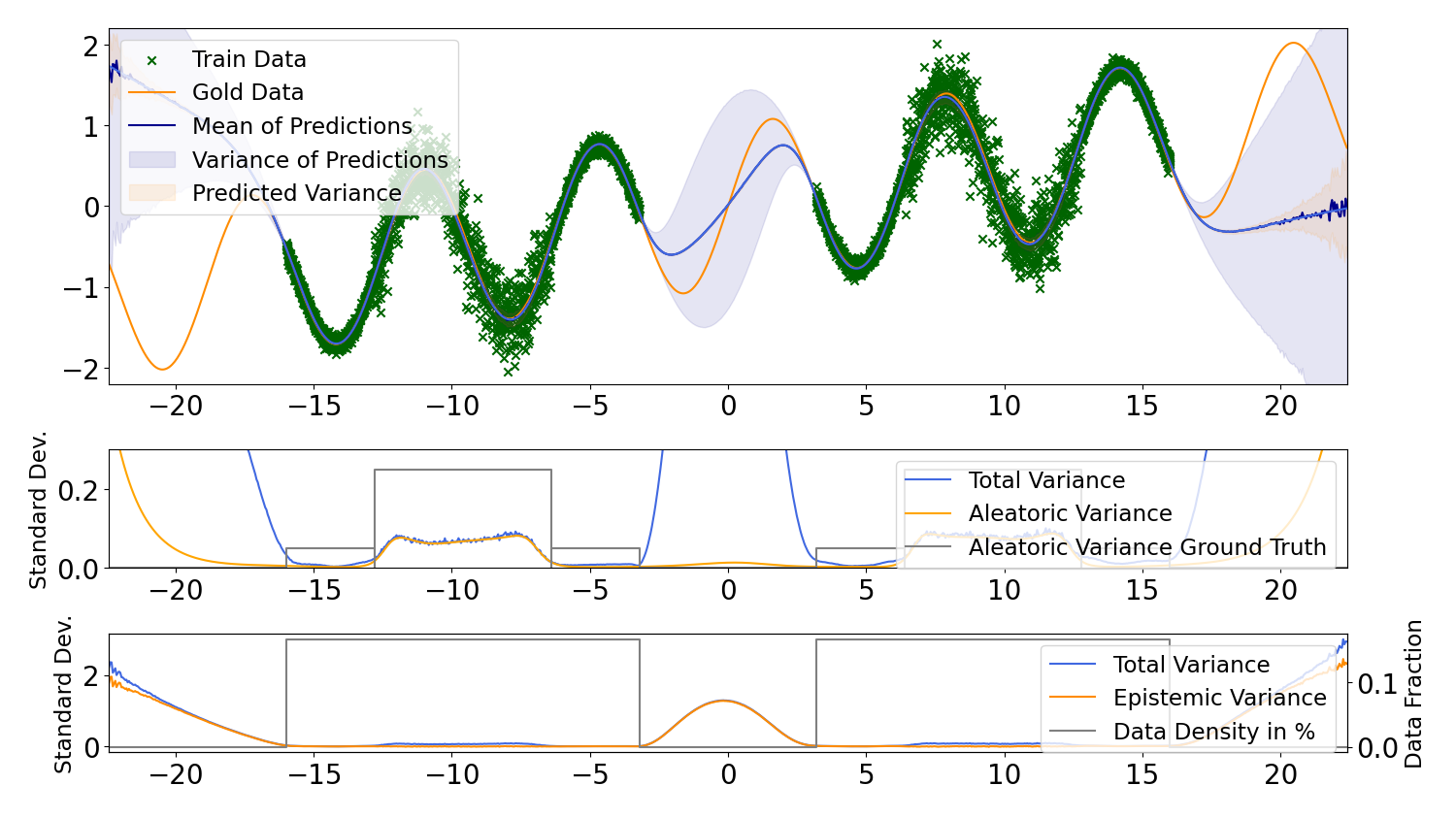}
        \caption{\acrlongpl{de}}
        \label{fig:noisy-sine:de}
    \end{subfigure}
    \hfill
    \begin{subfigure}[t]{0.49\textwidth}
        \centering
        \includegraphics[width=\textwidth]{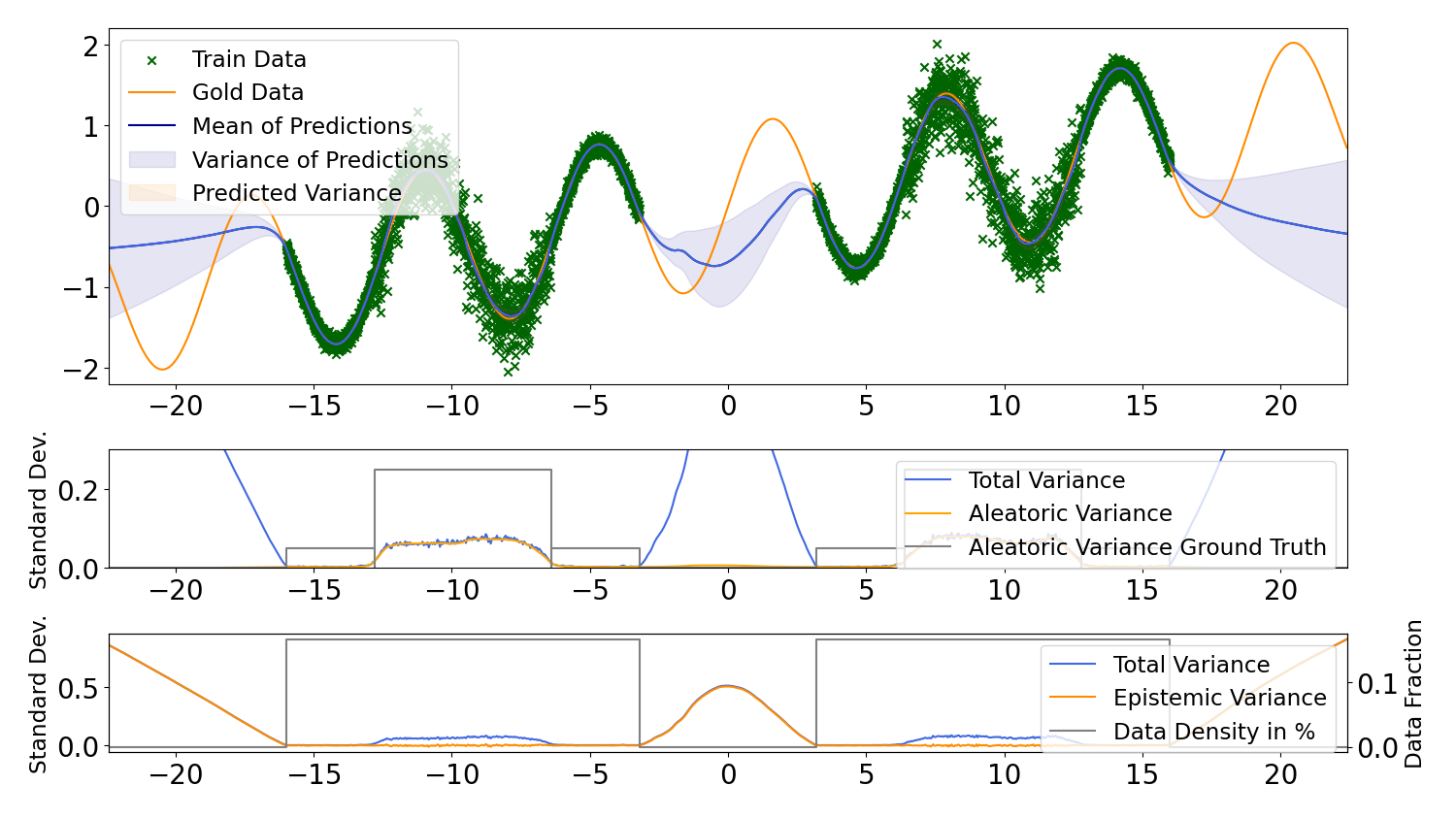}
        \caption{\acrshortpl{rlle}}
        \label{fig:noisy-sine:rlle}
    \end{subfigure}

    \caption{
    Comparison on the Noisy Sine regression task with each method configured to its best-performing activation function and hyperparameters.  
    \acrshort{mcdo}: SiLU activation, dropout rate $0.15$;  
    \acrshortpl{de}: $M{=}10$ members, SiLU activation;  
    \acrshortpl{rlle}: SELU activation with $\lambda_{\mathrm{var}}=0.25$ and $\gamma_{\mathrm{repulsion}}=1000$.  
    \acrshort{mcdo} systematically underestimates epistemic uncertainty, \acrshortpl{de} provide stable and well-calibrated estimates with moderate ensemble size, and \acrshortpl{rlle} achieve competitive performance when both activation and repulsion/variance parameters are carefully calibrated. \adjfrom{simonides2025ma}}
    \label{fig:noisy-sine-ensembles}
\end{figure}

\acrshortpl{rlle} demonstrated competitive overall performance, though their calibration proved more sensitive to hyperparameter settings than for \acrlongpl{de}.  
They also required substantially less training and storage cost than \acrlongpl{de}, while being only marginally more expensive than \acrshort{mcdo}.  
Figure~\ref{fig:noisy-sine-ensembles} illustrates these qualitative differences, showing the characteristic uncertainty profiles of all three methods in their best-performing configurations.  

While the Noisy Sine task highlights qualitative differences in uncertainty behavior under controlled conditions,  
Dirty-MNIST provides a more realistic image-classification benchmark to illustrate these trade-offs at scale.  

For illustration we reproduce results on the Dirty-MNIST benchmark (Table~\ref{tab:ensembles:rlle:dirtyMNIST}), whereas the original publication~\cite{steger2024rlle} reports a comprehensive evaluation across multiple datasets and baselines.  
\pgfplotstableread[col sep=comma]{BayesianNeuralNetworks/ensembles/imgs/data/OOD_net18_dirty_mnist_auroc.csv}\tableOOD
\pgfplotstableread[col sep=comma]{BayesianNeuralNetworks/ensembles/imgs/data/ID_net18_dirty_mnist_accuracy.csv}\tableIDaccuracy
\pgfplotstableread[col sep=comma]{BayesianNeuralNetworks/ensembles/imgs/data/ID_net18_dirty_mnist_nll.csv}\tableIDnll
\pgfplotstableread[col sep=comma]{BayesianNeuralNetworks/ensembles/imgs/data/ID_net18_dirty_mnist_ece.csv}\tableIDece

\pgfplotstableread[col sep=comma]{BayesianNeuralNetworks/ensembles/imgs/data/OOD_dirty_mnist_auroc_POVI_SGD.csv}\tableOODSGD
\pgfplotstableread[col sep=comma]{BayesianNeuralNetworks/ensembles/imgs/data/ID_dirty_mnist_accuracy_POVI_SGD.csv}\tableIDaccuracySGD
\pgfplotstableread[col sep=comma]{BayesianNeuralNetworks/ensembles/imgs/data/ID_dirty_mnist_nll_POVI_SGD.csv}\tableIDnllSGD
\pgfplotstableread[col sep=comma]{BayesianNeuralNetworks/ensembles/imgs/data/ID_dirty_mnist_ece_POVI_SGD.csv}\tableIDeceSGD

\pgfplotstableread[col sep=comma]{BayesianNeuralNetworks/ensembles/imgs/data/OOD_dirty_mnist_auroc_POVI_kde_WGD.csv}\tableOODWGD
\pgfplotstableread[col sep=comma]{BayesianNeuralNetworks/ensembles/imgs/data/ID_dirty_mnist_accuracy_POVI_kde_WGD.csv}\tableIDaccuracyWGD
\pgfplotstableread[col sep=comma]{BayesianNeuralNetworks/ensembles/imgs/data/ID_dirty_mnist_nll_POVI_kde_WGD.csv}\tableIDnllWGD
\pgfplotstableread[col sep=comma]{BayesianNeuralNetworks/ensembles/imgs/data/ID_dirty_mnist_ece_POVI_kde_WGD.csv}\tableIDeceWGD

\pgfplotstableread[col sep=comma]{BayesianNeuralNetworks/ensembles/imgs/data/OOD_dirty_mnist_auroc_POVI_kde_f_WGD_coll_emnist_128.csv}\tableOODemnist
\pgfplotstableread[col sep=comma]{BayesianNeuralNetworks/ensembles/imgs/data/ID_dirty_mnist_accuracy_POVI_kde_f_WGD_coll_emnist_128.csv}\tableIDaccuracyemnist
\pgfplotstableread[col sep=comma]{BayesianNeuralNetworks/ensembles/imgs/data/ID_dirty_mnist_nll_POVI_kde_f_WGD_coll_emnist_128.csv}\tableIDnllemnist
\pgfplotstableread[col sep=comma]{BayesianNeuralNetworks/ensembles/imgs/data/ID_dirty_mnist_ece_POVI_kde_f_WGD_coll_emnist_128.csv}\tableIDeceemnist

\pgfplotstableread[col sep=comma]{BayesianNeuralNetworks/ensembles/imgs/data/OOD_dirty_mnist_auroc_POVI_patches_std_16.csv}\tableOODPatchsixteen
\pgfplotstableread[col sep=comma]{BayesianNeuralNetworks/ensembles/imgs/data/ID_dirty_mnist_accuracy_POVI_patches_std_16.csv}\tableIDaccuracyPatchsixteen
\pgfplotstableread[col sep=comma]{BayesianNeuralNetworks/ensembles/imgs/data/ID_dirty_mnist_nll_POVI_patches_std_16.csv}\tableIDnllPatchsixteen
\pgfplotstableread[col sep=comma]{BayesianNeuralNetworks/ensembles/imgs/data/ID_dirty_mnist_ece_POVI_patches_std_16.csv}\tableIDecePatchsixteen

\pgfplotstableread[col sep=comma]{BayesianNeuralNetworks/ensembles/imgs/data/OOD_dirty_mnist_auroc_POVI_patches_std_8.csv}\tableOODPatcheight
\pgfplotstableread[col sep=comma]{BayesianNeuralNetworks/ensembles/imgs/data/ID_dirty_mnist_accuracy_POVI_patches_std_8.csv}\tableIDaccuracyPatcheight
\pgfplotstableread[col sep=comma]{BayesianNeuralNetworks/ensembles/imgs/data/ID_dirty_mnist_nll_POVI_patches_std_8.csv}\tableIDnllPatcheight
\pgfplotstableread[col sep=comma]{BayesianNeuralNetworks/ensembles/imgs/data/ID_dirty_mnist_ece_POVI_patches_std_8.csv}\tableIDecePatcheight

\pgfplotstableread[col sep=comma]{BayesianNeuralNetworks/ensembles/imgs/data/OOD_dirty_mnist_auroc_POVI_kde_f_WGD_patches_std_4.csv}\tableOODPatchfour
\pgfplotstableread[col sep=comma]{BayesianNeuralNetworks/ensembles/imgs/data/ID_dirty_mnist_accuracy_POVI_kde_f_WGD_patches_std_4.csv}\tableIDaccuracyPatchfour
\pgfplotstableread[col sep=comma]{BayesianNeuralNetworks/ensembles/imgs/data/ID_dirty_mnist_nll_POVI_kde_f_WGD_patches_std_4.csv}\tableIDnllPatchfour
\pgfplotstableread[col sep=comma]{BayesianNeuralNetworks/ensembles/imgs/data/ID_dirty_mnist_ece_POVI_kde_f_WGD_patches_std_4.csv}\tableIDecePatchfour

\pgfplotstableread[col sep=comma]{BayesianNeuralNetworks/ensembles/imgs/data/OOD_dirty_mnist_auroc_POVI_kde_f_WGD_coll_dirty_mnist_128.csv}\tableOODdirty
\pgfplotstableread[col sep=comma]{BayesianNeuralNetworks/ensembles/imgs/data/ID_dirty_mnist_accuracy_POVI_kde_f_WGD_coll_dirty_mnist_128.csv}\tableIDaccuracydirty
\pgfplotstableread[col sep=comma]{BayesianNeuralNetworks/ensembles/imgs/data/ID_dirty_mnist_nll_POVI_kde_f_WGD_coll_dirty_mnist_128.csv}\tableIDnlldirty
\pgfplotstableread[col sep=comma]{BayesianNeuralNetworks/ensembles/imgs/data/ID_dirty_mnist_ece_POVI_kde_f_WGD_coll_dirty_mnist_128.csv}\tableIDecedirty

\newcommand{\fetchAndStore}[4]{%
    \pgfplotstablegetelem{#1}{#2}\of{#3}
    \expandafter\xdef\csname #2Mean#1#4\endcsname{\pgfplotsretval}
    \pgfplotstablegetelem{#1}{#2_std}\of{#3}
    \expandafter\xdef\csname #2Std#1#4\endcsname{\pgfplotsretval}
}

\foreach \method in {softmax,logits,logits_GMM,softmax_laplace_MI,softmax_laplace_AU,softmax_SNGP_PE, logits_SNGP,softmax_en_AU,softmax_en_MI} {
    \foreach \intensity in {0,1,2} { %
        \fetchAndStore{\intensity}{\method}{\tableOOD}{OOD}
    }
}

\foreach \method in {softmax,logits,logits_GMM,softmax_laplace_MI,softmax_laplace_AU,softmax_SNGP_PE, logits_SNGP,softmax_en_AU,softmax_en_MI} {
    \fetchAndStore{0}{\method}{\tableIDaccuracy}{accuracy}
    \fetchAndStore{0}{\method}{\tableIDnll}{nll}
    \fetchAndStore{0}{\method}{\tableIDece}{ece}
}

\foreach \method in {softmax_POVI_MI, softmax_POVI_AU} {
    \foreach \intensity in {0,1,2} { %
        \fetchAndStore{\intensity}{\method}{\tableOODSGD}{OODSGD}
        \fetchAndStore{\intensity}{\method}{\tableOODWGD}{OODWGD}
        \fetchAndStore{\intensity}{\method}{\tableOODemnist}{OODemnist}
        \fetchAndStore{\intensity}{\method}{\tableOODPatchsixteen}{OODPatchsixteen}
        \fetchAndStore{\intensity}{\method}{\tableOODPatcheight}{OODPatcheight}
        \fetchAndStore{\intensity}{\method}{\tableOODPatchfour}{OODPatchfour}
        \fetchAndStore{\intensity}{\method}{\tableOODdirty}{OODdirty}
    
    }
}
\foreach \method in {softmax_POVI_MI, softmax_POVI_AU} {
    \fetchAndStore{0}{\method}{\tableIDaccuracySGD}{accuracySGD}
    \fetchAndStore{0}{\method}{\tableIDnllSGD}{nllSGD}
    \fetchAndStore{0}{\method}{\tableIDeceSGD}{eceSGD}

    \fetchAndStore{0}{\method}{\tableIDaccuracyWGD}{accuracyWGD}
    \fetchAndStore{0}{\method}{\tableIDnllWGD}{nllWGD}
    \fetchAndStore{0}{\method}{\tableIDeceWGD}{eceWGD}
    
    \fetchAndStore{0}{\method}{\tableIDaccuracyemnist}{accuracyemnist}
    \fetchAndStore{0}{\method}{\tableIDnllemnist}{nllemnist}
    \fetchAndStore{0}{\method}{\tableIDeceemnist}{eceemnist}

    \fetchAndStore{0}{\method}{\tableIDaccuracyPatchsixteen}{accuracyPatchsixteen}
    \fetchAndStore{0}{\method}{\tableIDnllPatchsixteen}{nllPatchsixteen}
    \fetchAndStore{0}{\method}{\tableIDecePatchsixteen}{ecePatchsixteen}
    
    \fetchAndStore{0}{\method}{\tableIDaccuracyPatcheight}{accuracyPatcheight}
    \fetchAndStore{0}{\method}{\tableIDnllPatcheight}{nllPatcheight}
    \fetchAndStore{0}{\method}{\tableIDecePatcheight}{ecePatcheight}
    
    \fetchAndStore{0}{\method}{\tableIDaccuracyPatchfour}{accuracyPatchfour}
    \fetchAndStore{0}{\method}{\tableIDnllPatchfour}{nllPatchfour}
    \fetchAndStore{0}{\method}{\tableIDecePatchfour}{ecePatchfour}
    
    \fetchAndStore{0}{\method}{\tableIDaccuracydirty}{accuracydirty}
    \fetchAndStore{0}{\method}{\tableIDnlldirty}{nlldirty}
    \fetchAndStore{0}{\method}{\tableIDecedirty}{ecedirty}
}

\begin{table}%
\centering
\caption{Comparison of uncertainty decomposition on Dirty-MNIST.  
Aleatoric and epistemic uncertainty are evaluated for detecting ambiguous and \acrshort{ood} samples.  
Best results are in bold, second best are underlined.
\acrshortpl{rlle} variants outperform \acrlongpl{de} (DE-5) clearly in \acrshort{ood} detection.
\reprofrom{steger2024rlle}}
\label{tab:my-table}
\resizebox{\textwidth}{!}{%
\begin{tabular}{@{}lcccccc@{}} %
\toprule
\multirow{2}{*}{\textbf{Method}} & \multirow{2}{*}{\textsc{Acc. $\uparrow$ [\%]}} & \multirow{2}{*}{\textsc{NLL $\downarrow$ [\%]}} & \multirow{2}{*}{\textsc{ECE $\downarrow$ [\%]}} & \multicolumn{3}{c}{\textsc{OOD Auroc $\uparrow$ [\%] }} \\ \cmidrule(l){5-7} 
                                 &                                                           &                                                               &                                                           & MNIST vs ambig. (AU) & MNIST vs. OOD (EU) & ambig. vs OOD (EU) \\ \midrule
MAP & ${\csname softmaxMean0accuracy\endcsname _{\pm \csname softmaxStd0accuracy\endcsname}}$ & ${\csname softmaxMean0nll\endcsname _{\pm \csname softmaxStd0nll\endcsname}}$ & ${\csname softmaxMean0ece\endcsname _{\pm \csname softmaxStd0ece\endcsname}}$ & ${\csname softmaxMean0OOD\endcsname _{\pm \csname softmaxStd0OOD\endcsname}}$ & ${\csname softmaxMean1OOD\endcsname _{\pm \csname softmaxStd1OOD\endcsname}}$ & ${\csname softmaxMean2OOD\endcsname _{\pm \csname softmaxStd2OOD\endcsname}}$ \\ \midrule
DDU & ${\csname softmaxMean0accuracy\endcsname _{\pm \csname softmaxStd0accuracy\endcsname}}$ & ${\csname softmaxMean0nll\endcsname _{\pm \csname softmaxStd0nll\endcsname}}$ & ${\csname softmaxMean0ece\endcsname _{\pm \csname softmaxStd0ece\endcsname}}$ & ${\csname softmaxMean0OOD\endcsname _{\pm \csname softmaxStd0OOD\endcsname}}$ & $\mathbf{\csname logits_GMMMean1OOD\endcsname _{\pm \csname logits_GMMStd1OOD\endcsname}}$ & $\mathbf{\csname logits_GMMMean2OOD\endcsname _{\pm \csname logits_GMMStd2OOD\endcsname}}$ \\
SNGP & $\underlineitalic{\csname softmax_SNGP_PEMean0accuracy\endcsname _{\pm \csname softmax_SNGP_PEStd0accuracy\endcsname}}$ & ${\csname softmax_SNGP_PEMean0nll\endcsname _{\pm \csname softmax_SNGP_PEStd0nll\endcsname}}$ & ${\csname softmax_SNGP_PEMean0ece\endcsname _{\pm \csname softmax_SNGP_PEStd0ece\endcsname}}$ & ${\csname softmax_SNGP_PEMean0OOD\endcsname _{\pm \csname softmax_SNGP_PEStd0OOD\endcsname}}$ & ${\csname softmax_SNGP_PEMean1OOD\endcsname _{\pm \csname softmax_SNGP_PEStd1OOD\endcsname}}$ & ${\csname softmax_SNGP_PEMean2OOD\endcsname _{\pm \csname logits_SNGPStd2OOD\endcsname}}$ \\
LL-Laplace & ${\csname softmax_laplace_MIMean0accuracy\endcsname _{\pm \csname softmax_laplace_MIStd0accuracy\endcsname}}$ & ${\csname softmax_laplace_MIMean0nll\endcsname _{\pm \csname softmax_laplace_MIStd0nll\endcsname}}$ & ${\csname softmax_laplace_MIMean0ece\endcsname _{\pm \csname softmax_laplace_MIStd0ece\endcsname}}$ & ${\csname softmax_laplace_AUMean0OOD\endcsname _{\pm \csname softmax_laplace_AUStd0OOD\endcsname}}$ & ${\csname softmax_laplace_MIMean1OOD\endcsname _{\pm \csname softmax_laplace_MIStd1OOD\endcsname}}$ & ${\csname softmax_laplace_MIMean2OOD\endcsname _{\pm \csname softmax_laplace_MIStd2OOD\endcsname}}$ \\ \midrule
\acrshort{LLPOVI} \emph{(ours)} & $\mathbf{\csname softmax_POVI_MIMean0accuracySGD\endcsname _{\pm \csname softmax_POVI_MIStd0accuracySGD\endcsname}}$ & $\mathbf{\csname softmax_POVI_MIMean0nllSGD\endcsname _{\pm \csname softmax_POVI_MIStd0nllSGD\endcsname}}$ & $\mathbf{\csname softmax_POVI_MIMean0eceSGD\endcsname _{\pm \csname softmax_POVI_MIStd0eceSGD\endcsname}}$ & $\mathbf{\csname softmax_POVI_AUMean0OODSGD\endcsname _{\pm \csname softmax_POVI_MIStd0OODSGD\endcsname}}$ & ${\csname softmax_POVI_MIMean1OODSGD\endcsname _{\pm \csname softmax_POVI_MIStd1OODSGD\endcsname}}$ & ${\csname softmax_POVI_MIMean2OODSGD\endcsname _{\pm \csname softmax_POVI_MIStd2OODSGD\endcsname}}$ \\ 
\acrshort{RLLPOVI} \emph{(ours)}  & $\mathbf{\csname softmax_POVI_MIMean0accuracyWGD\endcsname _{\pm \csname softmax_POVI_MIStd0accuracyWGD\endcsname}}$ & $\mathbf{\csname softmax_POVI_MIMean0nllWGD\endcsname _{\pm \csname softmax_POVI_MIStd0nllWGD\endcsname}}$ & $\mathbf{\csname softmax_POVI_MIMean0eceWGD\endcsname _{\pm \csname softmax_POVI_MIStd0eceWGD\endcsname}}$ & $\mathbf{\csname softmax_POVI_AUMean0OODWGD\endcsname _{\pm \csname softmax_POVI_MIStd0OODWGD\endcsname}}$ & ${\csname softmax_POVI_MIMean1OODWGD\endcsname _{\pm \csname softmax_POVI_MIStd1OODWGD\endcsname}}$ & ${\csname softmax_POVI_MIMean2OODWGD\endcsname _{\pm \csname softmax_POVI_MIStd2OODWGD\endcsname}}$ \\ 
\acrshort{fLLPOVI} \emph{(ours)}      &                                                           &                                                               &                                                           &                       &              &                     \\
\textit{~~~+ dirtyMNIST}  & ${\csname softmax_POVI_MIMean0accuracydirty\endcsname _{\pm \csname softmax_POVI_MIStd0accuracydirty\endcsname}}$ & ${\csname softmax_POVI_MIMean0nlldirty\endcsname _{\pm \csname softmax_POVI_MIStd0nlldirty\endcsname}}$ & ${\csname softmax_POVI_MIMean0ecedirty\endcsname _{\pm \csname softmax_POVI_MIStd0ecedirty\endcsname}}$ & ${\csname softmax_POVI_AUMean0OODdirty\endcsname _{\pm \csname softmax_POVI_MIStd0OODdirty\endcsname}}$ & ${\csname softmax_POVI_MIMean1OODdirty\endcsname _{\pm \csname softmax_POVI_MIStd1OODdirty\endcsname}}$ & ${\csname softmax_POVI_MIMean2OODdirty\endcsname _{\pm \csname softmax_POVI_MIStd2OODdirty\endcsname}}$ \\
\textit{~~~+ eMNIST}  & $\underlineitalic{\csname softmax_POVI_MIMean0accuracyemnist\endcsname _{\pm \csname softmax_POVI_MIStd0accuracyemnist\endcsname}}$ & ${\csname softmax_POVI_MIMean0nllemnist\endcsname _{\pm \csname softmax_POVI_MIStd0nllemnist\endcsname}}$ & ${\csname softmax_POVI_MIMean0eceemnist\endcsname _{\pm \csname softmax_POVI_MIStd0eceemnist\endcsname}}$ & ${\csname softmax_POVI_AUMean0OODemnist\endcsname _{\pm \csname softmax_POVI_MIStd0OODemnist\endcsname}}$ & ${\csname softmax_POVI_MIMean1OODemnist\endcsname _{\pm \csname softmax_POVI_MIStd1OODemnist\endcsname}}$ & $\underlineitalic{\csname softmax_POVI_MIMean2OODemnist\endcsname _{\pm \csname softmax_POVI_MIStd2OODemnist\endcsname}}$ \\
\textit{~~~+ Patches-16} & $\underlineitalic{\csname softmax_POVI_MIMean0accuracyPatchsixteen\endcsname _{\pm \csname softmax_POVI_MIStd0accuracyPatchsixteen\endcsname}}$ & $\underlineitalic{\csname softmax_POVI_MIMean0nllPatchsixteen\endcsname _{\pm \csname softmax_POVI_MIStd0nllPatchsixteen\endcsname}}$ & $\underlineitalic{\csname softmax_POVI_MIMean0ecePatchsixteen\endcsname _{\pm \csname softmax_POVI_MIStd0ecePatchsixteen\endcsname}}$ & $\underlineitalic{\csname softmax_POVI_AUMean0OODPatchsixteen\endcsname _{\pm \csname softmax_POVI_MIStd0OODPatchsixteen\endcsname}}$ & $\underlineitalic{\csname softmax_POVI_MIMean1OODPatchsixteen\endcsname _{\pm \csname softmax_POVI_MIStd1OODPatchsixteen\endcsname}}$ & ${\csname softmax_POVI_MIMean2OODPatchsixteen\endcsname _{\pm \csname softmax_POVI_MIStd2OODPatchsixteen\endcsname}}$ \\
\textit{~~~+ Patches-8} & $\underlineitalic{\csname softmax_POVI_MIMean0accuracyPatcheight\endcsname _{\pm \csname softmax_POVI_MIStd0accuracyPatcheight\endcsname}}$ & ${\csname softmax_POVI_MIMean0nllPatcheight\endcsname _{\pm \csname softmax_POVI_MIStd0nllPatcheight\endcsname}}$ & $\underlineitalic{\csname softmax_POVI_MIMean0ecePatcheight\endcsname _{\pm \csname softmax_POVI_MIStd0ecePatcheight\endcsname}}$ & ${\csname softmax_POVI_AUMean0OODPatcheight\endcsname _{\pm \csname softmax_POVI_MIStd0OODPatcheight\endcsname}}$ & ${\csname softmax_POVI_MIMean1OODPatcheight\endcsname _{\pm \csname softmax_POVI_MIStd1OODPatcheight\endcsname}}$ & ${\csname softmax_POVI_MIMean2OODPatcheight\endcsname _{\pm \csname softmax_POVI_MIStd2OODPatcheight\endcsname}}$ \\ 
\textit{~~~+ Patches-4} & $\underlineitalic{\csname softmax_POVI_MIMean0accuracyPatchfour\endcsname _{\pm \csname softmax_POVI_MIStd0accuracyPatchfour\endcsname}}$ & ${\csname softmax_POVI_MIMean0nllPatchfour\endcsname _{\pm \csname softmax_POVI_MIStd0nllPatchfour\endcsname}}$ & $\underlineitalic{\csname softmax_POVI_MIMean0ecePatchfour\endcsname _{\pm \csname softmax_POVI_MIStd0ecePatchfour\endcsname}}$ & ${\csname softmax_POVI_AUMean0OODPatchfour\endcsname _{\pm \csname softmax_POVI_MIStd0OODPatchfour\endcsname}}$ & ${\csname softmax_POVI_MIMean1OODPatchfour\endcsname _{\pm \csname softmax_POVI_MIStd1OODPatchfour\endcsname}}$ & ${\csname softmax_POVI_MIMean2OODPatchfour\endcsname _{\pm \csname softmax_POVI_MIStd2OODPatchfour\endcsname}}$ \\ \midrule

DE-5 & ${\csname softmax_en_MIMean0accuracy\endcsname _{\pm \csname softmax_en_MIStd0accuracy\endcsname}}$ & ${\csname softmax_en_MIMean0nll\endcsname _{\pm \csname softmax_en_MIStd0nll\endcsname}}$ & ${\csname softmax_en_MIMean0ece\endcsname _{\pm \csname softmax_en_MIStd0ece\endcsname}}$ & ${\csname softmax_en_AUMean0OOD\endcsname _{\pm \csname softmax_en_AUStd0OOD\endcsname}}$ & ${\csname softmax_en_MIMean1OOD\endcsname _{\pm \csname softmax_en_MIStd1OOD\endcsname}}$ & ${\csname softmax_en_MIMean2OOD\endcsname _{\pm \csname softmax_en_MIStd2OOD\endcsname}}$ \\ \bottomrule
\end{tabular}}
\label{tab:ensembles:rlle:dirtyMNIST}
\end{table}

Across the broader set of experiments, the repulsive variant (\acrshort{RLLPOVI}) emerged as the strongest among last-layer \acrshort{povi} methods, consistently outperforming plain and functional variants.  
Performance was particularly boosted by using auxiliary repulsion samples such as image patches, underlining that repulsion sample choice is critical for eliciting meaningful epistemic diversity.  
Overall, \acrshortpl{rlle} matched or exceeded the uncertainty quality of \acrshortpl{de} at a fraction of the computational and storage cost, and their compatibility with pretrained backbones makes them attractive for practical deployment.  
\section{Hardware Evaluation with \acrshort{tvm}}

\begin{table}
	\center
	\small
	\caption{Benchmark \acrshortpl{rlle} vs. \acrshortpl{de} performance on embedded ARM \acrshortpl{cpu}.}
	\label{tab:ensembles:benchmark}
	\begin{tabular}{l | c | c | c | c | r | r | r }
	\textbf{Method} &  \textbf{Dataset} &  \textbf{Arch.} &  $\boldsymbol{n}$ &  \textbf{CPU} &  \textbf{MACs [M]} & 
\multicolumn{2}{c}{\textbf{Latency [ms]}} \\
	\hline
		&   &      &       &     &      & naive & tuned       \\
	\hline

	Det. NN 	& Noisy Sine & MLP				&   - & A72 	&  2.6		&  9.14 &  0.99  \\
	RLLE 		& Noisy Sine & MLP				&  10 & A72 	&  7.8		& 25.93 &  2.93  \\
	DE-10 		& Noisy Sine & MLP				&  10 & A72 	& 26.0		& 91.43 &  9.92  \\
                             
	\hline                   
	                         
	Det. NN 	& Noisy Sine & MLP				&   - & A76 	&  2.6 		&  4.05 &  0.46  \\
	RLLE 		& Noisy Sine & MLP				&  10 & A76 	&  7.8 		& 11.31 &  1.22  \\
	DE-10 		& Noisy Sine & MLP				&  10 & A76 	& 26.0 		& 40.53 &  4.58  \\
                             
	\hline                   
	Det. NN 	& Dirty-MNIST& LeNet-5		 		&   - & A72 & 36.0		& 59.43 &  8.17  \\
	RLLE 		& Dirty-MNIST& LeNet-5		 		&  10 & A72 & 37.0		& 62.41 &  7.94  \\
	DE-10 		& Dirty-MNIST& LeNet-5		 		&  10 & A72 &360.5		&594.31 & 81.69  \\
	                                                                
	\hline                                                          
	                                                                
	Det. NN 	& Dirty-MNIST& LeNet-5		 		&   - & A76 & 36.0		& 26.50 &  2.55  \\
	RLLE 		& Dirty-MNIST& LeNet-5		 		&  10 & A76 & 37.0		& 27.88 &  2.96  \\
	DE-10 		& Dirty-MNIST& LeNet-5		 		&  10 & A76 &360.5		&265.04 & 25.49  \\

	\end{tabular}
\end{table}

The final step in our evaluation is to assess how well \acrlongpl{rlle} can be deployed on embedded hardware and what performance they achieve in practice.  
Because uncertainty-aware methods are only useful in constrained environments if they remain computationally efficient, we benchmark \acrshortpl{rlle} on ARM Cortex-A series processors and compare them against deterministic baselines and \acrlongpl{de}.

One advantage of \acrshortpl{rlle} is their compatibility with existing compiler toolchains.  
Unlike the \acrlong{pfp}, which required custom operators to be integrated into the \acrshort{tvm} stack, \acrshortpl{rlle} rely exclusively on standard deep learning components.  
This makes the compilation process straightforward: models can be imported into \acrshort{tvm} without modification, and performance tuning can be carried out using standard auto-scheduling methods.  
In our experiments we employed the \emph{Meta Scheduler}~\cite{Shao2022tvmMetaScheduler} from the \acrshort{tvm} compiler stack~\cite{chen2018tvm,Zheng2020ansor,Wu2023autotuning} to automatically optimize operator schedules for the specific \acrshort{arm} architectures, which proved essential for achieving efficient execution.  
Untuned schedules left substantial performance untapped, whereas auto-scheduling consistently reduced latency and produced efficient execution across all tested configurations.  

Table~\ref{tab:ensembles:benchmark} summarizes \acrshort{mac} counts and inference latencies for both tasks.  
For the Noisy Sine task, evaluated with a mini-batch size of $1000$, we used a two-layer \acrshort{mlp} with 50 hidden neurons per layer, while the more demanding Dirty-MNIST task, evaluated with a mini-batch size of $128$, was based on LeNet-5.  
The complexity of the backbone architecture determines the relative overhead of the ensemble heads.  
In the small \acrshort{mlp}, ten heads dominate the network, increasing the total size by roughly a factor of three in terms of \acrshortpl{mac}.  
In contrast, for LeNet-5 the additional heads introduce only a $2.7\%$ overhead.  
This highlights a central advantage of \acrshortpl{rlle}: their cost does not scale with the backbone architecture, making them suitable even for very large models.  

\acrlongpl{rlle} reduce training and storage requirements compared to \acrlongpl{de}, while retaining competitive predictive performance.  
On hardware, they achieved $3.4\times$ faster inference on the Noisy Sine task and $8.1\times$ faster on Dirty-MNIST.  
Compiler-level optimizations such as Meta Scheduler tuning further improved latency, but the core efficiency arises from the method itself: by sharing a backbone and introducing diversity only in the last layer, \acrshortpl{rlle} provide uncertainty estimation at a fraction of the computational cost.  
This makes them a practical and scalable choice for deploying uncertainty-aware models on embedded systems.

\subsection{Comparison of PFP and RLLEs}
\label{sec:ensembles:pfp-comparison}

\begin{table}
	\center
	\small
	\caption{Performance \acrshort{pfp} vs. \acrshortpl{rlle}.}
	\label{tab:ensembles:pfp_vs_rlle}
	\begin{tabular}{c|c|c|c|r|r}
		\multirow{2}{*}{\textbf{Dataset}} & 
		\multirow{2}{*}{\textbf{Arch.}} & 
		\multirow{2}{*}{\textbf{CPU}} & 
		\textbf{Batch} & 
		\multicolumn{2}{c}{\textbf{Latency [ms]}} \\ 
		\cline{5-6}
		& & &  \textbf{Size}& \textbf{PFP} & \textbf{RLLE} \\
		\hline                   
		\multirow{12}{*}{Dirty-MNIST} &
		\multirow{12}{*}{LeNet-5} &
		\multirow{6}{*}{A72} 
			& 1   & 1.23 	& 0.51 \\
		& &	& 8   & 7.77 	& 0.91 \\
		& &	& 16  & 15.58 	& 1.32 \\
		& &	& 32  & 32.91 	& 2.71 \\
		& &	& 64  & 70.82 	& 4.89 \\
		& &	& 128 & 150.72 	& 8.00 \\
		\cline{3-6}
		& & \multirow{6}{*}{A76} 
		 	& 1   & 0.50 	& 0.27  \\
		& &	& 8   & 3.03 	& 0.41 \\
		& &	& 16  & 6.33 	& 0.46 \\
		& &	& 32  & 13.60 	& 0.63 \\
		& &	& 64  & 27.46	& 1.11 \\
		& &	& 128 & 61.92 	& 2.93 \\
	\end{tabular}
\end{table}

Finally, we compare \acrshort{pfp} and \acrshortpl{rlle}, the two \acrshort{bnn} approximation methods examined in detail throughout this work.
Both were benchmarked on the Dirty-MNIST dataset with a LeNet-5 backbone to directly assess uncertainty quality and computational performance.  

In terms of uncertainty estimation, both methods perform strongly.  
For \acrlong{ood} detection, \acrshort{pfp} reached an \acrshort{auroc} of $0.966$, while \acrshortpl{rlle} achieved $0.995$, highlighting that both approximate Bayesian inference and ensemble-based approaches yield consistent uncertainty calibration across in- and out-of-distribution regimes.  

The computational comparison, summarized in Table~\ref{tab:ensembles:pfp_vs_rlle}, highlights the distinct performance characteristics of the two approaches.  
All reported measurements were obtained from \acrshort{tvm} builds tuned with the Meta Scheduler~\cite{Shao2022tvmMetaScheduler}, which substantially reduced latency compared to untuned execution for both methods.  
\acrshortpl{rlle} consistently outperform \acrshort{pfp}, and the performance gap widens with increasing batch size.  
For small mini-batches, both methods achieve low latency, with \acrshortpl{rlle} already showing a modest advantage that becomes increasingly pronounced as batch size grows.  
This difference arises from the underlying computational structure of the two methods.  
\acrshortpl{rlle} rely on standard dense operations, dominated by matrix multiplications that are natively supported by highly optimized hardware kernels and thus scale effectively with batch size through vectorization and parallelization.  
Their multi-head design further exposes an additional degree of parallelism: the output heads can be processed largely independently on separate processor cores, as they share the same backbone features but have distinct final layers.  
This enables partial parallel execution without inter-head communication, improving utilization on multi-core embedded platforms while maintaining a small memory footprint.  
In contrast, \acrshort{pfp} propagates both means and variances through each layer, which increases the number of intermediate tensors and memory accesses.  
While parts of the computation can be expressed as matrix multiplications, the frequent reading and writing of mean–variance pairs limits cache locality.  
Moreover, the inherently sequential propagation of stochastic moments offers fewer opportunities for parallel execution than the independent output heads of \acrshortpl{rlle}.  
As a result, \acrshort{pfp} achieves lower throughput at larger batch sizes, despite its efficient formulation at the algorithmic level.  

Conceptually, the two methods occupy complementary positions in the landscape of efficient Bayesian inference.  
\acrshort{pfp} remains within the variational framework and offers a principled, calibration-stable approximation that eliminates stochastic sampling entirely through closed-form propagation.  
\acrshortpl{rlle}, on the other hand, embrace an ensemble-based view, retaining the multi-modality of Deep Ensembles while remaining compatible with standard toolchains and hardware accelerators.  
This design yields remarkable speed and flexibility, though at the expense of increased sensitivity to calibration choices such as the repulsion strength and the selection of repulsive samples.  

Taken together, the results underline that neither approach dominates universally.  
The \acrlong{pfp} offers theoretical rigor and intrinsic stability, whereas \acrshortpl{rlle} achieve greater empirical efficiency and scalability.  
Both methods therefore represent viable and complementary pathways for enabling uncertainty-aware inference on embedded systems.  

\section*{Summary}
\label{sec:ensembles:takeaways}

This chapter has examined ensemble-based approximations for Bayesian neural networks, focusing on \acrshort{mcdo}, \acrshortpl{de}, and \acrshortpl{rlle}.  
Our controlled experiments on the Noisy Sine task confirmed the known weaknesses of \acrshort{mcdo}, the robustness and empirical strength of \acrshortpl{de}, and the potential of \acrshortpl{rlle} to achieve competitive uncertainty estimates at a fraction of the training and storage cost.  
While \acrshortpl{rlle} demand careful calibration of activation functions, hyperparameters and well chosen repulsive samples, their design provides explicit flexibility to balance epistemic and aleatoric uncertainty.  

On the hardware side, we showed that \acrshortpl{rlle} integrate seamlessly with standard compiler toolchains.  
Thanks to their reliance on standard operators, they can be compiled and tuned with \acrshort{tvm} directly, turning their theoretical efficiency into practical speedups on embedded CPUs.  
Because the additional heads add only marginal overhead compared to larger backbones, \acrshortpl{rlle} scale favorably to larger architectures and remain well suited for deployment under tight resource constraints.  

In summary, \acrshortpl{rlle} strike a pragmatic balance between theoretical grounding and practical deployability.  
They retain much of the uncertainty quality of full ensembles while reducing computational cost to levels compatible with embedded devices.  

The following chapter extends this perspective on probabilistic inference beyond digital accelerators, investigating photonic hardware where inherent device noise is harnessed directly as a stochastic source for probabilistic inference.

\chapter{Probabilistic Photonic Computing for Bayesian Neural Networks}
\label{ch:photonics}

\epigraph{Information is not a disembodied abstract entity;\\it is always tied to a physical representation.}{\textnormal{--- Rolf Landauer, \textit{The Physical Nature of Information} (1996)}}

\noindent
In the preceding chapters we investigated two complementary strategies to reduce the cost of deploying neural networks on resource-constrained devices.  
From the algorithmic side, we developed methods to reduce the number of costly computations, while from the hardware side, we considered approaches that lower the cost of the operations themselves.  
For deterministic \acrlongpl{dnn}, this two-pronged perspective was reflected by \emph{Galen} as an automatic compression method on the algorithmic level~\cite{krieger2023galen}, and by analog accelerators as a means to perform the underlying matrix multiplications more efficiently~\cite{klein2021bss2whitebox,kuhn2023nonassociativity,borras2024walkingnoiseECML,wang2025hardening,wang2025vant}.  

In this chapter we extend this perspective to \acrlongpl{bnn}.
Algorithmic approximations such as partial \acrshortpl{bnn}~\cite{kristiadi2020partialbnns,sharma2023partialbnns}, \acrshortpl{rlle}~\cite{steger2024rlle} and the \acrshort{pfp}~\cite{roth2016pfp,klein2025pfp} reduce the computational burden of probabilistic inference.  
At the hardware level, analog computing platforms promise to accelerate \acrshort{bnn} inference in a complementary way, by embedding probabilistic operations directly into physical processes.  

This perspective is particularly appealing because \acrshortpl{bnn} require stochasticity at their core.  
On digital processors, randomness must be simulated by pseudo-random number generators and integrated into multiple forward passes, which adds substantial computational overhead.  
Analog hardware, by contrast, is inherently noisy.  
While this noise is typically seen as a liability for deterministic inference, for probabilistic inference it can serve as a natural entropy source.  
If the noise can be shaped and controlled, the probabilistic sampling step of \acrshortpl{bnn} can be executed directly in hardware, making uncertainty estimation both faster and more energy efficient.  

Photonic computing offers a compelling platform for this idea.  
Light not only enables extremely high bandwidth and parallelism---for example through wavelength-division multiplexing---but also provides accessible entropy sources in the form of intensity fluctuations, phase noise, or quantum effects~\cite{brueckerhoff2025photonicAI}.  
These processes can be leveraged to realize the stochastic sampling required by Bayesian inference, aligning the algorithmic demands of probabilistic models with the physical properties of the hardware.

This chapter builds on two recent joint works that we conducted in collaboration with researchers specializing in photonic hardware.  
The position paper \emph{Probabilistic Photonic Computing for AI}~\cite{brueckerhoff2025photonicAI} and the proof-of-concept demonstration \emph{Probabilistic Photonic Computing with Chaotic Light}~\cite{brueckerhoff2024chaoticlight} emerged from this collaboration.  
The projects were carried out in a co-design effort, combining our expertise in probabilistic modeling and \acrlongpl{bnn} with our partners’ expertise in photonic device design and experimental realization.  
Accordingly, this chapter emphasizes the proof of concept as a hardware-accelerated \acrshort{bnn} inference engine and highlights the algorithmic adaptations required to map \acrlongpl{bnn} onto photonic hardware.  
For detailed discussions of the photonic design choices and device physics, we refer the reader to the original publications.  
\section{Photonic Neural Network Inference}

\acrshortpl{bnn} provide a principled framework for predictive uncertainty but incur high inference cost due to repeated sampling and multiple forward passes.  
On digital processors, pseudo-random number generation and sample-wise evaluation add latency and energy overhead.  
Algorithmic approaches reduce or sidestep repeated sampling, thereby lowering the computational burden, yet they rarely eliminate the fundamental need for stochasticity in probabilistic inference.

Analog computing platforms, including electrical accelerators~\cite{klein2021bss2whitebox,kuhn2023nonassociativity}, memristive crossbars~\cite{camus2019trainingaware,zhang2019precisionaware}, and photonic processors~\cite{Shen2017,Lin2018}, promise substantial energy efficiency compared to digital hardware.  
However, they are inherently subject to imperfections such as nonlinearities, device variations, and computational noise~\cite{borras2024walkingnoiseECML,wang2025hardening,wang2025vant}.  
For deterministic inference, such noise degrades accuracy and requires countermeasures like noisy or \acrlong{hil} training.  
For probabilistic inference, by contrast, noise is not necessarily a limitation: it can serve as an entropy source and thus as the very building block of Bayesian computation.  
This shift in perspective, from mitigating noise to exploiting it, motivates the exploration of analog hardware for accelerating probabilistic inference.  

Photonic computing offers properties that are particularly attractive for this idea.  
Light propagates at extremely high speed and enables massively parallel signal processing through \acrfull{wdm}.  
The available optical bandwidth in the telecom band alone exceeds several \acrshort{thz}, far surpassing electronic bandwidths~\cite{brueckerhoff2025photonicAI}.  
Moreover, photonic systems inherently exhibit entropy sources, including phase noise, intensity fluctuations in chaotic light, and quantum fluctuations at the single-photon level~\cite{Goodman2000,Vannucci1980,Shimoda1957,Pietralunga2003}.  
These noise processes can be harnessed as true random number generators, avoiding the overhead of digital pseudo-random generators and directly linking stochasticity to the physical information carrier.  

\vspace{0.5em}
\noindent
A wide range of photonic neural architectures has been proposed over the past decade.  
Coherent nanophotonic meshes based on \acrlongpl{mzi} (\acrshortpl{mzi}) enable programmable optical linear algebra~\cite{Shen2017}, while diffractive optical networks perform passive all-optical inference~\cite{Lin2018}.  
Microring weight banks and crossbar arrays based on \acrlongpl{pcm} (\acrshortpl{pcm}), in particular \acrfull{gst}, allow compact in-memory optical multiplication~\cite{Nandakumar2018pcm,Joshi2020}.  
These works primarily target deterministic inference, but they establish the device toolkit that can also support probabilistic computing.  

Entropy sources are a central ingredient of probabilistic computing.  
Classical statistical optics established the photon statistics of coherent and chaotic fields~\cite{Goodman2000,Vannucci1980,Shimoda1957}, while later studies investigated \acrlong{ase} (\acrshort{ase}) statistics in dense \acrlong{wdm} regimes relevant to telecom applications~\cite{Pietralunga2003}.  
Building on these foundations, chaotic light has been identified as a particularly practical entropy source, since it can be generated in the telecom band by erbium-doped fibers or waveguides.  
Unlike digital pseudo-random generators, chaotic-light entropy is broadband, spans multiple \acrshort{thz}, and is directly compatible with optical processors~\cite{brueckerhoff2024chaoticlight}.  

In parallel, analog non-volatile memories such as electronic memristors have been proposed as hardware substrates for probabilistic neural inference.  
Randomness in switching processes and conductance drift can be interpreted as tunable stochastic weights, and recent work has demonstrated Bayesian inference with memristor-based neural networks~\cite{Bonnet2023}.  
These works exploit stochasticity in electronic devices, reinforcing the broader trend of turning hardware noise from a liability into a computational resource.  

Our recent works on \emph{probabilistic photonic computing for AI}~\cite{brueckerhoff2025photonicAI} and on a prototype implementation using \emph{chaotic light}~\cite{brueckerhoff2024chaoticlight} extended this concept to photonic systems.  
We introduced chaotic light as a controllable entropy source and demonstrated its integration with \acrshort{gst}-based photonic crossbars, thereby establishing photonic processors as a viable platform for probabilistic inference.  

\section{Hardware Design Principles}

Photonic systems expose multiple noise sources that can be harnessed for probabilistic computing~\cite{brueckerhoff2025photonicAI}.  
Among the most relevant are phase noise in coherent lasers, intensity noise in chaotic light sources, and quantum fluctuations at the single-photon level.  
All three provide access to physical entropy, but they differ in ease of integration with large-scale photonic processors.  

Phase noise can be exploited through interferometric readout schemes and has been used for photonic random number generation.  
However, interferometers impose strict stability requirements, and integrating phase-sensitive components at scale remains challenging.  
Quantum fluctuations, such as those exploited in true photon-number–resolving detectors, provide intrinsic randomness but require cryogenic operation and highly sensitive components, making them impractical for near-term integrated photonic processors.  
In contrast, chaotic light generated by amplified spontaneous emission (\acrshort{ase}) is broadband, robust, and easily available in the telecom band.  
Its intensity fluctuations follow Bose–Einstein statistics and can be directly measured with standard photodetectors.  
For these reasons, chaotic light was chosen as the entropy source in the prototype of~\cite{brueckerhoff2024chaoticlight}.  

\begin{figure}
  \centering
  \begin{subfigure}{\linewidth}
    \includegraphics[width=\linewidth]{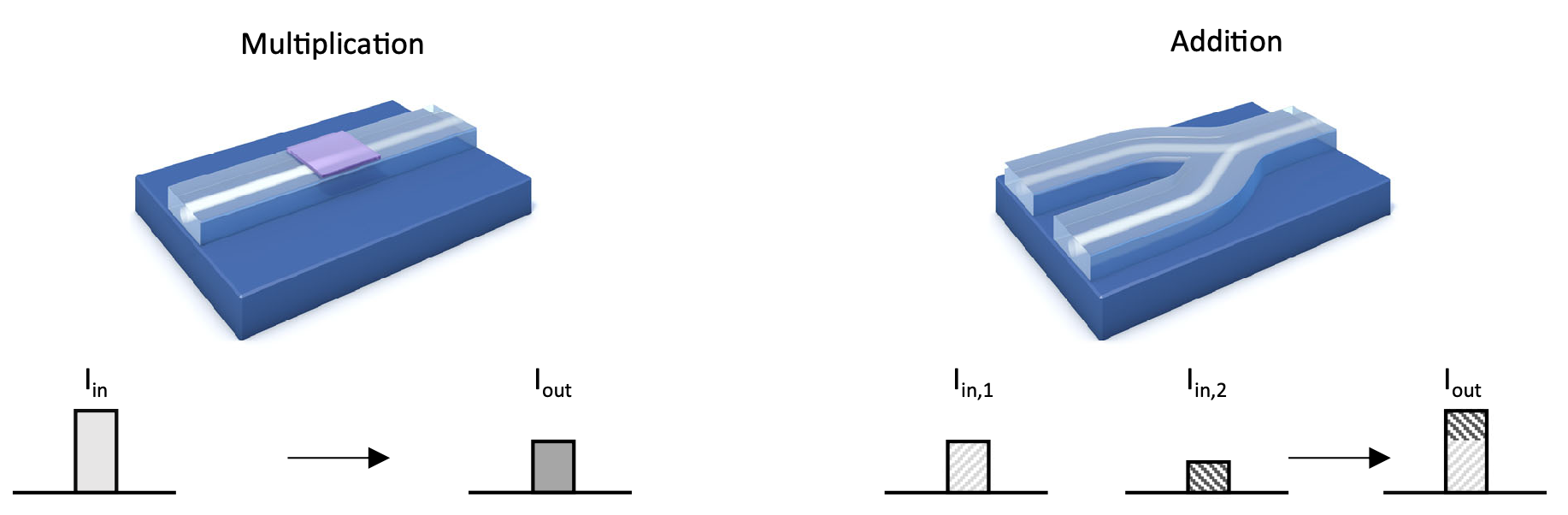}
    \caption{Multiplication and Addition}
    \label{fig:crossbar-overview-mul}
  \end{subfigure}

  \vspace{1em}
  \begin{subfigure}{0.4\linewidth}
    \includegraphics[width=\linewidth]{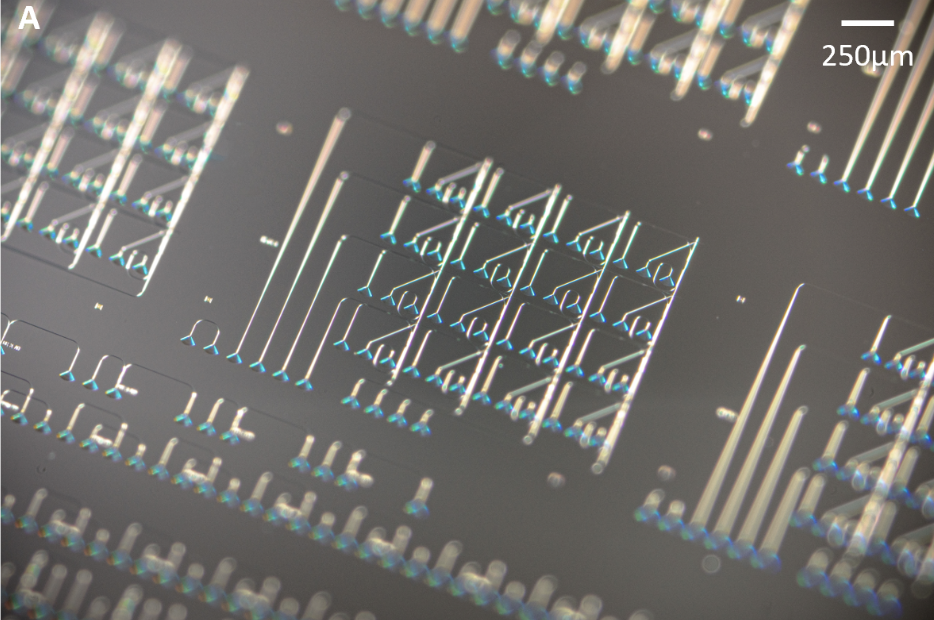}
    \caption{Chip Photograph}
    \label{fig:crossbar-overview-photo}
  \end{subfigure}

  \vspace{1em}

  \begin{subfigure}{\linewidth}
    \includegraphics[width=\linewidth]{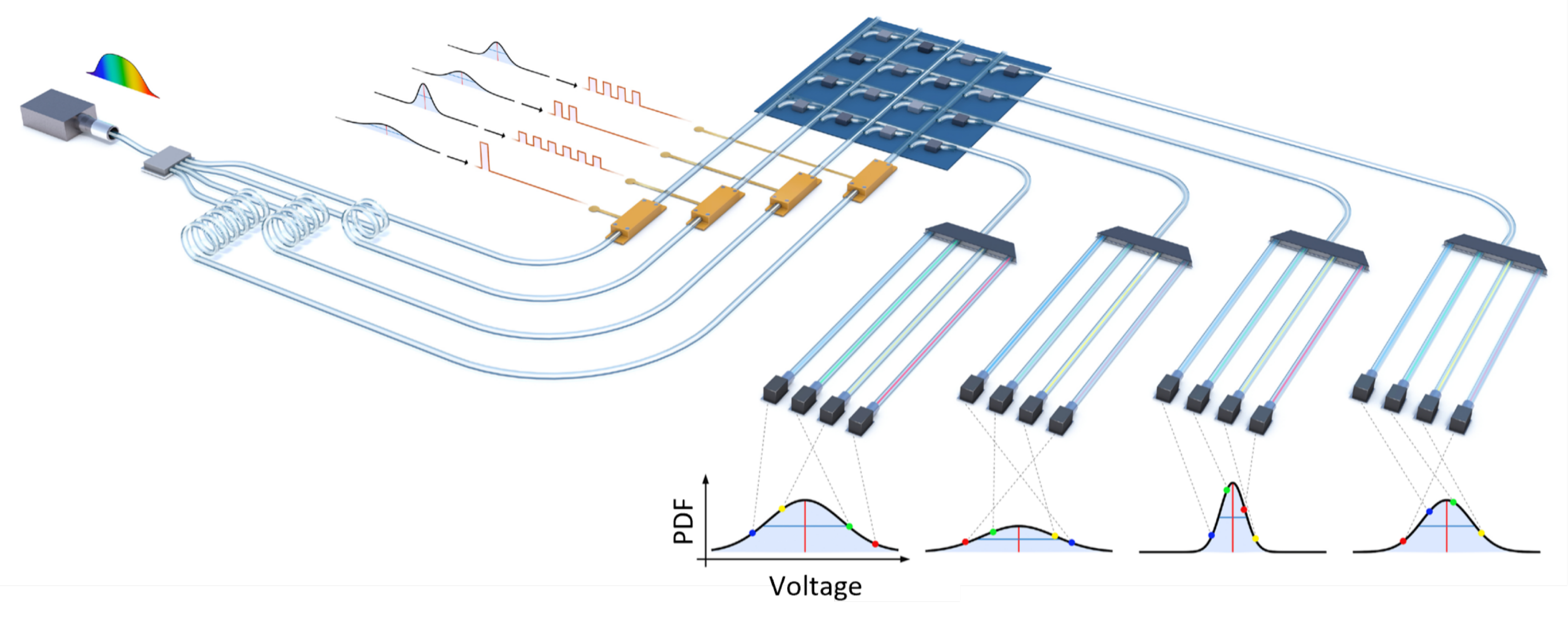}
    \caption{Crossbar Structure}
    \label{fig:crossbar-overview-3d}
  \end{subfigure}

  \caption{Photonic crossbar with \acrshort{gst} weights.  
  Subfigure~\subref{fig:crossbar-overview-mul} illustrates how multiplication is realized by attenuating the optical carrier according to the programmed \acrshort{gst} state, while addition is implemented by overlapping multiple optical fields in a waveguide, consistent with the statistical behavior of chaotic light distributions reported in~\cite{brueckerhoff2024chaoticlight}.  
  Subfigure~\subref{fig:crossbar-overview-photo} shows a microscope photograph of the fabricated photonic chip.  
  Subfigure~\subref{fig:crossbar-overview-3d} provides a three-dimensional overview of the full crossbar structure, where chaotic light from an \acrshort{ase} source is split and delayed to form multiple uncorrelated optical carriers.  
  Input distributions are encoded as waveforms on these carriers, processed through the \acrshort{gst}-based crossbar, and demultiplexed by \acrshort{wdm} into independent sampling channels.  
  \reprofrom{brueckerhoff2024chaoticlight}}
  \label{fig:crossbar-overview}
\end{figure}

To enable matrix multiplication, the processor employs a photonic crossbar array.  
Weights are stored in the non-volatile phase-change material \acrlong{gst}, which provides multiple stable transmission levels.  
Multiplication is realized by attenuating the optical carrier according to the programmed transmission state (Fig.~\ref{fig:crossbar-overview-mul}), and addition is performed by summing overlapping optical fields in a waveguide.  
A photograph of the fabricated chip is shown in Fig.~\ref{fig:crossbar-overview-photo}, while the full crossbar structure is depicted in Fig.~\ref{fig:crossbar-overview-3d}.

Parallelism is achieved by \acrlong{wdm}.  
The broadband spectrum of chaotic light spans several \acrshort{thz}, which can be demultiplexed into distinct wavelength channels.  
Each channel carries an independent realization of the chaotic fluctuations, enabling multiple uncorrelated random samples to be processed simultaneously.  
This property directly accelerates probabilistic inference, where many stochastic samples are typically required.  

Finally, several practical constraints shape the design.  
The independence of random numbers is limited by the correlation time of chaotic fluctuations and the electrical bandwidth of the detectors.  
Residual correlations between wavelength channels can occur due to imperfect demultiplexing.  
Moreover, photonic components such as waveguides and detectors remain larger than electronic counterparts, imposing constraints on integration density~\cite{brueckerhoff2025photonicAI}.  

In summary, chaotic light was selected as a pragmatic entropy source due to its controllability and ease of integration with \acrshort{gst}-based photonic crossbars.  
Combined with \acrshort{wdm} for parallelization, these design choices establish a physical platform for accelerating probabilistic inference on photonic hardware (Fig.~\ref{fig:crossbar-overview}).  

\section{Making Noise Controllable}

\begin{figure}[t]
  \centering
  \begin{subfigure}{0.59\linewidth}
    \includegraphics[width=\linewidth]{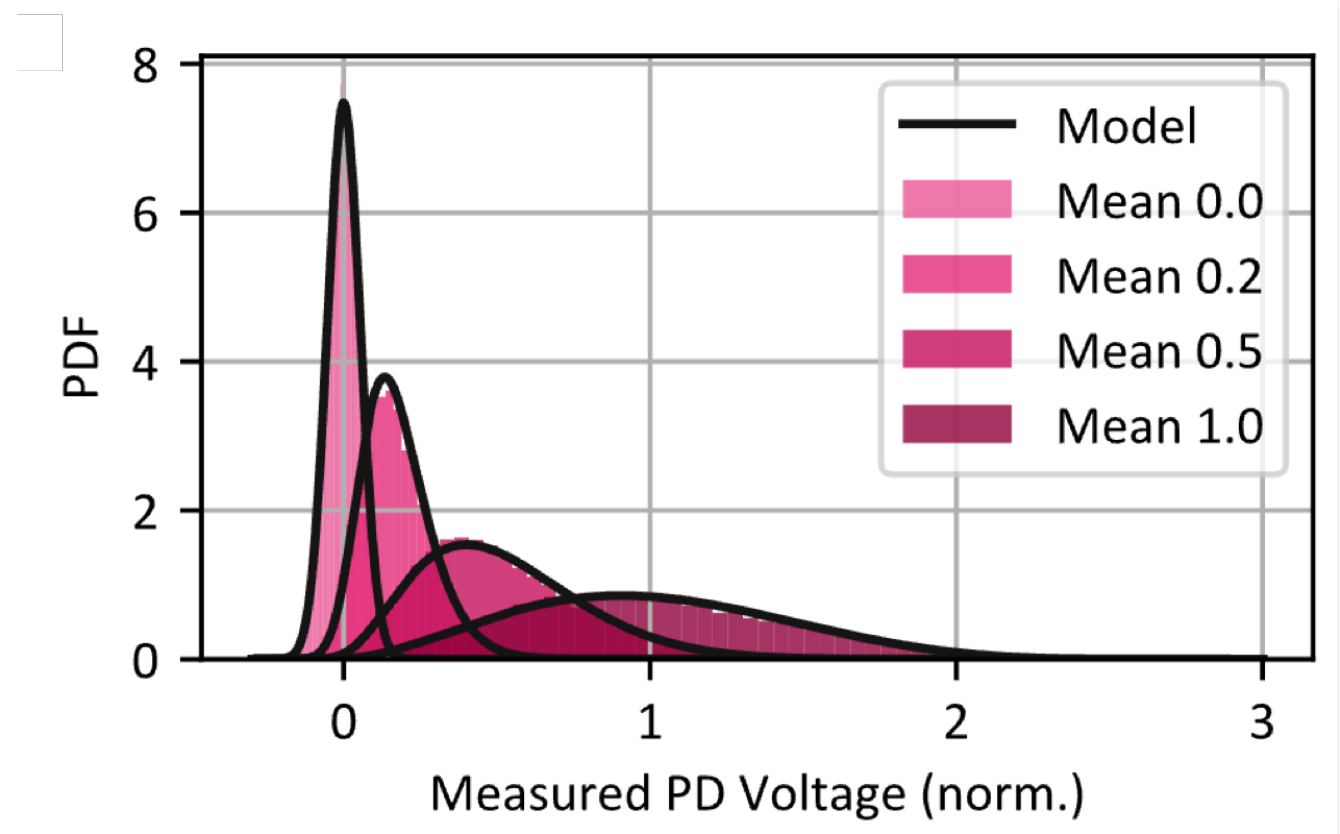}
	\caption{Measured \acrshortpl{pdf} at different mean intensities}
    \label{fig:photonics:mean-dist-hist}
  \end{subfigure}
  \hfill
  \begin{subfigure}{0.375\linewidth}
    \includegraphics[width=\linewidth]{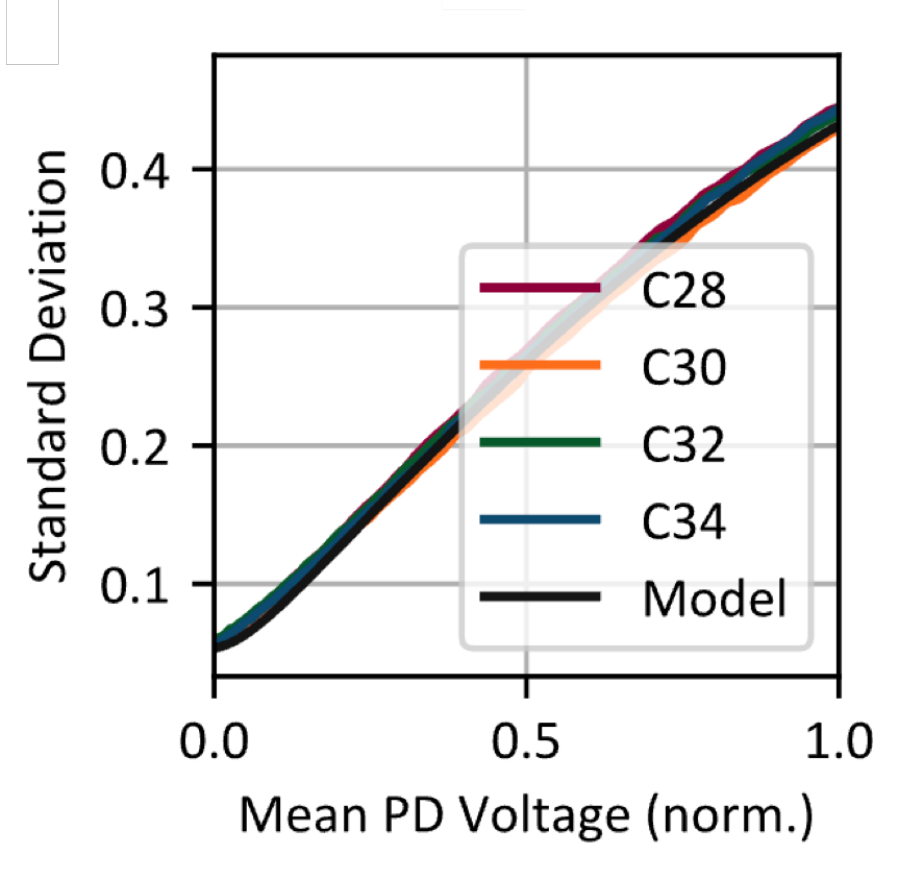}
    \caption{Variance–mean coupling}
    \label{fig:photonics:mean-noise-lin}
  \end{subfigure}

  \vspace{1em}

  \begin{subfigure}{0.92\linewidth}
    \includegraphics[width=\linewidth]{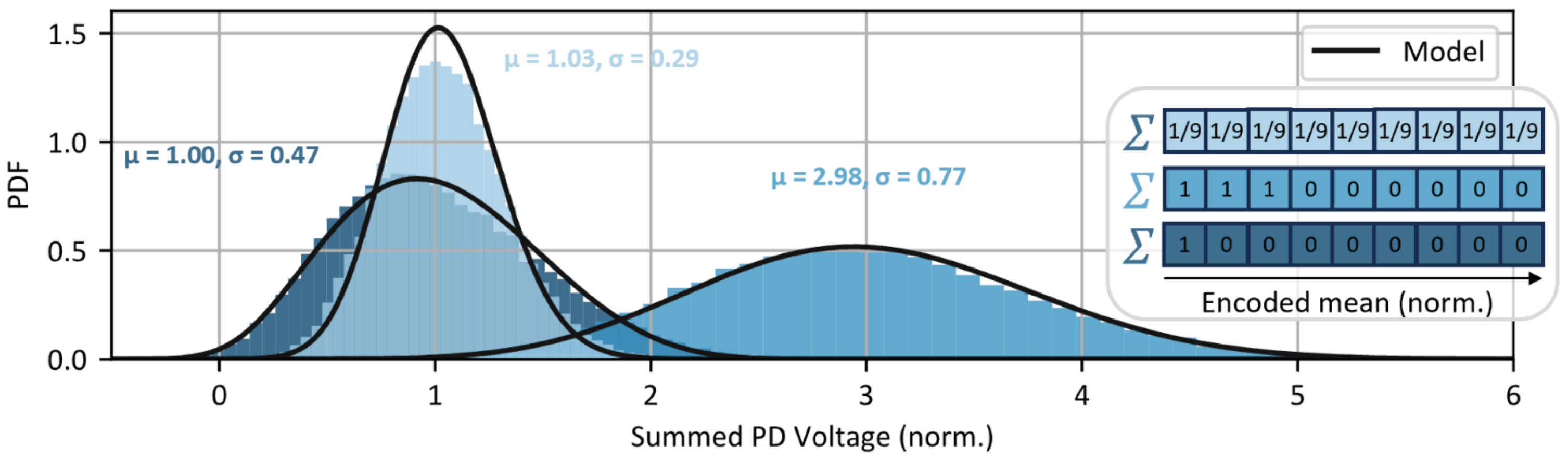}
    \caption{Variance control by encoding scheme}
    \label{fig:photonics:hist-encoding}
  \end{subfigure}

\caption{Statistical properties of chaotic-light noise.  
Subfigures~\subref{fig:photonics:mean-dist-hist} and~\subref{fig:photonics:mean-noise-lin} confirm Bose--Einstein statistics and the quadratic dependence of the variance on the mean intensity.  
Subfigure~\subref{fig:photonics:hist-encoding} illustrates controllable-noise encoding, where the same total intensity yields distributions with different variances.  
\reprofrom{brueckerhoff2024chaoticlight}}
  \label{fig:photonics:cl-fig2}
\end{figure}

A central challenge in harnessing physical noise for probabilistic computing lies not only in accessing it but in making it controllable.  
The measured signal at the photodetector combines two contributions: electronic noise from the readout circuit and intensity fluctuations of the chaotic-light carrier.  
At small optical intensities, the measured distribution is dominated by electronic noise, which is approximately Gaussian and independent of the mean intensity.  
With increasing intensity, photonic fluctuations dominate and the distributions evolve towards the Bose--Einstein form expected from chaotic light~\cite{Goodman2000,Vannucci1980,Shimoda1957,brueckerhoff2024chaoticlight}.  

The beating of frequency components in \acrlong{ase} leads to photon-number fluctuations described by an $M$-fold Bose--Einstein distribution~\cite{brueckerhoff2024chaoticlight}.  
Its variance grows quadratically with the mean intensity,  
\begin{equation}
\mathrm{Var}(n) = n_{\text{mean}} + \frac{n_{\text{mean}}^2}{M},
\end{equation}  
where $n_{\text{mean}}$ is the average photon number detected within the measurement interval and directly corresponds to the optical intensity,  
and $M$ is the degeneracy factor, approximated by the ratio of the measurement time $T$ to the coherence time $\tau_c$ of the source~\cite{brueckerhoff2024chaoticlight}.  
In the presented setup $M \approx 10.64$.  

As a consequence of this quadratic relation, mean and variance cannot be tuned independently.  
This intrinsic coupling poses a challenge when mapping probabilistic operations onto hardware, since \acrshortpl{bnn} require distributions with independently adjustable parameters.  

\begin{figure}[t]
  \centering
	\begin{subfigure}[b]{0.47\linewidth}
    \includegraphics[width=\linewidth]{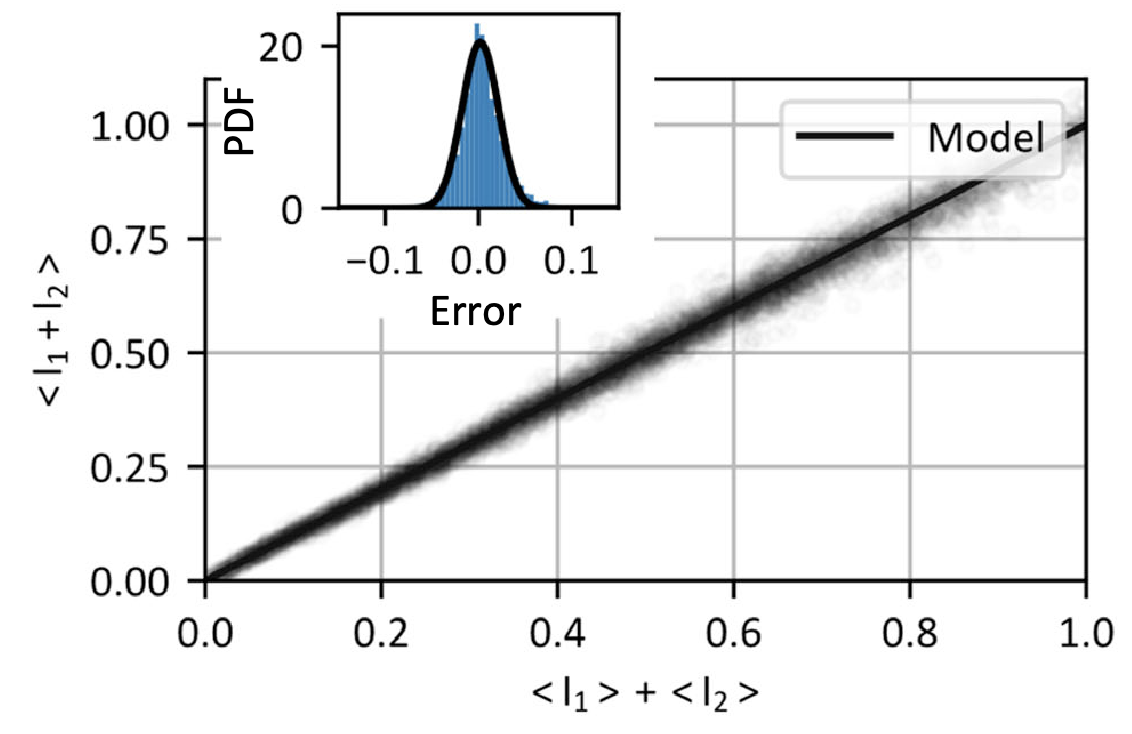}
    \caption{Mean}
    \label{fig:photonics:addition:mean}
  \end{subfigure}
  \hfill
	\begin{subfigure}[b]{0.47\linewidth}
    \includegraphics[width=\linewidth]{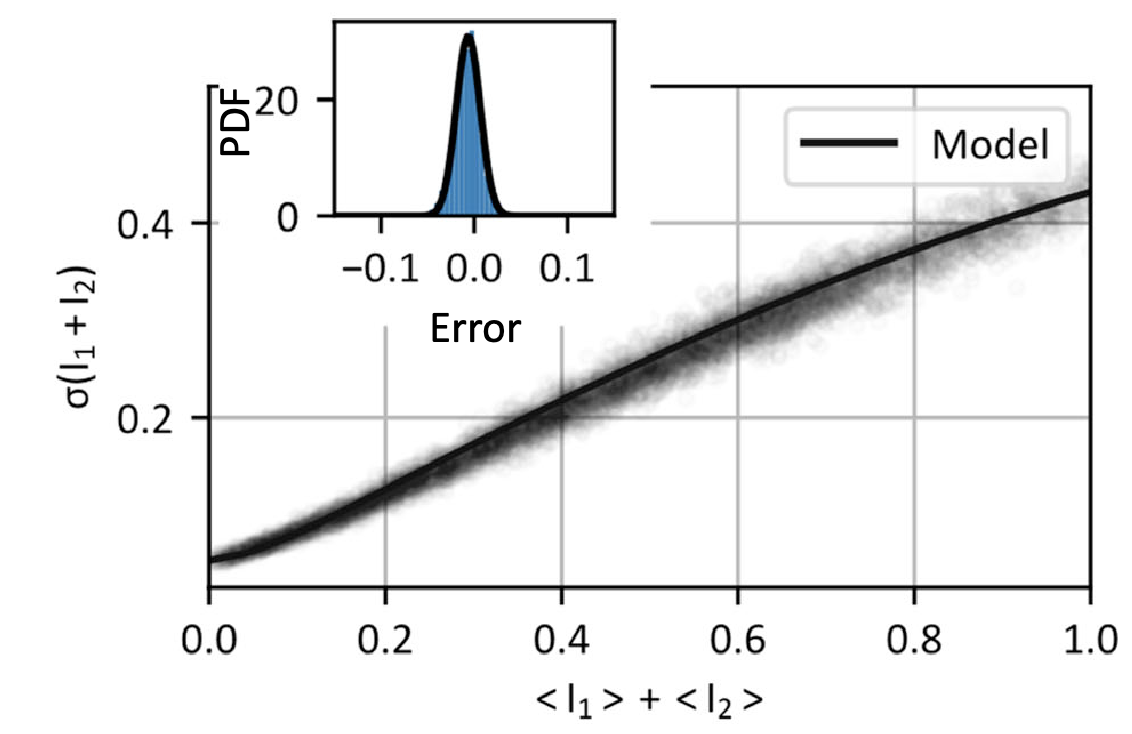}
    \caption{Standard Deviation}
    \label{fig:photonics:addition:std}
  \end{subfigure}

\caption{Experimental validation of addition with noisy signals.  
Subfigure~\subref{fig:photonics:addition:mean} shows that the means add linearly, as expected.  
In contrast, Subfigure~\subref{fig:photonics:addition:std} shows that the standard deviations scale nonlinearly due to the constant electronic ground noise contribution.  
\reprofrom{brueckerhoff2024chaoticlight}}
  \label{fig:photonics:cl-fig3}
\end{figure}

This behavior is confirmed experimentally (Fig.~\ref{fig:photonics:cl-fig2}): at low means the distributions are narrow and Gaussian-like, while at higher means they broaden and follow Bose--Einstein statistics (Fig.~\ref{fig:photonics:mean-dist-hist}, Fig.~\ref{fig:photonics:mean-noise-lin}).  
Detector saturation further limits the width at large means (Fig.~\ref{fig:photonics:mean-noise-lin}).  
Another property becomes evident when input distributions are accumulated (Fig.~\ref{fig:photonics:cl-fig3}).  
The means add linearly (Fig.~\ref{fig:photonics:addition:mean}), whereas the standard deviations follow a nonlinear curve due to the mean-invariant electronic noise contribution (Fig.~\ref{fig:photonics:addition:std})~\cite{brueckerhoff2024chaoticlight}.  

Building on this principle, we propose a scheme to control the variance independently of the mean.  
Where not only the number of active slots but also their amplitudes are adjusted.  
The same total intensity can either be represented by a few strong impulses, introducing proportionally more noise, or by many smaller impulses, where the noise averages out and becomes sub-proportional.  
As a result, the same intensity can yield output distributions with different variances (Fig.~\ref{fig:photonics:hist-encoding}).  
This controllable-noise encoding enables a decoupling of mean and variance, within the bounds set by the physical noise sources, and thus renders chaotic-light noise tunable.  
\section{Adapting \acrshortpl{bnn} to Photonic Hardware}

The feasibility of photonic probabilistic inference depends not only on the availability of tunable noise but also on adapting \acrlong{bnn} operators to the constraints of the hardware.  
Rather than assuming a generic architecture, the \acrlong{bnn} must be co-designed with the photonic device to exploit its strengths while compensating for its limitations.  

The first design choice is to realize stochasticity through probabilistic activations rather than probabilistic weights.  
Conventional \acrshortpl{bnn} often assign distributions to weights, but this strategy is infeasible in our setting for two reasons.  
The prototype crossbar provides only a limited number of programmable weight cells, and although \acrlong{pcm} devices based on \acrshort{gst} are well suited for compact in-memory multiplication, their transmittance can only be reprogrammed slowly compared to photonic processes.  
Rapidly varying weights for sampling is therefore not a viable option.  
In contrast, the chaotic-light carrier produces intrinsic intensity fluctuations that are directly detected at the photodiode readout.  
These fluctuations supply random samples at high rate, and their variance can be controlled by the proposed encoding scheme.  
It is thus more efficient to keep the weights deterministic and implement randomness at the activation level.  

A second design choice concerns numerical precision.  
The optical crossbar supports only a limited number of transmission levels, and the electrical readout is quantized.  
To maintain robustness under these conditions, \acrfull{qat}~\cite{jacob2018quantization,krishnamoorthi2018quantization} is required so that the \acrshort{bnn} representation matches the effective bit-width of the hardware.  
Quantization for \acrshortpl{bnn} remains largely unexplored, but our evaluation shows that the behavior is similar to conventional deep neural networks: once the bit-width falls below a critical threshold, accuracy and uncertainty estimation collapse together.  
This threshold depends on both the architecture and the dataset.  
Nevertheless, provided the bit-width remains above this limit, \acrshort{qat} enables reliable inference under reduced precision.  

A third design choice addresses the noise characteristics of the system.  
The chaotic-light fluctuations are tunable only within a fixed range, and on top of this comes a constant base noise floor mainly from the electronic components.  
As a result, the \acrshort{bnn} can adjust activation variances only within these bounds, while the residual base noise must be learned to tolerate during hardware-model training.  

A fourth design choice arises from the strictly positive nature of the optical signals.  
In this setup, light can only be added but not subtracted, so the hardware enforces a positive-only domain.  
As a consequence, negative weights and activations are excluded and must be compensated for at the algorithmic level.  

Finally, the limited chip area of the prototype constrains the number of weight cells.  
This rules out large-scale dense or convolutional layers.  
Instead, we developed a probabilistic average-pooling operator that requires only a few weights but exposes a large number of stochastic degrees of freedom by allowing noise parameters to be tuned on a per-activation level.  

Figures~\ref{fig:photonics:probpool-training} and~\ref{fig:photonics:probpool-simulation} illustrate the probabilistic photonic average-pooling operator in its two complementary forms: 
a training model and a physical simulation model, both implemented in Pyro.  
The coexistence of both implementations reflects the need to reconcile efficient gradient-based optimization in software 
with a faithful representation of the stochastic behavior observed on photonic hardware.  

The training model employs a Gaussian approximation of the Bose--Einstein distribution 
to avoid an additional sampling step and to accelerate training, while remaining sufficiently close to the physical statistics.  
This abstraction enables hardware-aware \acrshort{svi}-based training with realistic noise characteristics at a manageable computational cost.  
In contrast, the physical model samples directly from the exact Bose--Einstein \acrshort{pdf}, 
includes stochastic quantization, and explicitly accounts for photonic imperfections and limited precision.  
Together, these two operator variants ensure consistency between training and hardware-level simulation, 
allowing the \acrshort{bnn} to be optimized efficiently in software and evaluated accurately on the photonic prototype.  

In summary, adapting \acrlongpl{bnn} to photonic hardware requires several key design choices, as exemplified by our prototype implementation:  
stochasticity is implemented in the activations rather than the weights,  
representations are trained with reduced precision in mind,  
activation variances are adjusted only within constrained bounds,  
the positive-only nature of optical signals is considered,  
and operators are tailored to the limited number of available weights.  
Together, these adaptations enable a proof-of-concept demonstration of probabilistic inference on a photonic accelerator.  

\begin{figure}
  \centering
  \begin{tikzpicture}[
    node distance=8mm,
    every node/.style={font=\sffamily},
    block/.style={draw, minimum width=40mm, minimum height=8mm, align=center, rounded corners=2pt},
    detblock/.style={block, fill=gray!15},
    probblock/.style={block, fill=blue!15},
    arrow/.style={-Latex, thick, shorten >=2pt} %
  ]

  \node[detblock] (input) {Input};
  \node[detblock, below=of input] (quantrelu) {QuantReLU};
  \node[detblock, below=of quantrelu] (avgpool) {Avg.\ Pool 2$\times$2};

  \node[probblock, right=14mm of avgpool] (sigmabounds) {Compute $\sigma_{\min},\ \sigma_{\max}$};
  \node[probblock, below=of sigmabounds] (mixsigma) {$\sigma \;=\; L\,\sigma_{\max} + (1\!-\!L)\,\sigma_{\min}$};
  \node[probblock, below=of mixsigma] (gausssampler) {Gaussian Sampler\\$\mathcal{N}(0,\sigma)$};

  \node[detblock, below=38mm of avgpool] (sum) {$+$};

  \node[anchor=west] (Llabel) at ($(mixsigma.east)+(6mm,0)$) {noise $L$ (learnable)};
  \draw[arrow] (Llabel.west) -- (mixsigma.east);

  \draw[arrow] (input) -- (quantrelu);
  \draw[arrow] (quantrelu) -- (avgpool);

  \draw[arrow] (avgpool.east) -- (sigmabounds.west);
  \draw[arrow] (sigmabounds) -- (mixsigma);
  \draw[arrow] (mixsigma) -- (gausssampler);

  \draw[arrow] (avgpool.south) -| (sum.north);
  \draw[arrow] (gausssampler.south) |- (sum.east);

  \end{tikzpicture}

	\caption{Photonic probabilistic average-pooling operator (\emph{training model}).  
  	Deterministic blocks (gray) compute activations via QuantReLU and average pooling.  
	This model is used during training with Pyro; the Gaussian approximation of the Bose--Einstein distribution avoids an additional sampling step and significantly accelerates training while preserving the learned distributional characteristics.  
  	\reprofrom{brueckerhoff2024chaoticlight}}
  	\label{fig:photonics:probpool-training}
\end{figure}
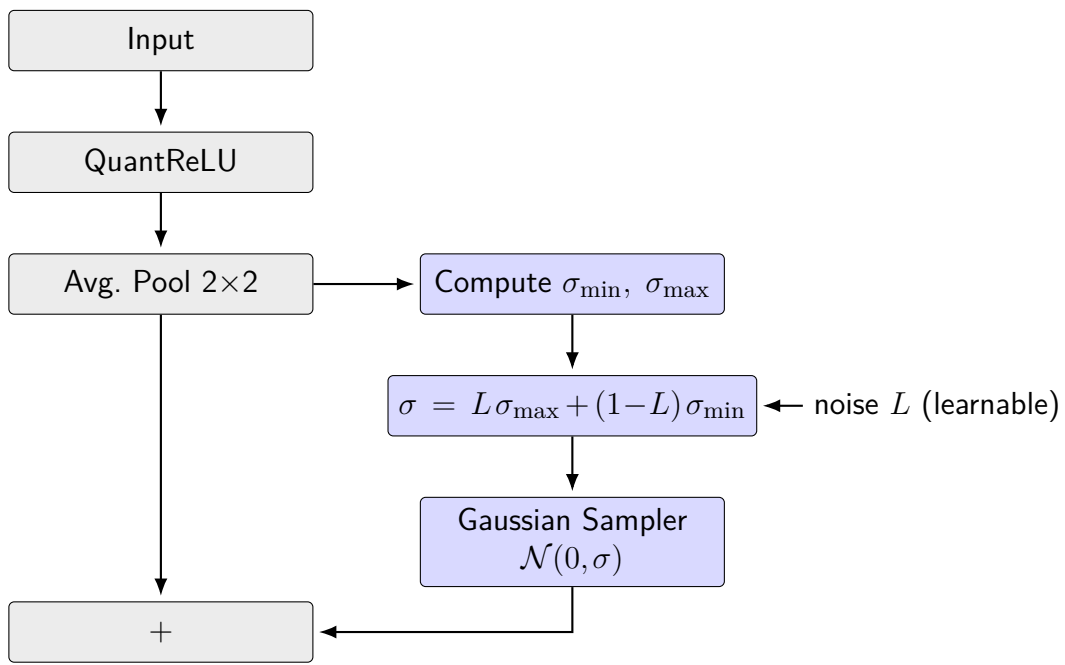

\begin{figure}[t]
  \centering
  \begin{tikzpicture}[
    node distance=7mm,
    every node/.style={font=\sffamily},
    block/.style={draw, minimum width=38mm, minimum height=8mm, align=center, rounded corners=2pt},
    detblock/.style={block, fill=gray!15},
    probblock/.style={block, fill=blue!15},
    noisesource/.style={block, fill=blue!15},
    arrow/.style={-Latex, thick, shorten >=2pt}
  ]

  \node[detblock] (input) {Input};
  \node[detblock, below=of input] (quantrelu) {QuantReLU};
  \node[detblock, below=of quantrelu] (pattern) {Pattern Generation};
  \node[noisesource, right=12mm of pattern] (noise) {Adjustable Noise};
  \node[probblock, below=of pattern] (stochquant) {Stochastic Quantization};
  \node[probblock, below=of stochquant] (pool) {Average Pool 2$\times$2};
  \node[probblock, below=of pool] (pdf) {PDF Sampling};
  \node[detblock, below=of pdf] (sum) {Summation of Samples};

  \draw[arrow] (input) -- (quantrelu);
  \draw[arrow] (quantrelu) -- (pattern);
  \draw[arrow] (pattern) -- (stochquant);
  \draw[arrow] (noise.west) -- (pattern.east);
  \draw[arrow] (stochquant) -- (pool);
  \draw[arrow] (pool) -- (pdf);
  \draw[arrow] (pdf) -- (sum);

  \end{tikzpicture}

  \caption{Photonic probabilistic average-pooling operator (\emph{physical simulation model}).  
  Deterministic preprocessing (gray) is followed by probabilistic operations (blue), where randomness is injected from the chaotic-light noise source.  
  The layer performs stochastic quantization, average pooling, and \acrshort{pdf}-based sampling using the exact Bose–Einstein distribution, with cached \acrshortpl{pdf} reused to reduce computational overhead.  
  This version models the hardware-level behavior including photonic imperfections and quantization effects.  
  \reprofrom{brueckerhoff2024chaoticlight}}
  \label{fig:photonics:probpool-simulation}
\end{figure}
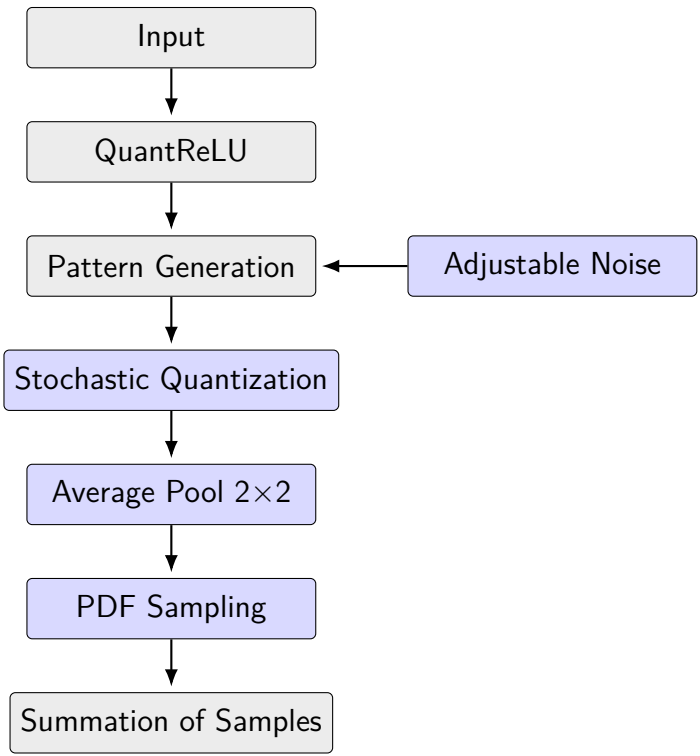

\section{Experimental Demonstration}

\begin{figure}[t]
  \centering
  \includegraphics[width=\linewidth]{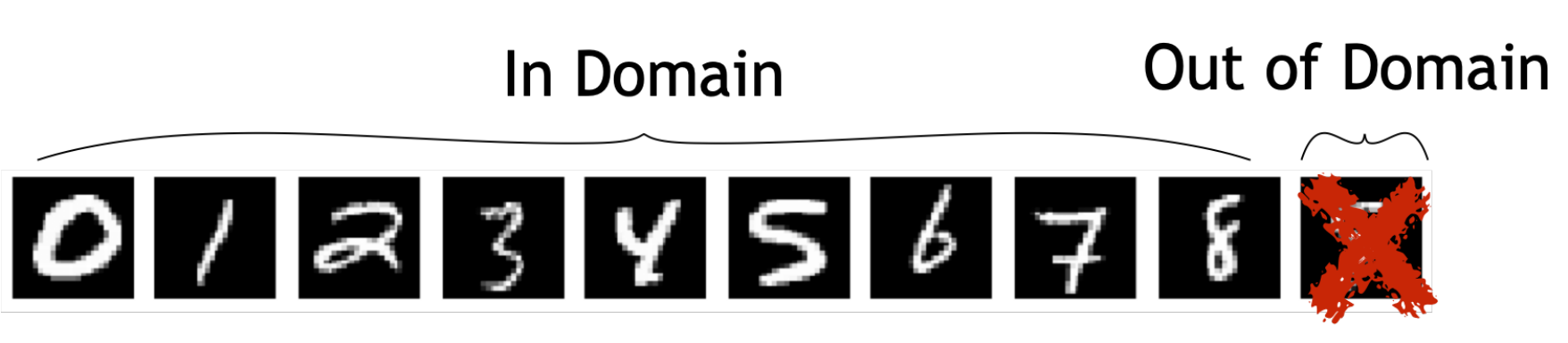}
  \caption{Illustration of the 9-class MNIST \acrshort{ood} detection task.  
  Digits $0$--$8$ are used for training and \acrshort{id} testing, while digit $9$ is held out as the \acrshort{ood} class.  
  \reprofrom{brueckerhoff2024chaoticlight}}
  \label{fig:photonics:mnist-9}
\end{figure}

To demonstrate the feasibility of probabilistic inference on photonic hardware, we performed a simple \acrshort{ood} detection experiment on a modified MNIST dataset.  
Figure~\ref{fig:photonics:mnist-9} illustrates the setup: digits $0$--$8$ were used for training and \acrlong{id} testing, while digit $9$ was held out as the \acrshort{ood} class.  
This task highlights both classification accuracy and the ability of the network to signal uncertainty when facing unseen inputs.  

\begin{figure}[t]
  \centering
  \includegraphics[width=\linewidth]{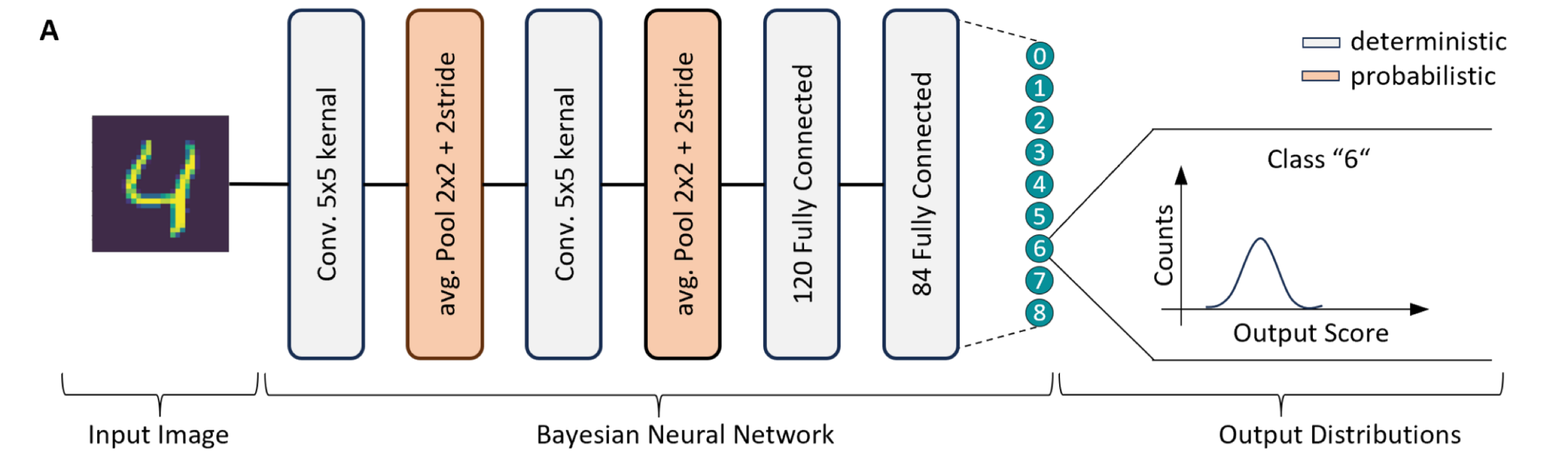}
	\caption{\acrshort{bnn} architecture for the MNIST \acrshort{ood} experiment.  
  Deterministic layers (gray) are combined with probabilistic average-pooling layers (orange), implemented using the probabilistic max-pool operator.  
  \reprofrom{brueckerhoff2024chaoticlight}}
  \label{fig:photonics:mnist-arch}
\end{figure}

The experiment was carried out with a compact \acrshort{bnn} tailored to the photonic prototype.  
Figure~\ref{fig:photonics:mnist-arch} shows the network architecture, which is based on a modified LeNet-5.  
Deterministic layers (gray) are combined with probabilistic average-pooling layers (orange) as introduced in the previous section.  
This design allows stochasticity to be injected into the activations through chaotic-light sampling while keeping the number of photonic weights small.  

\begin{figure}
  \centering
  \includegraphics[width=0.8\linewidth]{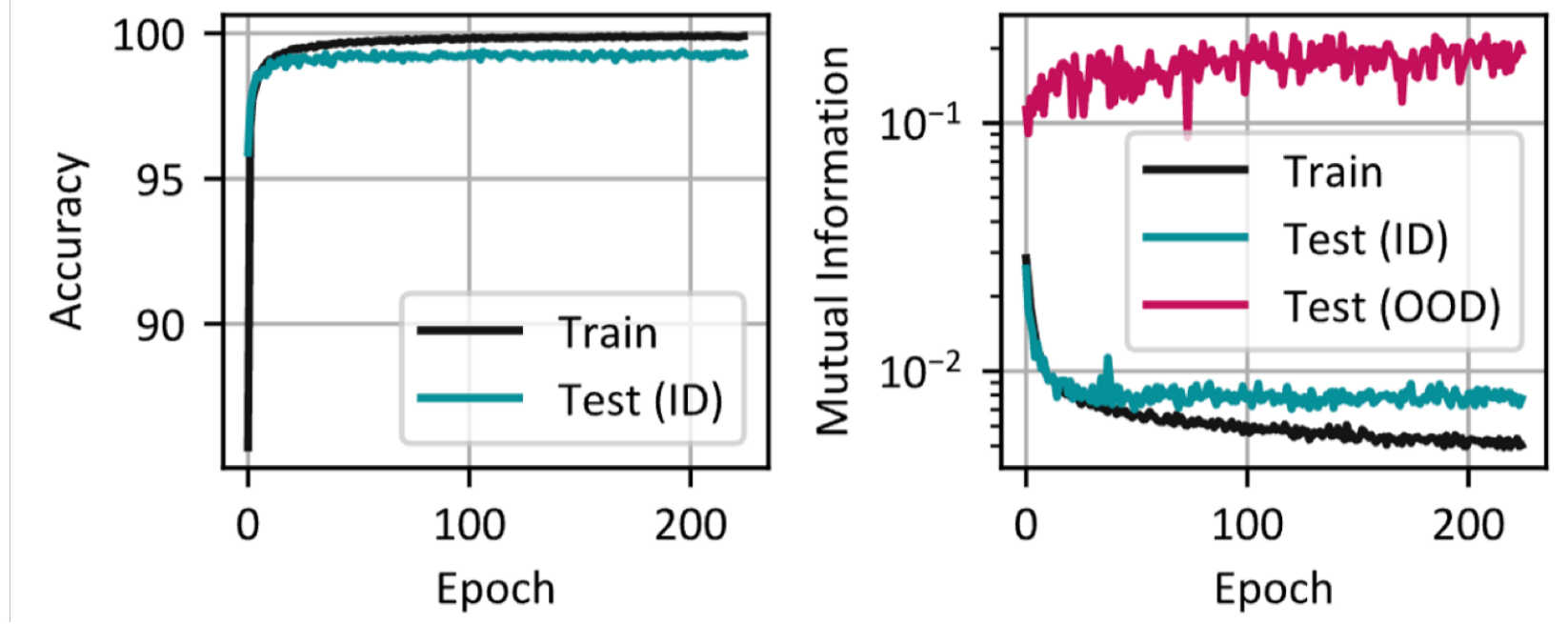}
  \caption{Training and test dynamics of the MNIST \acrshort{ood} experiment.  
	Left: accuracy on training and \acrshort{id} test data.  
	Right: separation between \acrshort{id} and \acrshort{ood} data in terms of \acrlong{mi}.  
  \reprofrom{brueckerhoff2024chaoticlight}}
  \label{fig:photonics:mnist-curves}
\end{figure}

Training dynamics further illustrate how the network learns both classification and uncertainty separation.  
As shown in Fig.~\ref{fig:photonics:mnist-curves}, accuracy on in-domain digits rises quickly, while the separation in \acrlong{mi} between \acrshort{id} and \acrshort{ood} inputs develops more gradually.  
This growing gap in \acrlong{mi} demonstrates that the photonic \acrshort{bnn} learns not only to classify seen digits but also to identify unseen ones through uncertainty estimates.  

\begin{figure}
  \centering
  \includegraphics[width=0.8\linewidth]{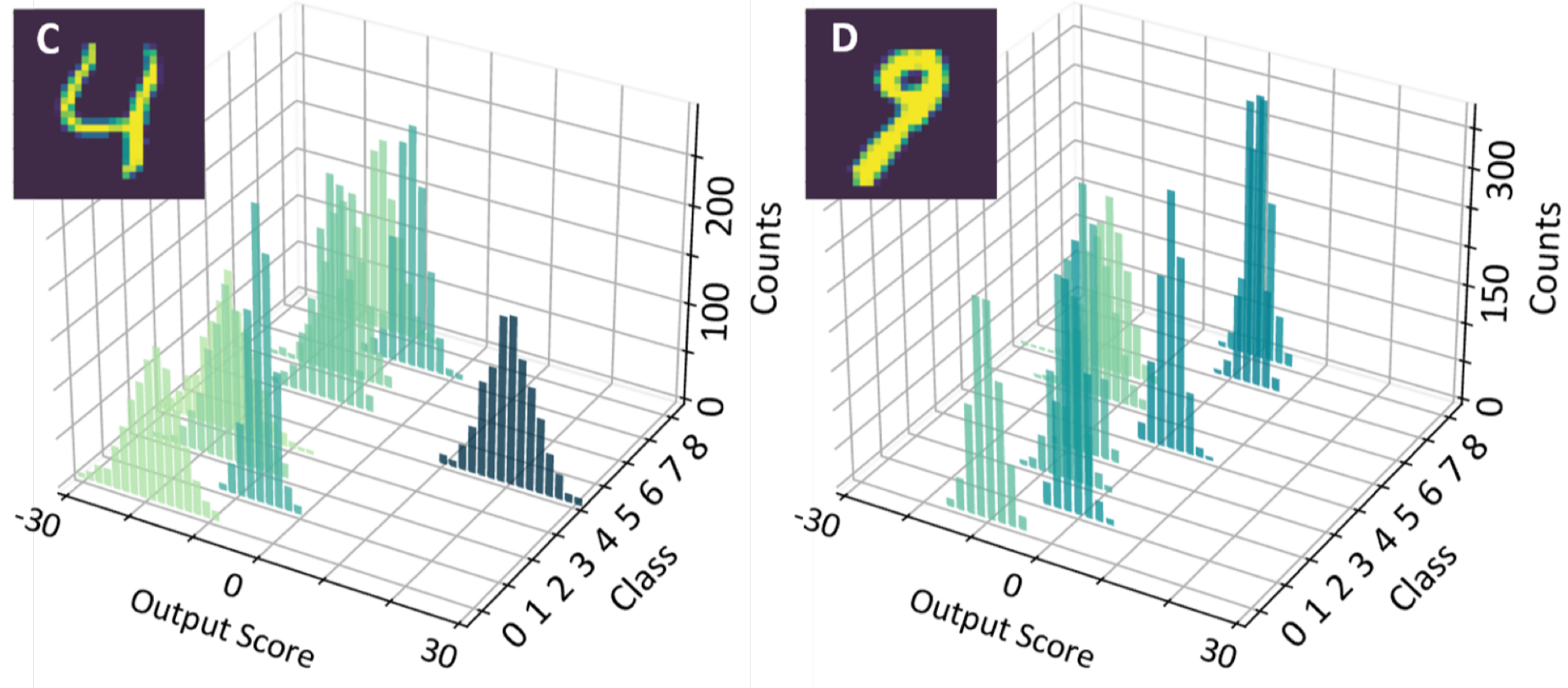}
	\caption{Predictive distributions for an \acrshort{id} (left) and an \acrshort{ood} (right) sample.  
	In-domain predictions are sharply peaked, while \acrshort{ood} predictions are spread out, reflecting uncertainty.  
  \reprofrom{brueckerhoff2024chaoticlight}}
  \label{fig:photonics:mnist-examples}
\end{figure}

To illustrate this effect qualitatively, Fig.~\ref{fig:photonics:mnist-examples} shows predictive distributions for a representative \acrlong{id} digit ($4$) and an \acrlong{ood} digit ($9$).  
The \acrlong{id} prediction is sharply peaked at the correct class, while the \acrshort{ood} prediction is broad and uncertain, consistent with the role of \acrlong{mi} as an \acrshort{ood} indicator.  

\begin{figure}
  \centering
  \includegraphics[width=0.4\linewidth]{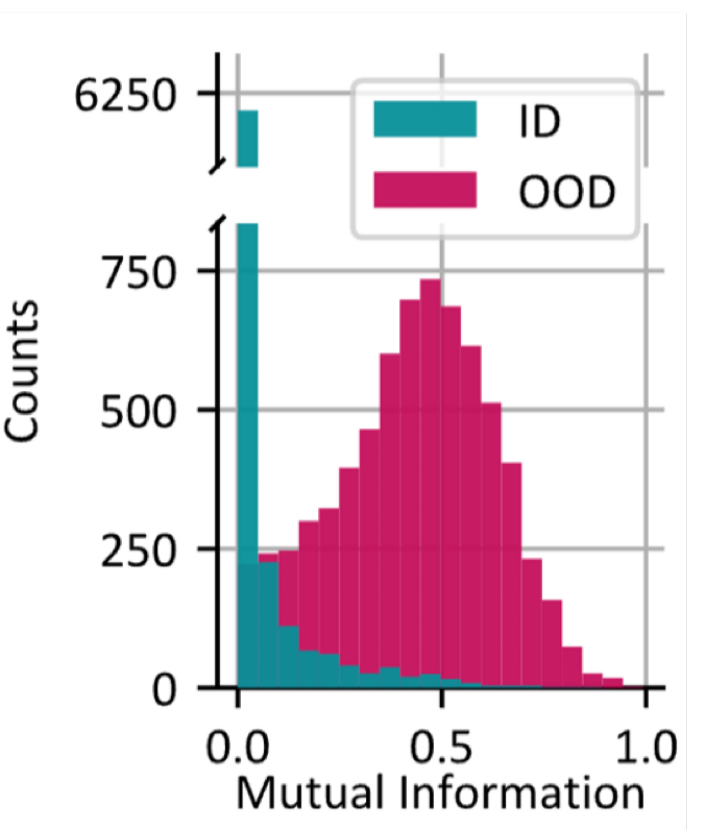}
  \caption{\Acrlong{mi} histogram for \acrshort{id} and \acrshort{ood} test samples.  
	The clear separation between the two distributions demonstrates the ability of the photonic \acrshort{bnn} to detect unseen data.  
  \reprofrom{brueckerhoff2024chaoticlight}}
  \label{fig:photonics:mnist-mi}
\end{figure}

The separation becomes most apparent when analyzing the full test set.  
Figure~\ref{fig:photonics:mnist-mi} demonstrates that \acrshort{id} and \acrshort{ood} examples form two clearly separated peaks in terms of \acrlong{mi}, enabling reliable \acrshort{ood} detection based on uncertainty.  

Together, these results confirm that probabilistic inference with photonic hardware is feasible.  
Even with the limitations of the prototype, the network achieves high accuracy on \acrlong{id} data and successfully identifies \acrlong{ood} samples via uncertainty estimates.  
The experiment demonstrates that chaotic-light-based activations provide a reliable entropy source for \acrlongpl{bnn} and that photonic accelerators can deliver uncertainty-aware inference.  

\section*{Summary}

Building on the algorithmic foundations of probabilistic inference and the modeling of analog noise in previous chapters, this chapter has explored the feasibility of implementing \acrlongpl{bnn} on photonic hardware.  
The motivation arises from reinterpreting the existing tension between the high computational cost of probabilistic computing in digital inference and the intrinsic, yet typically suppressed, noise present in analog accelerators.  

The core idea developed here is to invert this perspective and exploit the intrinsic stochasticity of photonic systems as a computational resource.  
In particular, chaotic-light intensity fluctuations, when made controllable, provide a natural entropy source for probabilistic inference.  
Coupled with a \acrshort{pcm}-based crossbar for in-memory multiplication enables a form of photonic probabilistic computing that directly supports \acrshortpl{bnn}.  

The specific challenges we faced were twofold.  
First, the noise needed to be statistically controllable; otherwise, the stochasticity of chaotic light would remain an unusable disturbance.  
We addressed this by exploiting the intrinsic mean–variance relationship of Bose--Einstein statistics and developing a controllable-noise encoding scheme, which allowed the variance of activations to be tuned within defined bounds.  
Second, the \acrshort{bnn} had to be adapted to the restrictions of the hardware, including limited precision, a constant base noise floor, strictly positive signals, and the small number of available weight cells.  
Prior to this work, it was unclear whether such a constrained \acrshort{bnn} would be feasible at all.  

We addressed these challenges through a series of design choices, which were consolidated in a probabilistic average-pooling operator implemented in Pyro.  
This operator realizes randomness at the level of activations rather than weights, incorporates \acrshort{qat} to align the network representation with the effective bit-width of the hardware, and uses a Gaussian approximation to the Bose--Einstein \acrshortpl{pdf} for efficient digital training.  
As a result, the developed operator captures the essential characteristics of the photonic prototype in a compact and trainable form, making it the central enabler to train \acrshortpl{bnn} for uncertainty-aware inference on photonic hardware.  

From this prototype, several key findings emerged.  
Chaotic light can be made into a tunable and practically useful entropy source.  
Photonic \acrshortpl{bnn} achieved both high in-domain accuracy and clear separation of in-domain and out-of-domain data on a 9-class MNIST task, demonstrating reliable uncertainty estimation.  
Most importantly, the study shows that careful algorithm–hardware co-design enables probabilistic inference even under the unconventional constraints of photonic devices.  

Looking ahead, this proof-of-concept suggests several promising directions for scaling photonic probabilistic computing.  
Larger crossbars and more complex operators, such as convolutions or attention mechanisms, could extend the scope beyond simple pooling layers.  
Closer integration of photonic and electronic domains may help to reduce interface bottlenecks and enable more comprehensive accelerator designs.  
More broadly, the results indicate that noise—long regarded as a limitation of analog computing—can be reframed as a computational resource, opening the possibility of future accelerators that combine speed, energy efficiency, and uncertainty awareness.

\clearpage{}%
\clearpage{}%
\chapter{Conclusion and Outlook}
\label{ch:conclusion}

\noindent
This work addresses two central challenges in modern machine learning:  
(i) how to enable \emph{resource-efficient inference} of neural networks on resource-constrained or emerging hardware platforms, and  
(ii) how to extend classical models towards \emph{trustworthy predictions} by incorporating uncertainty estimation.  

To this end, we study both \emph{deterministic and probabilistic neural networks}: classical \acrlongpl{dnn}, where efficiency is the primary goal, and their probabilistic counterpart, \acrlongpl{bnn}, which explicitly incorporate uncertainty quantification.  
For both paradigms, we pursue two strategies to improve efficiency.  
On the one hand, we reduce computational costs through algorithmic optimizations and code generation targeting digital processors.  
On the other hand, we explore analog accelerator prototypes, where operations can be realized more cheaply at the physical level, but robustness against nonidealities and tight hardware--algorithm co-design are essential.  

On \emph{digital processors for deterministic \acrshortpl{dnn}}, we advanced automatic model compression by combining latency measurements with sensitivity analysis to guide per-layer quantization and pruning.  
This work culminated in the \emph{Galen} framework, which establishes how guided policies can unify algorithmic compression and hardware-aware deployment, thereby making \acrshort{dnn} inference practical even on resource-constrained embedded devices.  

For \emph{analog accelerators executing deterministic \acrlongpl{nn}}, we investigated how nonidealities such as device-level noise affect inference.  
We developed simplified hardware models that were validated on \acrshort{bss2} prototypes and revealed partially surprising imperfections, such as non-associativity in dot products.  
These models enabled targeted training adaptations and also supported the development of diagnostic tools: in particular, Walking Noise proved highly effective at exposing layer-specific sensitivity.  
Building on these insights, we proposed hardening techniques to sustain accuracy under perturbations.  
In particular, \acrlong{vant} emerged as a simple yet impactful method, showing that neural networks can tolerate a wide range of hardware imperfections if these are explicitly incorporated during training.  

Turning to probabilistic models aimed at capturing uncertainty, namely \emph{\acrshortpl{bnn} on digital processors}, we addressed the high computational cost of Bayesian inference.  
This work developed scalable approximation and implementation techniques that enable efficient \acrshort{bnn} inference on embedded hardware.  
We began by analyzing \acrshortpl{bnn} and their core inference methods, \acrshort{mcmc} and \acrshort{svi}, evaluating their scalability and uncertainty estimation quality, and revealing how activation functions critically affect convergence and predictive behavior.  
Building on these insights, we introduced the \acrlong{pfp}, an extreme approximation of \acrlong{svi} that propagates distributions in closed form, thereby avoiding sampling and repeated forward passes.  
Within this work, we extended \acrshort{pfp} with compiler-based operator support and demonstrated its practical deployment on embedded processors.  
In addition, we investigated ensemble-based approaches such as Monte Carlo Dropout, Deep Ensembles, and \acrlongpl{rlle}, which integrate naturally with existing machine learning frameworks.  
Together, these methods illustrate how algorithmic approximations and compiler-based deployment can make uncertainty-aware inference feasible on embedded systems.

Finally, for \emph{\acrshortpl{bnn} on analog accelerators}, we combined probabilistic modeling with emerging photonic hardware.  
Here, we demonstrated how chaotic light can act as a physical entropy source, enabling ultrafast probabilistic computation and uncertainty estimation.  
This contribution establishes photonic processors as promising candidates for native \acrshort{bnn} inference, where hardware noise is harnessed as a computational resource rather than treated as an obstacle.  

Taken together, these contributions establish a unified perspective: resource-efficiency and trustworthiness can be advanced hand-in-hand across deterministic and probabilistic paradigms, by aligning algorithmic methods, compiler technology, and hardware-aware training for both digital and analog accelerators.

\section{Discussion of Key Insights}

The following discussion distills the key insights that emerged across the presented studies.  
Rather than reiterating individual contributions, it focuses on the overarching principles and trade-offs that shape the design space of resource-efficient inference.

A first overarching insight is that efficiency cannot be achieved by algorithms in isolation.  
While compression techniques such as pruning and quantization provide the levers, their impact depends critically on the way networks are mapped to hardware.  
Latency measurements and sensitivity analysis proved far more reliable than simple cost metrics, underscoring that the operating point of compressed models must be determined by actual hardware behavior.  
The broader lesson is that efficiency emerges from aligning algorithmic strategies with hardware mapping, while balancing the inevitable trade-off between accuracy and efficiency that bounds how far compression can be pushed.  

On analog accelerators, the central challenge is not computational cost but robustness against nonidealities.  
The key insight from modeling analog hardware is that neural networks can tolerate substantial imperfections---noise, drift or nonlinearities---if these effects are systematically embedded into the training process.  
Diagnostic tools such as Walking Noise proved effective in revealing sensitivity patterns, guiding adaptations, and, most importantly, in clarifying how \acrshortpl{dnn} learn to tolerate noise.  
A recurring theme was that robustness is best achieved through \emph{training with increasing complexity}, gradually exposing models to more realistic noise levels or hardware representations---an idea closely related to \emph{curriculum learning}~\cite{bengio2009curriculum}, where tasks are structured from simple to complex to facilitate stable and effective learning.
\acrshort{vant} exemplifies this principle: although simple in form, it provided a general and effective strategy for improving robustness in noisy settings.  
Taken together, these findings suggest that robustness on analog hardware depends less on eliminating imperfections and more on embracing them as part of the training process.  

For probabilistic models, the central challenge lies in reconciling uncertainty estimation with computational feasibility on digital processors.  
Sampling-based inference remains the quality benchmark, but its cost renders it impractical at scale, and even scalable variational methods, while viable with \acrshort{kl} annealing, continue to exhibit strong sensitivity to hyperparameters.  
The broader lesson is that \acrlongpl{bnn}, unlike their deterministic counterparts, remain bounded not only by efficiency constraints but also by the stability of their training dynamics.  

Beyond cost, however, two further insights proved critical for uncertainty quality itself.  
First, the choice of activation function has a surprisingly strong impact: our experiments showed that different nonlinearities can dramatically alter both the calibration and the out-of-distribution detection quality of \acrshortpl{bnn}, and that the optimal choice is often task dependent.  

Second, while single-mode variational inference remains among the most scalable approaches to Bayesian inference, its mean-field formulation imposes inherent limitations when confronted with the complex structure of real posterior distributions.  
Empirical and theoretical studies have shown that the true posteriors of \acrshortpl{bnn} are typically multi-modal, heavy-tailed, and strongly correlated~\cite{Wilson2020ProbPersp, izmailov2021whatBNNposteriors}, causing mean-field approximations to underestimate uncertainty or to miss multiple plausible explanations of the data—an effect that may also contribute to their pronounced sensitivity to hyperparameters and architectural design choices.  
Nevertheless, as demonstrated in Chapter~\ref{ch:bnns}, scalability can also provide a distinct advantage: by enabling the training of more expressive and deeper architectures, \acrshort{svi} allows representational power to partly offset the simplifying assumptions inherent to the inference scheme.  
Hence, although \acrshort{svi} may not fully capture the multi-modality of complex posteriors, suitably designed architectures can alleviate these shortcomings and still yield uncertainty estimates of high practical relevance at comparatively low computational cost.

Taken together, these observations illustrate that the practical value of \acrshort{svi} extends beyond its theoretical limitations.  
While mean-field inference restricts posterior expressiveness, its efficiency renders it particularly suitable for deployment on energy- and memory-constrained platforms.  
However, when moving from general-purpose accelerators to embedded systems, even relatively lightweight \acrshort{svi} variants become computationally demanding.  
To address this, two complementary yet fundamentally distinct strategies are examined: the \acrlong{pfp} as a minimalist variational approximation, and \acrshortpl{rlle} as ensemble-based alternatives offering comparable uncertainty quality at minimal computational cost.  

The \acrlong{pfp} remains within the variational framework and, by design, is single-modal.  
It provides a principled, calibration-efficient approximation but cannot capture the full expressiveness of multi-modal posteriors, and its specialized operations are not natively supported by current compiler stacks or hardware backends, which necessitated the implementation of a dedicated operator library within \acrshort{tvm}.  
By contrast, \acrshortpl{rlle} explicitly aim to preserve the multi-modality characteristic of Deep Ensembles through their repulsive multi-head formulation.  
They leverage dense operators that are already highly optimized in modern toolchains, enabling excellent speed and scalability, albeit at the expense of increased calibration sensitivity.  
Empirical comparisons reinforced this contrast: while \acrshortpl{rlle} consistently achieved lower latency and stronger \acrlong{ood} detection performance, they required careful tuning of repulsive samples, whereas \acrshort{pfp} remained more stable but fundamentally limited in posterior expressiveness and more challenging to deploy efficiently.

The broader principle is that uncertainty-aware inference on digital platforms is governed by a multi-dimensional trade-off: efficiency, calibration effort, posterior expressiveness, and theoretical grounding rarely align within a single method.
\acrshort{pfp} and \acrshortpl{rlle} thus occupy complementary positions in this design space, exemplifying the spectrum of viable pathways for embedding uncertainty estimation into resource-constrained systems.  

A further insight concerns the asymmetry between training and inference costs across Bayesian methods.  
At one extreme, \acrlong{mcdo} requires almost no additional probabilistic overhead during training, but typically demands many forward passes at inference, in practice exceeding the cost of \acrshort{mcmc} or \acrshort{svi}-based \acrshortpl{bnn}.  
At the other extreme, \acrshort{hmc} entails substantial training cost, yet inference can be reduced to evaluating a small, carefully chosen subsample of the chain.  
This can be interpreted as a theoretically grounded, well-selected ensemble.  
The comparison highlights that training and inference costs are not necessarily aligned.  
A method considered simple at training time, such as \acrshort{mcdo}, can become unexpectedly costly during inference, while a computationally expensive training method, such as \acrshort{hmc}, can yield relatively cheap predictions once samples are available.  
Which approach is preferable therefore depends on the specific constraints---whether efficiency during training or efficiency during deployment is the dominant bottleneck---underscoring that Bayesian methods must be evaluated across the entire pipeline rather than in a single phase.

When combining probabilistic models with analog accelerators, a striking insight emerges: hardware noise, traditionally seen as a limitation, can be harnessed as a useful source of stochasticity.  
In particular, photonic accelerators based on chaotic light demonstrate that entropy can be generated at extremely high speed directly in hardware, enabling native probabilistic computation driven by intrinsically rapid and low-cost photonic sampling rather than pseudo-random number generation.
This shifts the narrative from noise mitigation to noise exploitation, and highlights that in the analog domain, imperfections are not only unavoidable but can be productively integrated into the computational model.  

The broader lesson here is the necessity of hardware--algorithm co-design.  
The ability to use chaotic light as a physical entropy source did not arise from hardware in isolation, but from jointly adapting probabilistic models, training procedures, and experimental protocols to the specific statistical properties of the photonic device.  
Without such integration, leveraging noise for uncertainty estimation would not have been feasible.  

Taken together, these findings point to a clear principle.
Analog photonic devices illustrate that probabilistic inference need not be an add-on implemented in software but can emerge directly from the physics of computation.  
This suggests that future hardware platforms may move beyond deterministic acceleration of neural networks and instead provide native support for uncertainty estimation, blurring the boundary between algorithm and hardware.  

Taken together, these findings reveal two unifying themes that extend across deterministic and probabilistic models as well as digital and analog hardware.  
First, efficiency and robustness are consistently enabled by \emph{training with increasing complexity}: gradually introducing noise, hardware realism, or regularization such as \acrshort{kl} annealing proved essential for stability and generalization.  
Second, genuine progress requires \emph{algorithm--hardware co-design}, where compression strategies, probabilistic approximations, and robustness techniques are explicitly matched to the constraints and opportunities of the target platform.  
These principles illustrate that resource-efficiency and uncertainty-awareness cannot be addressed by isolated algorithmic innovation alone, but must emerge from a joint optimization of models, training schemes, and hardware execution.  
\section{Limitations}

While the results of this work demonstrate the feasibility and potential of resource-efficient and uncertainty-aware inference across digital and analog platforms, several limitations must be acknowledged.  

First, the scope of the models and datasets is deliberately restricted.  
Most experiments rely on relatively small networks and controlled benchmarks.  
The use of small models is motivated by both methodological and hardware constraints: for Bayesian studies, compact architectures are required to make baselines such as \acrshort{hmc} with \acrshort{nuts} and \acrshort{svi} computationally feasible, while for analog hardware, small networks are necessary to make underlying effects observable and to reflect the limitations of current prototypes.  
Although this enables careful experimentation and clear attribution of effects, it also limits the direct transfer of results to modern large-scale architectures, such as deep \acrlongpl{cnn} and transformers.  

Second, the conclusions regarding automatic model compression remain tied to specific hardware architectures.  
The experimental results presented here focus on embedded \acrshort{arm} \acrshortpl{cpu}, and the measured performance gains therefore reflect this class of processors in particular.  
While the absolute trade-offs observed are hardware-specific, the underlying approach is generic: built on the \acrshort{tvm} compiler stack, \emph{Galen} generalizes across the many architectures supported by \acrshort{tvm} and thus applies to a wide variety of accelerators beyond those explicitly studied in this work.  

Third, the robustness studies on analog accelerators relied on simplified hardware models to keep training efficient.  
This represents a trade-off: highly accurate models can capture device behavior more closely, but their complexity makes them too slow to be included in training loops, whereas simplified models strike a balance between realism and efficiency.  
These models, validated against \acrshort{bss2} prototypes, proved sufficient to reveal important nonidealities such as non-associativity in dot products and enabled targeted training adaptations.  

Complementing this, Walking Noise provided a powerful diagnostic that deepened our understanding of how noisy training affects learning and robustness.  
Its computational expense, however, prevents scaling to modern large architectures, highlighting the need for cheaper yet informative sensitivity metrics in future work.  

Fourth, while Bayesian inference methods were advanced in this work, their limitations must be carefully delineated.  
The \acrlong{pfp} is deployable in practice, as we provided a compiler-integrated library for embedded platforms, but its expressiveness remains bounded by the assumptions of \acrshort{svi}, in particular its single-mode nature.  
\acrshortpl{rlle} provide a pragmatic and highly efficient approximation, yet their reliance on calibration with repulsive samples---which are essentially artificial \acrlong{ood} data points---introduces certain limitations.  
The quality of the resulting uncertainty estimates depends critically on the expressiveness of these repulsive samples and on how well they resemble true \acrshort{ood} settings.  
This sensitivity highlights both the promise and the fragility of repulsion-based approximations in practical scenarios. 

More broadly, two general limitations of current \acrlongpl{bnn} became apparent.  
First, their uncertainty quality is highly sensitive to the choice of activation function, with different nonlinearities yielding substantially different calibration and \acrshort{ood} detection behavior.  
Second, approximating inherently multi-modal posteriors with single-mode methods, as in most variational inference approaches, remains fundamentally limited and highlights the need for richer yet efficient inference techniques.  

Finally, the photonic experiments represent proof-of-concept demonstrations.  
While chaotic light was successfully harnessed as a physical entropy source, scaling these approaches to larger networks and establishing true probabilistic weight representations remain open challenges--—but also areas of rapid progress as photonic devices continue to advance in scale and integration.  

Recognizing these limitations is essential for contextualizing the contributions of this work.  
They also provide a natural entry point into the outlook for future research, where the lessons learned here may be extended to larger models, more diverse hardware platforms, and more mature uncertainty estimation techniques.  
Especially transformer architectures and \acrlongpl{llm} would be highly interesting to investigate from a principled point of view, yet they also pose a substantial challenge for the probabilistic machine learning community.

\section{Outlook}

Acknowledging these challenges, a key direction for future work is to extend the evaluated methods—and \acrlongpl{bnn} more broadly—to large-scale architectures such as transformers, \acrlongpl{llm}, and modern diffusion-based generative models.
The experiments in this work were deliberately restricted to smaller networks to allow controlled evaluation and comparison with high-quality baselines, but scalability remains a central open question in the Bayesian research community.
Partial Bayesianization~\cite{kristiadi2020partialbnns,sharma2023partialbnns} offers one promising path: rather than modeling all parameters probabilistically, only selected layers or subsets of weights carry uncertainty, thereby balancing tractability with uncertainty quality.  
This idea was explored to through \acrshortpl{rlle}, which retain the efficiency of deterministic backbones while enriching the output distribution with repulsive multi-head structures.  
Still, fundamental questions remain: which components should be Bayesian, what fraction of parameters is sufficient, and how can such structures be identified automatically?  
For large language models, these challenges are compounded by their sequential autoregressive inference, where uncertainty must be propagated token by token and efficiency is dominated by long-context predictions.  
Tackling these questions would make probabilistic methods more tractable at scale and move them closer to practical deployment in the architectures that define modern deep learning.  

The role of activation functions provides another promising avenue.  
Our results showed that uncertainty quality in \acrlongpl{bnn} depends strongly on the chosen nonlinearity, with performance varying substantially across tasks.  
Rather than selecting from a fixed set of functions, future work could focus on learning activation functions directly, for instance with approaches inspired by \acrlongpl{kan}~\cite{liu2025kan}.  
Such methods could produce activation functions adapted to both model and task, and in the longer term may open the door to Bayesian architecture search, offering new insight into what architectural properties make networks naturally suited for probabilistic inference.  

At the algorithmic level, opportunities arise to overcome the single-mode limitations of variational inference.  
The \acrlong{pfp} provides an efficient route to closed-form propagation, but its expressiveness remains constrained.  
A promising extension is to combine it with ensemble methods, enriching their multi-modal character with local variational information.  
Similarly, \acrlongpl{rlle} demonstrated high efficiency, but their reliance on repulsive sample calibration remains a central limitation.  
A way to train these ensembles without artificial \acrlong{ood} data would be highly desirable, yet the appropriate approach remains an open question.
Maybe alternative strategies, such as independently training ensemble heads or designing more natural forms of repulsion, could help to reduce calibration sensitivity and improve robustness.  

On the model compression side, the \emph{Galen} framework illustrates how automatic compression policies can be aligned with measured hardware performance, but also suggests natural extensions.  
Incorporating per-layer latency measurements and low-level hardware information would provide even more detailed feedback to guide compression decisions.
Beyond embedded \acrshortpl{cpu} and \acrshortpl{gpu}, extending the framework to platforms such as \acrshortpl{fpga} or \acrshortpl{asic} will require improved cost models, since direct hardware-in-the-loop profiling becomes infeasible at scale.  
Such developments would broaden the reach of automatic compression and further integrate algorithmic techniques with architectural considerations.  

In the area of robustness, diagnostics such as Walking Noise provided unique insight into layer-wise sensitivity and helped to understand how noisy training improves robustness.  
Its computational cost, however, makes it unsuitable for large architectures.  
What is needed in general is a cheap but informative sensitivity metric.  
Two promising directions emerge:  
the first seeks to conserve the overall noise budget and learn its optimal distribution across layers, thereby capturing relative sensitivity within a single training run rather than through repeated evaluations.  
The second applies \acrshort{kl}-divergence-based perturbations, analogous to the sensitivity analysis employed in \emph{Galen}, where the divergence between perturbed and baseline predictions serves as an efficient proxy for layer-wise robustness.  
Developing such metrics would combine interpretability with scalability, making robustness analysis more practical for modern neural networks.  

For analog accelerators, the statistical properties of device noise align naturally with \acrlong{svi}, since many noise processes follow Gaussian or other well-characterized parametric distributions.  
This makes variational methods a well-matched candidate for probabilistic inference on analog hardware.  
At the same time, enriching \acrshort{svi} with ensemble ideas could help to overcome its single-mode limitations, enabling richer multi-modal posteriors while retaining the efficiency advantages that make \acrshort{svi} attractive in this setting.  

More broadly, progress in Bayesian inference may depend on building bridges between methods.  
High-quality \acrshort{mcmc} posteriors can serve as a reference or teacher model to regularize or initialize \acrshort{svi}, while priors learned by \acrshort{svi} can accelerate \acrshort{mcmc} by providing informed starting points.  
First results suggest that knowledge distillation can be effective in transferring information from \acrshort{mcmc} to \acrshort{svi} \acrshortpl{bnn}.  
Exploring such transfers opens a wider design space in which different inference methods complement each other, combining the quality of \acrshort{mcmc} with the scalability of \acrshort{svi}.  

Finally, photonic accelerators represent one of the most promising frontiers for probabilistic machine learning.  
The demonstrations in this work established that chaotic light can serve as a physical entropy source, providing a fast and high-quality basis for native probabilistic computation.  
Future generations of photonic hardware are expected to offer greater scale, finer controllability, and tighter integration with mainstream machine learning frameworks.  
At the same time, they will raise new challenges, including the tolerance of accumulating base noise in deeper architectures and the development of training methods that can adapt to device-specific noise processes.  
Sustained progress in this area will depend on close hardware–algorithm co-design, ensuring that advances in photonic devices and probabilistic modeling reinforce each other as the technology matures.

\section*{Concluding Remarks}

The questions that motivated this work lie at the core of modern machine learning:  
how to make neural networks more resource-efficient, and how to make their predictions more trustworthy.  
These challenges are far from solved and will continue to occupy the field for years to come.  
Yet, the studies presented here provide concrete steps toward these goals, demonstrating how efficiency and robustness can be advanced together when hardware, compilers, and algorithms are considered as parts of a unified system.  

A recurring theme throughout this work is that the most effective strategies tend to embrace, rather than avoid, complexity and imperfection.  
Training with increasing realism---through progressive noise schedules, hardware-model refinement, or \acrshort{kl} annealing---consistently enabled models to adapt and generalize under challenging conditions.  
Similarly, hardware--algorithm co-design proved essential: performance improvements arose not merely from better models, but from aligning those models with the computational structure and operational constraints of the hardware that realizes them.  
These insights suggest a broader view of neural network design in which computation, approximation, and physical realization are treated as a single, integrated system.  

Looking forward, the principles established here provide a foundation for the next generation of machine learning systems---systems that are both efficient and trustworthy, and that blur the boundary between software and hardware.  
Whether applied to large-scale models or to emerging analog accelerators, the same guiding ideas hold: leverage the structure of the hardware, expose imperfections during training, and design algorithms that learn to thrive within their physical constraints.  
By following these principles, the field moves closer to neural networks that not only compute efficiently, but also reason with confidence about what they do not know.

\clearpage{}%

\newpage
\thispagestyle{empty}
\null
\newpage

\pagestyle{plain}
\clearpage{}%

\begin{acknowledgements}      

\begingroup
\setstretch{1.25}

First and foremost, I would like to express my deepest gratitude to my supervisor, Professor Dr. Holger Fröning.  
I have been very fortunate to benefit from his guidance over the past years and from the opportunity to work on several novel and exciting research topics.  
At every stage, Holger provided excellent advice, generous support, and sincere as well as constructive feedback.  
I am deeply indebted to him for his continuous encouragement and for the many opportunities he offered throughout my doctoral studies.  

\vspace{0.8\baselineskip}

I would also like to thank my co-advisor, Professor Dr. Franz Pernkopf, for his professional expertise and invaluable advice.  
Franz enabled me to collaborate with his group members Dr. Wolfgang Roth and Sophie Steger, whose outstanding research and insightful discussions greatly enriched this work and deepened my understanding of probabilistic machine learning.  

\vspace{0.8\baselineskip}

It has been a true pleasure to work within the \emph{HAWAII} research group.  
I am sincerely grateful to Dr. Günther Schindler and Dr. Lorenz Braun for their mentorship, their openness to share expertise, and for the many insightful discussions that shaped the early stages of my research.  
I would also like to thank Andrea Seeger, Dr. Felix Zahn, Dr. Vahdaneh Kiani, and Dr. Jonas Dann for their valuable feedback, friendly collaboration, and for creating an inspiring and supportive research atmosphere during this time.  
Over the years, our group evolved, and I feel genuinely fortunate to now work side by side with my current colleagues---Hendrik Borras, Xiao Wang, Daniel Barley, Dr. Kazem Shekofteh, Aleksandra Poreba, Alexandra Stehle, and Robin Janssen.  
The close collaboration, open exchange of ideas, and the joyful daily interactions within this team have made my recent years both deeply rewarding and truly enjoyable.  
I am especially thankful to Hendrik for his reliability and inspiring spirit in our joint projects.  

\vspace{0.8\baselineskip}

Throughout my time at the institute, I had the pleasure of collaborating with several excellent students---Lisa Kuhn, Torben Krieger, Falk Selker, Christian Simonides, and Jonathan Bernhard---whose master's theses contributed significantly to this dissertation.  
Their enthusiasm and commitment made these collaborations both inspiring and enjoyable.  

\vspace{0.8\baselineskip}

I gratefully acknowledge the financial support provided under the scope of the COMET program within the K2 Center “Integrated Computational Material, Process and Product Engineering (IC-MPPE)” (Project No.~886385).  
This program is funded by the Austrian Federal Ministries for Climate Action, Environment, Energy, Mobility, Innovation and Technology (BMK) and for Labour and Economy (BMAW), represented by the Austrian Research Promotion Agency (FFG), as well as by the federal states of Styria, Upper Austria, and Tyrol.  

\vspace{0.8\baselineskip}

Finally, I would like to express my heartfelt gratitude to my family and friends for their unwavering support, patience, and belief in me throughout this demanding and exciting journey.  
Your encouragement has been a constant source of motivation.

\endgroup

\end{acknowledgements}
\clearpage{}%
\clearpage{}%

\begin{declaration}

This declaration is made in accordance with Heidelberg University's guidelines on the responsible and transparent use of generative artificial intelligence in academic writing.

\vspace{1.5em}

In the preparation of this dissertation, large language model (LLM)–based generative AI tools---specifically \textit{ChatGPT-5} (OpenAI, San Francisco, CA, USA)---were employed solely to support linguistic refinement and the enhancement of readability, as well as to assist in the formulation and structuring of the text. 
All AI-assisted passages have been critically reviewed, revised, and approved by the author.

\vspace{1.5em}

All scholarly and scientific contributions---including the conception of ideas, development of methodology, design and execution of experiments, data analyses, and formulation of conclusions---are entirely the author's own work, unless explicitly 
stated otherwise.

\vspace{1.5em}

\end{declaration}

\clearpage{}%
\printglossary[type=\acronymtype,style=long,nogroupskip,nonumberlist]
\begin{spacing}{1}

\cleardoublepage

\printbibliography[heading=bibintoc, title={References}]

\end{spacing}

\end{document}